\documentclass[twoside,11pt]{article}

\usepackage{jmlr2e}

\usepackage{amsmath, amssymb, graphicx, algorithm2e, algorithmic, bm}
\usepackage[]{natbib}
\usepackage{times}  

\usepackage[english]{babel}
\usepackage[utf8]{inputenc}

\usepackage[T1]{fontenc}    % use 8-bit T1 fonts
\usepackage{url}            % simple URL typesetting
\usepackage{booktabs}       % professional-quality tables
\usepackage{amsfonts}       % blackboard math symbols
\usepackage{nicefrac}       % compact symbols for 1/2, etc.
\usepackage{microtype}      % microtypography

\usepackage{color}
\usepackage{subfig}% use \subfigure, \subtable instead

\usepackage{mathtools}
\usepackage{tcolorbox}%for grey box

\usepackage{floatrow,fr-subfig}
\usepackage{dsfont}  %for big one
\usepackage{framed}

\usepackage[makeroom]{cancel} %to cross cancel terms that simplify

\newcommand{\R}{\mathbb{R}}  
\newcommand{\E}{\mathbb{E}}
\renewcommand{\P}{\mathbb{P}}
\newcommand{\N}{\mathbb{N}}
\newcommand{\Q}{\mathbb{Q}}

\newcommand{\bigo}{\mathcal{O}}

\newcommand{\ONE}{\mathds{1}}

\usepackage{subfig}
\newsavebox{\measurebox}

\usepackage{lastpage}
\jmlrheading{21}{2020}{1-\pageref{LastPage}}{12/17; Revised
	4/20}{9/20}{17-720}{Stefano Trac\`a, Cynthia Rudin, Weiyu Yan}
\ShortHeadings{Regulating Greed Over Time in Multi-Armed Bandits}{Trac\`a, Rudin, Yan}

\begin{document}

\title{Regulating Greed Over Time in Multi-Armed Bandits}

\author{
    \name Stefano Trac\`a \email stet@alum.mit.edu \\
    \addr Operations Research Center\\
    Massachusetts Institute of Technology\\
     Cambridge, MA 02139, USA
    \AND
    \name Cynthia Rudin \email cynthia@cs.duke.edu \\
    \addr Department of Computer Science\\
        Duke University\\
        Durham, NC 27708, USA
    \AND 
    \name Weiyu Yan \email weiyu.yan@alumni.duke.edu \\
    \addr Department of Electrical and Computer Engineering\\
        Duke University\\
        Durham, NC 27708, USA
}
	
\editor{Shie Mannor}

\maketitle

\begin{abstract}
In retail, there are predictable yet dramatic time-dependent patterns in customer behavior, such as periodic changes in the number of visitors, or increases in customers just before major holidays. The current paradigm of multi-armed bandit analysis does not take these known patterns into account. This means that for applications in retail, where prices are fixed for periods of time, current bandit algorithms will not suffice.
This work provides a remedy that takes the time-dependent patterns into account, and we show how this remedy is implemented for the UCB, $\varepsilon$-greedy, and UCB-L algorithms, and also through a new policy called the variable arm pool algorithm. 
In the corrected methods, exploitation (greed) is regulated over time, so that more exploitation occurs during higher reward periods, and more exploration occurs in periods of low reward. In order to understand why regret is reduced with the corrected methods, we present a set of bounds that provide insight into why we would want to exploit during periods of high reward, and discuss the impact on regret. Our proposed methods perform well in experiments, and were inspired by a high-scoring entry in the Exploration and Exploitation 3 contest using data from Yahoo$!$ Front Page. That entry heavily used time-series methods to regulate greed over time, which was substantially more effective than other contextual bandit methods.
\end{abstract}


\section{Introduction}

\begin{figure}
	\caption{Trend of English users shopping online in the weeks prior to Christmas. Retailers can observe clear weekly patterns in sales. In particular, customers tend not to shop right before Christmas and on Christmas day, but they delay their purchases to Boxing Day. When using a multi-armed bandit approach, a retailer would not want to be in exploration phase when there is a peak in visits like the one observed on Boxing Day. Source: ispreview.co.uk.}
	\centering
	\includegraphics[width=12cm,height=7.5cm]{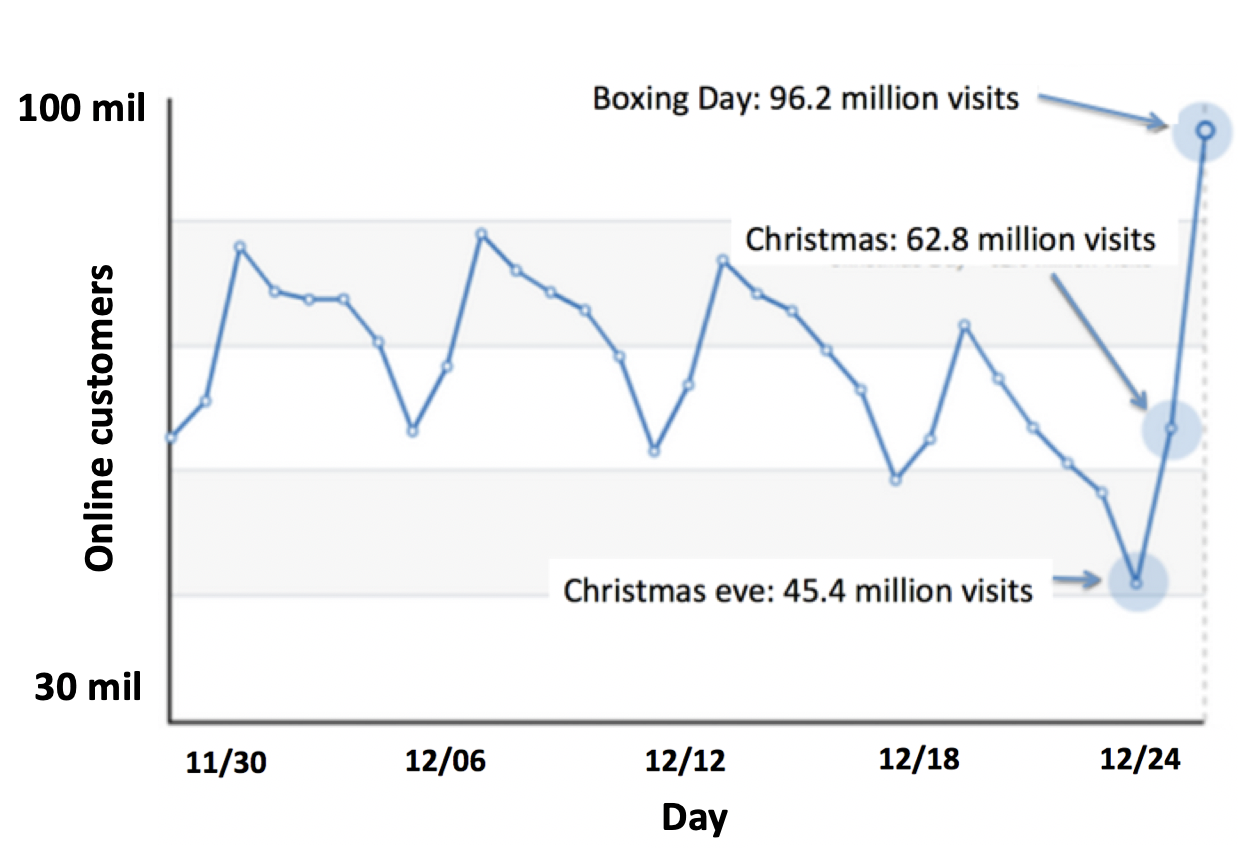}
	\label{boxing_day}
\end{figure}

\begin{figure}
	\caption{Google searches for the word ``scarf'' and the word ``strawberry'' over a ten-year period. Retailers can observe clear yearly patterns in the quantity of searches. When using a multi-armed bandit approach, a retailer could use the periods with low searches to explore new strategies (such as price, or location of the items in a store, or coupons), and then exploit the information gained during high seasons. Source: Google Trends.}
	\centering
	\includegraphics[width=12cm,height=7.5cm]{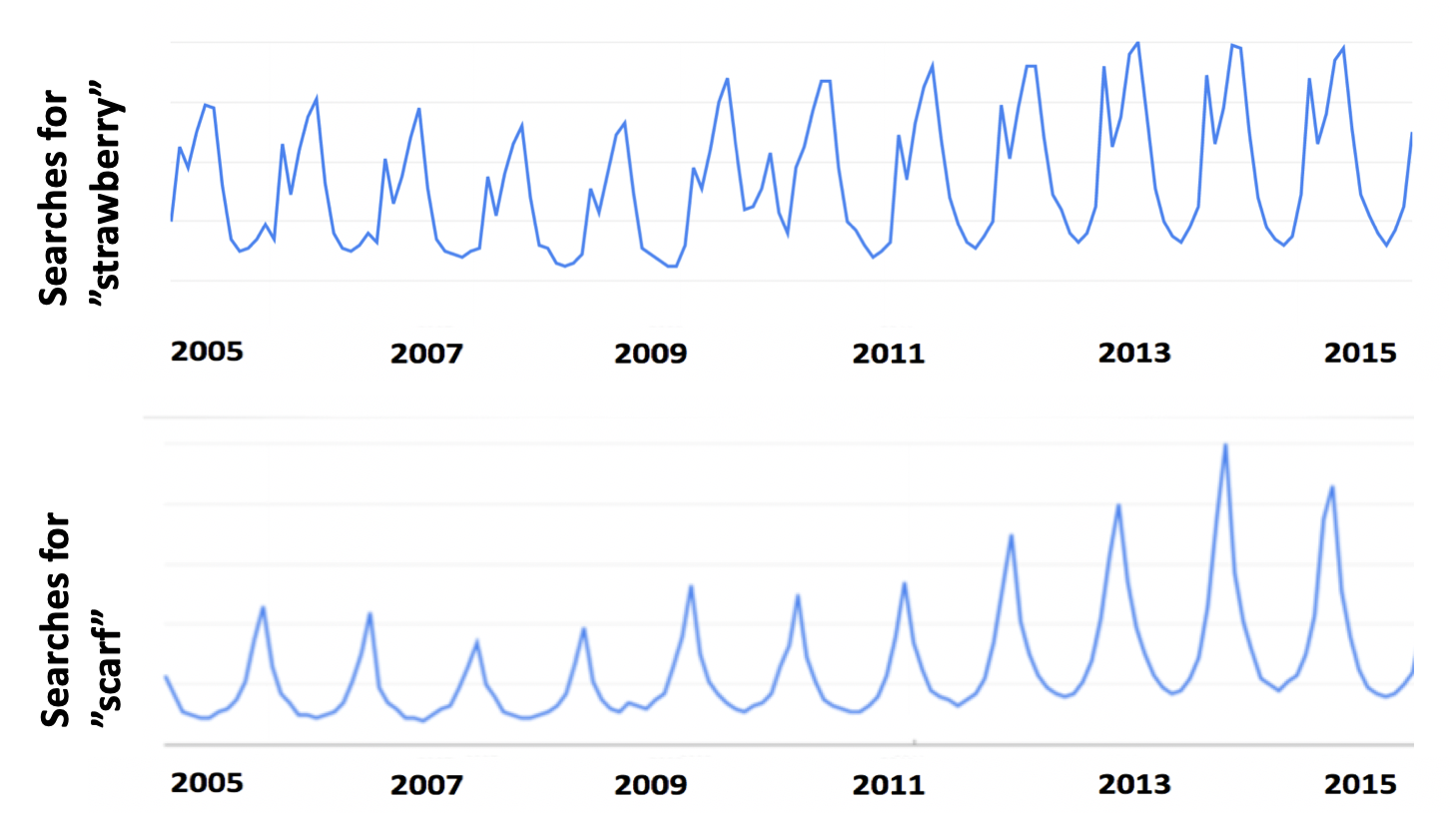}
	\label{google_trends}
\end{figure}

Consider a classic pricing problem faced by retailers, where the price of a new product on a given day is chosen to maximize the expected profit. The optimal price is learned asymptotically through a mix of exploring various pricing choices and exploiting those known to yield higher profits, potentially through the use of a multi-armed bandit (MAB). The retailer is not permitted to change the price for the day once it has been set. The demand (the number of customers) is approximately known in advance, since we assume the retailer knows the daily trend of the number of customers over time that visit the store. This information can be leveraged in order produce a better exploration/exploitation scheme. For instance, if we know that many customers will come to the store on
the week before Christmas, we would not want to explore new prices on those days. We might even stop exploring all together during that week. Our setting violates the classic assumptions of random rewards with a static probability distribution that is typically considered in multi-armed bandits. Since the retailer cannot change the price more than once each day, daily rewards are correlated through the trends in customer behavior; a large number of customers on a given day means a possible larger reward for that day (but also a larger possible regret for that day if the price is chosen sub-optimally). If one uses a standard MAB algorithm in the case where trends are dramatic, the result could be bad; an example is the case where the number of customers at the store will have a predictably large spike on a given day (e.g., for boxing day in England, shown in Figure \ref{boxing_day}), where the classic MAB algorithm could choose a suboptimal price on that particular day for the purpose of exploration. 
For retailers, there are almost always clear trends in customer arrivals, and they are often periodic or otherwise predictable. Some examples are in Figure \ref{google_trends}. These dramatic trends might have a substantial impact on which policy we would use to price products.

%The main contributions of this work are: (i) A framework that illustrates when it is beneficial to stop exploration to favor exploitation. (ii) Algorithms that show how to adapt existing policies to regulate greed over time. These are: the  $\varepsilon$-z greedy algorithm with regulating threshold (Section \ref{Section::epsilon_threshold}, Algorithm \ref{Algorithm::epsilon_threshold}); 
%the $\varepsilon$-soft greedy algorithm (Section \ref{Section::soft_espilon}, Algorithm \ref{Algorithm::epsilon_soft}); 
%the UCB-z algorithm with regulating threshold (Section \ref{Section::Algorithm::UCB_threshold}, Algorithm \ref{Algorithm::UCB_threshold});  the UCB-soft algorithm (Section \ref{Section::Algorithm::UCB_soft}, Algorithm \ref{Algorithm::UCB_soft}); and the variable pool algorithm,  which regulates greed by creating a suitable pool of available arms based on the value of the multiplier function (Section \ref{Section::z_pool}, Algorithm \ref{Algorithm::z_pool}).
%(iii) Theoretical regret bounds for the above algorithms.
%(iv) Numerical comparisons (in Appendix \ref{Section::Experimantal_Results}). We compare to ``smarter" versions of the classic $\varepsilon$-greedy algorithm (Algorithm \ref{Algorithm::epsilon_slightly_smarter}) and UCB algorithm (Algorithm \ref{Algorithm::UCB_slightly_smarter}).
The main contributions of this work are: i. A new framework that illustrates when it is beneficial to stop or limit exploration to favor exploitation; ii. Algorithms that show how to adapt existing policies to regulate greed over time, and a new algorithm (variable arm pool) that regulates greed over time; iii. Theoretical regret bounds for the algorithms;
iv. Numerical comparisons, both in a simulated environment (Section \ref{Section::Experimental_Results} and Appendix \ref{Appendix::Experiments}) and in a real-data environment  (Section \ref{Section::Experimental_Results_real_data} and Appendix \ref{Appendix::Experiment details}).
To help the reader navigate the algorithms and their theoretical results, we listed the detailed contributions in Table \ref{Table::contents}. 
We compare the performance of our algorithms to ``smarter''  versions of the classic $\varepsilon$-greedy algorithm (Algorithm \ref{Algorithm::epsilon_slightly_smarter}) and UCB algorithm (Algorithm \ref{Algorithm::UCB_slightly_smarter}). Since cumulative rewards are impacted by trends, the standard algorithms incorrectly estimate the mean rewards of the arms. The ``smarter''  versions fix this issue, and thus are a reasonable baseline to compare with. However, the ``smarter''  algorithms do not regulate greed over time and their performance is worse than the algorithms that do this regulation. We also show experiments when the reward multiplier is not known but is estimated using time series analysis.

In our setting, the behavioral information about customers is distilled so that it takes the form of a \textit{reward multiplier}  $G(t)$, where we assume $G(t)$ is known or can be well-estimated before the decision is made at time $t$. $G(t)$ should be thought of as the number of customers in the store on day $t$. If $G(t)$ is not known but could be well-approximated, the regret bounds weaken accordingly.

The new algorithms, that take advantage of knowing $G(t)$, are not a simple extension of the $\varepsilon$-greedy algorithm and the UCB algorithm. They anticipate the number of customers and choose how much exploration to allow at that timestep. Some multiplier functions will work better than others for regulating greed over time, but the theorems provided will reveal when the cumulative regret will be high due to the form of $G(t)$, which (as we mentioned) is known in advance.

As a result of the reward multiplier function, theoretical regret bound analysis of the multi-armed bandit problem becomes more complicated, because now the distribution of rewards depends explicitly on time. We not only care \textit{how many} times each suboptimal arm is played, but exactly \textit{when} they are played. For instance, if suboptimal arms are played only when the reward multiplier is low, intuitively it should not hurt the overall regret. A strength of our theorems is that they tell us when and when not to use our algorithms, depending on the multiplier function; for some multiplier functions we can determine in advance when the algorithms are likely to produce large regret.

%{\bf  ?? ADD SENTENCE THAT INTRODUCES THE MORTAL CASE TOO AND WHAT WE CAMPARE OUR ALG TO}

A Python implementation of the algorithms is available online.\footnote{
	The implementation of the algorithms used in the simulated environment is available at\\ \url{https://github.com/5tefan0/Regulating-Greed-Over-Time}\\
	The implementation of the algorithms used in the real data setting is available at \\ \url{https://github.com/ShrekFelix/Regulating-Greed-Over-Time}.}.

\normalsize
\begin{table}%[h!]     H.1Theorem::epsilon_threshold_estimatedG
	\makebox[\linewidth][c]{
		\begin{tabular}{| c | c | l | c |}%
			\hline
			%& & & \\
			{\bf Section} & {\bf Algorithm} & {\bf Related Theoretical Results} & {\bf Proof} \\ 
			%& & & \\   
			\hline 
			%& & &\\               
			\ref{Section::epsilon_threshold} &  $\varepsilon$-z greedy algorithm & Theorem \ref{Theorem::epsilon_z}: Regret bound for Alg. \ref{Algorithm::epsilon_threshold} & \ref{proof_epsilon1}\\
			%& & &\\
			&  & Theorem \ref{Theorem::epsilon_z_oreder_of_magnitude}: Reformulation of Theorem \ref{Theorem::epsilon_z} & \\
			%& & &\\
			&  & Corollary \ref{corollary_epsilon_threshold}: Corollary of Theorem \ref{Theorem::epsilon_z} & \\
			%& & &\\
			\hline  
			%& & &\\
			\ref{Section::soft_espilon} &  Soft $\varepsilon$-greedy algorithm & Theorem \ref{Theorem::regret_soft_epsilon_greedy}: Regret bound for Alg. \ref{Algorithm::epsilon_soft} &  \ref{proof_epsilon2}\\
			%& & &\\
			&  & \shortstack{Theorem \ref{Theorem::regret_soft_epsilon_greedy_order_of_magnitude}: Reformulation of Theorem \ref{Theorem::regret_soft_epsilon_greedy}} & \\
			%& & &\\
			\hline
			%& & &\\
			\ref{Section::Algorithm::UCB_threshold} &  UCB-$z$ algorithm & Lemma \ref{Lemma::minimum_pulls_UCB}: Minimum pulls with UCB &  \ref{Appendix::UCBz}\\
			%& & &\\
			&  & Theorem \ref{Theorem::lower_bound_x_n}: Lower bound for Lemma \ref{Lemma::minimum_pulls_UCB} & \\
			%& & &\\
			&  & Theorem \ref{Theorem::regulated_UCB}: Regret bound for Alg. \ref{Algorithm::UCB_threshold} & \\
			%& & &\\
			&  & Corollary \ref{Lemma::minimum_pulls_UCB_different_ranges}: Extension of Lemma \ref{Lemma::minimum_pulls_UCB} & \\
			%& & &\\
			&  & Corollary \ref{Theorem::lower_bound_x_n_different_range}: Lower bound for Corollary \ref{Lemma::minimum_pulls_UCB_different_ranges} & \\
			%& & &\\
			& & Corollary \ref{Corollary::P-UCB_x_n}: Corollary of Corollary \ref{Theorem::lower_bound_x_n_different_range} &\\
			%& & &\\
			\hline
			%& & &\\
			\ref{Section::Algorithm::UCB_soft} & Soft UCB algorithm & Theorem \ref{Theorem::soft_UCB}: Regret bound for Alg. \ref{Algorithm::UCB_soft} &  \ref{proof_UCB2}\\
			%& & &\\
			&  & Theorem \ref{Theorem::soft_UCB_order_of_magnitude}: Reformulation of Theorem \ref{Theorem::soft_UCB} & \\
			%& & &\\
			&  & Lemma \ref{Lemma::minimum_pulls_UCB_soft}: Minimum pulls  with Soft UCB & \\
			%& & &\\
			&  & Lemma \ref{Lemma::minimum_pulls_UCB_soft_different_ranges}: Extension of Lemma \ref{Lemma::minimum_pulls_UCB_soft} & \\
			%& & &\\
			\hline
			%& & &\\
			\ref{Section::z_pool} &  Variable pool algorithm & Theorem \ref{Theorem::regret_variable_pool}: Regret bound for Alg. \ref{Algorithm::z_pool} &  \ref{Appendix::variable_pool}\\
			%& & &\\
			\hline
			%& & &\\
			\ref{Section::UCB_soft_mortal} &  Soft UCB mortal algorithm & Theorem \ref{Theorem::G_mortal}: Regret bound for Alg. \ref{Algorithm::UCB-LG} &  \ref{Appendix::UCB_mortal_regret_bound}\\
			%& & &\\
			\hline
		\end{tabular}
		\caption{Location of algorithms and theoretical results. The proofs are in the Appendix.}
		\label{Table::contents}
	}
\end{table}
\normalsize
%The new algorithms perform substantially better than the ``smarter'' versions of the $\varepsilon$-greedy algorithm and the UCB algorithm (see Algorithm \ref{Algorithm::epsilon_slightly_smarter} and Algorithm \ref{Algorithm::UCB_slightly_smarter} in Appendix \ref{Section::Experimantal_Results}) where the rewards obtained at each round are properly calculated. This is because the algorithms we propose regulate greed over time.\\
%The multiplier function $G(t)$ is assumed to be known and bounded. If it is not, it can be estimated using should another source of data; e.g., Figures 1b and 1c show periodicity that is easy to predict for ``strawberries" and ``scarf" can be predicted at time $t$ using weather information. If $G(t)$ is not known precisely, similar results hold and we show how to modify the bounds.\\

\section{Related Work}
Multi-armed bandit algorithms were introduced by \citet[]{lai1985asymptotically}. There are several surveys and books that cover the many streams of work on multi-armed bandit problems \cite[e.g.,][]{bergemann2006bandit, cesa2006prediction, gittins2011multi, bubeck2012regret}.
%, see \citet[]{even2006action}, \citet[]{kleinberg2008multi},  \citet[]{audibert2010best}, \citet[]{mannor2011bandits}, \citet[]{seldin2011pac}, \citet[]{hazan2011better}, \citet[]{chu2011contextual}, \citet[]{slivkins2011contextual}, \citet[]{abbasi2011improved},  \citet[]{bubeck2012best},  \citet[]{kaufmann2012thompson},  \citet[]{arora2012online}, \citet[]{agarwal2012contextual},    \citet[]{bubeck2013bounded},  \citet[]{badanidiyuru2013bandits}, \citet[]{maillard2014latent}, \citet[]{agrawal2014bandits}, \citet[]{agarwal2014taming}, \citet[]{yahyaa2015thompson}, \citet[]{auer2016algorithm}, \citet[]{rakhlin2016bistro}, \citet[]{seldin2016lower}, \citet[]{syrgkanis2016efficient}.  

The setup of this work differs from other works considering time-dependent multi-armed bandit problems --  we do not assume the mean rewards of the arms exhibit random changes over time (they are static in our setting, as in the non-time-dependent problem), and we assume that the reward multiplier is known in advance.
% in accordance with what we observe in real scenarios. 
%No previous works that we know of consider regulating greed over time based on known reward trends. %Let  works that consider multi-armed bandits where rewards are non-stationary, all of them assuming the mean rewards for the arms can change non-deterministically over time (unlike our assumption). 
Other works consider different scenarios where reward distributions can change over time, but in a way that is not known in advance. For these settings, the algorithm needs to compensate for changes in the reward distribution after the change, rather than altering their strategy in advance of the change. 
Along these lines, \citet[]{liu2013learning} consider a problem where each arm transitions in an unknown Markovian way to a different reward state when it is played, and evolves according to an unknown random process when it is not played. \citet[]{garivier2008upper} presented an analysis of a discounted version of the UCB and a sliding window version of the UCB, where the distribution of rewards can have abrupt changes and stays stationary in between. \citet[]{besbes2014optimal}  considers the case where the mean rewards for each arm can change, where the variation of that change is bounded. \citet[]{slivkins2007adapting} consider an extreme case where the rewards exhibit Brownian motion, leading to regret bounds that scale differently than typical bounds (linear in $T$ rather than logarithmic). One of the works that is relevant to ours is that of \citet[]{chakrabarti2009mortal} that considers ``mortal bandits,'' where arms disappear or appear. In the mortal setting, we compare the performance of Algorithm \ref{Algorithm::UCB-LG} with the UCB-L Algorithm introduced in \citet[]{TracaRuYa2019}, which regulates exploration based on the remaining lifespans of the available arms.

A particularly interesting setting is discussed by \citet[]{komiyama2013multi}, where there are lock-up periods when one is required to play the same arm several times in a row. There is a similarity of that problem to the one studied here. In our setting, we fix the price of the product for an entire day, each day is a timestep, and there is not a separate timestep for each customer. If we were instead to take a timestep for each customer, we would need to lock the arm over the course of the day in order to keep the same price throughout the entire day. In other words, in our scenario, the micro-lock-up periods occur at each step of the game, and their effective lengths are given by $G(t)$. (The difference between these situations is that playing the same arm $G(t)$ times is not equivalent to playing an arm once and then multiplying the reward by $G(t)$; if we multiply a single step's reward by $G(t)$, we do not learn the same amount as if we had taken $G(t)$ pulls and viewed the regret $G(t)$ separate times.)
In the work of \citet[]{komiyama2013multi}, lock periods are present but there is no regulating greed based on the size of the lock periods.

The setting studied in \citet[]{bouneffouf2016multi} also consider the case of rewards that follow a known trend, but it is fundamentally different because in our setting the multiplier function is exogenous (in the sense that it does not depend on the arms). In \citet[]{bouneffouf2016multi}, the trend of the rewards of an arm depends on how many times the arm has been played (for example, people may not like a song the first time they hear it, but they may like it more after a few times), so there are no particular periods where it is more important to regulate greed and stop exploration, in order to exploit more during high-reward periods. 

%In some of the algorithms that we propose, when the rewards multiplier are above a certain threshold, we create lock-up periods ourselves, where exploration is stopped, but the best arm is allowed to change during high-reward zones as we gather information over rounds. %The algorithm we propose for step-function reward multipliers is reminiscent, since we propose to stop exploration all together during the ``high reward zone" and to simply play the arm with the highest expected reward.  
%The methods proposed in this work are contextual, in the sense that we consider externally available information in the form of time dependent trends. 

%Exploration-exploitation problems occur not only in retail (both in stores and online), but in marketing, on websites such as Yahoo$!$ Front Page, where the goal is to choose which of a set of articles to display to the user, on other webpages where ads are shown to the user on a sidebar (e.g., Facebook, Slashdot), and even on websites like YouTube that recommend the next video (and relative targeted ad) to watch. In all of these applications, 

The ideas in this paper were inspired by a high scoring entry in the Exploration and Exploitation 3 Phase 1 data mining competition, where the goal was to build a better recommendation system for Yahoo$!$ Front Page news articles. 
%\cite{RudinChSu15} 
At each time, several articles were available to choose from, and these articles would appear only for short time periods and would never be available again. One of the main ideas in this entry was simple yet effective: if any article gets more than 9 clicks out of the last 100 times we show the article, and keep displaying it until the clickthrough rate goes down. This alone increased the clickthrough rate by almost a quarter of a percent. In the Yahoo$!$ advertising problem, the high reward period was created by the availability of an article (an arm), which is different than the retail store case, but the same effect is present, where regulating the rate of exploitation (i.e., \emph{greed}) over time is beneficial to overall rewards. Here also, it is useful to stop exploring during times when a function like $G(t)$ is high.
%For retailers, the number of customers is much larger on certain days than others, and for these days we should exploit by choosing the best prices, and not explore. 
For Yahoo$!$ Front Page, articles have a short lifespan and some articles are much better than others, in which case, if we find a particularly good article, we should exploit by repeatedly showing that one, and not explore new articles. 
%For retail or online advertising, there are certain periods where customers are more likely to make a purchase, so prices should be set to optimal values during those times, and we should not be exploring suboptimal choices.
The framework here distills the problem, allowing us to isolate and study this effect of a time dependent function that we can use to regulate greed over time.

\section{Algorithms for regulating greed over time}
This section illustrates the problem, the proposed algorithms to regulate greed over time, and theoretical results on the bound on the expected regret of each policy. 

\subsection{Problem setup}\label{Section::problem-setup}
Formally, the stochastic multi-armed bandit problem with regulated greed is a game played in $n$ rounds. At each round $t$ the player chooses an action among a finite set of $m$ possible choices called \emph{arms}. For retail, these arms could be prices set once per day, coupons offered for a fixed time period, or what products to put on sale for the day (these are all actions chosen at time $t$ that will affect all $G(t)$ customers arriving within that round). Note that prices, coupons, and sales can not be chosen per customer, so the standard setting of the multi-armed bandit does not apply here. When arm $j$ is played ($j \in \{1, \cdots, m\}$) an \emph{unscaled} random reward $X_j(t)$ is drawn from an unknown distribution and the player receives the \emph{scaled} reward $X_j(t)G(t)$ where $G(t)$ is the \emph{multiplier function}. The distribution of $X_j(t)$ does not change with time (the index $t$ is just used to indicate the turn in which the reward was drawn), while $G(t)$ is a known function of time assumed to be bounded (this is, for instance, the number of customers in the store on day $t$). At each turn, the player suffers also a possible regret from not having played the best arm. The unscaled mean regret for having played arm $j$ is given by $\Delta_j=\mu_*-\mu_j$, where $\mu_*$ is the mean reward of the best arm (indicated by ``*'') and $\mu_j$ is the mean reward obtained when playing arm $j$; the scaled mean regret is given by $G(t)\Delta_j$. At the end of each turn the player can update their estimate of the mean reward $\mu_j$ and use it in the next turn $t$:
\begin{equation}\label{mean_estimator}
	%\widehat{X}_{j} = \frac{1}{T_j(t-1)}\sum_{s=1}^{T_j(t-1)} X_j(s),
	\widehat{X}_{j} = \frac{1}{T_j(t-1)}\sum_{s=1}^{t-1} X_j(s)\ONE_{\{I_s=j\}},
\end{equation}
where $T_j(t-1)$ is the number of times arm $j$ has been played before round $t$ starts, and $\ONE_{\{I_s=j\}}$ is an indicator function equal to $1$ if arm $j$ has been played at time $s$ (otherwise its value is $0$). This update will help the player in choosing a good arm in the next round. The total regret at the end of the game is given by
\begin{equation*}
	\text{Total regret} = \sum_{t=1}^n \sum_{j=1}^m \left(X_*(t) - X_j(t)\right) G(t) \ONE_{\{I_t=j\}}. 
\end{equation*}  
If the choice of the arm played at time $t$ is fixed a priori by the sequence $\mathcal{I} = \{I_t\}_{t=1}^n$, the total mean regret of a game $\mathcal{I}$ is defined as
\begin{equation*}
	R_n  =  \E_{X_1,\,\dots \,,X_m}\left[ \text{Total regret} \right] 
	=  \sum_{t=1}^n \sum_{j=1}^m \Delta_j G(t) \ONE_{\{I_t=j\}},
\end{equation*}
where the expectation is taken over the distribution of the rewards of each arm.
When the choice of the arm at time $t$ is not fixed but given a policy $\pi$, the expected cumulative regret is defined as $\E_{\pi}[R_n]$,
%\begin{equation*}
%	\E_{\pi}[R_n] = \sum_{t=1}^n \sum_{j=1}^m \Delta_j G(t) \P(I_t=j), 
%\end{equation*}
where now the expectation is taken over the distribution induced by the policy on the choice of the arms to play. When there is not confusion on the policy used, we simply indicate this quantity by $\E[R_n]$ and we simply call this quantity ``expected regret'' or ``mean regret''.
The strategies presented in the following sections aim to minimize the expected cumulative regret $\E[R_n]$ by regulating \emph{exploitation} (i.e., \emph{greed}) of the best arm found so far, and \emph{exploration}, based on the values of the multiplier function $G(t)$. In general, when the multiplier function is high, the player risks incurring a high regret if a bad arm is played. We show that it is beneficial to stop exploration in this situation  and resume exploration when rewards and regrets are lower. A complete list of the symbols used throughout the paper can be found in Appendix \ref{Section::notation}.

For any strategy (not just the ones in this paper), it is problematic if $G(t)$ is high early in the game. This would be analogous to opening a new store just before the sales peak at Christmas, and being expected to have already optimized the price at that time. In that case, no strategy could have explored enough to perform well. The strategies presented in this work are useful for cases where we can explore enough before $G(t)$ becomes large to determine what the optimal arm should be. When $G(t)$ is large, our strategies exploit. In the exposition that follows, we present the strategy as is, without imposing additional exploration to deal with high $G(t)$ at early times. This serves a useful purpose: it allows one to see directly in the bounds that forcing exploitation when $G(t)$ is high may not work early on. But in fact, \textit{any} algorithm, whether exploring or exploiting, will have a poor regret guarantee when $G(t)$ is high early on. Even early on, it still may be better in practice to exploit the best arm so far, rather than to try a risky arm when the stakes are high.

\subsection{Regulating greed with threshold using an $\varepsilon$-greedy algorithm}\label{Section::epsilon_threshold}

\begin{figure}
	\makebox[\textwidth][c]{ %to center figures!
		\begin{subfloatrow}
			\subfloat[\small{The \emph{Wave Greed}}]{{\includegraphics[width=4.0cm,height=4.5cm]{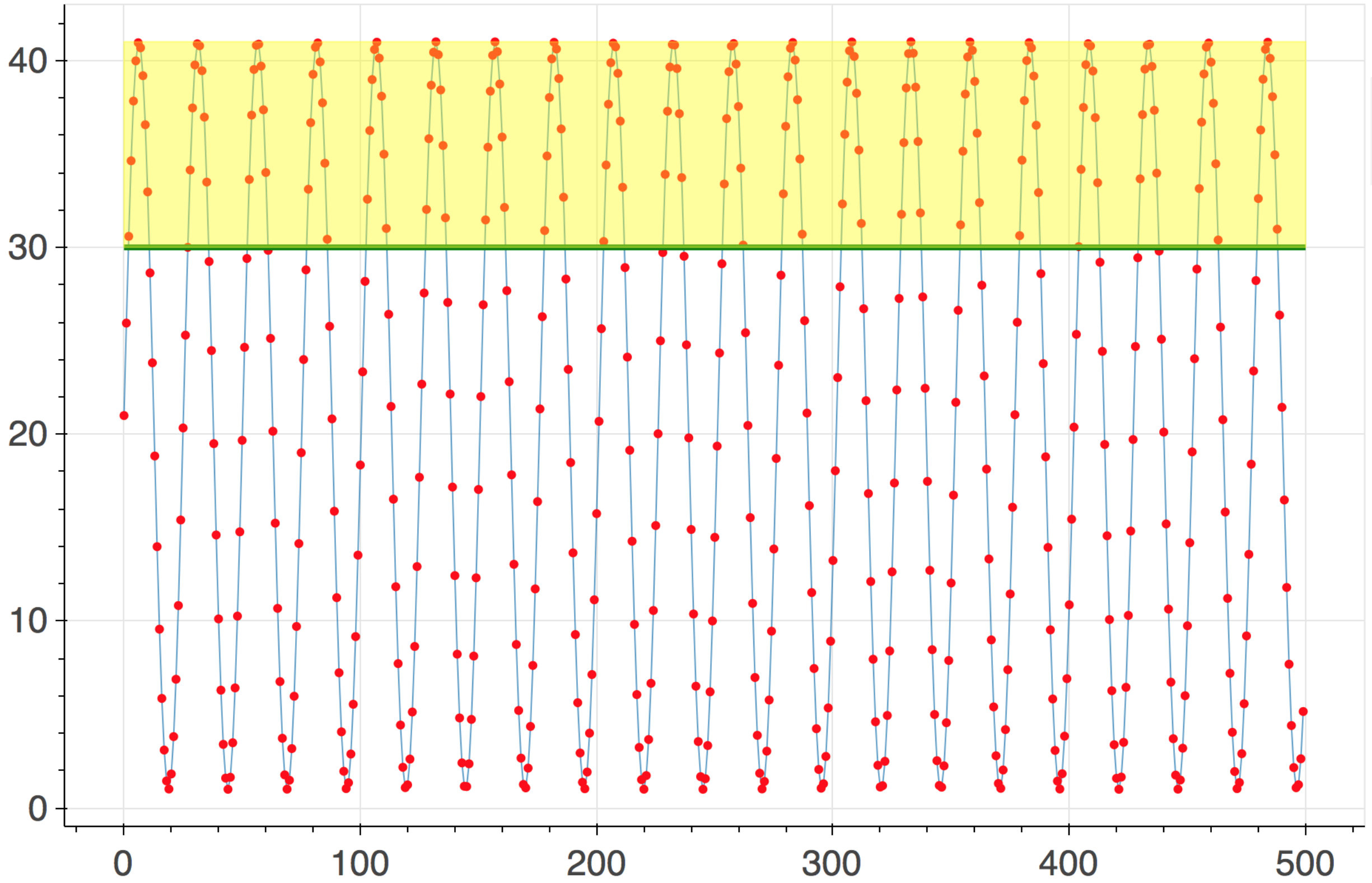} }\label{Figure::wave_z}}%
			\;\;
			\qquad
			\subfloat[\small{The \emph{Christmas Greed}}]{{\includegraphics[width=4.0cm,height=4.5cm]{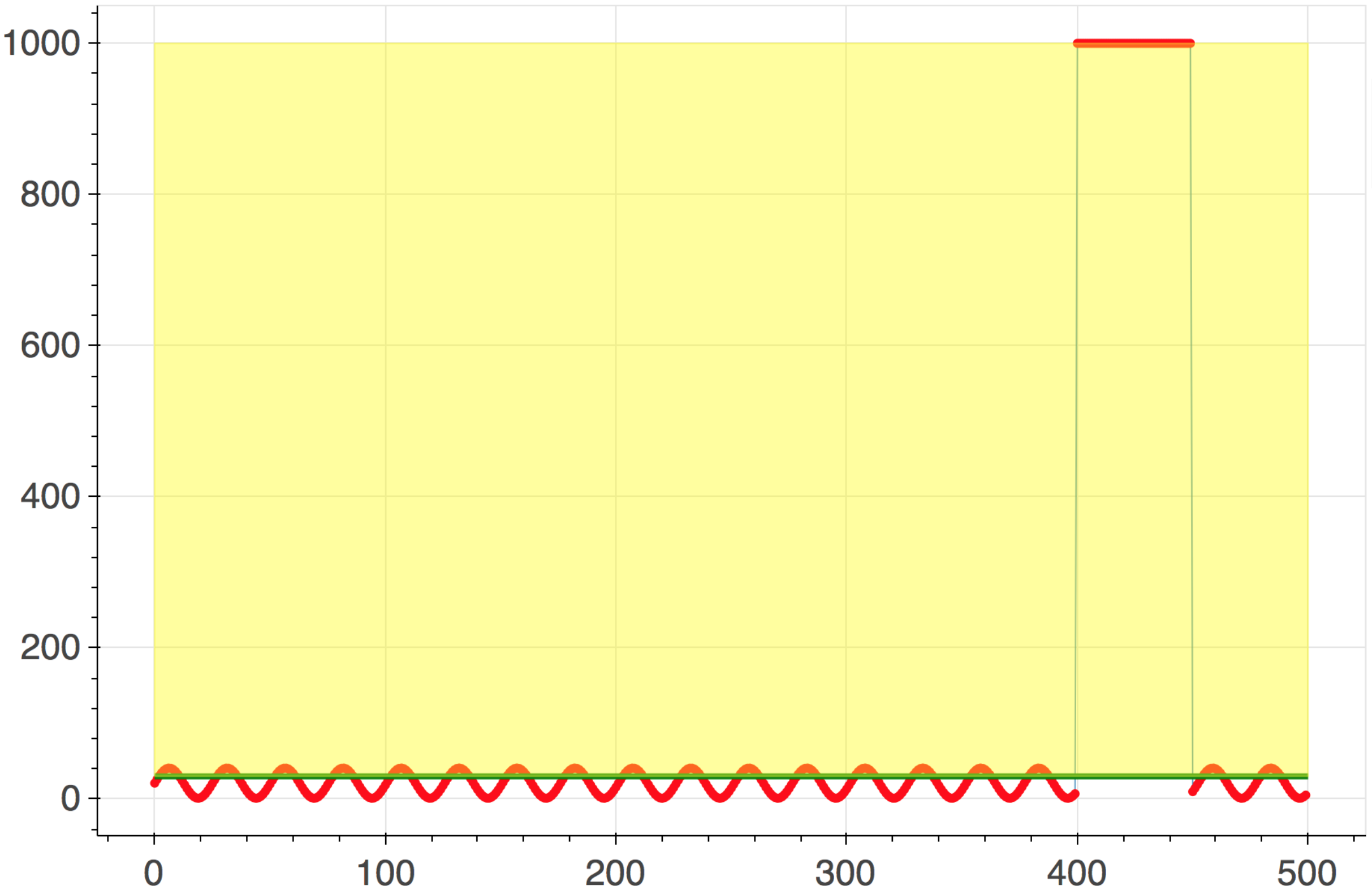} }\label{Figure::christmas_z}}\;\;
			\qquad
			\subfloat[\small{The \emph{Step Greed}}]{{\includegraphics[width=4.0cm,height=4.5cm]{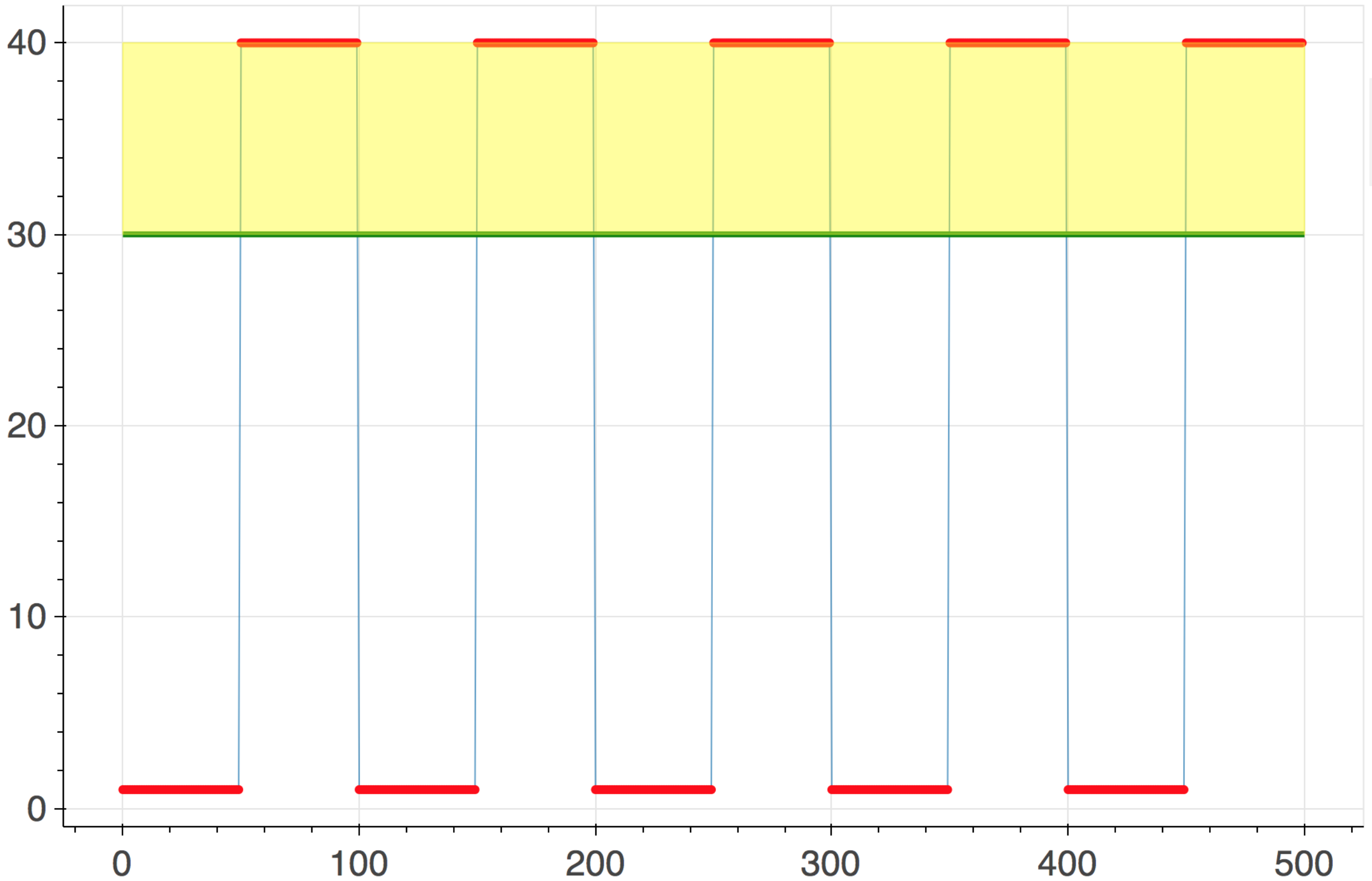} }\label{Figure::step_z}}\;\;
		\end{subfloatrow}
	}
	\caption{Examples of how a multiplier function $G(t)$ (in red) can be divided into high reward periods and low reward periods by the threshold $z = 30$. Refer to Section \ref{Section::Experimental_Results} for the analytical expression of the ``Wave Greed'', ``Christmas Greed'', and ``Step Greed'' multiplier functions.}
	\label{Figure::z_examples}
\end{figure}

In Algorithm \ref{Algorithm::epsilon_threshold} we present the $\varepsilon$-z greedy algorithm, a variation of the $\varepsilon$-greedy algorithm of \citet[]{auer2002finite}, in which a threshold $z$ has been introduced in order to regulate greed. This is the simplest method we know that would allow regulating greed over time. It has the disadvantage of adding one more parameter $z$, though we usually choose the heuristic of $z$ being 75\% of the maximum of $G(t)$. A good value for $z$ can also be estimated by running the algorithm on past data and by finding the one that gives the lowest regret. At each turn $t$, when the rewards are ``high'' (i.e., the $G(t)$ multiplier is above the threshold $z$) the algorithm exploits the best arm found so far, that is, arm $j$ with the highest mean estimate as defined in Equation \eqref{mean_estimator}. When the rewards are ``low'' (i.e., the $G(t)$ multiplier is under the threshold $z$), the algorithm will explore an arm at random (each arm having probability $1/m$ of being selected) with probability $\varepsilon_t = \min\left\{1,km / \tilde{t}\right\}$, where $\tilde{t}$ counts how many times the multiplier function has been under the threshold up to time $t$, and $k$ is a constant greater than $10$ and such that $k > 4/(\min_j \Delta_j^2)$. The examples in Figure \ref{Figure::z_examples} show how the threshold $z$ (in green) determines which turns are in the high reward periods (region in yellow, when $G(t)\geq z$ we want to be greedy) or in the low reward periods ($G(t)$ < $z$, where we balance exploration and exploitation). %$\varepsilon_t = \min\left\{1,\frac{km}{\tilde{t}}\right\}$

%\begin{framed}
%\RestyleAlgo{style}

\RestyleAlgo{boxruled}
\begin{algorithm}
	\SetAlgoLined
	\caption{{\color{black}$\varepsilon$-z greedy algorithm}}\nllabel{Algorithm::epsilon_threshold}
	\SetKwInOut{Input}{Input}
	\SetKwInOut{Output}{Output}
	\SetKwInOut{Loop}{Loop}
	\SetKwInOut{Initialization}{Initialization}
	\Input{number of rounds $n$, number of arms $m$, threshold $z$, a constant $k>10$ such that $k > 4/(\min_j \Delta_j^2)$, sequences $\{\varepsilon_t\}_{t=1}^n = \min\left\{1,km/\tilde{t}\right\}$ and $\{G(t)\}_{t=1}^n$ }
	\Initialization{play all arms once and initialize $\widehat{X}_j$ (defined in \eqref{mean_estimator}) for each $j=1,\dots,m$}
	\For{$t=m+1$ \KwTo $n$}{
		\eIf{$G(t)<z$}{with probability $\varepsilon_t$ play an arm uniformly at random (each arm with probability $1/m$), otherwise (with probability $1-\varepsilon_t$) play arm $j$ such that $ \widehat{X}_j > \widehat{X}_i \; \forall i$\;}{play arm $j$ such that $ \widehat{X}_j > \widehat{X}_i \; \forall i$\;} 
		%	{{\bf if }($G(t)<z$)}{ with probability $\varepsilon_t$ play an arm uniformly at random (each arm has probability $\frac{1}{m}$ of being selected),\\ otherwise (with probability $1-\varepsilon_t$) play arm $j$ such that \[ \widehat{X}_j > \widehat{X}_i \; \forall i\]
		%	{{\bf else }play arm $j$ such that \[ \widehat{X}_j > \widehat{X}_i \; \forall i\]}
		
		{Get reward $G(t)X_j$\;}
		{Update $\widehat{X}_j$\;}   
	}
\end{algorithm}

\noindent The following theorem provides a finite-time bound on the mean regret of the $\varepsilon$-z greedy algorithm (proof in Appendix \ref{proof_epsilon1}) in the form of:
\begin{eqnarray*}
	\E[R_n] & \leq & \text{Regret during initialization phase}\\
	& + & \text{Regret under the threshold}\\
	& + & \text{Regret above the threshold.}
\end{eqnarray*}

\tcbset{colback=blue!2!white}
\begin{tcolorbox}
	\begin{theorem}[{\color{black} Regret bound of the $\varepsilon$-z greedy algorithm}]\label{Theorem::epsilon_z}
		The bound on the mean regret $\E[R_n]$ at time $n$ is given by
		\begin{eqnarray}
			\E[R_n]\displaystyle & \leq & \displaystyle\sum_{j=1}^m G(j) \Delta_j  \label{initialization_epsilon_threshold}\\
			& + & \displaystyle\sum_{t=m+1}^n G(t)\ONE_{\{G(t)< z\}}\sum_{j : \mu_j<\mu_*} \Delta_j \left( \varepsilon_t\frac{1}{m}+(1-\varepsilon_t) \beta_j(\tilde{t}) \right) \label{low_epsilon_threshold}\\
			& + &\displaystyle\sum_{t=m+1}^n G(t)\ONE_{\{G(t)\geq z\}} \sum_{j : \mu_j<\mu_*} \Delta_j \beta_j(\tilde{t}) , \label{high_epsilon_threshold}
			%	& + &\displaystyle\sum_{t\in B} \sum_{j : \mu_j<\mu_*} \Delta_j \beta_j(\tilde{t}) \left[ \sum_{s=t}^{\tau_t} G(s)\right], \label{high_epsilon_threshold}
		\end{eqnarray}
		where
		\begin{equation}\label{beta_threshold}
			\beta_j(\tilde{t})=k\left( \frac{\tilde{t}}{m k e }\right) ^{-\frac{k}{10}}\log \left( \frac{\tilde{t}}{m k e } \right)   + \frac{4}{\Delta_j^2} \left( \frac{\tilde{t}}{m k e } \right)^{-\frac{k \Delta_j^2 }{4}}.
		\end{equation}
	\end{theorem}
\end{tcolorbox}

The sum in \eqref{initialization_epsilon_threshold} is the exact mean regret during the initialization phase of Algorithm \ref{Algorithm::epsilon_threshold}. Addend \eqref{low_epsilon_threshold} is a bound on the expected regret for turns that present low values of $G(t)$, where the quantity in the parenthesis is an upper bound on the probability of playing arm $j$: $\beta_j(\tilde{t})$ in \eqref{beta_threshold} is an upper bound on the probability that arm $j$ is considered to be the best arm at round $t$, and $1/m$ is the probability of choosing arm $j$ when the choice is made at random. The reason for the choice of $k > 4/(\min_j \Delta_j^2)$ is to make $\beta_j(\tilde{t})$ a quantity that is $o(1/\tilde{t})$. %which is a bound on the probability of incorrectly considering a suboptimal arm $j$ to be the best choice. 
By setting the parameter $k$ accordingly, we can ensure the logarithmic bound on the expected cumulative regret over the number of rounds (because the $\varepsilon_t$ are $\theta\left(1/t\right)$ and their sum over time is logarithmically bounded, and $\beta_j(\tilde{t})$ is $o\left(1/\tilde{t}\right)$). Finally, in \eqref{high_epsilon_threshold} we have a bound on the expected regret for turns with high values of $G(t)$, and in this case we consider only the upper bound $\beta_j(\tilde{t})$ on the probability that arm $j$ is considered to the best arm since we do not explore during high reward periods. The usual $\varepsilon$-greedy algorithm as introduced by \citet[]{auer2002finite} is a special case when $G(t)=1$ $\forall t$ and $z > 1$. %This is useful for comparing the two algorithms and see why this new policy performs better. 
Notice that $\varepsilon_t$ is a quantity $\theta\left( 1/\tilde{t}\right)$, while $\beta_j(\tilde{t})$ is $o\left( 1/\tilde{t}\right)$, so that an asymptotic logarithmic bound in $n$ holds for $\E[R_n]$ if $\tilde{t}$ grows at the same rate as $t$ (because of the logarithmic upper bound of the harmonic series\footnote{Logarithmic upper bound of the harmonic series: $\sum_{t=1}^n \frac{1}{t} \leq \log(n) + 1$}).

The following version of Theorem \ref{Theorem::epsilon_z} shows the order of magnitude of the regret at the end of the game.
\begin{tcolorbox}
	\begin{theorem}[{\color{black} Regret bound of the $\varepsilon$-z greedy algorithm}]\label{Theorem::epsilon_z_oreder_of_magnitude}
		The bound on the mean regret $\E[R_n]$ at time $n$ is given by
		\begin{eqnarray*}\footnotesize
			\E[R_n]\displaystyle  \leq  \bigo(1)
			+  \displaystyle\sum_{t=m+1}^n G(t)\ONE_{\{G(t)< z\}} \left(\theta\left(\frac{1}{t}\right) + o\left(\frac{1}{t}\right) \right)
			+ \displaystyle\sum_{t=m+1}^n G(t)\ONE_{\{G(t)\geq z\}} o\left(\frac{1}{t}\right).
		\end{eqnarray*}\normalsize
	\end{theorem}
\end{tcolorbox}
Intuitively, this bound is better than the standard $\varepsilon$-greedy bound because, when $G(t)$ is low, the regret of choosing a suboptimal arm is multiplied by a quantity that is of the order $\theta\left(1/t\right)+o\left(1/t\right)$, while when $G(t)$ is high it is multiplied by a $o\left(1/t\right)$ quantity. In contrast, in the standard $\varepsilon$-greedy algorithm, the regret of choosing a suboptimal arm is always multiplied by a quantity that is of the order $\theta\left(1/t\right)+o\left(1/t\right)$.

\subsubsection{Comparison with a smarter version of the standard $\varepsilon$-greedy algorithm}

We want to compare this bound with the one of the standard version of the $\varepsilon$-greedy algorithm but, since it is not well suited for the setting in which the rewards are altered by the multiplier function, we discount the rewards obtained at each round (by dividing them by $G(t)$) so that it can also produce accurate estimates of the mean reward for each arm. This ``smarter'' version of the $\varepsilon$-greedy algorithm is presented in Algorithm \ref{Algorithm::epsilon_slightly_smarter} (Section \ref{Section::Experimental_Results}).
The bound on the probability of playing a suboptimal arm $j$ for the standard $\varepsilon$-greedy algorithm is given by $\beta_j(t)$ (i.e., $\beta_j(\tilde{t})$ when $\tilde{t}=t$) and we refer to it as $\beta^{\text{old}}_j(t)$.
%The usual $\varepsilon$-greedy algorithm has $\tilde{t}=t$, thus its bound
In general, $\beta^{\text{old}}_j(t)$ is lower than $\beta_j(\tilde{t})$ (since $\tilde{t}\leq t$). Intuitively, this reflects the fact that the new algorithm performs fewer exploration steps. Moreover, in the standard $\varepsilon$-greedy algorithm, the probability of choosing a suboptimal arm $j$ at time $t$ is given by 
\[ \P\left(I_t^{\text{old}}=j\right)= \varepsilon_t\frac{1}{m}+(1-\varepsilon_t) \beta^{\text{old}}_j(t), \]
which is less than the probability of the new algorithm in case of low $G(t)$ (even if the rate of decay is the same),
\[ \P\left(I_t^{\text{new}}=j\right)= \varepsilon_t\frac{1}{m}+(1-\varepsilon_t) \beta_j(\tilde{t}), \]
but can easily be higher than the probability of the new algorithm in case of high rewards (which is given by only $\beta_j(\tilde{t})$). In fact, for suboptimal arm $j$, when $G(t)>z$, we have \small
\begin{eqnarray}
	\P\left(I_t^{\text{old}}=j\right) - \P\left(I_t^{\text{new}}=j\right) &=& \varepsilon_t\frac{1}{m}+(1-\varepsilon_t) \beta^{\text{old}}_j(t) - \beta_j(\tilde{t})\nonumber\\ 
	&=& \frac{1}{m} \min\left\{1,\frac{km}{t}\right\} -  \beta^{\text{old}}_j(t) \min\left\{1,\frac{km}{t}\right\} + \beta^{\text{old}}_j(t) - \beta_j(\tilde{t}). \label{difference}
\end{eqnarray}\normalsize
If $t>km$ we get
\begin{equation}\label{comp1}
	\eqref{difference} = \frac{1}{m}\frac{km}{t} +\beta^{\text{old}}_j(t)\left( 1- \frac{km}{t}\right)  - \beta_j(\tilde{t}) ,
\end{equation}
if $t\leq km$ we get
\begin{equation}\label{comp2}
	\eqref{difference} = \frac{1}{m} -\cancel{\beta^{\text{old}}_j(t)} +\cancel{\beta^{\text{old}}_j(t)}  - \beta_j(\tilde{t}),
\end{equation}
and for $t$ large enough both expressions are positive since $\beta_j(\tilde{t})$ is $o\left(1/\tilde{t}\right)$.  % and we assume that $\tilde{t}$ is $\theta(t)$.
Having \eqref{comp1} and \eqref{comp2} positive means that when we are in a high-rewards period the probability of choosing a suboptimal arm decreases faster when using Algorithm \ref{Algorithm::epsilon_threshold}. In that case, Algorithm \ref{Algorithm::epsilon_threshold} would have lower regret than the standard $\varepsilon$-greedy algorithm.

%In practice, the threshold $z$ should be defined as $\argmin(\E[R_n])$, but this can be computationally challenging, in which case 
If past data are available, a good value for $z$ can be chosen using cross validation techniques, i.e., by trying different thresholds with the available data and by choosing the one that yields the best performance, or using the heuristic of the 75\% of the maximum value of $G(t)$ as mentioned earlier (we used this heuristic in the real-data experiment in Section \ref{Section::Experimental_Results_real_data}).

\subsubsection{Regret bound for a simple scenario}
The following corollary illustrates the benefits of the bound in a simple scenario (shown in Figure \ref{Figure::step_z}) when the multiplier function can only take two values and the regulating threshold divides the higher value from the lower one (this is the case of the ``Step'' multiplier function used in the numerical experiments in Section \ref{Section::Experimental_Results}).
\begin{tcolorbox}{\color{black}
		\begin{corollary}\label{corollary_epsilon_threshold}
			Suppose the greed function $G(t)$ takes only two values: $g_{\text{low}}$ and $g_{\text{high}}$. It takes the value $g_{\text{low}}$ for a fraction $q$ of the turns played, and the value $g_{\text{high}}$ for the remaining $n-qn$ turns. %for example, if $q=1/2$, $G(t)$ alternates at each turn between $g_{\text{low}}$ and $g_{\text{high}}$). 
			Then, the bound in Theorem \ref{Theorem::epsilon_z} of the expected regret at turn $n$ reduces to
			\begin{eqnarray*}
				\E[R_n]\displaystyle  \leq  \bigo(1) 
				+  g_{\text{low}} \, (qn)  \,   \left( \theta\left(\frac{1}{t}\right)+ o\left(\frac{1}{t}\right) \right) 
				+ g_{\text{high}}\, (n-qn) \,    o\left(\frac{1}{t}\right).
			\end{eqnarray*}
		\end{corollary}
	}
\end{tcolorbox}\normalsize
The term $\theta(1/t)$ that hurts regret most is multiplied only by $g_{\text{low}}$, and not by $g_{\text{high}}$. When the rewards are high (and so is the possible regret), only terms of order $o(1/t)$ are present. If exploration were permitted during the high reward zone (as is the case when using the standard $\varepsilon$-greedy algorithm), large terms $g_{\text{high}}$ would affect the regret, which is what Algorithm \ref{Algorithm::epsilon_threshold} is designed to avoid. In this case it does not matter where we put the threshold as long as it is above $g_{\text{low}}$ and below $g_{\text{high}}$.
In Appendix \ref{Appendix::Unknown_G_th} we generalize Theorem \ref{Theorem::epsilon_z} to the case when the function $G(t)$ is not known but estimated by $H(t)$. (The method used in Appendix \ref{Appendix::Unknown_G_th} is easily applied to the following theorems too when $G(t)$ is not known).

\subsection{Soft $\varepsilon$-greedy algorithm}\label{Section::soft_espilon}
We present in Algorithm \ref{Algorithm::epsilon_soft} a ``soft version'' of the $\varepsilon$-greedy algorithm where greed is regulated gradually (in contrast with the hard threshold $z$ of the previous section). This algorithm has the advantage of an adaptive threshold which the user does not need to choose. Again, in high reward zones, exploitation will be preferred, while in low reward zones the algorithm will explore the arms more. Let us define the following function
\begin{equation}%\label{psi}
	\psi(t)=\frac{\log\left(1+\frac{1}{G(t)}\right)}{\log\left(1+\frac{1}{ \min_{s\in\{m+1,\cdots,n\}} G(s)}\right)},
\end{equation}
and let $\gamma=\min_{s\in\{m+1,\cdots,n\}} \psi(s)$. 
Notice that $0<\psi(t)\leq 1$ $\forall t$ and that its values are close to $0$ when $G(t)$ is high, while they are close to $1$ for low values of $G(t)$. The new probabilities of exploration during the game are given at each turn $t$ by $\varepsilon_t= \min\left\{\psi(t),km/t\right\}$. In this way, we still maintain the decay of the probabilities of exploration, but we push them closer to zero when the multiplier function $G(t)$ is high to avoid high regrets. We generally  assume that $\min_{s\in \{m+1,\cdots,n\}} G(s)$ is not smaller than $1$. This particular functional form of $\psi(t)$ has been chosen because it is well-suited to be used for determining probabilities (of exploration), the standard $\varepsilon$-greedy algorithm of \citet[]{auer2002finite} is recovered when $G(t)=1$ for all $t$, and it makes the regret bound easier to prove. 

%\RestyleAlgo{boxruled}
%\begin{algorithm}[]
%	\SetAlgoLined
%	\caption{Soft $\varepsilon$-greedy algorithm}\nllabel{Algorithm::epsilon_soft}
%	\SetKwInOut{Input}{Input}
%	\SetKwInOut{Output}{output}
%	\SetKwInOut{Loop}{Loop}
%	\SetKwInOut{Initialization}{Initialization}
%	%	\SetKwData{Left}{left}\SetKwData{This}{this}\SetKwData{Up}{up}
%	%	\SetKwFunction{Union}{Union}\SetKwFunction{FindCompress}{FindCompress}
%	\Input{number of rounds $n$, number of arms $m$, a constant $k>10$, such that $k > 4/(\min_j \Delta_j^2)$, sequences $\{\varepsilon_t\}_{t=1}^n = \min\left\{\psi(t),km/t\right\}$ and $\{G(t)\}_{t=1}^n$ }
%	\Initialization{play all arms once and initialize $\widehat{X}_j$ (defined in \eqref{mean_estimator}) for each $j=1,\dots,m$}
%	\For{$t=m+1$ \KwTo $n$}{ 
%		{With probability $\varepsilon_t$ play an arm uniformly at random (each arm with probability $\frac{1}{m}$),\\ otherwise (with probability $1-\varepsilon_t$) play arm $j$ such that $ \widehat{X}_j > \widehat{X}_i \; \forall i$ \;}
%		{Get reward $G(t)X_j$\;}
%		{Update $\widehat{X}_j$\;}
%	}
%\end{algorithm}

\RestyleAlgo{boxruled}
\begin{algorithm}[]
	\SetAlgoLined
	\caption{Soft $\varepsilon$-greedy algorithm}\nllabel{Algorithm::epsilon_soft}
	\SetKwInOut{Input}{Input}
	\SetKwInOut{Output}{output}
	\SetKwInOut{Loop}{Loop}
	\SetKwInOut{Initialization}{Initialization}
	%	\SetKwData{Left}{left}\SetKwData{This}{this}\SetKwData{Up}{up}
	%	\SetKwFunction{Union}{Union}\SetKwFunction{FindCompress}{FindCompress}
	\Input{number of rounds $n$, number of arms $m$, a constant $k>10$, such that $k > 4/(\min_j \Delta_j^2)$, sequences $\{\varepsilon_t\}_{t=1}^n = \min\left\{\psi(t),km/t\right\}$ and $\{G(t)\}_{t=1}^n$ }
	\Initialization{play all arms once and initialize $\widehat{X}_j$ (defined in \eqref{mean_estimator}) for each $j=1,\dots,m$}
	\For{$t=m+1$ \KwTo $n$}{ 
		{With probability $\varepsilon_t$ play an arm uniformly at random (each arm with probability $\frac{1}{m}$),\\ otherwise (with probability $1-\varepsilon_t$) play arm $j$ such that $ \widehat{X}_j > \widehat{X}_i \; \forall i$\;}
		{Get reward $G(t)X_j$\;}
		{Update $\widehat{X}_j$\;}
	}
\end{algorithm}

The following theorem (proved in Appendix \ref{proof_epsilon2}) shows that a logarithmic bound holds in this case too (because the $\varepsilon_t$ are $\theta\left(1/t\right)$ and their sum over time is logarithmically bounded, while the $\beta_j^S(t)$ term is $o\left(1/t\right)$). 
\begin{tcolorbox}
	\begin{theorem}[Regret-bound for Soft $\varepsilon$-greedy algorithm]\label{Theorem::regret_soft_epsilon_greedy}
		The bound on the mean regret $\E[R_n]$ at time $n$ is given by
		\begin{eqnarray}
			\E[R_n]\displaystyle & \leq & \displaystyle\sum_{j=1}^m G(j) \Delta_j \label{initialization_epsilon_soft_A}  \\
			& + & \displaystyle\sum_{t=m+1}^n G(t)\sum_{j : \mu_j<\mu_*} \Delta_j \left( \varepsilon_t\frac{1}{m}+(1-\varepsilon_t) \beta_j^S(t) \right) \label{bound_epsilon_soft_B}
		\end{eqnarray}
		where
		\begin{equation}%\label{beta_soft}
			\beta^S_j(t) =  k\left( \frac{\gamma t}{mke } \right)^{-\frac{k}{10} } \log \left( \frac{\gamma t}{mke } \right)   + \frac{4}{\Delta_j^2} \left( \frac{\gamma t}{mke } \right)^{-\frac{k\Delta_j^2 }{4} }.
		\end{equation}
	\end{theorem}
\end{tcolorbox}

The sum in \eqref{initialization_epsilon_soft_A} is the exact mean regret during the initialization of Algorithm \ref{Algorithm::epsilon_soft}. For the rounds after the initialization phase, the quantity in the parenthesis of \eqref{bound_epsilon_soft_B} is the upper bound on the probability of playing arm $j$ (where $\beta_j^S(t)$ is the bound on the probability that arm $j$ is the best arm at round $t$, and $1/m$ is the probability of choosing arm $j$ when the choice is made uniformly at random). 

Theorem \ref{Theorem::regret_soft_epsilon_greedy} can be stated in a simpler form that shows the order of magnitude of the bounding quantity:
\begin{tcolorbox}
	\begin{theorem}[Regret bound of the Soft $\varepsilon$ greedy algorithm]\label{Theorem::regret_soft_epsilon_greedy_order_of_magnitude}
		The bound on the mean regret $\E[R_n]$ at time $n$ is given by
		\begin{eqnarray}
			\E[R_n]\displaystyle & \leq & \bigo(1)  
			+  \displaystyle\sum_{t=m+1}^n G(t)\left[  \min\left( \psi(t), \theta\left(\frac{1}{t}\right)\right) + o\left(\frac{1}{t}\right) \right]. \label{interpretable_soft}
		\end{eqnarray}
	\end{theorem}
\end{tcolorbox}
Intuitively, the advantage of Algorithm \ref{Algorithm::epsilon_soft} over the standard $\varepsilon$-greedy algorithm is that, when $G(t)$ is high, only the $o\left(1/t\right)$ term contributes significantly to the regret (because $\psi(t)$ pushes the term $ \min( \psi(t), \theta(1/t))$ to zero).

\subsubsection{Comparison with a smarter version of the standard $\varepsilon$-greedy algorithm}

\begin{figure}[]
	\caption{Comparison of probabilities of exploration over the number of rounds. Before $n'$, $\varepsilon_t^{\text{old}}$ is $1$ and always greater than $\psi(t)$. After $w$, $\varepsilon_t^{\text{old}}$ is always less than $\psi(t)$. When $G(t)$ is high, $\psi(t)$ pushes the probability of exploration towards zero. }
	\centering
	\includegraphics[width=0.90\textwidth]{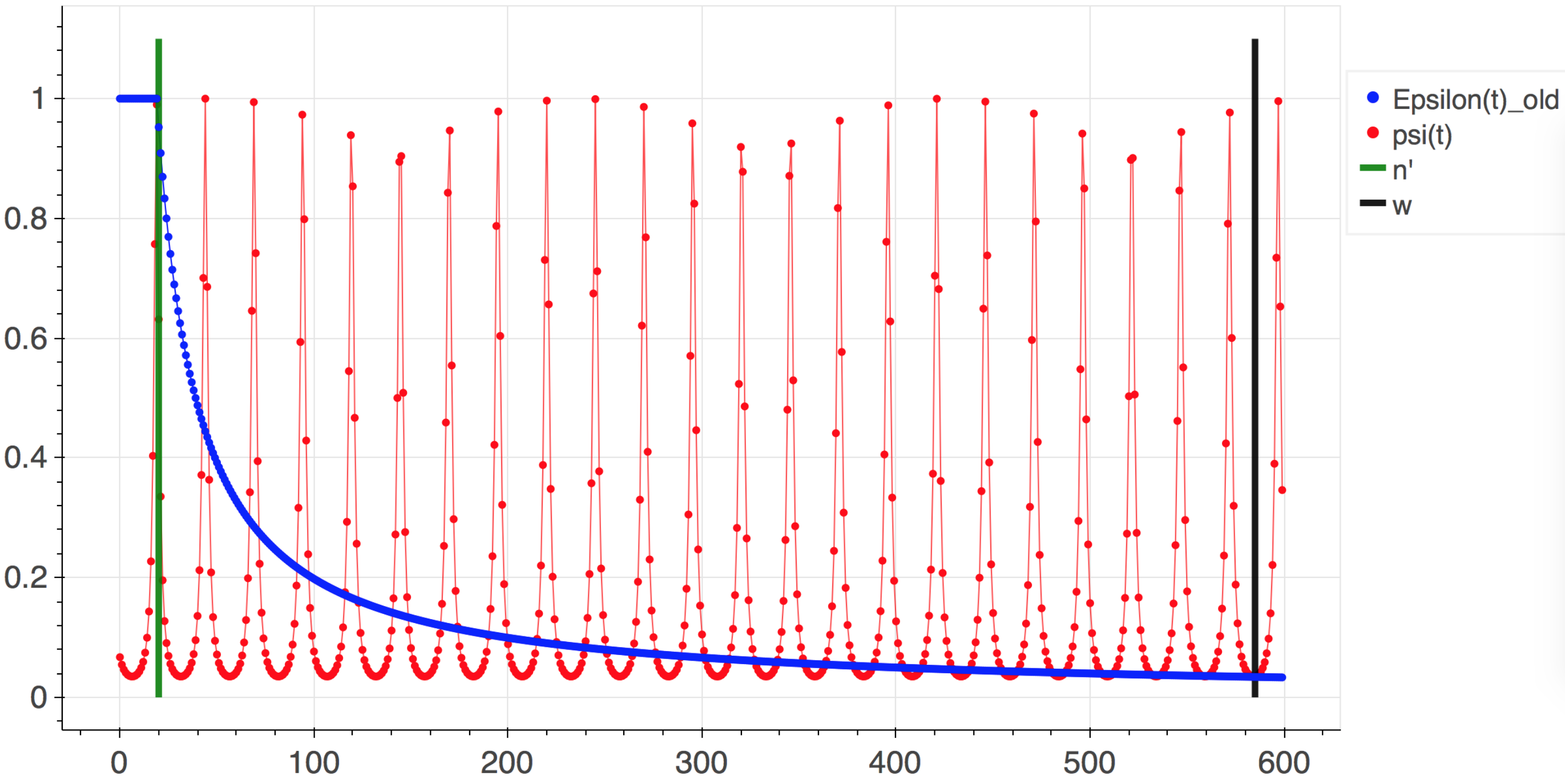}
	\label{plot_probabilities}
\end{figure}

As before, we want to compare this bound with the ``smarter'' version of the $\varepsilon$-greedy algorithm presented in Algorithm \ref{Algorithm::epsilon_slightly_smarter}.
%In order to compare the soft-$\varepsilon$-greedy algorithm with the usual $\varepsilon$-greedy algorithm, a closer look at the proof for the regret bound is required. 
In the usual $\varepsilon$-greedy algorithm, after the ``critical time'' $n'=km$, the probability $\P( \widehat{X}_{j,T_j(t-1)} \geq \widehat{X}_{i,T_i(t-1)})$ of arm $j$ being considered as the current best arm can be bounded by a quantity $\beta^{\text{old}}_j(t)$ that is $o\left(1/t\right)$. Before time $n'$, the decay of $\P( \widehat{X}_{j,T_j(t-1)} \geq \widehat{X}_{i,T_i(t-1)})$ is exponential: $\theta(e^{-t})$ (see Remark \ref{Remark1} in Appendix \ref{proof_epsilon1}).  %, namely $o\left(1/t^\lambda\right)$, $\forall \lambda$ (see Remark \ref{Remark1} in Appendix \ref{proof_epsilon1}). 
The probability of choosing a suboptimal arm $j$ changes as follows:
\begin{itemize}
	\item if $t<n'$, $\P(I_t=j) = \frac{1}{m}$;
	\item if $t \geq n'$,  $\P(I_t=j) = \frac{k}{t} + \left(1- \frac{k m}{t}\right)\beta^{\text{old}}_j(t) \;$, which is $\theta\left(\frac{1}{t}\right)$ as $t$ grows. %+o(1/t)+o(1/t^2)
\end{itemize}
In the Soft $\varepsilon$-greedy algorithm, before time $w = \min\{s: f(s) <\gamma\, f(s)=km/s\}$, %$w=\min_{s\in\{1,\cdots,n\}}\frac{c m}{d^2 s} < \gamma$, 
we have that $\beta^S_j(t)$, which is the bound on the probability $\P( \widehat{X}_{j,T_j(t-1)} \geq \widehat{X}_{i,T_i(t-1)})$ of arm $j$ being the current best arm, is a quantity that decays exponentially with rate $\theta(e^{-\gamma t})$ (see Remark \ref{Remark::soft-epsilon} in Appendix \ref{proof_epsilon2}). %$o\left(1/(\gamma t)^\lambda\right)$, $\forall \lambda$ as $t$ grows (the argument is similar to the Remark \ref{Remark1} in Appendix \ref{proof_epsilon1}). 
After $w$, $\beta^S_j(t)$ can be bounded by a quantity that is $o\left(1/(\gamma t)\right)$ as $t$ grows. 
The bound on the probability of choosing a suboptimal arm $j$ changes as follows:
\begin{itemize}
	\item if $t<n'$, $\P(I_t=j) = \frac{1}{m} \psi(t)+ (1-\psi(t))\beta^S_j(t)$;
	\item if $n' \leq t \leq w$, $\P(I_t=j) = \frac{1}{m}\min\left\{\psi(t),\frac{km}{t}\right\} + \left(1-\min\left\{\psi(t),\frac{km}{t}\right\}\right)\beta^S_j(t)$;
	\item if $t>w$, $\P(I_t=j) =\frac{k}{t} + \left(1-\frac{k}{t}\right) \beta^S_j(t) $.
\end{itemize}
In order to interpret these quantities, let us see what happens for high or low values of the multiplier $G(t)$ as $t$ grows in Table \ref{soft_comparison}. For brevity, we abuse notation when using Landau's symbols, because in some cases $t$ is not allowed to go to infinity; it is convenient to still use this notation to compare the decay rates of the probabilities of choosing a suboptimal arm, which also gives a qualitative explanation of what happens when using the algorithms. Let $\P(I_t=j)^{\text{soft}}$ be the probability of choosing a suboptimal arm for the Soft $\varepsilon$-algorithm. When comparing to the standard $\varepsilon$-greedy algorithm, the rate of decay of $\P(I_t=j)^{\text{soft}}$ is faster when $G(t)$ is high and $t \leq w$ (for the other cases, the rate of decay is the same as $\P(I_t=j)^{\text{old}}$, the probability of choosing a suboptimal arm for the standard $\varepsilon$-greedy algorithm). %For the Soft $\varepsilon$-algorithm, the rate at which the probability of choosing a suboptimal arm decays is faster when $G(t)$ is high and $t \leq w$ (for the other cases, the rate of decay is the same as the standard $\varepsilon$-greedy algorithm).  
Notice that the parameter $\gamma$ slows down the decay. %with respect to the usual $\varepsilon$-greedy algorithm. 
This is a direct consequence of the slower exploration. An example of a typical behavior of $\psi(t)$ and $\varepsilon_t^{\text{old}}$ is shown in Figure \ref{plot_probabilities}, where $G(t)=20+19\sin(t/2)$. The blue curve shows the probability of exploration under the usual $\varepsilon$-greedy algorithm that does not regulate greed, while the red curve shows how the function $\psi(t)$ oscillates depending on the value of the multiplier function. If $\psi(t)< km/t$, then $\psi(t)$ is the probability of exploration at time $t$ which drops when $G(t)$ is high (which means higher rewards but also higher regrets), while it is bounded by $km/t$ when $G(t)$ is low (which means lower rewards and regrets).

\small
\begin{table}[]
	\makebox[\linewidth][c]{
		
		\begin{tabular}{ | c |  c | c | c | c |}%
			\hline
			& &  & & \\
			round $t$ & $G(t)$ & $\P(I_t=j)^{\text{old}}$ &    $\P(I_t=j)^{\text{soft}}$ & $\P(I_t=j)^{\text{soft}} < \P(I_t=j)^{\text{old}}$ ?   \\ 
			& &  & & \\   
			\hline                
			& &   &&\\
			&  high  &   $\frac{1}{m}$  &   $\theta(e^{-\gamma t})$  & yes, faster decay\\  %$o\left(\frac{1}{(\gamma t)^\lambda}\right)$, $\forall \lambda$
			$t<n'$	& & &  &  \\
			&  low   &   $\frac{1}{m}$   &  close to $\frac{1}{m}$ & no, but not by much \\
			& &  & & \\
			\hline
			& &  & &\\
			&	 high  & $\theta\left(\frac{1}{t}\right) + o\left(\frac{1}{t}\right)$ &   $\theta(e^{-\gamma t})$  & yes, faster decay\\ %$o\left(\frac{1}{(\gamma t)^\lambda}\right)$, $\forall \lambda$
			$n'\leq t\leq w$	& & &  &  \\
			&   low        & $\theta\left(\frac{1}{t}\right) + o\left(\frac{1}{t}\right)$ &   $\theta\left(\frac{1}{t}\right)  +  \theta(e^{-\gamma t})$  & yes, but same rate of decay\\
			& &  & & \\ %+ o\left(\frac{1}{(\gamma t)^\lambda}\right)$, $\forall \lambda$
			\hline
			& &  & &\\
			&	 high  & $\theta\left(\frac{1}{t}\right) + o\left(\frac{1}{t}\right)$ &   $\theta\left(\frac{1}{t}\right) + o\left(\frac{1}{\gamma t}\right)$  &  no, but same rate of decay\\
			$t>w$	& & & &  \\
			&   low   & $\theta\left(\frac{1}{t}\right) + o\left(\frac{1}{t}\right)$ &    $\theta\left(\frac{1}{t}\right) + o\left(\frac{1}{\gamma t}\right)$  & no, but same rate of decay \\
			& &  & & \\
			\hline  
		\end{tabular}
		\caption{Summary of the {\bf decay rate of the bound on the probabilities of choosing a suboptimal arm} for the Soft $\varepsilon$-greedy algorithm and the standard $\varepsilon$-greedy algorithm (supposing it is taking in account the time-patterns, i.e., a ``smarter'' version). The decay depends on the turn number of the game  (see Figure \ref{plot_probabilities}). The probability of choosing a suboptimal arm decays much faster for the Soft $\varepsilon$-greedy algorithm when $G(t)$ is high. For the other cases, the rate of decay is the same.}
		\label{soft_comparison}
	}
\end{table}
\normalsize

\subsection{Regulating greed with threshold in the UCB algorithm}\label{Section::Algorithm::UCB_threshold}

%Before presenting the results on the UCB algorithm with threshold, we introduce some novel results on the standard UCB algorithm that are useful for the proof of the regret bound. 
Before we provide results for our algorithm, we need to introduce some novel preliminary results on the standard UCB algorithm. 
Specifically, we need the minimum number of times that each arm will be pulled on or before turn $t$ (Lemma \ref{Lemma::minimum_pulls_UCB}). This quantity is required each time $G(t)$ exceeds the threshold. It is used for computing an upper bound on the probability that the algorithm thinks arm $j$ is the best. (We cannot use the typical result of the expected number of times that each arm is pulled, as is used in standard UCB analysis.)

\subsubsection{Preliminary results on the UCB algorithm}

The following lemma provides the minimum number of times each arm is pulled when using the standard UCB algorithm.
\begin{tcolorbox}
	\begin{lemma}\label{Lemma::minimum_pulls_UCB}
		Suppose the rewards of the arms are bounded in $[a,b]$ and let us define $r = b-a$ the range of the possible rewards. When using the UCB policy for a game consisting of $n$ turns, each arm will be pulled at least $x_n$ times, where 
		\begin{eqnarray*}
			x_n &=& \max\left\{ y\in \N \, : (m-1)\phi(r,y) \leq n    \right\} + 1,\\
			\phi(r,y)  &=&  \min\left\{ t \geq \tau(r,y), \, t \in \N \,: t >  \frac{2\, \log(t)}{r^2 + \frac{2 \, \log(t)}{y} -2r\sqrt{\frac{2 \, \log(t)}{y} }}\right\} ,\\
			\tau(r,y) &=& \min\left\{t \in \N :  \sqrt{\frac{2\log(t)}{y} } \geq r\right\}.
		\end{eqnarray*}
	\end{lemma}
\end{tcolorbox}
Corollary \ref{Lemma::minimum_pulls_UCB_different_ranges} in Appendix \ref{Appendix::UCBz} considers the case where random rewards of each arm take values in different ranges.
Theorem \ref{Theorem::lower_bound_x_n} gives a lower bound on the previous quantity, showing that the minimum number of times the algorithm pulls an arm grows at least logarithmically.
\begin{tcolorbox}
	\begin{theorem}\label{Theorem::lower_bound_x_n}
		Suppose the rewards of the arms are bounded in $[a,b]$ and let us define $r = b-a$ the range of the possible rewards. Let $x_n$ be the  minimum number of times each arm will be pulled when using the UCB policy for a game consisting of $n$ turns ($n>m$). Then,
		\[x_n \in \Omega(\log(n)).\]
		%In other words, the minimum number of times the algorithm pulls an arm in a game of $n$ turns grows at least logarithmically in $n$. %other UCB formulation would replace 2 with the UCB parameter
	\end{theorem}
\end{tcolorbox}
Corollary \ref{Theorem::lower_bound_x_n_different_range} in Appendix \ref{Appendix::UCBz} considers the case where random rewards of each arm take values in different ranges.

\subsubsection{UCB with threshold}

Following what has been presented to improve the $\varepsilon$-greedy algorithm in the setting with multiplier function $G(t)$, we introduce in Algorithm \ref{Algorithm::UCB_threshold} a modification of the UCB algorithm. We again set a threshold $z$ and, if the multiplier of the rewards $G(t)$ is above this level, the new algorithm exploits the best arm. When $G(t)$ is under the threshold, the algorithm is going to play the arm with the highest upper confidence bound on the mean estimate. The threshold $z$ can be chosen as suggested in Section \ref{Section::epsilon_threshold}. 

\begin{algorithm}[]
	\SetAlgoLined
	\caption{UCB-$z$ algorithm}\nllabel{Algorithm::UCB_threshold}
	\SetKwInOut{Input}{Input}
	\SetKwInOut{Output}{Output}
	\SetKwInOut{Loop}{Loop}
	\SetKwInOut{Initialization}{Initialization}
	\Input{number of rounds $n$, number of arms $m$, threshold $z$, sequence $\{G(t)\}_{t=1}^n$ }
	\Initialization{play all arms once and initialize $\widehat{X}_j$ (as defined in \eqref{mean_estimator}) for each $j=1,\cdots,m$}
	\For{$t=m+1$ \KwTo $n$}{ 
		\eIf{$G(t)<z$}{play arm $j$ with the highest %upper confidence bound on the mean estimate 
			$\widehat{X}_{j,T_j(t-1)}+\sqrt{\frac{2\log t}{T_j(t-1)}}$\;}{play arm $j$ such that $ \widehat{X}_j > \widehat{X}_i \; \forall i $\;}	
		{Get reward $G(t)X_j$\;}
		{Update $\widehat{X}_j$\;}
	}
\end{algorithm}

%It is possible to prove that also in this case the regret can be bounded logarithmically in $n$. 
Let $B=\{t: G(t-1)<z, G(t)\geq z\}$ be the set of rounds where the high-reward zone is entered. Let us call $y_1, y_2, \cdots, y_{|B|}$ the elements of $B$ and order them in increasing order such that $y_1 < y_2 < \dots < y_{|B|}$. Let us also define for every $k \in \{1,\dots, |B|\}$ the set $Y_k=\{t: t\geq y_k, G(t)\geq z, t<y_{k+1}, y_{|B|+1}=n\}$ of times in the high-reward period entered at time $y_k$. % and let $\Lambda_k=\max_{t\in Y_k} G(t)$ the highest value of $G(t)$ on $Y_k$. Finally, for every $k$, let $R_k= \Lambda_k |Y_k| $.% the maximum regret of the $k$th high reward zone. \\
Now, given a game of $n$ total rounds, let us call $\tilde{n}$ the number of rounds played under the threshold $z$ by the end of the game. The regret bound showed in Theorem \ref{Theorem::regulated_UCB} is the sum of the regret in the high reward zone and the regret in the low reward zone and is of the form
\begin{eqnarray*}
	\E[R_n] & \leq & \text{Regret during initialization phase}\\
	& + & \text{Regret under the threshold}\\
	& + & \text{Regret above the threshold.}
\end{eqnarray*} 
(When $G(t) = 1$ for all $t$ the usual regret bound for the UCB policy is recovered.) The regret bound also uses Lemma \ref{Lemma::minimum_pulls_UCB} which counts the number of times arm $j$ has been pulled when using an UCB policy under the threshold. Usually, for applications of the UCB algorithm, the expected number of pulls is sufficient for theoretical results, but our framework is different, since we need to compute the probability of playing an arm at each turn $t$ the threshold is exceeded.

\begin{tcolorbox}
	\begin{theorem}[Regret-bound for the UCB-$z$ algorithm]\label{Theorem::regulated_UCB}
		The bound on the mean regret $\E[R_n]$ at time $n$ is given by
		\begin{eqnarray}
			\E[R_n]\displaystyle & \leq & \displaystyle\sum_{j=1}^m G(j) \Delta_j \label{initialization_Algorithm::UCB_threshold}  \\
			& + & z \left[ 8 \sum_{j : \mu_j<\mu_*} \left(\frac{\log{\tilde{n}}}{\Delta_j}\right) + \left(1 + \frac{\pi^2}{3}\right)\left(\sum_{j=1}^m \Delta_j\right) \right]  \label{classic_UCB_bound}   \\
			& + & \displaystyle \sum_{j=1}^m  \Delta_j \sum_{k=1}^{|B|}  \sum_{t\in Y_k}G(t) 2\beta_j^U(t),\label{Algorithm::UCB_threshold_constant}
		\end{eqnarray}
		where 
		\[\beta_j^U(t) = \frac{2}{\Delta_j^2}e^{-\frac{\Delta_j^2}{2} (x_t-1)}\]
		and $x_t$ is the minimum amount of pulls for each arm at time $t$ (see Lemma \ref{Lemma::minimum_pulls_UCB}.)
	\end{theorem}
\end{tcolorbox}
The sum in \eqref{initialization_Algorithm::UCB_threshold} represents the exact regret coming from the initialization phase, in \eqref{classic_UCB_bound} we have a logarithmic regret (in the number of turns played under the threshold) that comes from the classic UCB policy, where $z$ is the upper bound on $G(t)$ for low reward turns and the square brackets is the sum of the regret of each arm multiplied by an upper bound on the expected number of times that arm is played. Finally, in \eqref{Algorithm::UCB_threshold_constant} we have the regret that comes from turns in the high reward zone, where $2\beta_j^U(t)$ is an upper bound on the probability of playing arm $j$ at turn $t$. 
Theorem \ref{Theorem::lower_bound_x_n} guarantees that $\beta_j^U$ decreases fast enough since $x_t$ grows at least logarithmically.
For the proof of Theorem \ref{Theorem::regulated_UCB}, Theorem \ref{Theorem::lower_bound_x_n}, and Lemma \ref{Lemma::minimum_pulls_UCB} see Appendix \ref{Appendix::UCBz}.
%The $\varepsilon$-greedy methods are more amenable to this type of analysis than UCB methods, because the proofs require bounds on the probability of choosing the wrong arm \textit{at each turn}. The UCB proof instead requires us to bound the expected number of times the suboptimal arms are played, without regard to \textit{when} those arms were chosen. We were able to avoid using the maximum of the $G(t)$ values in the $\varepsilon$-greedy proofs, but this is unavoidable in the UCB proofs without leaving terms in the bound that cannot be explicitly calculated or simplified (an alternate proof would use weaker Central Limit Theorem arguments).

%The upper bound $\beta_j^U(t)$ converges exponentially to a constant. In this case, if you want to ensure a logarithmic bound on the regret, the set cardinality of $B$ needs to eventually decrease with a rate of $o(1/t)$ (which is not a problem since after many turns the algorithm will figure out the best arm and whether we are in the high reward and low reward zone is irrelevant). 
%Theorem \ref{Theorem::regulated_UCB} can be stated in a simpler form that shows the order of magnitude of the bounding quantity.

\subsection{The Soft UCB algorithm}\label{Section::Algorithm::UCB_soft}

In Algorithm \ref{Algorithm::UCB_soft}, present a ``soft version" of the UCB algorithm where greed is regulated gradually (in contrast with the hard threshold of the previous section). Again, in high reward zones, exploitation will be preferred, while in low reward zones the algorithm will explore the arms.\\% $\\xi(t)$ can be a real number or a sequence $\\xi(t)=\{\\xi(t)_t\}_{t=1}^n$ that can be used to mitigate or enhance the effects that the multiplier function $G(t)$ has in the exploitation rate.\\
Let us define the following function:
\begin{equation}\label{csi_function}
	\xi(t) = \left(1 + \frac{t}{G(t)}\right).
\end{equation}
At each turn $t$ of the game, the algorithm plays the arm with the highest upper confidence bound on the mean estimate, but, with the introduction of $\xi(t)$, the confidence interval around $\widehat{X}_{j,T_j(t-1)}$ is built in a way such that, when $G(t)$ is high, it collapses upon the mean reward estimate, forcing the player to choose the arm with the highest mean estimate (thus, leading to a pure exploitation policy). In contrast, when the multiplier $G(t)$ is low, the confidence interval around $\widehat{X}_{j,T_j(t-1)}$ stretches out, allowing the player to explore arms with more uncertainty.

One of the main difficulties of the formulation of these bounds is to define a correct functional form for the upper confidence bounds so that it is possible to obtain smoothness in the arm decision, reasonable Hoeffding's inequality bounds while working out the proof (see Appendix \ref{proof_UCB2}), and a convergent series (the second summation in \eqref{convSeries_}). This particular choice for the functional form of $\xi(t)$ was chosen because of the following:
\begin{itemize}
	\item $\xi(t)$ correctly stretches or shrinks the upper confidence bound based on the multiplier function $G(t)$;
	\item when $G(t)$ is a constant equal to $1$, the standard UCB algorithm is almost recovered (``almost'' because the upper confidence bound, when $G(t)=1$, would have a $\log(t)$ factor for standard UCB and a $\log(1+t)$ factor for soft UCB, where we have an extra $1$ to avoid a negative argument of the logarithm);
	\item $\xi(t)$ is well-suited to be the argument of the logarithm that appears under the square root of the upper confidence bound (because it is positive);
	\item we can easily use Hoeffding’s inequality at the beginning of the proof of the regret bound (see Appendix \ref{proof_UCB2}).
\end{itemize}
%comes from the fact that $\xi(t)$ correctly stretches or shrinks the upper confidence bound based on the multiplier function; when $G(t)$ is a constant equal to $1$, the standard UCB algorithm is almost recovered (``almost'' because the upper confidence bound, when $G(t)=1$, would have a $\log(t)$ factor for standard UCB and a $\log(1+t)$ factor for soft UCB, where we have an extra $1$ to avoid a negative argument of the logarithm); and it is well-suited to be the argument of the logarithm that appears under the square root of the upper confidence bound (because it is positive). In this way, we can easily use Hoeffding’s inequality at the beginning of the proof of the regret bound (see Appendix \ref{proof_UCB2}).
\RestyleAlgo{boxruled}
\begin{algorithm}[]
	\SetAlgoLined
	\caption{Soft UCB algorithm}\nllabel{Algorithm::UCB_soft}
	\SetKwInOut{Input}{Input}
	\SetKwInOut{Output}{output}
	\SetKwInOut{Loop}{Loop}
	\SetKwInOut{Initialization}{Initialization}
	\Input{number of rounds $n$, number of arms $m$, sequence $\{G(t)\}_{t=1}^n$ }
	\Initialization{play all arms once and initialize $\widehat{X}_j$ (as defined in \eqref{mean_estimator}) for each $j=1,\cdots,m$}
	\For{$t=m+1$ \KwTo $n$}{ 
		{play arm $j$ with the highest  $\widehat{X}_{j,T_j(t-1)}+\sqrt{\frac{2 \,\log\xi(t)}{T_j(t-1)}}$\;}
		{Get reward $G(t)X_j$\;}
		{\hspace{0.1cm}Update $\widehat{X}_j$\;}
	}
\end{algorithm}
Also in this case, it is possible to achieve a bound that grows logarithmically in $n$. 
\begin{tcolorbox}
	\begin{theorem}[Regret-bound for soft-UCB algorithm]\label{Theorem::soft_UCB}
		Let $S=\{m+1, \dots, n\}$. The bound on the mean regret $\E[R_n]$ at time $n$ is given by
		%\begin{equation*}%\label{softbound_UCB}
		%\E[R_n] \leq \displaystyle \sup_{t\in\{1, \cdots, n\}}G(t) \left(   \sum_{j : \mu_j<\mu_*}  \frac{8}{\Delta_j}\log \left( \max_{t\in\{1, \cdots, n\} }\xi_{\\xi(t)}(t) \right)  + 2\sum_{j=1}^m \Delta_j \sum_{t=1}^n \xi(t)^{-4}t^2\right).
		%\end{equation*}
		\footnotesize
		\begin{eqnarray}
			\E[R_n]\displaystyle & \leq & \displaystyle\sum_{j=1}^m G(j) \Delta_j  \label{initialization_Algorithm::UCB_soft_} \\
			& + &\displaystyle \max_{t\in S}G(t) \left[   \sum_{j : \mu_j<\mu_*}  \frac{8}{\Delta_j}\log \left( \displaystyle \max_{t\in S} \xi(t) \right)  + \sum_{j=1}^m \Delta_j \left( 1 + \sum_{t=m+1}^n \frac{2(t-1-m)^2}{\xi(t)^{4}} \right)\right]. \label{convSeries_}
		\end{eqnarray}\normalsize
	\end{theorem}
\end{tcolorbox}
Theorem \ref{Theorem::soft_UCB} can be stated in a simpler form that shows the order of magnitude of the bounding quantity.
\begin{tcolorbox}
	\begin{theorem}[Regret-bound for soft-UCB algorithm]\label{Theorem::soft_UCB_order_of_magnitude}
		Let $S=\{m+1, \dots, n\}$. The bound on the mean regret $\E[R_n]$ at time $n$ is given by
		%\small
		\begin{eqnarray}
			\E[R_n]\displaystyle & \leq & \bigo(1) \\
			& + & \displaystyle \max_{t\in S}G(t) \left[ \bigo\left(\log \left( \displaystyle \max_{t\in S} \xi(t) \right)\right) + \bigo(1)\right].
		\end{eqnarray}
	\end{theorem}
\end{tcolorbox}
The first sum in \eqref{initialization_Algorithm::UCB_soft_} is the exact mean regret of the initialization phase of Algorithm \ref{Algorithm::UCB_soft}. For the rounds after the initialization phase, the mean regret is bounded by the quantity in \eqref{convSeries_}, which is almost identical to the bound of the usual UCB algorithm if we assume $G(t)=1$ (i.e., rewards are not modified by the multiplier function). Similarly to the UCB algorithm with threshold, it is possible to compute the minimum number of times an arm will be pulled before or on turn $t$ (see Lemma \ref{Lemma::minimum_pulls_UCB_soft} and Lemma \ref{Lemma::minimum_pulls_UCB_soft_different_ranges} in Appendix \ref{proof_UCB2}).
\normalsize

\subsection{Regulating greed with variable arm pool}\label{Section::z_pool}%\vspace{-0.25cm}

\begin{figure}[]
	\makebox[\textwidth][c]{ %to center figures!
		\begin{subfloatrow}
			\subfloat[\small{The arm pool when $G(t)$ is low: most arms are included, and the algorithm is in exploration mode. }]{{\includegraphics[width=6.5cm,height=5.0cm]{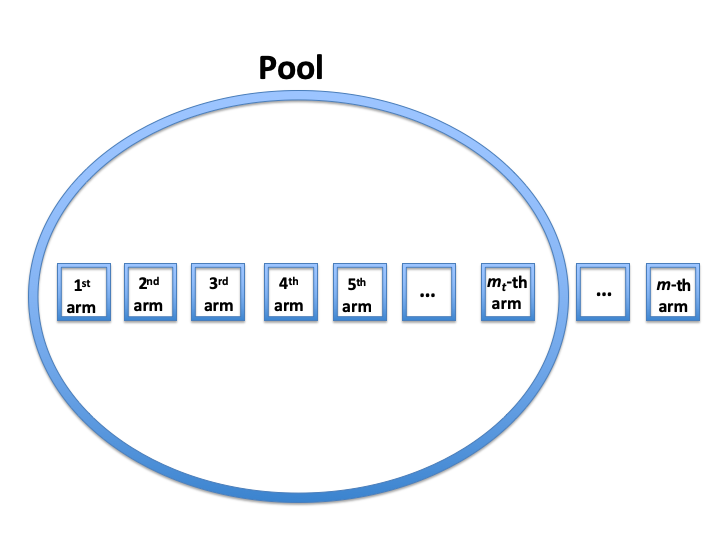} }\label{Figure::VP_Glow}}%
			\;\;
			\qquad
			\subfloat[\small{The arm pool when $G(t)$ is high: only arms with the highest estimated mean reward are included, and the algorithm is in exploitation mode.}]{{\includegraphics[width=6.5cm,height=5.0cm]{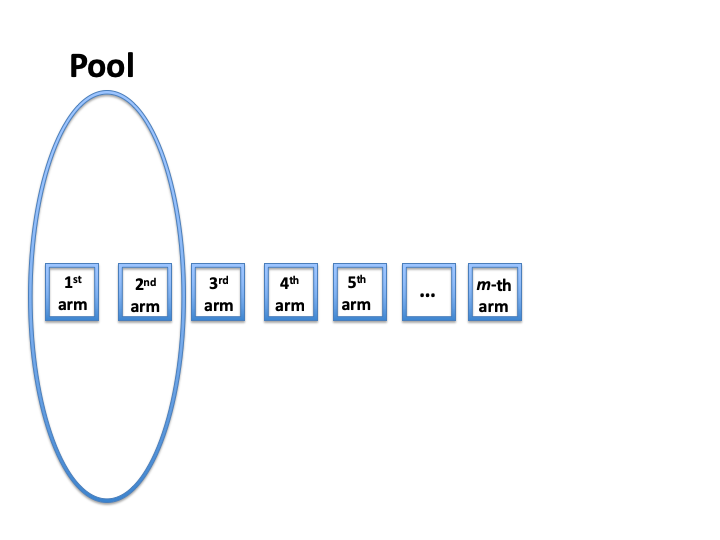} }\label{Figure::VP_Ghigh}}\;\;
		\end{subfloatrow}
	}
	\caption{}
	\label{Figure::variable_pool}
\end{figure}

In Algorithm \ref{Algorithm::z_pool} we present a policy that regulates greed by varying the size $m_t$ of the pool of arms from which we are allowed to choose (uniformly at random). When the greed function is high, the pool size of arms $m_t$ shrinks (possibly to just one arm: the one with the highest mean reward so far), so that we choose randomly among the arms that performed best. When the greed function is low we choose randomly among a larger pool (which could possibly contain all the arms). Unlike $\varepsilon$-greedy algorithms that can go back to very bad arms when exploring, this algorithm only explores among all arms when $G(t)$ is very small. The size of the pool is given by
\begin{equation}
	m_t = \min\left( m, \max\left( 1, \left\lfloor\frac{c m}{t G(t)} \right\rfloor\right) \right),
\end{equation} 
where $c>1$.
Figure \ref{Figure::VP_Glow} and \ref{Figure::VP_Ghigh} show an example of the pool behavior when $G(t)$ is low or when it is high.

\RestyleAlgo{boxruled}
\begin{algorithm}
	\caption{variable arm pool algorithm}\nllabel{Algorithm::z_pool}
	\SetKwInOut{Input}{Input}
	\Input{number of rounds $n$, number of arms $m$, a constant $c > 1$, and $\{G(t)\}_{t=1}^n$\;} %ADD INITIALIZATION
	\For{$t=m+1$ \KwTo $n$}{ 
		{Set pool size to $m_t = \min\left( m, \max\left( 1, \left\lfloor\frac{c m}{t G(t)}\right\rfloor \right) \right)$\;}
		{}
		{Play arm $j$ at random from the pool \;}
		{Get reward $G(t)X_j$\;}
		{Update $\widehat{X}_j$\;} 
	}
\end{algorithm}

Let us define $$
\lambda_t=%\frac{1}{2m}\sum_{s=1}^{t} \ONE\left\{\min\left( m, \max\left( 1, \frac{c m}{s G(s)}\right) \right) = m\right\} =
\frac{1}{2m}\sum_{s=1}^{t} \ONE\left\{m_s = m\right\},$$
where $\sum_{s=1}^{t} \ONE\left\{m_s = m\right\}$ is the number of times that the pool contained all the arms up to turn $t$. 
For the following theorem we require that the multiplier function $G(t)$ is such that $\lambda_t > \gamma\log(t)$ for some $\gamma > 5$.
\tcbset{colback=white}
\begin{tcolorbox}
	\begin{remark}\label{Remark::G_prime}
		If $G(t)$ does not satisfy the requirement that $\lambda_t \geq \gamma \log(t)$ it is easy to construct a new multiplier function $G'(t)$ by first finding the set $S=\{ t: \lambda_t > \lceil \gamma \log(t-1) \rceil \}$  and then by defining 
		\begin{equation*}
			G'(s) = \begin{cases}
				(c-1)/s & \mbox{for } s \in \{t, t+1, \cdots, t+2m\} \mbox{ if }  s\in S  \\
				G(s)    & \mbox{otherwise}.
			\end{cases}
		\end{equation*}
		If we do use $G'(t)$ instead of $G(t)$ in the algorithm, the bound that we present below holds for $G'(t)$ instead of $G(t)$.
	\end{remark}
\end{tcolorbox}
%If $G(t)$ does not satisfy the requirement that $\lambda_t \geq \gamma \log(t)$, it is easy to construct a new function that can be used in the algorithm instead of the multiplier function. The new function is called $G'(t)$, and we construct it by first finding the set $S=\{ t: \lambda_t > \lceil \gamma \log(t-1) \rceil \}$  and then by defining 
%\begin{equation*}
%G'(s) = \begin{cases}
%(c-1)/s & \mbox{for } s \in \{t, t+1, \cdots, t+2m\} \mbox{ if }  s\in S  \\
%G(s)    & \mbox{otherwise}.
%\end{cases}
%\end{equation*} 
%If we do use $G'(t)$ instead of $G(t)$ in the algorithm, the bound that we present below holds for $G'(t)$ instead of $G(t)$.

The following theorem provides a finite-time bound on the regret after $n$ rounds.
\tcbset{colback=blue!2!white}
\begin{tcolorbox}
	\begin{theorem}[variable pool algorithm]\label{Theorem::regret_variable_pool}
		The bound on the mean regret $\E[R_n]$ at time $n$ is given by
		\begin{equation}
			\E[R_n]\displaystyle  \leq  \sum_{t=1}^n \sum_{j=1}^m \Delta_j G(t) \frac{2}{m_t}\beta_j^{VP}(t) ,
		\end{equation}	
		where
		\begin{equation}%\label{beta::z_pool}
			\beta_j^{VP}(t)=\gamma\log(t)(t)^{-\gamma/5} + \frac{2}{(\Delta_j^{*-m_t})^2} t^{-\frac{\gamma(\Delta_j^{*-m_t})^2}{2}}.
		\end{equation}
	\end{theorem}
\end{tcolorbox}
Here $\beta_j^{VP}$ is the upper bound on the probability of choosing arm $j$. This algorithm tends to perform well in the experiments. Like previous algorithms, it also regulates greed based on the multiplier function (i.e., if $G(t)$ is high your pool may just contain the arm with the highest mean reward and that translates into exploitation of the best arm), but it goes back to play bad arms less often than $\varepsilon$-greedy or UCB policies.

%Algorithm: After initialization, if $G(t)<z$ play arm $j$ with probability
%$$ \frac{\varepsilon_t}{m} + (1-\varepsilon_t)\P(\widehat{X}_j \geq \widehat{X}_i \; \forall i ), $$
%where 
%$$ \varepsilon_t = \min\left\{ 1,\frac{c m }{d^2 \tilde{t}}\right\}, $$
%$t$ is the current round, $\tilde{t}$ is the number of rounds under the threshold $z$, $0<d< \min_{1\leq j \leq m} \Delta_j$ and $m$ is the number of arms. If $G(t)\geq z$ play arm $j$ with probability $\P(\widehat{X}_j \geq \widehat{X}_i \; \forall i )$.\\\\

\normalsize

\section{Experimental results: simulated environment}\label{Section::Experimental_Results}

\begin{figure}[]%
	\makebox[\textwidth][c]{ %to center figures!
		\begin{subfloatrow}
			\subfloat[\small{The \emph{Wave Greed}}]{{\includegraphics[width=4.0cm,height=4.5cm]{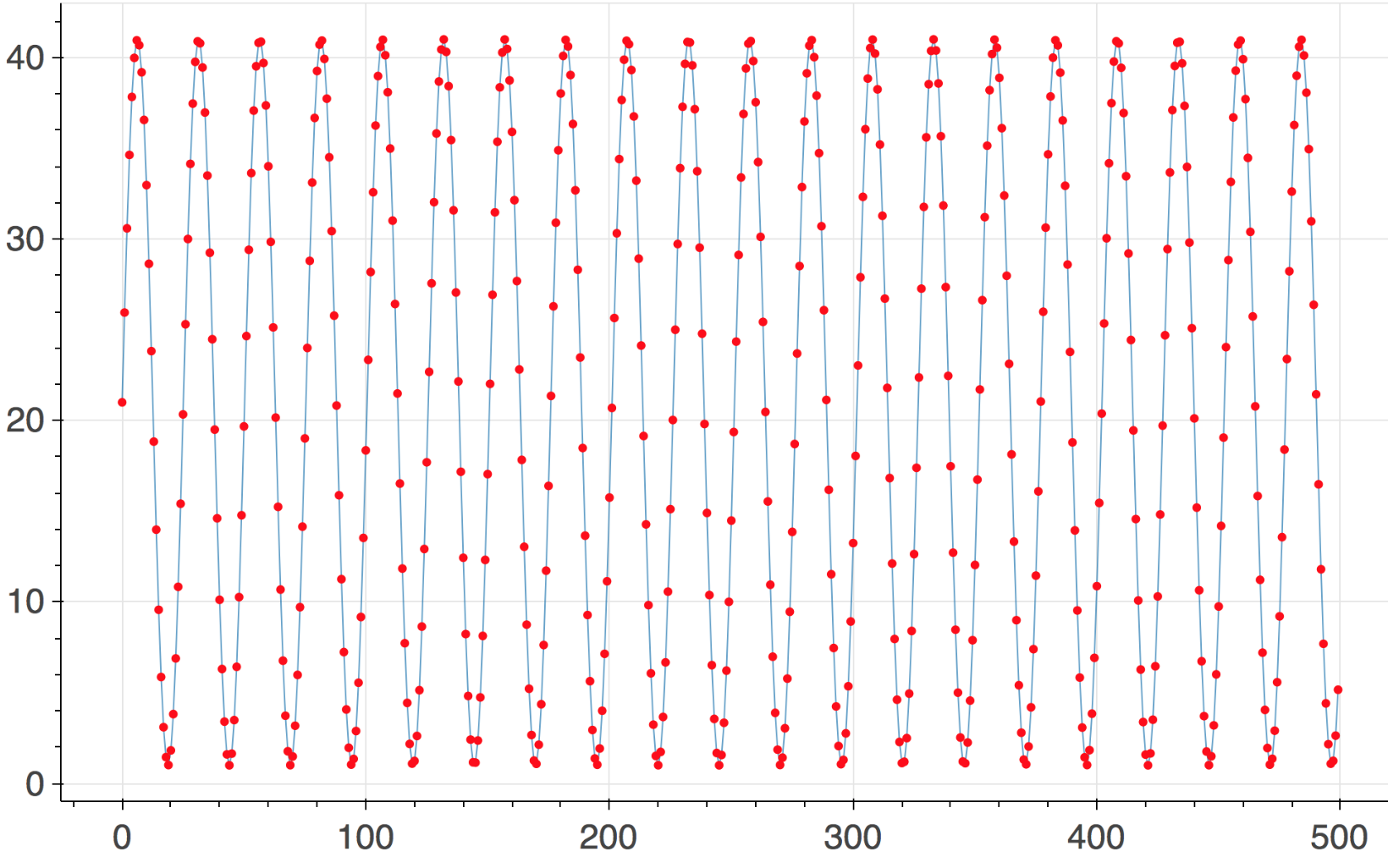} }\label{Wave_greed}}%
			\;\;
			\qquad
			\subfloat[\small{The \emph{Christmas Greed}}]{{\includegraphics[width=4.0cm,height=4.5cm]{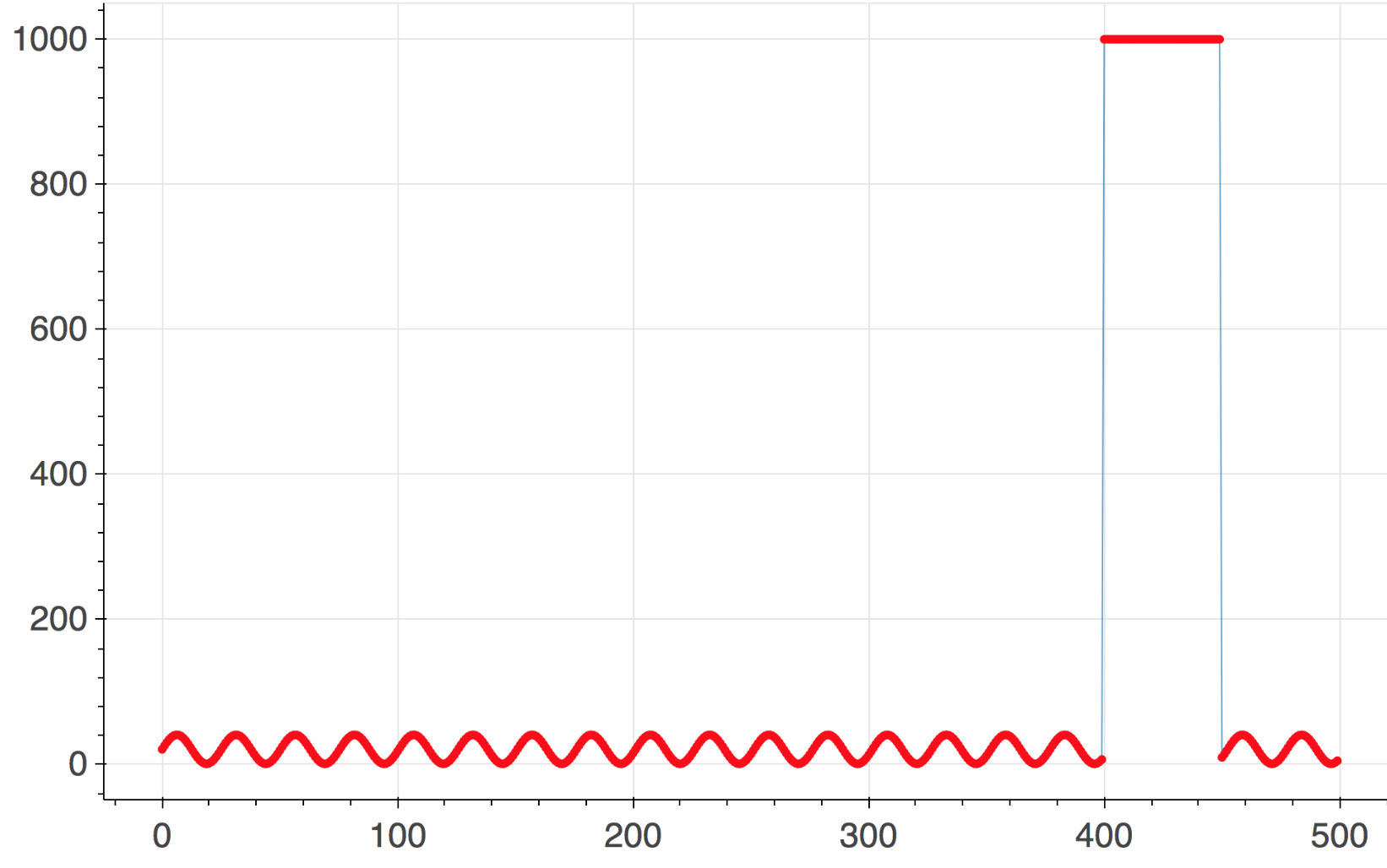} }\label{Christmas_greed}}\;\;
			\qquad
			\subfloat[\small{The \emph{Step Greed}}]{{\includegraphics[width=4.0cm,height=4.5cm]{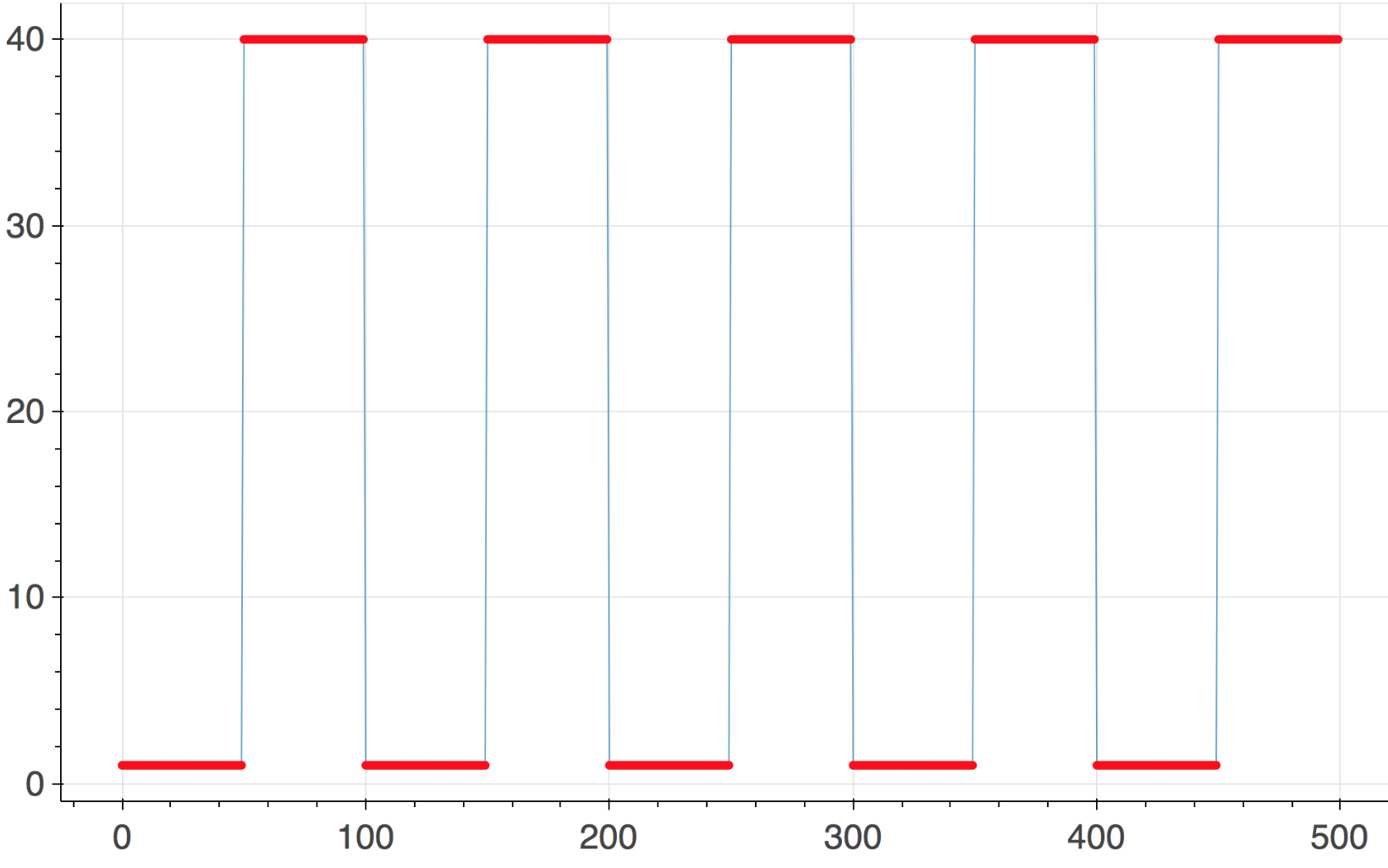} }\label{Step_greed}}\;\;
		\end{subfloatrow}
	}
	\caption{Shapes of the multiplier functions used in the experiments.}
\end{figure}

We consider three types of multiplier function $G(t)$:
\begin{itemize}
	\item The \emph{Wave Greed} (Figure \ref{Wave_greed}): in a Wave-type greed function, rewards are multiplied following the trend of a periodic wave: $G(t)=21+20\sin(0.25t)$ for $t \in \{1, \cdots, n\}$. We aim to exploit the best arm found so far during the peaks, while balancing exploration and exploitation during low-rewards periods. This behavior mimics weekly patterns.
	\item The \emph{Christmas Greed} (Figure \ref{Christmas_greed}): similarly to the Wave greed, rewards are multiplied following the trend of a wave, but a big peak (which we call ``Christmas", in analogy to the phenomenon of the boom of customers during the Christmas holidays) appears towards the end of the game. Formally,  $G(t)=21+20\sin(0.25t)$ if $t \in [1,0.8n) \cup (0.9n,n]$, and $G(t)=1000$ if $t \in [0.8n,0.9n]$.
	\item The \emph{Step Greed} (Figure \ref{Step_greed}): a Step-type greed function assumes only two values: one low and one high. In our case, we choose $G(t)= 40$ (high value) if $t \in [0.1n,0.2n) \cup [0.3n,0.4n) \cup [0.5n,0.6n) \cup [0.7n,0.8n) \cup [0.9n, n]$, otherwise $G(t)=1$ (low value). We aim to exploit the best arm so far when the greed function assumes its high value, while balancing exploration and exploitation when it assumes its low value.
\end{itemize}
In this section we report the mean results (over 50 games) of the final cumulative rewards of games consisting of 1500 turns and 200 arms. A complete set of results with different number of arms (25, 50, 100, and 200) and different game-lengths (500, 1000, and 1500 turns) can be found in 
Appendix \ref{Appendix::Experiments}. 

We consider rewards coming from two different distributions:
\begin{itemize}
	\item Bernoulli distribution: when rewards come from Bernoulli distributions, each arm is assigned a probability of success $p_j$ drawn randomly from an uniform distribution on $[0,1]$;
	\item Truncated-Normal distribution: when rewards come form Truncated-Normal distributions, each arm is assigned a mean reward $\mu_j$ drawn randomly from an uniform distribution on $[0,1]$, and a standard deviation $\sigma = 0.1$. Rewards for arm $j$ are then drawn from a Normal $N_j \sim \mathcal{N}(\mu_j,1)$, but are bounded in $[0,1]$ so that the reward at time $t$ is given by $X_j(t) = \max( 0 , \min( 1, N_j  ) )$.
\end{itemize}
In the algorithms, it is also possible to use Normal distributed rewards, but since the theorems about UCB algorithms require the rewards to be bounded, we preferred to keep this assumption also in the experiments. 

The standard UCB and $\varepsilon$-greedy algorithms are not suitable for the setting in which the rewards are altered by the multiplier function. Thus, in their current form, it would not be fair to compare directly with them. The fact that rewards are multiplied by $G(t)$ would irremediably bias all the estimations of the mean rewards, leading standard UCB and $\varepsilon$-greedy to choose arms whose rewards were artificially inflated because they happened to be played in a high reward period. For example, suppose we show an ad on a website at lunch time: many people will see it because at that time the web-surfing is at its peak (i.e., the $G(t)$ multiplier is high). So even if the ad was bad, we may register more clicks than a good ad showed at 3:00AM (i.e., when the $G(t)$ multiplier is low). To obtain a fair comparison, we created ``smarter'' versions of the UCB and $\varepsilon$-greedy algorithms in which the rewards are discounted at each round by $G(t)$; then, the old version of the algorithms can be smarter in that they can produce accurate estimates of the mean reward for each arm. The smarter version of the usual UCB algorithm is presented in Algorithm \ref{Algorithm::UCB_slightly_smarter} and the one for the $\varepsilon$-greedy algorithm is shown in Algorithm \ref{Algorithm::epsilon_slightly_smarter}. 

\begin{figure}[]%
	\makebox[\textwidth][c]{ %to center figures!
		\begin{subfloatrow}
			\subfloat[\small{Rewards from Bernoulli distributions. }]{{\includegraphics[width=6.7cm,height=5.0cm]{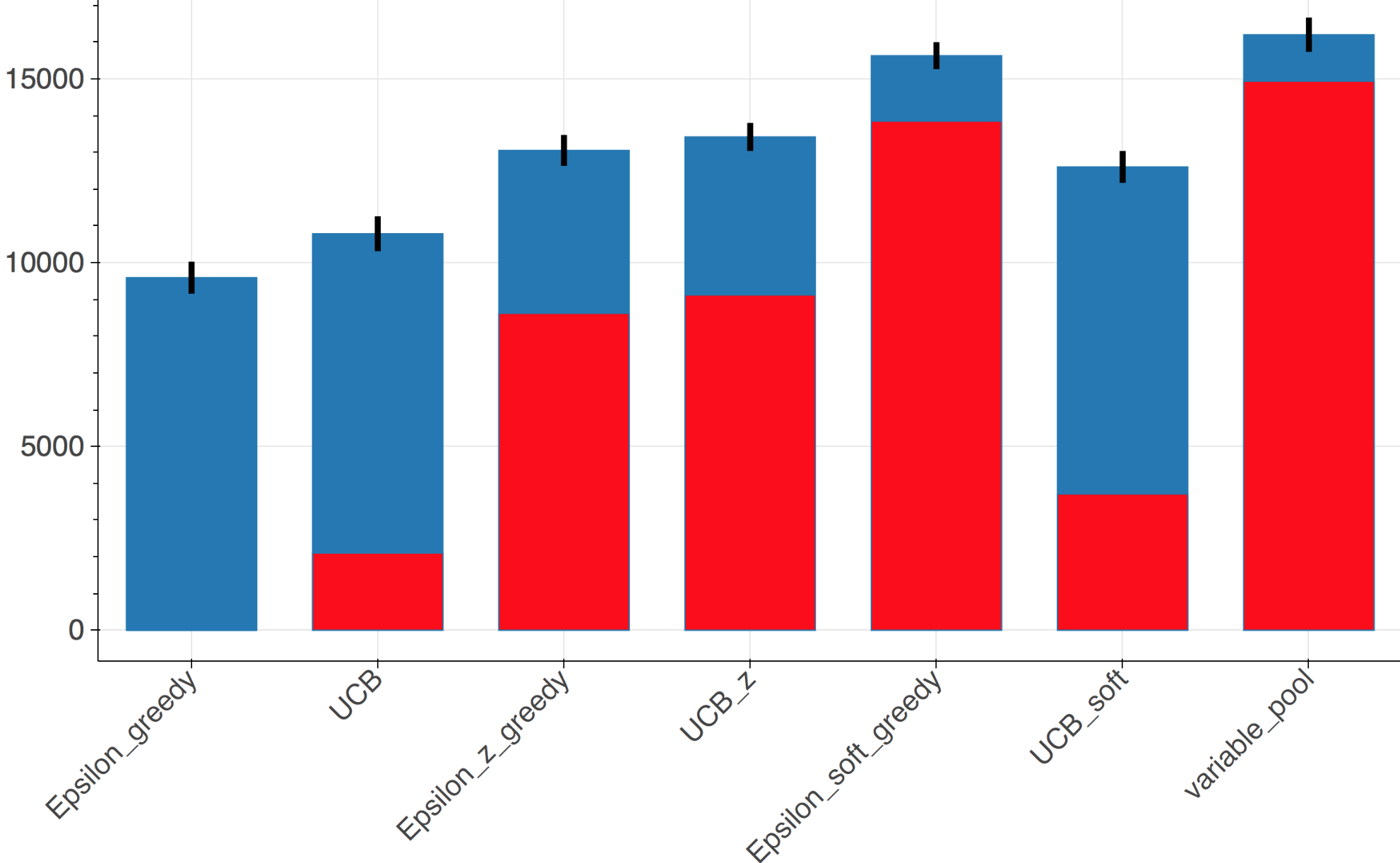} }\label{Bernoulli_Wave_200a_1500t_h}}%
			\;\;
			\qquad
			\subfloat[\small{Rewards from truncated Normal distributions.
			}]{{\includegraphics[width=6.7cm,height=5.0cm]{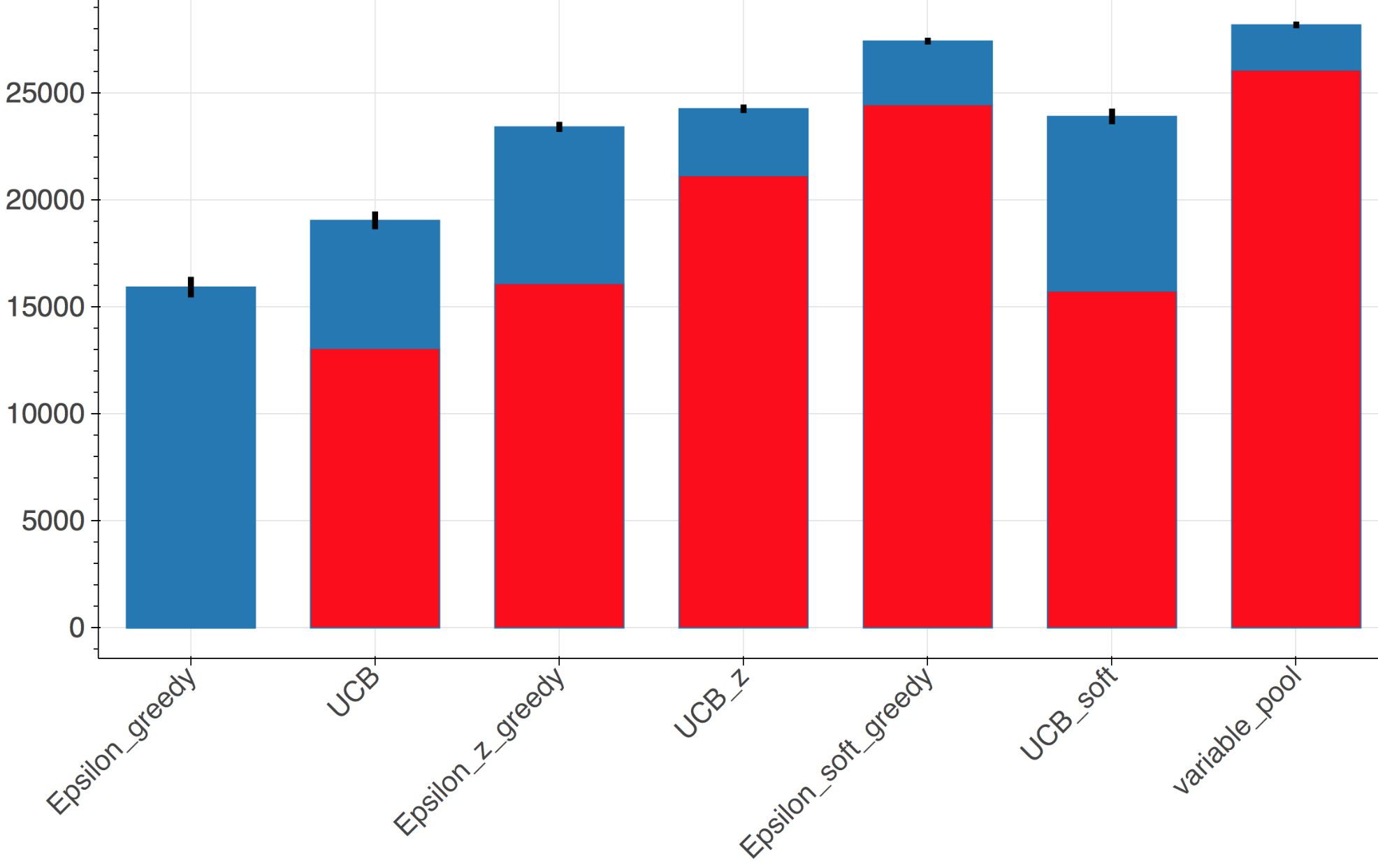} }\label{Truncated_Normal_Wave_200a_1500t_h}}\;\;
		\end{subfloatrow}
	}
	\caption{Comparison of average final rewards in games with 200 arms, 1500 turns, and a Wave-type greed function. The first two bars refer to the smarter version of the $\varepsilon$-greedy and UCB algorithms, while the other five bars refer to our algorithms that regulate greed over time. }\label{Figure::200a_1500t_Wave_h}
\end{figure}

\begin{figure}[]%
	\makebox[\textwidth][c]{ %to center figures!
		\begin{subfloatrow}
			\subfloat[\small{Rewards from Bernoulli distributions. }]{{\includegraphics[width=6.7cm,height=5.0cm]{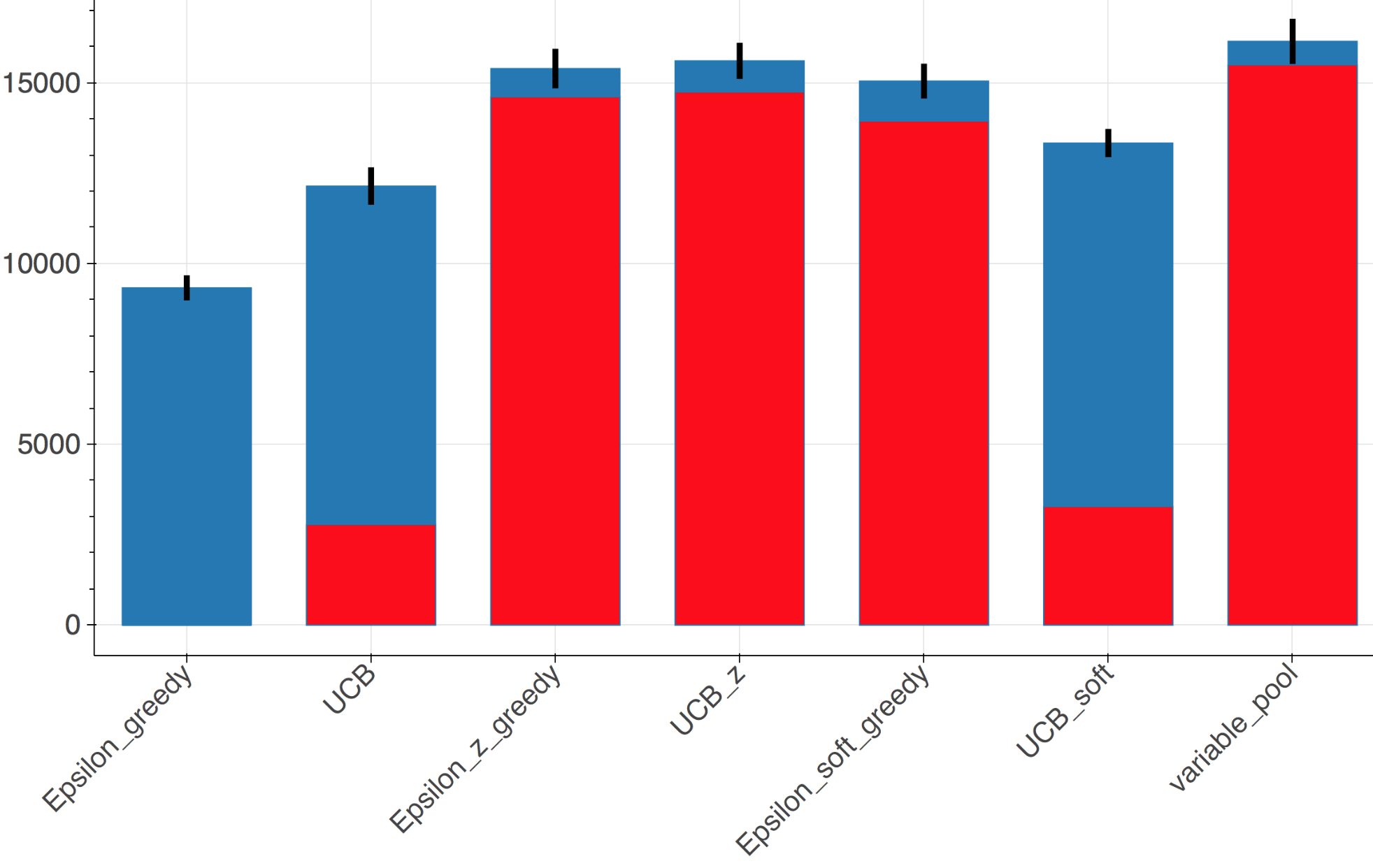} }\label{Bernoulli_Step_200a_1500t_h}}%
			\;\;
			\qquad
			\subfloat[\small{Rewards from truncated Normal distributions.
			}]{{\includegraphics[width=6.7cm,height=5.0cm]{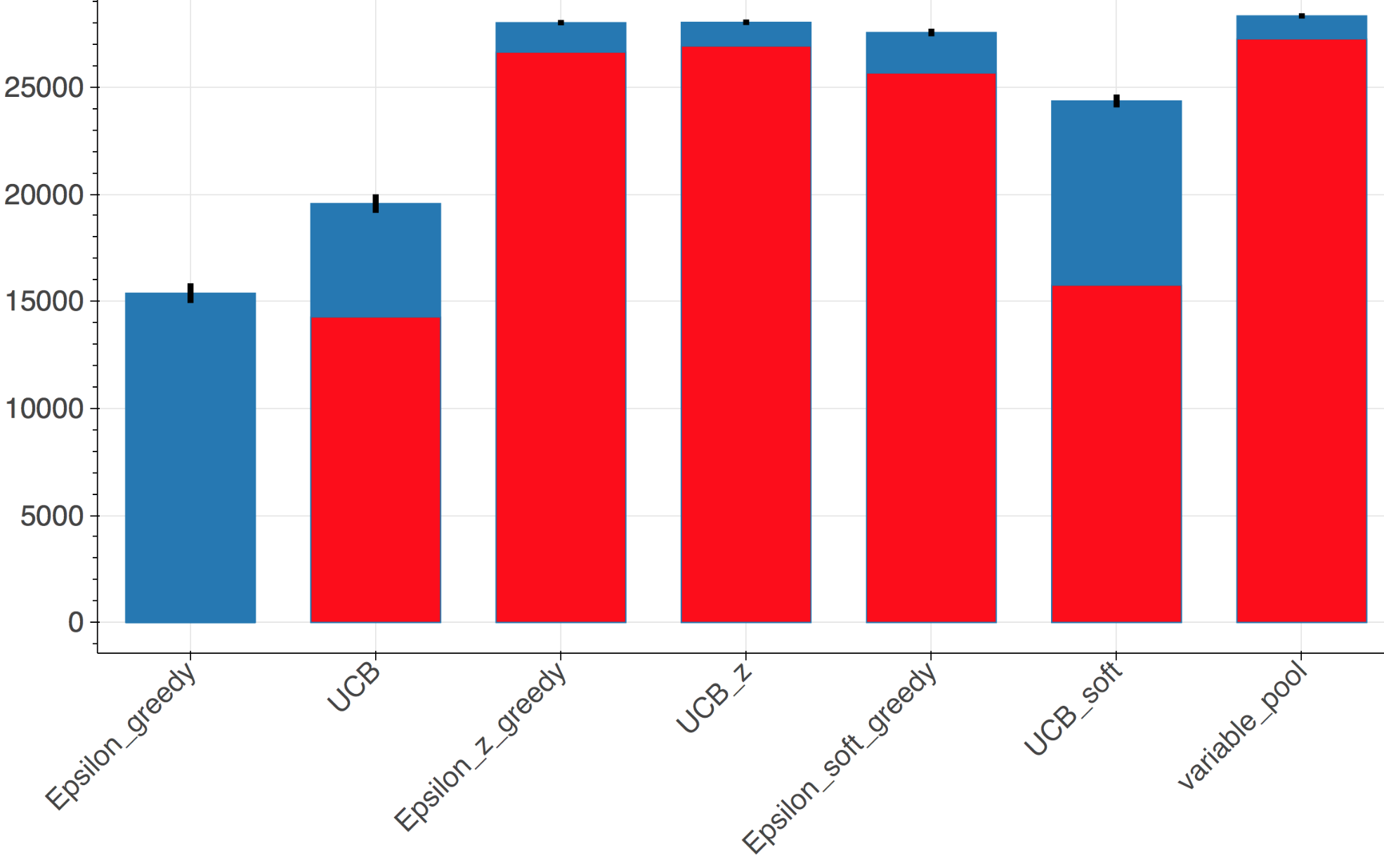} }\label{Truncated_Normal_Step_200a_1500t_h}}\;\;
		\end{subfloatrow}
	}
	\caption{Comparison of average final rewards in games with 200 arms, 1500 turns, and a Step-type greed function.}\label{Figure::200a_1500t_Step_h}
\end{figure}

\begin{figure}[]%
	\makebox[\textwidth][c]{ %to center figures!
		\begin{subfloatrow}
			\subfloat[\small{Rewards from Bernoulli distributions. }]{{\includegraphics[width=6.7cm,height=5.0cm]{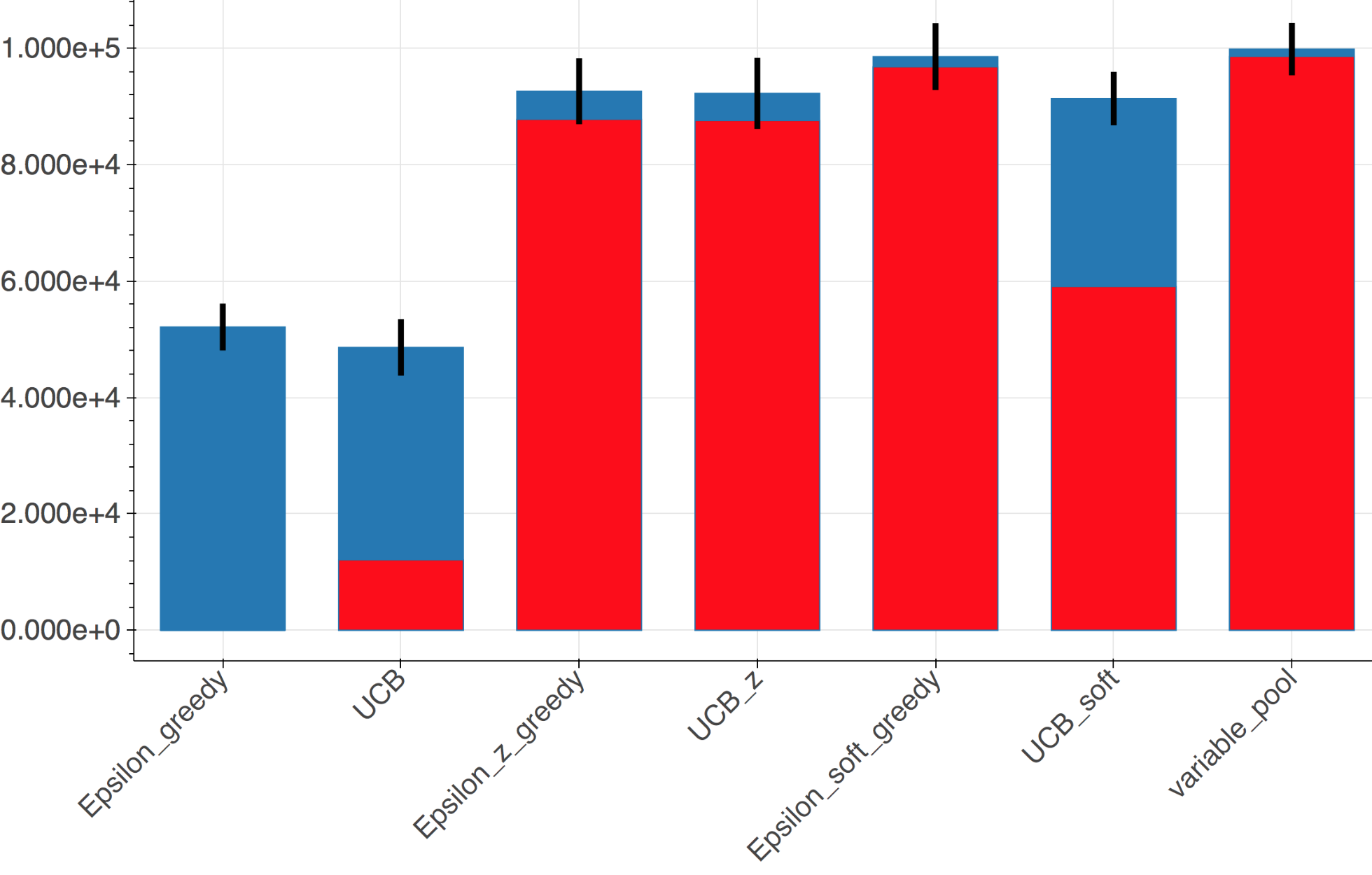} }\label{Bernoulli_Christmas_200a_1500t_h}}%
			\;\;
			\qquad
			\subfloat[\small{Rewards from truncated Normal distributions.
			}]{{\includegraphics[width=6.7cm,height=5.0cm]{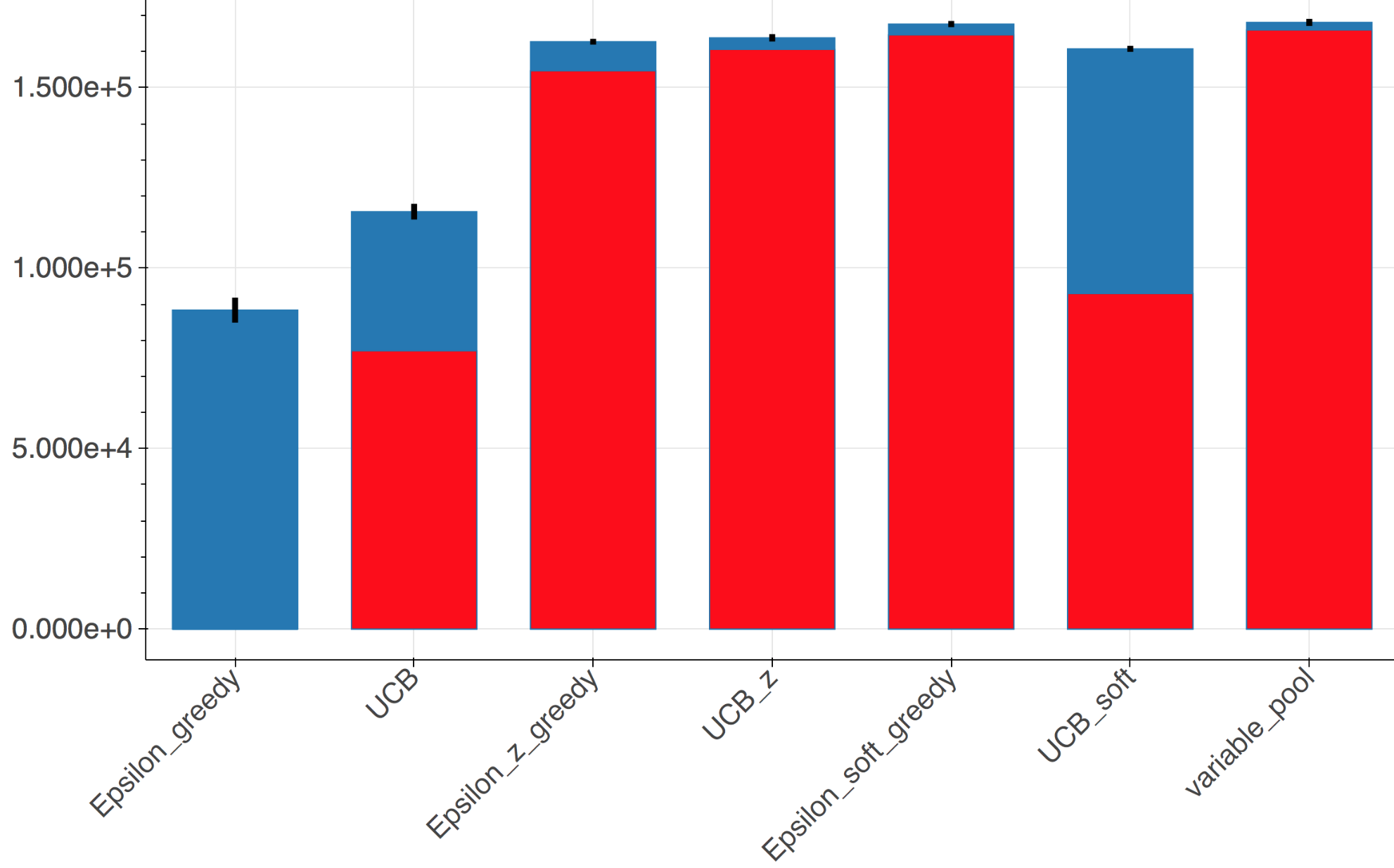} }\label{Truncated_Normal_Christmas_200a_1500t_h}}\;\;
		\end{subfloatrow}
	}
	\caption{Comparison of average final rewards in games with 200 arms, 1500 turns, and a Christmas-type greed function.}\label{Figure::200a_1500t_Christmas_h}
\end{figure}

In Figure \ref{Figure::200a_1500t_Wave_h}, \ref{Figure::200a_1500t_Step_h}, and \ref{Figure::200a_1500t_Christmas_h} we show the average cumulative rewards at the end of the game. The red part of the bar indicates what portion of the rewards came from ``pure exploitation.'' The definition of ``pure exploitation'' depends on the algorithm used:
\begin{itemize}
	\item pure exploitation in $\varepsilon$-greedy algorithms: when the algorithms decide to exploit. This is forced in algorithms with threshold when the greed function is above that threshold;% In the simulations, all the $\varepsilon$-greedy algorithms are initialized with the same parameters;
	\item pure exploitation in UCB algorithms: when the arm played has the highest estimated mean reward. This is forced in algorithms with threshold when the greed function is above that threshold;
	\item pure exploitation in the variable-pool algorithm: when the pool size is 1.
\end{itemize}
.
%\newpage

\begin{figure}[]%
	\makebox[\textwidth][c]{ %to center figures!
		\begin{subfloatrow}
			\subfloat[\small{Bernoulli rewards. }]{{\includegraphics[width=3in]{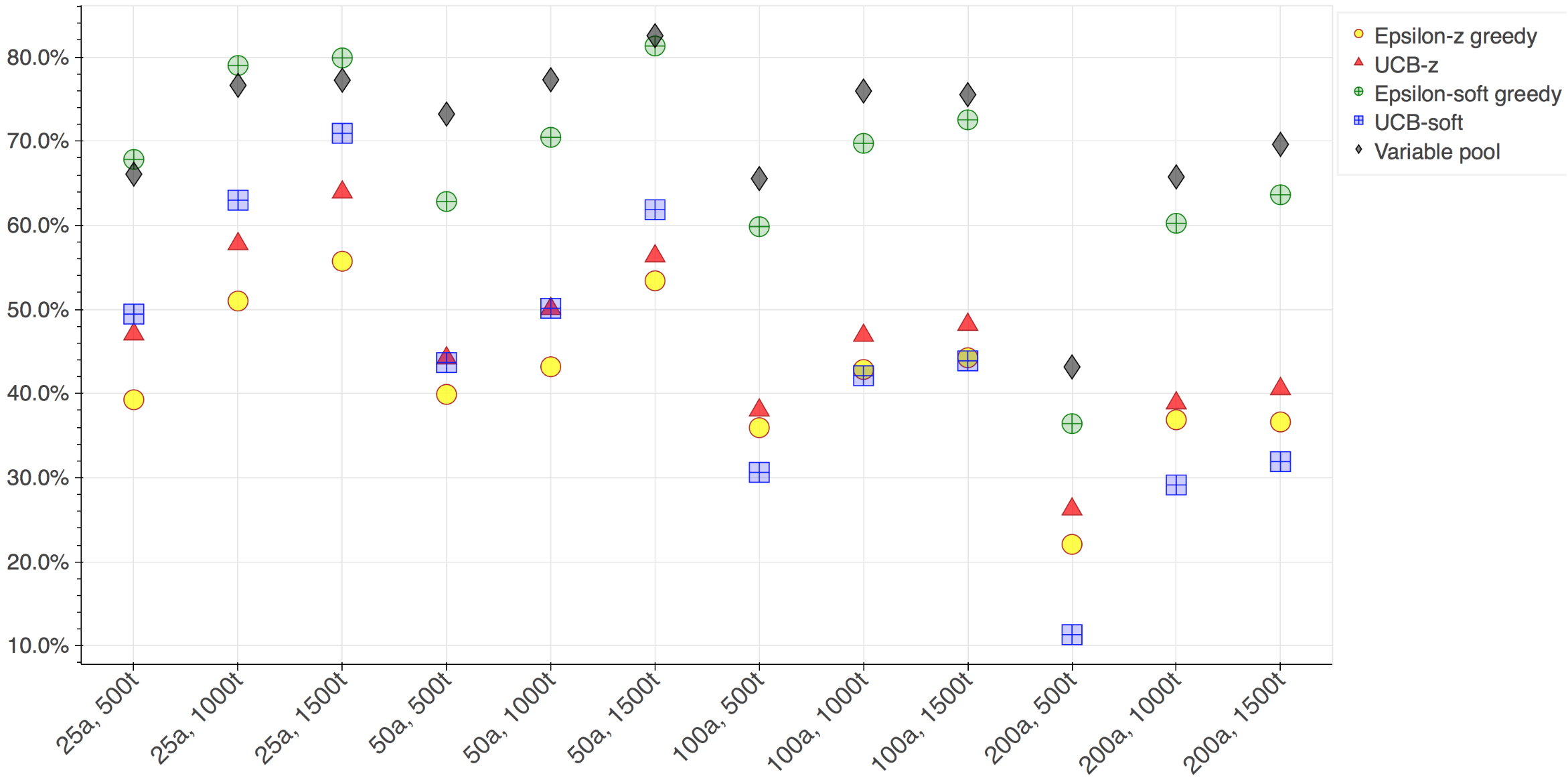} }\label{BEW_h}}%
			%\qquad
			\subfloat[\small{Truncated-Normal rewards.
			}]{{\includegraphics[width=3in]{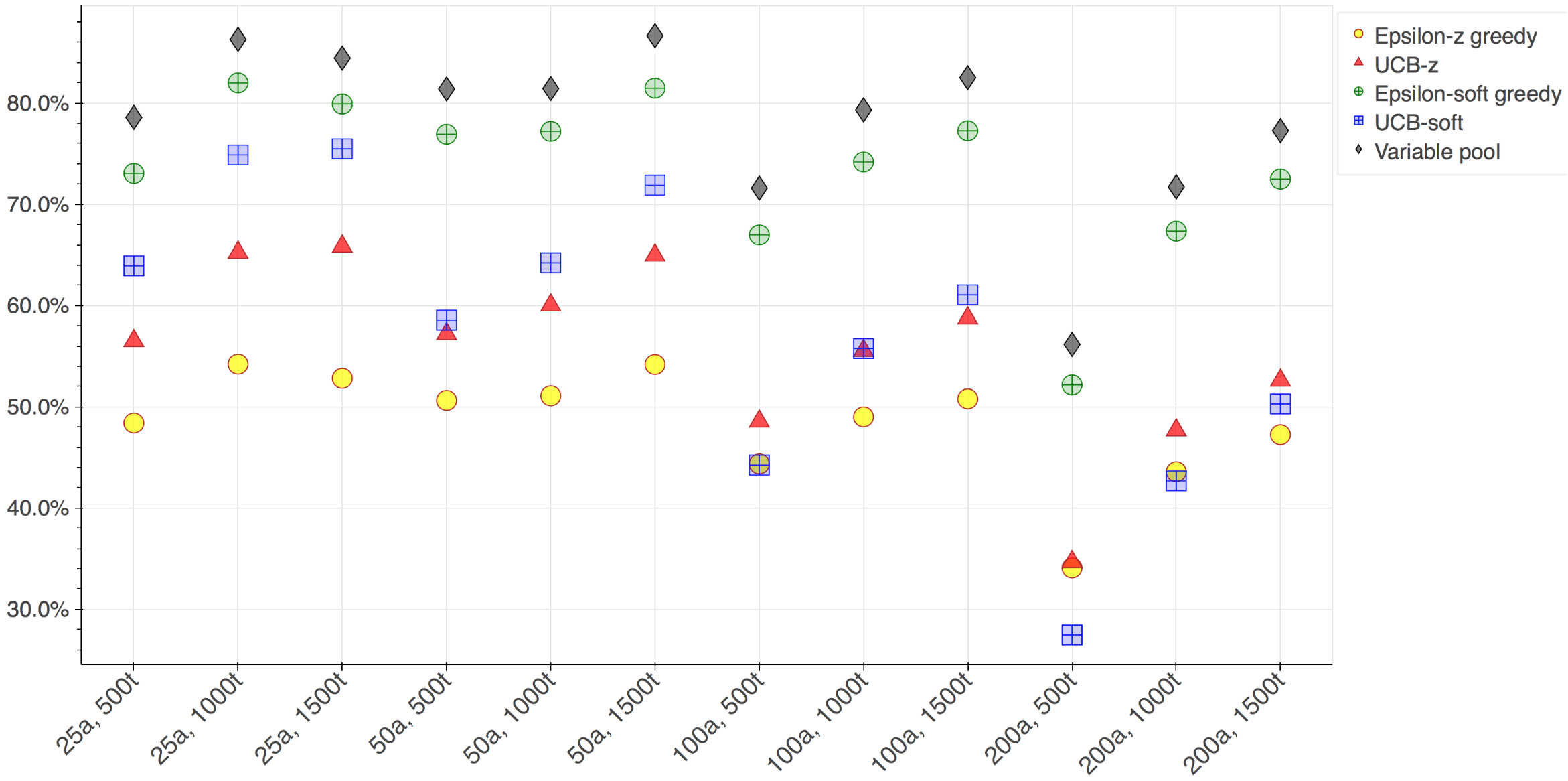} }\label{NEW_h}}\;\;
		\end{subfloatrow}
	}
	\caption{Cumulative Reward increase when regulating greed over time compared to the (smarter) $\varepsilon$-greedy algorithm (Wave-greed function).}\label{EW}
\end{figure}

\begin{figure}[]%
	\makebox[\textwidth][c]{ %to center figures!
		\begin{subfloatrow}
			\subfloat[\small{Bernoulli rewards. }]{{\includegraphics[width=3in]{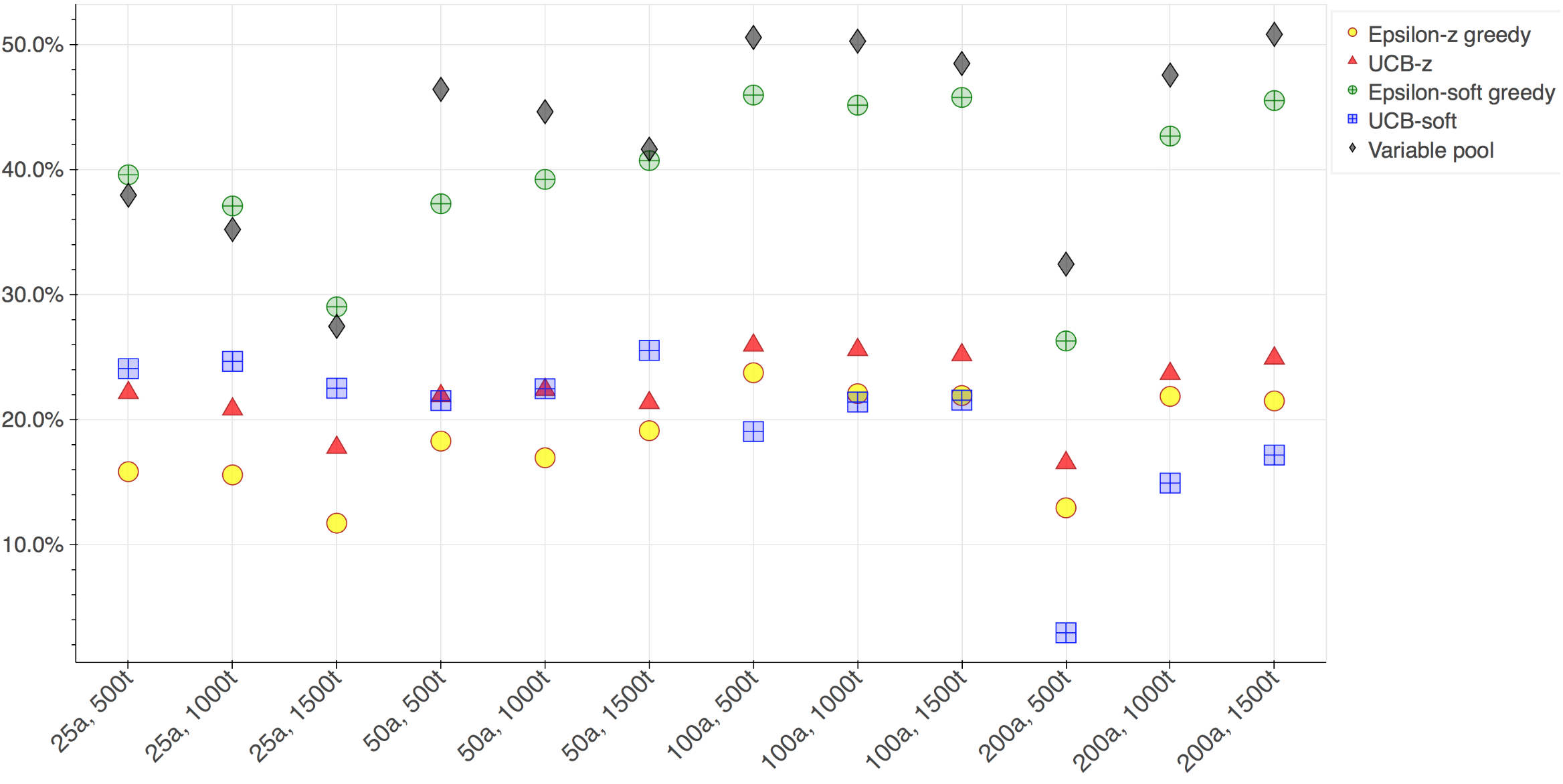} }\label{BUW_h}}%
			\subfloat[\small{Truncated-Normal rewards.
			}]{{\includegraphics[width=3in]{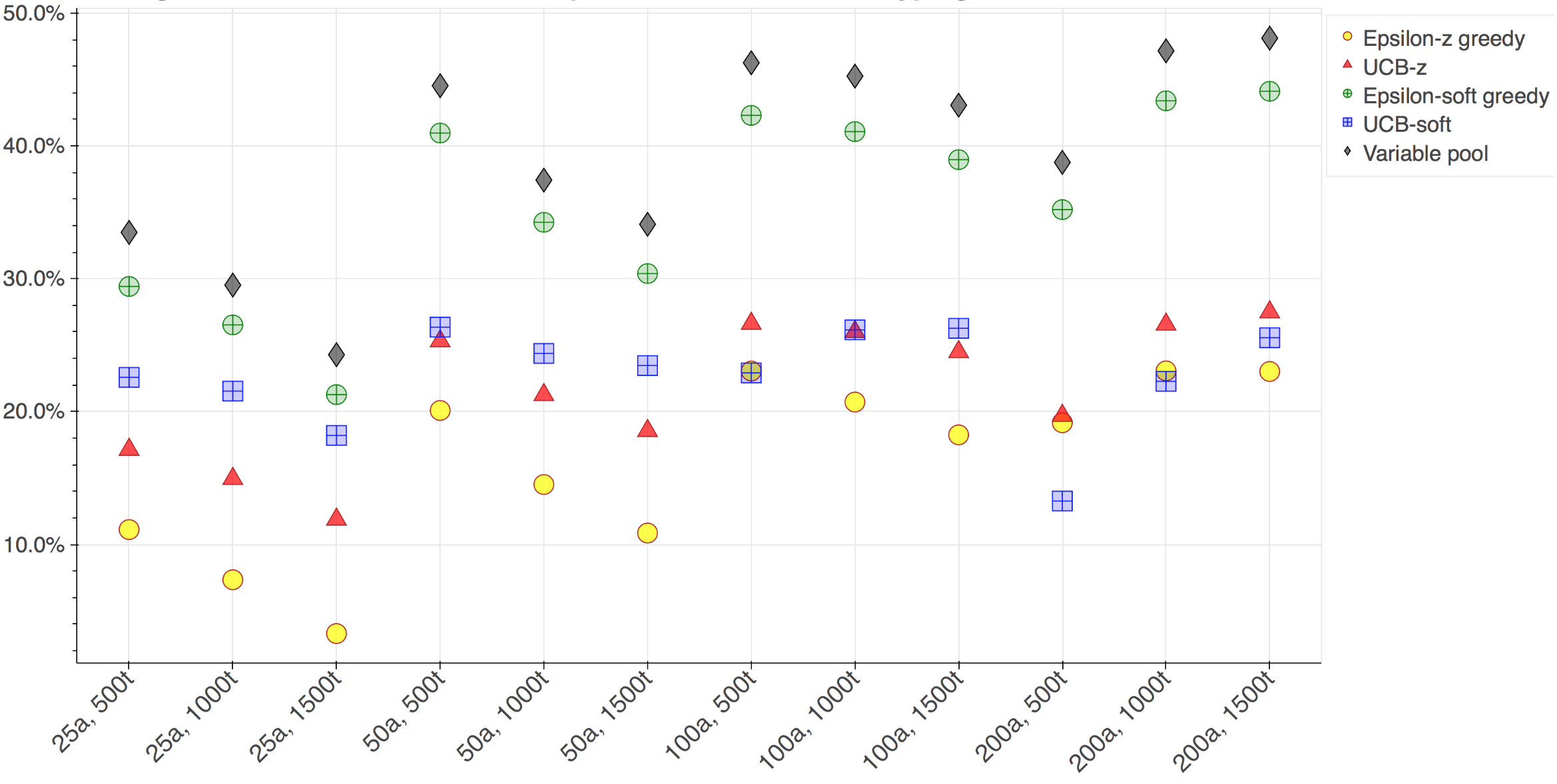} }\label{NUW_h}}\;\;
		\end{subfloatrow}
	}
	\caption{Cumulative Reward increase when regulating greed over time compared to the (smarter) UCB algorithm (Wave-greed function).}\label{UW}
\end{figure}

In Figure \ref{BEW_h} and \ref{NEW_h} we show the average increase in final rewards when regulating greed over time. The comparison is with respect to the (smarter) $\varepsilon$-greedy algorithm when the greed function is of the Wave type. In Figure \ref{BUW_h} and \ref{NUW_h} we show the average increase in final rewards when regulating greed over time. This time the comparison is with respect to the (smarter) UCB algorithm when the greed function is of the Wave type. The results show the power of regulating greed over time: by exploiting more when the greed function is high and exploring more when it is low, the final cumulative rewards are much higher than those of algorithms that do not regulate greed over time. Refer to the full set of experiments in Appendix \ref{Appendix::Experiments} for the comparisons with the other types of greed functions. \\

%In Figure \ref{BEW_h} and \ref{NEW_h} we show the average increase in rewards with respect to the (smarter) $\varepsilon$-greedy algorithm when the greed function is the Wave greed. In Figure \ref{BUW_h} and \ref{NUW_h} we show the average increase in rewards with respect to the (smarter) UCB algorithm when the greed function is the Wave greed. Refer to Appendix \ref{Appendix::Experiments} for the comparisons with the other types of greed functions. \\
%The results show the power of regulating greed over time: by exploiting more when the greed function is high and exploring more when it is low, the final cumulative rewards are much higher than the ones of algorithms that do not regulate greed over time.

\RestyleAlgo{boxruled}
\begin{algorithm}[]
	\caption{Smarter version of the standard $\varepsilon$-greedy algorithm}\nllabel{Algorithm::epsilon_slightly_smarter}
	\SetKwInOut{Input}{Input}
	\SetKwInOut{Output}{output}
	\SetKwInOut{Loop}{Loop}
	\SetKwInOut{Initialization}{Initialization}
	%	\SetKwData{Left}{left}\SetKwData{This}{this}\SetKwData{Up}{up}
	%	\SetKwFunction{Union}{Union}\SetKwFunction{FindCompress}{FindCompress}
	\Input{number of rounds $n$, number of arms $m$, a constant $c>10$, a constant $d$ such that $d<\min_j \Delta_j$ and $0<d<1$, sequences $\{\varepsilon_t\}_{t=1}^n = \min\left\{1,c m/d^2 t\right\}$ and $\{G(t)\}_{t=1}^n$ }
	\Initialization{play all arms once and initialize $\widehat{X}_j$ (as defined in \eqref{mean_estimator}) for each $j=1,\cdots,m$}
	\For{$t=m+1$ \KwTo $n$}{ 
		{with probability $\varepsilon_t$ play an arm uniformly at random (each arm with probability $\frac{1}{m}$), otherwise (with probability $1-\varepsilon_t$) play arm $j$ such that $ \widehat{X}_j > \widehat{X}_i \; \forall i$\;}
		{Get reward $G(t)X_j(t)$\;}
		{\hspace{0.1cm}Update $\widehat{X}_j$\;}
	}
\end{algorithm}

\RestyleAlgo{boxruled}
\begin{algorithm}[]
	\caption{Smarter version of the standard UCB algorithm}\nllabel{Algorithm::UCB_slightly_smarter}
	\SetKwInOut{Input}{Input}
	\SetKwInOut{Output}{output}
	\SetKwInOut{Loop}{Loop}
	\SetKwInOut{Initialization}{Initialization}
	%	\SetKwData{Left}{left}\SetKwData{This}{this}\SetKwData{Up}{up}
	%	\SetKwFunction{Union}{Union}\SetKwFunction{FindCompress}{FindCompress}
	\Input{number of rounds $n$, number of arms $m$, sequence $\{G(t)\}_{t=1}^n$ }
	\Initialization{play all arms once and initialize $\widehat{X}_j$ (as defined in \eqref{mean_estimator}) for each $j=1,\cdots,m$}
	\For{$t=m+1$ \KwTo $n$}{ 
		{play arm $j$ with highest $\widehat{X}_{j,T_j(t-1)}+\sqrt{\frac{2 \,\log(t)}{T_j(t-1)}}$\;}
		{Get reward $G(t)X_j(t)$\;}
		{\hspace{0.1cm}Update $\widehat{X}_j$\;}
	}
\end{algorithm}

\newpage
\section{Soft UCB mortal algorithm}\label{Section::UCB_soft_mortal}
Since the dataset from the Yahoo! Webscope program has arms that become available and unavailable at different times, we also introduce the mortal bandit setting and how the UCB algorithm can be adapted to this scenario.

In the mortal case, the algorithm chooses in turn $t$ an arm among a set of available arms $M_t$. The set $M_t$ can change: arms may become unavailable (they ``die'') and can never be played again, or new arms may arrive (they ``are born''). Since the pool of available arms varies, the optimal arm to pull is not fixed anymore during the duration of the game. Let us denote the best arm in turn $t$ by $i^*_t$. Let $L_j$ be the set of turns during which arm $j$ is available, and let $s_j$ and $l_j$ be the first and last turns in $L_j$. We assume that the lifespan of each arm is given (for example, an arm could be a coupon that has a known expiration date). Algorithm \ref{Algorithm::UCB-LG} is a modification of the UCB algorithm that has two main advantages.
The first is to encourage exploration of arms that will be available over turns with high multiplicative reward $G(t)$. Each arm $j$ has a score given by the sum of the values of $G(t)$ over its lifespan:
$$\text{score}_j =\sum_{t \in L_j} G(t).$$ At each turn $t$, the upper confidence bound for arm $j$ will be scaled by the fraction of arms available that have score below that of $j$: $$\psi_{\text{future}}(j,t) =\frac{c}{|M_t|} \sum_{i \in M_t}\ONE{\{\text{score}_i \leq \text{score}_j\}}, $$ where $c\geq1$ is a positive constant (in the experiments, $c=2.0$). The function $\psi_{\text{future}}(j,t)$ compares the scores of the available arms and it is high if arm $j$ will be alive during high values of $G(t)$.

The second main advantage of Algorithm \ref{Algorithm::UCB-LG} is that it encourages exploitation over turns with high multiplicative reward $G(t)$. Similarly to the Soft UCB algorithm, the function $\xi_{\text{present}}(t)$ will shrink the upper confidence bound in the presence of high values of $G(t)$:
\begin{equation}\label{csi_mortal_function}
	\xi_{\text{present}}(t) = \left(1 + \frac{t}{G(t)}\right).
\end{equation}
In the initialization phase, a subset of arms $M_I \subset M_1$ are played. $M_I$ can be chosen by selecting arms with the highest values of $\psi_{\text{future}}(j,1)$ (for simplicity, in the experiments we set $M_I = M_1$). Theorem \ref{Theorem::G_mortal} provides a finite-time regret bound for Algorithm \ref{Algorithm::UCB-LG}. 
%the function $\psi_{\text{future}}(j,t)$ will be high if arm $j$ will be alive during high values of $G(t)$. 
%To compute $\psi_{\text{future}}(j,t)$ we calculate for each arm $j$ $$\text{score}_j =\sum_{t \in L_j} G(t). $$ Then, once we have the value of $\text{score}_j$ for all arms, we compute $$ \psi_{\text{future}}(j,t) =\frac{c}{|M_t|} \sum_{i \in M_t}\ONE{\{\text{score}_i \leq \text{score}_j\}}  ,$$
%where $c>1$ is a positive constant (in the experiments, $c=2.0$).
%For example,
%\begin{equation}\label{psi_function}
%\psi_{\text{future}}(j,t) = \log(l_j - t) \left( \sum_{t \in L_j} G(t)\right)^{\frac{1}{\alpha}},
%\end{equation}
%where, $\alpha >1$, $s_j$ and $l_j$ are the first and last turn that arm $j$ is alive.
%Another function can be
%\begin{equation}\label{psi_function2}
%\psi_{\text{future}}(j,t) = \log(l_j - t) \left( \sum_{t \in L_j, G(t) > z} G(t)\right),
%\end{equation}
%where $z$ is a chosen threshold.
%In high reward zones, exploitation will be preferred, while in low reward zones the algorithm will explore the arms.\\
%Let us define the following function:
%\begin{equation}\label{csi_function}
%\xi_{\text{present}}(t) = \left(1 + \frac{t}{G(t)}\right).
%\end{equation}
%(It should also be possible to define a new function that combines $\psi_{\text{future}}$ and $\xi_{\text{present}}$.)

\RestyleAlgo{boxruled}
\begin{algorithm}[]
	\SetAlgoLined
	\caption{Soft UCB mortal algorithm}\nllabel{Algorithm::UCB-LG}
	\SetKwInOut{Input}{Input}
	\SetKwInOut{Output}{output}
	\SetKwInOut{Loop}{Loop}
	\SetKwInOut{Initialization}{Initialization}
	%	\SetKwData{Left}{left}\SetKwData{This}{this}\SetKwData{Up}{up}
	%	\SetKwFunction{Union}{Union}\SetKwFunction{FindCompress}{FindCompress}
	\Input{number of rounds $n$, set $M_t$ of available arms at turn $t$, rewards range $[a,b]$, with $r = b-a$ }
	\Initialization{play all arms in $M_I$ once, and initialize $\widehat{X}_{j}$ for each $j=1,\cdots,|M_I|$}
	\For{$t=m_I+1$ \KwTo $n$}{ 
		{Play arm with highest $\widehat{X}_{j} + \psi_{\text{future}}(j,t)\sqrt{\frac{2 \log \xi_{\text{present}}(t-s_j)}{T_{j}(t-1)}} $\;}
		{Get reward $G(t)X_j(t)$\;}
		{ \hspace{0.1cm}Update $\widehat{X}_{j}$\;}
	}
\end{algorithm}

\tcbset{colback=blue!2!white}
\begin{tcolorbox}
	\begin{theorem}\label{Theorem::G_mortal}
		%{\bf Theorem}\textit{ 
		Let $\bigcup_{z=1}^{E_j}L_j^z$ be a partition of $L_j$ into epochs with different best available arm, $s_j^z$ and $l_j^z$ be the first and last step of epoch $L_j^z$, and for each epoch let $u_{j,z}$ be defined as
		\begin{equation}
			u_{j,z} = \max_{t\in\{s_j^z,\cdots,l_j^z\}}\left\lceil \frac{8 \psi_{\text{\rm future}}(j,t) \log\xi_{\text{\rm present}}(t-s_j)}{\Delta_{j,z}^2} \label{ujz} \right\rceil,
		\end{equation}
		where
		\begin{equation}
			\Delta_{j,z} = \Delta_{j,i^*_{t}} \;\;\text{for}\;t \in L_j^z. \label{Deltajz}
		\end{equation}
		Then, the bound on the mean regret $\E[R_n]$ at time $n$ is given by
		%\footnotesize
		\begin{eqnarray}
			\E[R_n] &\leq& \sum_{j \in M_I}G(j)\Delta_{j,i^*_{t}} \label{init_UCB_soft_mortal} \\
			&+&  \sum_{j \in M}\; \sum_{z =1}^{E_j}  \left(\max_{t \in E_j} G(t)\right)\Delta_{j,z}     \min\left(l_j^z-s_j^z \;,\; \beta_j^{M}\right) , \label{ub_UCB_soft_mortal}
		\end{eqnarray}	%\normalsize
		where 
		\footnotesize
		\begin{equation}
			\beta_j^{M} =  u_{j,z} + \displaystyle\sum_{\substack{t \in L_j^z \\ t>m_I}}\; (t-s_{i^*_{t}}) (t-s_j-u_{j,z}+1)\left[  \xi_{\text{\rm present}}(t-s_j)^{-\frac{4}{r^2}\psi_{\text{\rm future}}(j,t)} +  \xi_{\text{\rm present}}(t-s_{i^*_{t}})^{-\frac{4}{r^2} \psi_{\text{\rm future}}(i^*_{t},t)}  \right].  \label{beta_UCB_soft_mortal}
		\end{equation}\normalsize
		%	}
	\end{theorem}
\end{tcolorbox}
\tcbset{colback=white}

In Theorem \ref{Theorem::G_mortal}, we split each lifespan of each arm $j$ into $E_j$ epochs that have a different best arm. For each epoch $L^z_j$ $(z = 1, \dots, E_j)$, we denote with $\Delta_{j,z}$ the difference of the mean reward of the best available arm during epoch $z$ and the mean reward of arm $j$.  
The quantity $u_{j,z}$ in \eqref{ujz} intuitively represents the number of times we need to have pulled a suboptimal arm $j$ to be able to lower its probability of being chosen. The first summation in \eqref{init_UCB_soft_mortal} is the regret suffered by the algorithm during the initialization phase, while the quantity in \eqref{ub_UCB_soft_mortal} bounds the regret for the rest of the game. The bound is computed by considering each epoch partition of each arm lifespan. If an epoch is too short (i.e., $ \min(l_j^z-s_j^z \;,\; \beta_j^{M}) = l_j^z-s_j^z$), then the bound is achieved simply by multiplying the largest value of the multiplier function ($\max_{t \in E_j} G(t)$) by the length of the epoch ($l_j^z-s_j^z$) and the mean regret of choosing arm $j$ (which is $\Delta_{j,z}$). If an epoch is not too short (i.e., $ \min(l_j^z-s_j^z \;,\; \beta_j^{M}) = \beta_j^{M}$), then we can use the quantity defined in \eqref{beta_UCB_soft_mortal} instead of $l_j^z-s_j^z$. The proof of Theorem \ref{Theorem::G_mortal} is in Appendix \ref{Appendix::UCB_mortal_regret_bound}.

Algorithm \ref{Algorithm::UCB-LG} differs from the UCB-L algorithm (see Algorithm \ref{Algorithm::UCB-L}) introduced in \citet[]{TracaRuYa2019} in that UCB-L regulates exploration based on the remaining life of the arms (i.e., if arms have a short lifespan or are close to their expiration then UCB-L favors exploration of arms that have longer lifespan) while the Soft UCB mortal algorithm encourages exploration of arms that are available when $G(t)$ is high (even if their lifespan might be short). If the arms never expire then the Soft UCB mortal algorithm reduces to the Soft UCB algorithm by setting $c = 1$ (because the score of all the arms is the same).

\section{Experimental Results: real data environment}\label{Section::Experimental_Results_real_data}

In this section, we show the performance of the proposed algorithms along with the widely used standard $\varepsilon$-greedy and UCB algorithms, using the event log data from the Yahoo! Webscope program. The dataset consists of a stream of recommendation events that display articles to users. Each entry of the dataset contains information on:
\begin{itemize}
	\item the arm pulled (which is the article shown to the human viewing articles on Yahoo!);
	\item the outcome (whether the article was clicked or not);
	\item the pool of arms (articles) available at that time and the associated timestamp.
\end{itemize}
We preprocessed the original text file into a structured data frame (for a small extract of the data frame see Table \ref{dataframe}). We created time bins of the duration of one second. We set $G(t)$ as the number of customers who appear during time bin $t$.

\begin{table}[]
	\centering
	\caption{An extract of the Yahoo! Webscope program dataset used in the simulation.}
	\label{dataframe}
	\begin{tabular}{ | c | c | c | c | c | c | } 
		\hline
		event index & timestamp & displayed article & clicked? & article pool & $G(t)$\\ 
		\hline
		1 & 1317513291 & id-560620 & 0 & id-560620,... & 32\\
		2 &            & id-565648 & 0 & id-565648,... & \\
		3 &            & id-563115 & 0 & id-563115,... & \\
		\hline
		4 & 1317513292 & id-552077 & 0 & id-552077,... & 13\\
		5 &            & id-564335 & 0 & id-564335,... & \\
		\hline
	\end{tabular}
\end{table}

%These data are unusual in that they allow for unbiased offline evaluations of multi-armed bandit algorithms \citep{li2010contextual}. 
A unique property of this dataset is that the articles shown to the visitors were chosen uniformly at random from the pool of available articles, so that it is possible to use an unbiased offline evaluation methodology for our algorithms  \citep[see][]{Li:2011:UOE:1935826.1935878}: at each time, the reward is calculated only when the article that was displayed to the human user matches the article chosen by the algorithm (otherwise, the record is discarded). %(We do not know the outcome of actions not recorded in the dataset.) 
More details on the offline evaluation can be found in Appendix \ref{Appendix::Experiment details}.
% In each game, the algorithms are tested on the same data. The recommender algorithms play for a fixed number of turns. We recorded the accumulated rewards of each algorithm.
% At each time, a reward can be calculated only when the article that was displayed to the human user matches the action of the algorithm. (We do not know the outcome of actions not recorded in the dataset.) This means rewards can only be calculated at a fraction of times that the algorithm is playing. Therefore, while still playing the same number of turns, some algorithms will have more evaluations than others. In particular, an algorithm can be unlucky, in that most of its actions are discarded by chance. However, when the dataset was constructed, articles were shown uniformly at random to the human user, and overall, the difference between the number of evaluations per algorithm is small. More details on the experiment can be found in Supplement D.

\begin{figure}[]
	\caption{The 652 little black horizontal bars represent the lifespans of the articles that appear in the dataset. Since the articles come and go, in order to recreate a non-mortal setting, we selected the longest slice with the most amount of arms always available (arms between the two vertical lines). }\label{Figure::lifespan}
	\centering
	\includegraphics[scale=0.7]{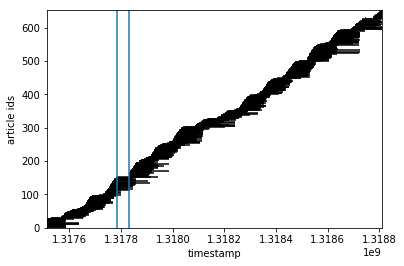}
	\label{lifespan}
\end{figure}

Note that articles in this dataset appear in the candidate pool at some time and often become unavailable shortly afterward. For this reason, the dataset creates a natural setting for mortal bandit algorithms. In Figure \ref{lifespan}, the 652 little black horizontal bars represent the lifespans of the 652 articles that appear in the dataset.
Since the $\varepsilon$-z greedy, Soft $\varepsilon$-greedy, UCB-$z$, and Soft UCB algorithms do not consider mortal bandits, we can run these algorithms only on a slice of the dataset where the candidate pool is invariant, so that we can create a non-mortal setting for the algorithms. 

We also conducted experiments with variations of the UCB algorithm that are well-suited for a mortal setting in order to show the value of regulating greed over time: 
\begin{itemize}
	\item Soft UCB mortal (Section \ref{Section::UCB_soft_mortal});
	\item A ``smarter'' version of UCB-L: Algorithm \ref{Algorithm::UCB-L}. The UCB-L algorithm regulates exploration based on the remaining lifespan of the arms, i.e., if an arm has short lifespan or will expire soon then the algorithm will prefer to play other arms that have longer life in order to gain information that can be used in the future  \citep[see][where UCB-L was one of the algorithms with the best performance]{TracaRuYa2019}.
\end{itemize} 
We also included results of the standard UCB algorithm, since it is used in industry for mortal settings (even though it is not meant for a mortal setting) and to show the disadvantage of using an algorithm not tailored for the mortal setting. 

We ran the algorithms 100 times to derive an empirical distribution of the rewards. For the $\varepsilon$-greedy algorithms and the  variable pool algorithm, different articles are chosen during each run of the algorithms, because during exploration phases they are chosen at random. For those, we use one slice of the dataset (the time period between the two blue lines in Figure \ref{lifespan}), which is the longest slice with the largest number of available arms. Also, since the offline evaluation consumes records very quickly (all records for which the algorithm did not choose the article recommended by the website must be discarded because no label is available), 
%due to the discard of records that do not match articles recommended by the website with articles recommended by the algorithms, 
we duplicated this slice to increase the size of the dataset. %Even when we reuse the the same slice, there are different evaluations because of the randomness of when the algorithm's recommendation agrees with the website's random selection of articles shown to the user.
The UCB algorithms are deterministic in their choices because they always pick the arm with the best upper confidence bound, i.e., running the UCB family many times on the same dataset will always give the same result. Therefore, we used a sliding window to cover different portions of the dataset to evaluate the performance of the UCB algorithms for each game. The randomness of their final rewards arises from the different portions of dataset used.

\begin{figure}[]
	\caption{$\varepsilon$-greedy algorithms playing 100 games with 10000 turns per game. The best algorithm is the Soft $\varepsilon$-greedy algorithm, followed by variable pool and $\varepsilon$-z greedy. The standard $\varepsilon$-greedy algorithm has the lowest average reward.}
	\label{Figure::EGs_immortal}
	\centering
	\includegraphics[scale=0.8]{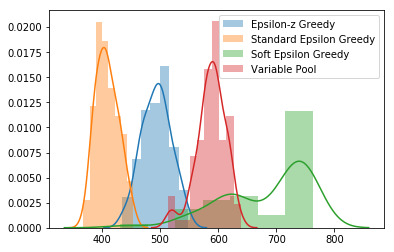}
\end{figure}

\begin{figure}[]
	\caption{UCB algorithms playing a set of 100 slightly different games with 10000 turns per game under immortal circumstances (invariant arms pool). The best algorithm is the UCB-$z$ algorithm, followed by Soft UCB. The standard UCB algorithm has the lowest average reward.}
	\label{Figure::UCB_immortal}
	\centering
	\includegraphics[scale=0.8]{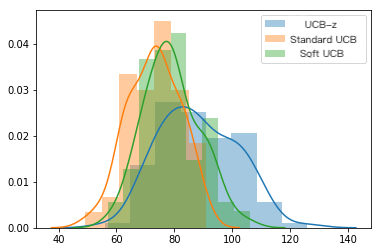}
\end{figure}

\begin{figure}[]
	\caption{UCBs playing a set of 100 slightly different games with 10000 turns per game under mortal circumstances (variable arms pool). The best algorithm is the Soft UCB mortal algorithm, followed by UCB-L. The standard UCB algorithm has the lowest average reward.}
	\label{Figure::UCB_mortal}
	\centering
	\includegraphics[scale=0.8]{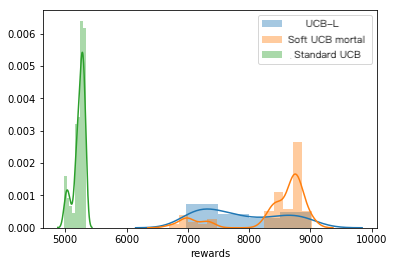}
\end{figure}

Figures \ref{Figure::EGs_immortal}-\ref{Figure::UCB_mortal} show the distribution of the rewards for each algorithm discussed above. We computed mean reward and conducted $t$-test to verify that the average reward of the algorithms that regulate exploration based on the values of the multiplier function $G(t)$ are significantly higher than the average rewards of the standard algorithms (a summary of the results is in Tables \ref{EG performance}, \ref{immortal UCB performance}, and \ref{mortal UCB performance}). The algorithms that regulate greed outperform the standard ones.

\begin{table}
	\centering
	\caption{Performance comparison of the $\varepsilon$-greedy algorithms in immortal circumstances. The mean reward in each turn is computed as the ratio of the total cumulative rewards and the number of turns.}
	\label{EG performance}
	\begin{tabular}{ | c | c | c | } 
		\hline
		Algorithm & Mean reward in each turn & p-value (vs. standard $\varepsilon$-greedy) \\
		\hline
		%        Standard Epsilon Greedy & 406.17 & - \\
		%        Epsilon-z Greedy & 492.81 & 3e-61 \\
		%        Soft Epsilon Greedy & 588.0 &  4e-88\\
		%        Variable Pool & 688.04 & 3e-117 \\
		Standard $\varepsilon$-greedy & $0.0406$ & - \\
		$\varepsilon$-z greedy & $0.0493$ & $<0.0001$ \\
		Soft $\varepsilon$-greedy & $0.0588$ &  $<0.0001$\\
		Variable Pool & $0.0688$ & $<0.0001$ \\
		\hline
	\end{tabular}
\end{table}

\begin{table}
	\centering
	\caption{Performance comparison of the UCB algorithms in immortal circumstances. The mean reward in each turn is computed as the ratio of the total cumulative rewards and the number of turns.}
	\label{immortal UCB performance}
	\begin{tabular}{ | c | c | c | } 
		\hline
		Algorithm & Mean reward in each turn & p-value (vs. standard UCB) \\
		\hline
		%        Standard UCB & 0.03651 & - \\
		%        UCB-z & 0.044085 & 2e-17 \\
		%        Soft UCB-z & 0.03955 & 7e-06 \\
		Standard UCB & $0.0365$ & - \\
		UCB-$z$ & $0.0441$ & $<0.0001$ \\
		Soft UCB & $0.0396$ & $<0.0001$ \\
		\hline
	\end{tabular}
\end{table}

\begin{table}
	\centering
	\caption{Performance comparison of variations of the UCB algorithm in mortal circumstances. The mean reward in each turn is computed as the ratio of the total cumulative rewards and the number of turns.}
	\label{mortal UCB performance}
	\begin{tabular}{ | c | c | c | } 
		\hline
		Algorithm & Mean reward in each turn & p-value (vs. standard UCB) \\
		\hline
		standard UCB & $0.0523$ & - \\
		UCB-L & $0.0787$ & $<0.0001$\\
		UCB soft mortal & $0.0840$ &  $<0.0001$\\
		%        standard UCB & 0.0522756 & - \\
		%        UCB-L & 0.0786927 & 1.4777379601322776e-96\\
		%        UCB soft mortal & 0.0840055 &  1.7956403679189202e-113\\
		\hline
	\end{tabular}
\end{table}

%The classic UCB doesn't consider mortal bandits. The usual practice in the industry is just ignore dying arms. For comparison we also included the standard UCB in mortal circumstances.  Not using the information of articles' lifespan, its performance is far behind UCB soft mortal and UCB-L.

%new pics

\RestyleAlgo{boxruled}
\begin{algorithm}[]
	\SetAlgoLined
	\caption{Smarter version of the UCB-L algorithm}\nllabel{Algorithm::UCB-L}
	\SetKwInOut{Input}{Input}
	\SetKwInOut{Output}{output}
	\SetKwInOut{Loop}{Loop}
	\SetKwInOut{Initialization}{Initialization}
	%	\SetKwData{Left}{left}\SetKwData{This}{this}\SetKwData{Up}{up}
	%	\SetKwFunction{Union}{Union}\SetKwFunction{FindCompress}{FindCompress}
	\Input{number of rounds $n$, set $M_t$ of available arms at turn $t$, $c>0$, rewards range $[a,b]$, $\{G(t)\}_{t=1}^n$}
	\Initialization{play all arms in $M_I$ once, and initialize $\widehat{X}_{j}$ for each $j=1,\cdots,|M_I|$}
	\For{$t=m_I+1$ \KwTo $n$}{ 
		{Play arm with highest $\widehat{X}_{j} + \psi(j,t)\sqrt{\frac{2 \log \xi(t-s_j)}{T_{j}(t-1)}}, \;\;\text{where}\,\, \psi(j,t)= c\log(l_j-t+1)$\;}
		{Get reward $G(t)X_j(t)$\;}
		{ \hspace{0.1cm}Update $\widehat{X}_{j}$\;}
	}
\end{algorithm}

\subsection{Discussion on cases when the algorithms do not apply or reduce to standard cases}
We have introduced several algorithms, each with their own regret bound. The regret bounds should guide us on the choice of algorithm, but our experiments indicate that the variable pool algorithm seems to perform consistently well in our experiments.

It is possible to construct pathological cases where bandit algorithms can perform badly in our new setting. Cases, for instance, where $G(t)$ is very high at the beginning and then decrease rapidly are examples where no bandit strategy would perform well, since the important decisions need to be made very early. In that case, regulating greed may make the problem worse, but this is transparent from the regret bounds and thus known beforehand. Note that any bandit algorithm will also perform badly if rewards are very high at the beginning when little information is available.

Note also that if the number of turns is very high, regulating greed algorithms and standard MAB algorithms will all start to behave similarly because eventually they all figure out the best arms to play; experiments on bandits are only relevant in the non-asymptotic regime.

%For both classical MAB algorithms and our algorithms, there are problem-dependent parameters controlling exploration that cannot easily be set a priori without further knowledge of the problem. Parameters could be tuned using past data if available, or can be tuned by looking at the behavior of the algorithm (for example, if there is a step-type multiplier, and the methods are not exploring when the multiplier is low, then we need to rethink the parameter value that was chosen.)

When $G(t)=1$, all of the algorithms introduced here, except for the variable pool algorithm, reduce to standard MAB algorithms with same regret bound. For the variable pool algorithm,  
%and the Soft UCB mortal algorithms
when $G(t)=1$, the pool starts with all of the arms, and gradually eliminates arms from the pool as $t$ increases.

\subsection{Discussion on the Yahoo! contest}

The motivation of this work comes from a high scoring entry in the Exploration and Exploitation 3 contest, where the goal was to build a better recommender system for Yahoo! Front Page news articles. The competition data is the same data used in our experiments above. These data and its competition setup had several challenging characteristics, including mortal arms (as studied here), broad trends over time in click-through-rate, the inability to access data from the correct distribution in order to cross-validate during the competition, and other complexities. This paper does not aim to handle all of these, but it does describe the key insight that led to increased performance for one of the leading teams, and that insight is precisely the regulation of greed over time \citep{RudinChSu15}. Although there were features available for each timestep, this team was not able to successfully use the features to substantially boost performance, and the exploration/exploitation aspects turned out to be more important. Here are the main insights leading to large performance gains, all involving regulating greed over time:
\begin{itemize}
	\item ``Peak grabber'': Stop exploration when a good arm appears. Specifically, when the article was clicked 9/100 times, keep showing it and stop exploration all together until the arm's click through rate drops below that of another arm. Since this strategy does not handle the massive global trends we observed in the data, it needed to be modified to the dynamic peak grabber strategy described next.
	\item ``Dynamic peak grabber'': Stop exploration when the click through rate of one arm is at least 15\% above that of the global click through rate. The global rate was estimated by exponentially weighted moving averages.
	\item Stop exploring old articles: We can determine approximately how long the arm is likely to stay, and we reduce exploration gradually as the arm gets older.
	\item Do not fully explore new arms: When a new arm appears, do not use 1 as the upper confidence bound for the probability of click, which would force a UCB algorithm to explore it, use .88 instead. This allows the algorithm to continue exploiting the arms that are known to be good rather than exploring new ones.
\end{itemize}
The peak grabber strategies inspired the abstracted setting here, where one can think of a good article appearing during periods of high $G(t)$, where we would want to limit exploration; however, the other strategies are also relevant cases where the exploration/exploitation tradeoff is regulated over time. There were no ``lock-up'' periods in the contest dataset, though as discussed earlier, the $G(t)$ function is also relevant for modeling that setting. The large global trends we observed in the contest data click through rates are very relevant to the $G(t)$ model, since one might want to explore less when the click rate is high in order to get more clicks overall.

\section{Conclusions}
%Time series patterns in customer behavior may be crucial in order to optimize rewards in the contest of the multi-armed bandit problem. We show that regulating greed over time (that is, exploiting when the rewards are high, but exploring when they are low) is intuitive and beneficial. Moreover, theoretical bounds show that the mean regret is logarithmically bounded in the number of rounds played, which is that do not take into account time patterns. We show that with some modifications it is easy to adapt well-known algorithms (namely, the $\varepsilon$-greedy and UCB policies) to this setting while still preserving qualitative properties. 
The dynamic trends we observe in most retail and marketing settings are dramatic. It is possible that understanding these dynamics and how to take advantage of them is central to the success of multi-armed bandit algorithms in a very large class of situations that occur in practice (such as retail). We showed in this work how to adapt algorithms and regret bound analysis to this setting, where we now need to consider not only the average number of times that an arm was pulled in the past, but precisely when the arm was pulled. The key element of our algorithms is that they regulate greed (exploitation) over time, where during high reward periods, less exploration is performed.
The algorithms that can regulate greed outperform significantly standard ones, and the results are supported by simulations and a study on the Yahoo! event log dataset that was created specifically for evaluating the performance of multi-armed bandit algorithms.
We chose to show how to adapt very well known algorithms to this setting (in theory and in practice), but there are many possible extensions to this work to many other important multi-armed bandit algorithms. 
%In particular, if $G(t)$ is not known or approximated in advance, it may be easy to estimate from data in real time, as in the dynamic peak grabber strategy. The analysis of the algorithms in this paper could be extended to other important multi-armed bandit algorithms besides $\varepsilon$-greedy and UCB. Further, future work will consider the connection of mortal bandits (with appearing/disappearing arms) with the $G(t)$ setting, since for mortal bandits, each bandit's $G(t)$ function can change at a different rate.

\section*{Acknowledgments}
We thank team members Ed Su and Virot Ta Chiraphadhanakul for their work on the ICML Exploration and Exploitation 3 contest that inspired the work here. 
%Thank you also to Philippe Rigollet for helpful comments and encouragement. 
This project was partially funded by the Ford-MIT alliance. 

% \nocite{*}
% \bibliographystyle{plainnat}

\newpage

\appendix

\section{Regret-bound for $\varepsilon$-z greedy algorithm with hard threshold}\label{proof_epsilon1}

Proposition \ref{Proposition::inclusion_epsilon_z} is also used in \citet[]{auer2002finite} for the proof of the regret bound for the $\varepsilon$-greedy algorithm. %Propositions \ref{Proposition::inclusion_UCB} and \ref{Proposition::inclusion_UCB_xi} are useful in our setting because when using UCB policies we do not only care about average number of time the algorithms pull an arm by the end of the game (as in \citet[]{auer2002finite}), but we care about the probability that an arm is pulled in a particular turn. 
\tcbset{colback=blue!2!white}
\begin{tcolorbox}
	\begin{proposition}\label{Proposition::inclusion_epsilon_z}
		Let us define the following events:\small
		\begin{eqnarray*}
			A &=& \left\{ \widehat{X}_{j,T_j(t-1)} > \widehat{X}_{*,T_*(t-1)} \right\},\\
			B &=& \left\{ \widehat{X}_{*,T_*(t-1)}  <  \mu_* - \frac{\Delta_j}{2} \right\},\\
			C &=& \left\{ \widehat{X}_{j,T_j(t-1)} > \mu_j + \frac{\Delta_j}{2}\right\}.
		\end{eqnarray*}
		\normalsize Then,
		\begin{equation}\label{____inclusion1}
			A \subset \left(  B \cup  C  \right).
		\end{equation}
	\end{proposition}
\end{tcolorbox}
%\begin{itemize}
%	\item $\widehat{X}_{*,T_*(t-1)} \leq \mu_* - \frac{\Delta_j}{2}$;
%	\item $\widehat{X}_{j,T_j(t-1)} \geq \mu_j + \frac{\Delta_j}{2}$.
%\end{itemize}
Intuitively, inclusion \eqref{____inclusion1} means that we play arm $j$ when we underestimate the mean reward of the best arm, or when we overestimate that of arm $j$.
Assume for the sake of contradiction that there exists an element $\omega \in A$ that does not belong to $B \cup C$. Then, we have that $\omega \in \left(B \cup C\right)^{\mathcal{C}}$\footnotesize
\begin{eqnarray}
	\Rightarrow \;\;\omega & \in & \left(  \left\{ \widehat{X}_{*,T_*(t-1)}  <  \mu_* - \frac{\Delta_j}{2}  \right\} \cup 
	\left\{ \widehat{X}_{j,T_j(t-1)} > \mu_j + \frac{\Delta_j}{2} \right\}  \right)^{\mathcal{C}} \label{____toContradict1} \nonumber\\
	\Rightarrow \;\;\omega &\in&   \left\{ \widehat{X}_{*,T_*(t-1)}  \geq  \mu_* - \frac{\Delta_j}{2}  \right\} \cap 
	\left\{ \widehat{X}_{j,T_j(t-1)} \leq \mu_j + \frac{\Delta_j}{2}  \right\}. \label{____NONinclusion1}
\end{eqnarray}\normalsize
By definition we have $\mu_* - \frac{\Delta_j}{2} = \mu_* - \frac{\mu_*-\mu_j}{2} = \frac{\mu_*+\mu_j}{2} =  \mu_j + \frac{\Delta_j}{2}$. From the inequalities given in \eqref{____NONinclusion1} it follows that  \footnotesize
\begin{eqnarray*}
	\widehat{X}_{*,T_*(t-1)} \geq \mu_* - \frac{\Delta_j}{2} = \mu_j + \frac{\Delta_j}{2}  
	\geq \widehat{X}_{j,T_j(t-1)},
\end{eqnarray*}\normalsize
but this contradicts our assumption that $ \omega \in A = \left\{ \widehat{X}_{j,T_j(t-1)}  > \widehat{X}_{*,T_*(t-1)}  \right\}$.\\ 
Therefore, all elements of $A$ belong to $B \cup C$.%$\hfill\qed$

\tcbset{colback=blue!2!white}
\begin{tcolorbox}
	{\bf Theorem \ref{Theorem::epsilon_z}}\textit{ The bound on the mean regret $\E[R_n]$ at time $n$ is given by
		\begin{eqnarray*}
			\E[R_n]\displaystyle & \leq & \displaystyle\sum_{j=1}^m G(j) \Delta_j  \label{initialization_epsilon_threshold_A}\\
			& + & \displaystyle\sum_{t=m+1}^n G(t)\ONE_{\{G(t)< z\}}\sum_{j : \mu_j<\mu_*} \Delta_j \left( \varepsilon_t\frac{1}{m}+(1-\varepsilon_t) \beta_j(\tilde{t}) \right) \label{low_epsilon_threshold_A}\\
			& + &\displaystyle\sum_{t=m+1}^n G(t)\ONE_{\{G(t)\geq z\}} \sum_{j : \mu_j<\mu_*} \Delta_j \beta_j(\tilde{t}) , \label{high_epsilon_threshold_A}
			%	& + &\displaystyle\sum_{t\in B} \sum_{j : \mu_j<\mu_*} \Delta_j \beta_j(\tilde{t}) \left[ \sum_{s=t}^{\tau_t} G(s)\right], \label{high_epsilon_threshold}
		\end{eqnarray*}
		where
		\begin{equation*}\label{beta_threshold_A}
			\beta_j(\tilde{t})=k\left( \frac{\tilde{t}}{m k e }\right) ^{-\frac{k}{10}}\log \left( \frac{\tilde{t}}{m k e } \right)   + \frac{4}{\Delta_j^2} \left( \frac{\tilde{t}}{m k e } \right)^{-\frac{k \Delta_j^2 }{4}}.
		\end{equation*}
	}
\end{tcolorbox}

\tcbset{colback=white}
\begin{tcolorbox}
	{\bf First step:} Decomposition of $\E[R_n]$.
\end{tcolorbox}
The total mean regret of a game $\mathcal{I} = \{I_t\}_{t=1}^n$ at round $n$ is given by
\begin{equation}
	R_n=\sum_{t=1}^n \sum_{j=1}^m \Delta_j G(t) \ONE_{\{I_t=j\}},
\end{equation}
where $G(t)$ is the greed function evaluated at time $t$, $\ONE_{\{I_t=j\}}$ is an indicator function equal to $1$ if arm $j$ is played at time $t$ (otherwise its value is $0$) and $\Delta_j=\mu^*-\mu_j$ is the difference between the mean of the best arm reward distribution and the mean of the $j$'s arm reward distribution. By considering the threshold $z$ which determines which rule is applied to decide what arm to play, we can rewrite the regret as 
\begin{eqnarray*}
	R_n&=\displaystyle \sum_{t=1}^n \sum_{j=1}^m \Delta_j G(t) \ONE_{\{G(t)<z\}} \ONE_{\{I_t=j\}}+ \\
	&\displaystyle+\sum_{t=1}^n \sum_{j=1}^m \Delta_j G(t) \ONE_{\{G(t)\geq z\}} \ONE_{\{I_t=j\}}.
\end{eqnarray*}
By taking the expectation with respect to the policy, we have that 
\begin{eqnarray*}
	\E[R_n]&=\displaystyle \sum_{t=1}^n \sum_{j=1}^m \Delta_j G(t) \ONE_{\{G(t)<z\}} \P(\{I_t=j\})+ \nonumber \\
	&\displaystyle+\sum_{t=1}^n \sum_{j=1}^m \Delta_j G(t) \ONE_{\{G(t)\geq z\}} \P(\{I_t=j\}),\nonumber%\label{ExpectedRegretA}
\end{eqnarray*}
which can be rewritten as
\begin{eqnarray}
	\E[R_n]&=&\displaystyle \sum_{t=1}^n \sum_{j=1}^m \Delta_j G(t) \ONE_{\{G(t)<z\}} \left[ \varepsilon_t\frac{1}{m}+(1-\varepsilon_t) \P\left( \widehat{X}_{j,T_j(t-1)} > \widehat{X}_{i,T_i(t-1)} \;\; \forall i \right) \right] \nonumber \\
	&+&\displaystyle\sum_{t=1}^n \sum_{j=1}^m \Delta_j G(t) \ONE_{\{G(t)\geq z\}} \P\left( \widehat{X}_{j,T_j(t-1)} > \widehat{X}_{i,T_i(t-1)} \;\; \forall i \right). \label{ExpectedRegretA}
\end{eqnarray}
For the rounds of the algorithm where $G(t)<z$, we are in the standard setting, so for those times, we follow the standard proof of \cite{auer2002finite}. For the times that $G(t)$ is over the threshold, we need to create a separate bound.
\tcbset{colback=white}
\begin{tcolorbox}
	{\bf Second step:} Upper bound for $\P\left( \widehat{X}_{j,T_j(t-1)} > \widehat{X}_{i,T_i(t-1)} \;\; \forall i \right)$.
\end{tcolorbox}
Let us now bound the probability of playing the sub-optimal arm $j$ at time $t$ when the greed function is above the threshold $z$. From Proposition \ref{Proposition::inclusion_epsilon_z} we have that\footnotesize
\begin{eqnarray}
	\P\left( \widehat{X}_{j,T_j(t-1)} > \widehat{X}_{i,T_i(t-1)} \;\; \forall i \right)
	&\leq& \P\left(\widehat{X}_{j,T_j(t-1)} > \widehat{X}_{*,T_*(t-1)}\right) \nonumber\\
	&\leq& \P\left(\widehat{X}_{j,T_j(t-1)} > \mu_j + \frac{\Delta_j}{2} \right) + \P\left(\widehat{X}_{*,T_*(t-1)} <  \mu_* - \frac{\Delta_j}{2} \right) \label{___FirstTerm1}.
\end{eqnarray}\normalsize
Let us consider the first term of \eqref{___FirstTerm1} (the computations for the second term are similar),\small
\begin{eqnarray}
	\P\left(\widehat{X}_{j,T_j(t-1)} > \mu_j + \frac{\Delta_j}{2} \right) 
	&= & \sum_{s=1}^{t-1} \P\left(T_j(t-1)=s , \widehat{X}_{j,s} > \mu_j + \frac{\Delta_j}{2}\right) \nonumber \\
	&= & \sum_{s=1}^{t-1} \P\left(T_j(t-1)=s \,\bigg|\, \widehat{X}_{j,s} > \mu_j + \frac{\Delta_j}{2}\right)\P\left(\widehat{X}_{j,s} > \mu_j + \frac{\Delta_j}{2}\right) \nonumber\\
	&\leq & \sum_{s=1}^{t-1} \P\left(T_j(t-1)=s \,\bigg|\, \widehat{X}_{j,s} > \mu_j + \frac{\Delta_j}{2}\right)e^{-\frac{\Delta_j^2}{2}s}, \label{ToCont1}
\end{eqnarray}\normalsize
where in the last inequality we used Hoeffding's\footnote{{\bf Hoeffding's bound:} Let $X_1, \cdots, X_n$ be r.v. bounded in $[a_i,b_i]$ $\forall i$. Let $\widehat{X} = \frac{1}{n}\sum_{i=1}^n X_i$ and $\mu = \E[\widehat{X}]$. \\Then, $\P\left(\widehat{X} - \mu \geq \varepsilon\right) \leq \exp\left\{ -\frac{2n^2\varepsilon^2}{\sum_{i=1}^n (b_i-a_i)^2 }  \right\}$.} bound. 
\begin{tcolorbox}
	{\bf Third step:} Upper bound for $\sum_{s=1}^{t-1}\P\left(T_j(t-1)=s \,\bigg|\, \widehat{X}_{j,s} > \mu_j + \frac{\Delta_j}{2}\right)e^{-\frac{\Delta_j^2}{2}s}$.
\end{tcolorbox}
Let us define $T_j^R(t-1)$ as the number of times arm $j$ is played at random when exploring (note that $T_j^R(t-1) \leq T_j(t-1)$ and that $T_j^R(t-1) = \sum_{s=1}^{t-1} B_s$ where $B_s$ is a Bernoulli r.v. with parameter $\varepsilon_s/m$), and let us define $$ \lambda_t=\frac{1}{2m}\sum_{s=1}^{\tilde{t}} \varepsilon_s,   $$ where $\tilde{t}$ is the number of rounds played under the threshold $z$ up to time $t$.  Then,
\begin{eqnarray}
	\eqref{ToCont1} &\leq& \sum_{s=1}^{ \lfloor \lambda_t \rfloor } \P\left(T_j(t-1)=s \,\bigg|\, \widehat{X}_{j,s} > \mu_j + \frac{\Delta_j}{2}\right) + \sum_{s=\lfloor \lambda_t \rfloor + 1}^{t-1}e^{-\frac{\Delta_j^2}{2}s}  \nonumber\\
	&\leq&  \sum_{s=1}^{ \lfloor \lambda_t \rfloor } \P\left(T_j(t-1)=s \,\bigg|\, \widehat{X}_{j,s} > \mu_j + \frac{\Delta_j}{2}\right) + \frac{2}{\Delta_j^2}e^{-\frac{\Delta_j^2}{2} \lfloor \lambda_t \rfloor} \nonumber\\
	&\leq & \sum_{s=1}^{ \lfloor \lambda_t \rfloor } \P\left(T^R_j(t-1) \leq s \,\bigg|\, \widehat{X}_{j,s} > \mu_j + \frac{\Delta_j}{2}\right) + \frac{2}{\Delta_j^2}e^{-\frac{\Delta_j^2}{2} \lfloor \lambda_t \rfloor}\nonumber\\
	&\leq & \lfloor \lambda_t \rfloor \P\left(T^R_j(t-1) \leq \lfloor \lambda_t \rfloor\right) + \frac{2}{\Delta_j^2}e^{-\frac{\Delta_j^2}{2} \lfloor \lambda_t \rfloor}\label{Boundx0}
\end{eqnarray}
%%
%i) for s=1 to \lambda_t I upper bound exp(-Delta^2/2s) by 1, and I just rewrite the summation without the exponential factor.
%
%ii) for s=\lambda_t+1 to (t-1) I upper bound the probability in the sum by 1 so that I have the sum for s=\lambda_t+1 to (t-1) of exp(-Delta^2/2s). Then I use the fact I explain right after (bound on the exponential sum from x+1 to k) to upperbound  the sum for s=\lambda_t+1 to (t-1) of exp(-Delta^2/2s)
%
%%
where for the first $\lfloor \lambda_t \rfloor$ terms of the sum we used $1$ as upper bound on $e^{-\frac{\Delta_j^2}{2}s}$, and for the remaining terms we used the fact that $ \sum_{s=x+1}^{\infty} e^{-ks} \leq \frac{1}{k}e^{-kx}$, where in our case $k=\frac{\Delta_j^2}{2}$.
\begin{tcolorbox}
	{\bf Fourth step:} Upper bound for $\P\left(T^R_j(t-1) \leq \lfloor \lambda_t \rfloor\right)$.
\end{tcolorbox}
We have that 
$$ \E[T_j^R(t-1)] = \frac{1}{m} \sum_{s=1}^{\tilde{t}} \varepsilon_s, \;\;\;\; Var(T_j^R(t-1))=  \sum_{s=1}^{\tilde{t}} \frac{\varepsilon_s}{m}\left( 1- \frac{\varepsilon_s}{m} \right) \leq \frac{1}{m} \sum_{s=1}^{\tilde{t}} \varepsilon_s= \E[T_j^R(t-1)],$$ 
and, using the Bernstein inequality $\P(S_n \leq \E[S_n] - a) \leq \exp\{-\frac{a^2/2}{\sigma^2  +  a/2}\}$ with $S_n = T_j^R(t-1)$ and $a=\frac{1}{2}\E[T_j^R(t-1)]$,
\begin{eqnarray}
	\P(T_j^R(t-1) \leq \lfloor \lambda_t \rfloor) &=& \P\left( T_j^R(t-1) \leq  \E[T_j^R(t-1)] - \frac{1}{2}\E[T_j^R(t-1)] \right) \nonumber\\
	&\leq & \exp \left\{ -\frac{  \frac{1}{8}(\E[T_j^R(t-1)])^2   }{\E[T_j^R(t-1)] + \frac{1}{4}\E[T_j^R(t-1)]   } \right\} \nonumber\\
	&= & \exp  \left\{ -\frac{4}{5}\frac{1}{8}\E[T_j^R(t-1)] \right\}  = \exp \left\{ -\frac{1}{5} \lfloor \lambda_t \rfloor  \right\}.\label{BoundPx0}
\end{eqnarray}
\begin{tcolorbox}
	{\bf Fifth step:} Lower bound for $\lfloor \lambda_t \rfloor$.
\end{tcolorbox}
To get an upper bound for \eqref{BoundPx0}, we need a lower bound on $\lfloor \lambda_t \rfloor$. Let us define $n'= \left\lfloor km \right\rfloor$ and consider the case when $\tilde{t}\geq n$. The case when $\tilde{t} < n$ leads to an exponential decay of the bound (intuitively, at the beginning of the game the values of $\varepsilon_s$ that compose $\lambda_t$ are still high enough, see Remark \ref{Remark1}). Then,
\begin{eqnarray}
	\lambda_t&=&\frac{1}{2m}\sum_{s=1}^{\tilde{t}} \varepsilon_s  \nonumber \\
	&	= & \frac{1}{2m}\sum_{s=1}^{\tilde{t}} \min\left\{ 1  ,\frac{k m}{s}\right\} \nonumber\\
	&	=& \frac{1}{2m}\sum_{s=1}^{n'} 1 + \frac{1}{2m}\sum_{s=n'+1}^{\tilde{t}} \frac{k m}{s} \nonumber\\
	&	= & \frac{n'}{2m} + \frac{1}{2m}\left( \sum_{s=1}^{\tilde{t}} \frac{k m}{s} - \sum_{s=1}^{n'} \frac{k m}{s} \right)  \nonumber\\
	&	\geq & \frac{n'}{2m} + \frac{k}{2} \left( \log(\tilde{t}+1) - (\log(n') + \log(e) \right) \nonumber\\ 
	&	\geq & \frac{k}{2} \log\left(\frac{n'}{m}\frac{1}{k}\right) +  \frac{k}{2}  \log\left( \frac{\tilde{t}}{n' e}  \right)\nonumber\\
	&	=& \frac{k}{2} \log \left( \frac{\tilde{t}}{mke } \right)\label{Lowerx0}.
\end{eqnarray}
\tcbset{colback=white}
\begin{tcolorbox}
	\begin{remark}\label{Remark1}
		Note that if $\tilde{t}$ (or $t$ in the usual $\varepsilon$-greedy algorithm) was less than $n'$, then we would have $\lambda_t=\tilde{t}/2m$, yielding an exponential decay of the bound on the probability of $j$ being the best arm. To see this, $\tilde{t}<n'$ would imply that, using \eqref{Boundx0} and \eqref{BoundPx0}, $$\eqref{Boundx0} \leq \frac{\tilde{t}}{2m} \exp\left\{ -\frac{1}{5}\frac{\tilde{t}}{2m} \right\} + \frac{2}{\Delta_j^2}\exp\left\{-\frac{\Delta_j^2}{2}\frac{\tilde{t}}{2m}\right\} .$$
		Intuitively, if $\tilde{t}<n'$ then $\lambda_t$ is already big enough to guarantee an exponential decay. The interesting case is instead when $\tilde{t}\geq n'$, for which we need to provide a lower bound for $\lambda_t$ to prove that it will still be large enough so that the bound on the probability of choosing a suboptimal arm $j$ is of order $o(1/t)$.
	\end{remark}
\end{tcolorbox}
\begin{tcolorbox}
	{\bf Sixth step:} Bringing together all bounding quantities.
\end{tcolorbox}
Continuing the proof, we obtain a bound on the first term in \eqref{___FirstTerm1} as follows. Using \eqref{Lowerx0} combined with \eqref{BoundPx0} in \eqref{Boundx0}, we get that 
\begin{equation}
	\text{first addend of }\eqref{___FirstTerm1} \leq \frac{k}{2}\left( \frac{\tilde{t}}{m k e }\right) ^{-\frac{k}{10}}\log \left( \frac{\tilde{t}}{m k e } \right)   + \frac{2}{\Delta_j^2} \left( \frac{\tilde{t}}{m k e } \right)^{-\frac{k \Delta_j^2 }{4}}.
\end{equation}
Since the computations for the second term in \eqref{___FirstTerm1} are similar, a bound on $\P( \widehat{X}_{j,T_j(t-1)} > \widehat{X}_{i,T_i(t-1)} \;\; \forall i )$ is given by 
\begin{equation}
	\beta_j(\tilde{t})=k\left( \frac{\tilde{t}}{m k e }\right) ^{-\frac{k}{10}}\log \left( \frac{\tilde{t}}{m k e } \right)   + \frac{4}{\Delta_j^2} \left( \frac{\tilde{t}}{m k e } \right)^{-\frac{k \Delta_j^2 }{4}}.
\end{equation}
We can use this to easily bound $\P( \widehat{X}_{j,T_j(t-1)} \geq \widehat{X}_{i,T_i(t-1)} \;\; \forall i )$ in \eqref{ExpectedRegretA} which yields the following bound on the mean regret at time $n$ (recall that the first $m$ turns are used in the initialization phase, each yielding a regret of $G(j) \Delta_j$):
\begin{eqnarray*}
	\E[R_n]\displaystyle & \leq & \displaystyle\sum_{j=1}^m G(j) \Delta_j  \\
	& + & \displaystyle\sum_{t=m+1}^n G(t)\ONE_{\{G(t)< z\}}\sum_{j : \mu_j<\mu_*} \Delta_j \left( \varepsilon_t\frac{1}{m}+(1-\varepsilon_t) \beta_j(\tilde{t}) \right) \\
	& + &\displaystyle\sum_{t=m+1}^n G(t)\ONE_{\{G(t)\geq z\}} \sum_{j : \mu_j<\mu_*} \Delta_j \beta_j(\tilde{t}).%\hfill\qed
	%	& + &\displaystyle\sum_{t\in B} \sum_{j : \mu_j<\mu_*} \Delta_j \beta_j(\tilde{t}) \left[ \sum_{s=t}^{\tau_t} G(s)\right], \label{high_epsilon_threshold}
\end{eqnarray*}

%where $B=\{t: G(t-1)<z, G(t)\geq z\}$ is the set of rounds where the high-reward zone is entered, and $\tau_t$ is the last round of high-reward zone entered at time $t$. 
%This, combined with the bound $\beta_j(\tilde{t})$ above, proves the theorem. 

\newpage

\section{Regret bound for Soft $\varepsilon$-greedy algorithm}\label{proof_epsilon2}
\tcbset{colback=blue!2!white}
\begin{tcolorbox}
	{\bf Theorem \ref{Theorem::regret_soft_epsilon_greedy}}\textit{
		The bound on the mean regret $\E[R_n]$ at time $n$ is given by
		\begin{eqnarray*}
			\E[R_n]\displaystyle & \leq & \displaystyle\sum_{j=1}^m G(j) \Delta_j \label{initialization_epsilon_soft}  \\
			& + & \displaystyle\sum_{t=m+1}^n G(t)\sum_{j : \mu_j<\mu_*} \Delta_j \left( \varepsilon_t\frac{1}{m}+(1-\varepsilon_t) \beta_j^S(t) \right) \label{bound_epsilon_soft}
		\end{eqnarray*}
		where
		\begin{equation*}\label{beta_soft}
			\beta^S_j(t) =  k\left( \frac{\gamma t}{mke } \right)^{-\frac{k}{10} } \log \left( \frac{\gamma t}{mke } \right)   + \frac{4}{\Delta_j^2} \left( \frac{\gamma t}{mke } \right)^{-\frac{k\Delta_j^2 }{4} }.
		\end{equation*}
	}
\end{tcolorbox}
\tcbset{colback=white}
\begin{tcolorbox}
	{\bf First step:} Derivation of $\E[R_n]$.
\end{tcolorbox}
The total mean regret of a game $\mathcal{I} = \{I_t\}_{t=1}^n$ at round $n$ is given by
\begin{equation}
	R_n=\sum_{t=1}^n \sum_{j=1}^m \Delta_j G(t) \ONE_{\{I_t=j\}},
\end{equation}
where $G(t)$ is the greed function evaluated at time $t$, $\ONE_{\{I_t=j\}}$ is an indicator function equal to $1$ if arm $j$ is played at time $t$, and $\Delta_j=\mu^*-\mu_j$ is the difference between the mean of the best arm reward distribution and the mean of the $j$'s arm reward distribution. 
By taking the expectation over the policy, we have that 
\begin{eqnarray*}
	\E[R_n]&=\displaystyle \sum_{t=1}^n \sum_{j=1}^m \Delta_j G(t) \P(\{I_t=j\})
\end{eqnarray*}
\tcbset{colback=white}
\begin{tcolorbox}
	{\bf Second step:} Upper bound for $\P(\{I_t=j\})$.
\end{tcolorbox}
At each round $t$, arm $j$ is played with probability
$$ \frac{\varepsilon_t}{m} + (1-\varepsilon_t)\P\left(\widehat{X}_j > \widehat{X}_i \; \forall i \right), $$
where $\varepsilon_t=\min\left\{\psi(t),\frac{km}{t}\right\}$ and
\begin{equation*}\label{psi}
	\psi(t)=\frac{\log\left(1+\frac{1}{G(t)}\right)}{\log\left(1+\frac{1}{\min_{s\in\{m+1,\cdots,n\}} G(s)}\right)}.
\end{equation*}
Recall that $\gamma=\min_{1\leq t\leq n} \psi(t)$. \\
%$$ \varepsilon_t = \min\left\{  \frac{\log\left(1+\frac{1}{G(t)}\right)}{\log\left(1+\frac{1}{\min G(t)}\right)}  ,\frac{c m }{d^2 \tilde{t}}\right\}, $$
Let us bound the probability $\P(\{I_t=j\})$ of playing the sub-optimal arm $j$ at time $t$. We have that\footnotesize
\begin{eqnarray}
	\P\left( \widehat{X}_{j,T_j(t-1)} > \widehat{X}_{i,T_i(t-1) \;\; \forall i} \right)&	\leq &  \P\left(\widehat{X}_{j,T_j(t-1)} > \widehat{X}_{*,T_*(t-1)}\right) \nonumber\\
	&	\leq & \P\left(\widehat{X}_{j,T_j(t-1)} > \mu_j + \frac{\Delta_j}{2} \right) + \P\left(\widehat{X}_{*,T_*(t-1)} <  \mu_* - \frac{\Delta_j}{2} \right). \label{cons2}
\end{eqnarray}\normalsize
\tcbset{colback=white}
\begin{tcolorbox}
	{\bf Third step:} Upper bound for \eqref{cons2}.
\end{tcolorbox}
For the bound on the two addends in \eqref{cons2}, we have identical steps to the proof for Theorem 1, and thus 
\begin{eqnarray}
	\P\left(\widehat{X}_{j,T_j(t-1)} > \mu_j + \frac{\Delta_j}{2} \right) &\leq& \lfloor \lambda_t \rfloor \P\left(T_j^R(t-1) \leq \lfloor \lambda_t \rfloor\right) + \frac{2}{\Delta_j^2}e^{-\frac{\Delta_j^2}{2} \lfloor \lambda_t \rfloor} \label{first2}\\
	\P\left(\widehat{X}_{*,T_*(t-1)} <  \mu_* - \frac{\Delta_j}{2} \right)  &\leq& \lfloor \lambda_t \rfloor \P\left(T_j^R(t-1) \leq \lfloor \lambda_t \rfloor\right) + \frac{2}{\Delta_j^2}e^{-\frac{\Delta_j^2}{2} \lfloor \lambda_t \rfloor}\label{second2}
\end{eqnarray}
and, similarly to the proof of the $\varepsilon$-greedy algorithm with hard threshold (Appendix \ref{proof_epsilon1}), we have 
\begin{equation}
	\P\left(T_j^R(t-1) \leq \lfloor \lambda_t \rfloor\right) \leq \exp \left\{ -\frac{1}{5} \lfloor \lambda_t \rfloor  \right\}. \label{PBoundx02}
\end{equation}
\tcbset{colback=white}
\begin{tcolorbox}
	{\bf Fourth step:} Lower bound for $\lfloor \lambda_t \rfloor$.
\end{tcolorbox}
Now we need a lower bound on $\lfloor \lambda_t \rfloor$.
Let %$ t >  km $ and 
$t > w $ (see Remark \ref{Remark::soft-epsilon} for the case $t\leq w$) where $w=\min\{s : \frac{c m}{d^2 s} < \gamma\}$. Then,
\begin{eqnarray}
	\lambda_t &=&\frac{1}{2m}\sum_{s=1}^{t} \varepsilon_s \nonumber\\
	& = & \frac{1}{2m}\sum_{s=1}^{t} \min\left\{  \psi(s) ,\frac{k m}{s}\right\} \nonumber\\
	%&  =& \frac{1}{2m}\sum_{s=1}^{w} \psi(s) + \frac{1}{2m}\sum_{s=w+1}^{t} \frac{k m}{s} \\
	& \geq & \frac{1}{2m}\sum_{s=1}^{w} \gamma + \frac{1}{2m}\left( \sum_{s=1}^{t} \frac{k m}{s} - \sum_{s=1}^{w} \frac{k m}{s} \right) \nonumber\\
	& \geq & \frac{w \gamma}{2m} + \frac{k}{2} \left( \log(t+1) - (\log(w) + \log(e) \right) \nonumber\\ 
	& \geq & \frac{k}{2} \log\left(\frac{w \gamma}{m}\frac{1}{k}\right) +   \log\left( \frac{t}{w e}  \right)\nonumber\\
	& =& \frac{k}{2} \log \left( \frac{\gamma t}{mke } \right) .\label{Boundx02}
\end{eqnarray}
\tcbset{colback=white}
\begin{tcolorbox}
	\begin{remark}\label{Remark::soft-epsilon}
		Simalarly to what noted in Remark \ref{Remark1}, if $t\leq w$ then we would have $\lambda_t=\gamma t/2m$, yielding an exponential decay of the bound on the probability of $j$ being the best arm. In fact, for $t\leq w$, using \eqref{PBoundx02} combined with \eqref{first2} and \eqref{second2}, we have that 
		\begin{equation*}
			\eqref{cons2} \leq \frac{\gamma t}{2m}\exp\left\{ -\frac{1}{5}\frac{\gamma t}{2m} \right\} + \frac{2}{\Delta_j^2}\exp\left\{-\frac{\Delta_j^2}{2}\frac{\gamma t}{2m}\right\} .
		\end{equation*}
		Intuitively, if $t\leq w$ then $\lambda_t$ is already big enough to guarantee an exponential decay. The interesting case is instead when $t>w$, for which we need to provide a lower bound for $\lambda_t$ to prove that it will still be large enough so that the bound on the probability of choosing a suboptimal arm $j$ is of order $o(1/t)$.
	\end{remark}
\end{tcolorbox}
\begin{tcolorbox}
	{\bf Fifth step:} Bringing together all bounding quantities.
\end{tcolorbox}
Using the upper bound for $\lambda_t$ given in \eqref{PBoundx02} and the lower bound in \eqref{Boundx02}, from  \eqref{cons2} the bound on $\P\left( \widehat{X}_{j,T_j(t-1)} > \widehat{X}_{i,T_i(t-1)} \;\; \forall i \right)$ is given by 
\[
\beta_j^S(t) =  k \log \left( \frac{\gamma t}{mke } \right)  \left( \frac{\gamma t}{mke } \right)^{-\frac{k}{10} } + \frac{4}{\Delta_j^2} \left( \frac{\gamma t}{mke } \right)^{-\frac{k\Delta_j^2 }{4} }.
\]
Since the mean regret is given by
\begin{equation}
	\E[R_n]=\sum_{t=1}^n \sum_{j=1}^m \Delta_j G(t) \P\left(\{I_t=j\}\right),
\end{equation} 
the bound on the mean regret at time $n$ is given by (recall that the first $m$ turns are used in the initialization phase, each yielding a regret of $G(j) \Delta_j$):
\begin{eqnarray*}
	\E[R_n]\displaystyle &\leq& \displaystyle\sum_{j=1}^m G(j) \Delta_j  \\
	&+& \displaystyle\sum_{t=m+1}^n G(t)\sum_{j=1}^m \Delta_j \left( \varepsilon_t\frac{1}{m}+(1-\varepsilon_t) \beta_j^S(t)\right).
\end{eqnarray*}

\newpage

\section{Regret bound for the UCB algorithm with hard threshold}\label{Appendix::UCBz}
\tcbset{colback=blue!2!white}
\begin{tcolorbox}
	{\bf Theorem \ref{Theorem::regulated_UCB}}\textit{ 
		The bound on the mean regret $\E[R_n]$ at time $n$ is given by
		\begin{eqnarray*}
			\E[R_n]\displaystyle & \leq & \displaystyle\sum_{j=1}^m G(j) \Delta_j   \\
			& + & z \left[ 8 \sum_{j : \mu_j<\mu_*} \left(\frac{\log{\tilde{n}}}{\Delta_j}\right) + \left(1 + \frac{\pi^2}{3}\right)\left(\sum_{j=1}^m \Delta_j\right) \right]     \\
			& + & \displaystyle \sum_{j=1}^m  \Delta_j \sum_{k=1}^{|B|}  \sum_{t\in Y_k}G(t) 2\beta_j^U(t).
		\end{eqnarray*}
		where 
		\[\beta_j^U(t) = \frac{2}{\Delta_j^2}e^{-\frac{\Delta_j^2}{2} (x_t-1)}\]
		and $x_t$ is the minimum amount of pulls for each arm at time $t$ (see Lemma \ref{Lemma::minimum_pulls_UCB}.)
	}
\end{tcolorbox}

\tcbset{colback=white}
\begin{tcolorbox}
	{\bf First step:} Decomposition of $\E[R_n]$.
\end{tcolorbox}
The total mean regret of a game $\mathcal{I} = \{I_t\}_{t=1}^n$ at round $n$ is given by
\begin{equation}
	R_n=\sum_{t=1}^n \sum_{j=1}^m \Delta_j G(t) \ONE_{\{I_t=j\}},
\end{equation}
where $G(t)$ is the greed function evaluated at time $t$, $\ONE_{\{I_t=j\}}$ is an indicator function equal to $1$ if arm $j$ is played at time $t$ (otherwise its value is $0$) and $\Delta_j=\mu^*-\mu_j$ is the difference between the mean of the best arm reward distribution and the mean of the $j$'s arm reward distribution. By considering the threshold $z$ which determines which rule is applied to decide what arm to play, we can rewrite the regret as 
\begin{eqnarray*}
	R_n&\leq& \displaystyle\sum_{j=1}^m G(j) \Delta_j  \\
	&+&\displaystyle \sum_{t=m+1}^n \sum_{j=1}^m \Delta_j G(t) \ONE_{\{G(t)<z\}} \ONE_{\{I_t=j\}} \\
	&+&\displaystyle\sum_{t=m+1}^n \sum_{j=1}^m \Delta_j G(t) \ONE_{\{G(t)\geq z\}} \ONE_{\{I_t=j\}}.
\end{eqnarray*}
Let $B=\{t: G(t-1)<z, G(t) \geq z\}$ be the set of rounds where the high-reward zone is entered. 
Let us call $y_1, y_2, \cdots, y_{B}$ the elements of $B$ and order them in increasing order such that $y_1 < y_2 < \cdots < y_{B}$. Let us also define for every $k \in \{1,\cdots, |B|\}$ the set $Y_k=\{t: t\geq y_k, G(t)\geq z, t<y_{k+1}\}$ (where $y_{B+1}=n$) of times in the high-reward period entered at time $y_k$.
%By eliminating the set $B$ from the game, we have transformed the original game into a shorter one, with $\eta$ steps, where $G(t)$ is bounded by $z$ and the usual UCB algorithm is played. 
By taking the expectation over the policy we have that 
\begin{eqnarray}
	\E[R_n]&\leq& \displaystyle\sum_{j=1}^m G(j) \Delta_j \nonumber \\
	&+&\displaystyle z \sum_{j=1}^m   \Delta_j  \E\left[T_j(\tilde{n}) \right] \nonumber \label{A}\\
	&+&\displaystyle \sum_{j=1}^m  \Delta_j \sum_{k=1}^{|B|}  \sum_{t\in Y_k}G(t)  \P(I_{t}=j),%  \label{ExpectedRegretA}
\end{eqnarray}
where $\tilde{n}$ is the number of turns played when $G(t)$ is under the threshold $z$ at the end of the game, and $\E\left[T_j(\tilde{n}) \right] = \E \left[\sum_{t=m+1}^n \ONE_{\{G(t)<z\}} \ONE_{\{I_t=j\}} \right]$ is the expected number of time arm $j$ is played when $G(t)$ is under the threshold $z$.
For the rounds of the algorithm where $G(t)<z$, we are in the standard setting, so for those times, we follow the standard proof of \cite{auer2002finite}. For the times that $G(t)$ is over the threshold, we need to create a separate bound.
Let us now bound the probability of playing the sub-optimal arm $j$ at time $t$ when the greed function is above the threshold $z$.
\tcbset{colback=white}
\begin{tcolorbox}
	{\bf Second step:} Upper bound for $\P(\{I_{t}=j\})$ when $G(t)\geq z$.
\end{tcolorbox}
From Proposition \ref{Proposition::inclusion_epsilon_z}, we have that\footnotesize
\begin{eqnarray}
	\P( \widehat{X}_{j,T_j(t-1)} > \widehat{X}_{i,T_i(t-1)} \;\; \forall i )
	&\leq& \P(\widehat{X}_{j,T_j(t-1)} > \widehat{X}_{*,T_*(t-1)})\nonumber\\
	&\leq& \P\left(\widehat{X}_{j,T_j(t-1)} > \mu_j + \frac{\Delta_j}{2} \right) + \P\left(\widehat{X}_{*,T_*(t-1)} <  \mu_* - \frac{\Delta_j}{2} \right) \label{FirstTerm1}.
\end{eqnarray}\normalsize
Let us consider the first term of \eqref{FirstTerm1} (the computations for the second term are similar), and let us call $x_t$ the minimum number of times an arm is pulled at turn $t$ when $G(t)$ is under the threshold (see Lemma \ref{Lemma::minimum_pulls_UCB}). We have that
\begin{eqnarray}
	\P\left(\widehat{X}_{j,T_j(t-1)} > \mu_j + \frac{\Delta_j}{2} \right) 
	&= & \sum_{s=1}^{t-1} \P\left(T_j(t-1)=s , \widehat{X}_{j,s} > \mu_j + \frac{\Delta_j}{2}\right) \nonumber \\
	&= & \sum_{s=1}^{t-1} \P\left(T_j(t-1)=s \,\bigg|\, \widehat{X}_{j,s} > \mu_j + \frac{\Delta_j}{2}\right)\P\left(\widehat{X}_{j,s} > \mu_j + \frac{\Delta_j}{2}\right) \nonumber\\
	&\leq & \sum_{s=1}^{t-1} \P\left(T_j(t-1)=s \,\bigg|\, \widehat{X}_{j,s} > \mu_j + \frac{\Delta_j}{2}\right)e^{-\frac{\Delta_j^2}{2}s} \label{T_j_s_sum}\\
	& =  & \sum_{s=x_t}^{t-1} \P\left(T_j(t-1)=s \,\bigg|\, \widehat{X}_{j,s} > \mu_j + \frac{\Delta_j}{2}\right)e^{-\frac{\Delta_j^2}{2}s} \label{T_j_s_sum2} \\
	&\leq & \frac{2}{\Delta_j^2}e^{-\frac{\Delta_j^2}{2} (x_t-1)} := \beta_j^U(t), \label{beta_U}
\end{eqnarray}
%&\leq &  \frac{e^{-\frac{\Delta_j^2}{2}} - e^{-\frac{\Delta_j^2}{2}t}}{1 - e^{-\frac{\Delta_j^2}{2}}} . 
where, from \eqref{T_j_s_sum} to \eqref{T_j_s_sum2}, we used the fact that, for $s < x_t$, $T_j(t-1)$ can not be equal to $s$ since $x_t$ is the minimum number of pulls, and for the remaining terms we used the fact that $ \sum_{s=x}^{\infty} e^{-ks} \leq \frac{1}{k}e^{-k(x-1)}$, where in our case $k=\frac{\Delta_j^2}{2}$. By symmetric argument, we can bound $\P(\{I_{t}=j\})$ by $2\beta_j^U(t)$.
\tcbset{colback=white}
\begin{tcolorbox}
	{\bf Third step:} Upper bound for $\E[T_j(\tilde{n})]$ when $G(t)<z$.
\end{tcolorbox}
When $G(t)<z$, we balance exploration and exploitation by classic UCB.
%Since, when $G(t)<z$, we balance exploration and exploitation by classic UCB, we can bound $\E\left[T_j(\tilde{n}) \right]$ by $\frac{8 \log(\tilde{n})}{\Delta_j^2} + \left(1 + \frac{\pi^2}{3}\right) $ which is the quantity found in the bound of the usual UCB policy.  %(for $n$ rounds the UCB algorithm has a mean regret bounded by $\sum_{j=1}^m \frac{\log{n}}{\Delta_j} + \left(1 + \frac{\pi^2}{3}\right)\sum_{j=1}^m \Delta_j$).
%
%
%\begin{eqnarray*}
%	\E[R_n]&\leq& \displaystyle\sum_{j=1}^m G(j) \Delta_j  \\
%	&+&\displaystyle z \sum_{j=1}^m   \Delta_j  \E\left[T_j(\tilde{n}) \right] \nonumber \\
%	&+&\displaystyle \sum_{j=1}^m  \Delta_j \sum_{k=1}^{|B|}  \sum_{t\in Y_k}G(t) 2\beta_j^U(t).\label{ExpectedRegretA}
%\end{eqnarray*} 
The bound of $\E\left[T_j(\tilde{n}) \right]$ follows from the usual bound on the UCB algorithm: for $n$ rounds the UCB algorithm has a mean regret bounded by $$\sum_{j=1}^m \frac{8\log{n}}{\Delta_j} + \left(1 + \frac{\pi^2}{3}\right)\sum_{j=1}^m \Delta_j.$$
\begin{tcolorbox}
	{\bf Fourth step:} Bringing together all bounding quantities.
\end{tcolorbox}
The bound on the mean regret $\E[R_n]$ at time $n$ is given by (recall that the first $m$ turns are used in the initialization phase, each yielding a regret of $G(j) \Delta_j$):
\begin{eqnarray*}
	\E[R_n]\displaystyle & \leq & \displaystyle\sum_{j=1}^m G(j) \Delta_j  \\
	& + & z \left[ 8 \sum_{j : \mu_j<\mu_*} \left(\frac{\log{\tilde{n}}}{\Delta_j}\right) + \left(1 + \frac{\pi^2}{3}\right)\left(\sum_{j=1}^m \Delta_j\right) \right]     \\
	& + & \displaystyle \sum_{j=1}^m  \Delta_j \sum_{k=1}^{|B|}  \sum_{t\in Y_k}G(t) 2\beta_j^U(t).
\end{eqnarray*}
%When the size of set $B$ decreases with $n$, 
%(is of order $\theta(1/t)$ after an arbitrary time), 
%the total regret has a logarithmic bound in $n$. 

%\newpage
\tcbset{colback=blue!2!white}
\begin{tcolorbox}
	{\bf Lemma \ref{Lemma::minimum_pulls_UCB}}\textit{ 
		Suppose the rewards of the arms are bounded in $[a,b]$ and let us define $r = b-a$ the range of the possible rewards. When using the UCB policy for a game consisting of $n$ turns, each arm will be pulled at least $x_n$ times, where 
		\begin{eqnarray*}
			x_n &=& \max\left\{ y\in \N \, : (m-1)\phi(r,y) \leq n    \right\} + 1,\\
			\phi(r,y)  &=&  \min\left\{ t \geq \tau(r_{ji},y), \, t \in \N \,: t >  \frac{2\, \log(t)}{r^2 + \frac{2 \, \log(t)}{y} -2r\sqrt{\frac{2 \, \log(t)}{y} }}\right\} ,\\
			\tau(r,y) &=& \min\left\{t \in \N :  \sqrt{\frac{2\log(t)}{y} } \geq r\right\}.
		\end{eqnarray*}
	}
\end{tcolorbox}
During the initialization phase, each arm is pulled once, so by turn $t=m$ we have that $T_i(m) = 1$ $\forall i$. Let us consider arm $j$, and $t>m$. After arm $j$ has been played $T_j(t-1)$ times, the algorithm will play arm $j$ for the $(T_j(t-1)+1)$-th time when the following condition is met:
\begin{equation*}\label{eq1_lemma}
	\widehat{X}_{j} + \sqrt{\frac{2 \, \log(t)}{T_j(t-1)} } > 
	\widehat{X}_{i} + \sqrt{\frac{2\, \log(t)}{T_i(t-1)} } \,\,\,\forall i\neq j.  
\end{equation*}
Since rewards are bounded in $[a,b]$, we have that $\widehat{X}_{j}\geq a$ and $\widehat{X}_{i} \leq b$. Therefore,
\begin{equation*}\label{eq1.5_lemma}
	a + \sqrt{\frac{2 \, \log(t)}{T_j(t-1)} } > 
	b + \sqrt{\frac{2\, \log(t)}{T_i(t-1)} } \,\,\,\forall i\neq j
	\,\,\Rightarrow\,\,
	\widehat{X}_{j} + \sqrt{\frac{2 \, \log(t)}{T_j(t-1)} } > 
	\widehat{X}_{i} + \sqrt{\frac{2\, \log(t)}{T_i(t-1)} } \,\,\,\forall i\neq j.  
\end{equation*}
The algorithm is forced to play arm $j$ at time $t$ when each of the $T_i(t-1)$'s are large enough so that the following holds:
\begin{equation}\label{eq2_lemma}
	a + \sqrt{\frac{2 \, \log(t)}{T_j(t-1)} } > 
	b + \sqrt{\frac{2\, \log(t)}{T_i(t-1)} } \,\,\,\forall i\neq j.  
\end{equation}
We want to solve \eqref{eq2_lemma} for $T_i(t-1)$, to see how many times the algorithm is required to pull each arm $i\neq j$ to satisfy \eqref{eq2_lemma}, which can be rewritten as:
\begin{equation}\label{lemma_condition2}
	\sqrt{\frac{2\, \log(t)}{T_i(t-1)} } < \sqrt{\frac{2 \, \log(t)}{T_j(t-1)} } - r\,\,\,\forall i\neq j.
\end{equation}
Let us define
\begin{equation*}\label{lemma_tau}
	\tau(r,T_j(t-1)) = \min\left\{t \in \N \,:  \sqrt{\frac{2\log(t)}{T_j(t-1)} } \geq r\right\}.
\end{equation*}
For $t<\tau(r,T_j(t-1))$, there is no solution to inequality \eqref{lemma_condition2} for $T_i(t-1)$ (because the right hand side of \eqref{lemma_condition2} is negative), and therefore arm $j$ is not necessarily being pulled in turns $t<\tau(r,T_j(t-1))$. \\
For $t\geq \tau(r,T_j(t-1))$, for \eqref{lemma_condition2} to hold, we need that $\forall i \neq j$
\begin{eqnarray*}
	&& \frac{2\, \log(t)}{T_i(t-1)} < r^2 + \frac{2 \, \log(t)}{T_j(t-1)} -2r\sqrt{\frac{2 \, \log(t)}{T_j(t-1)} } \nonumber\\
	&\Leftrightarrow& T_i(t-1) > \frac{2\, \log(t)}{r^2 + \frac{2 \, \log(t)}{T_j(t-1)} -2r\sqrt{\frac{2 \, \log(t)}{T_j(t-1)} }}.%\label{eq3_lemma}
\end{eqnarray*} 
Let us define 
\begin{equation*}\label{phi}
	\phi(r,T_j(t-1)) =  \min\left\{ t \geq \tau(r,T_j(t-1)), \, t\in \N\,: t >  \frac{2\, \log(t)}{r^2 + \frac{2 \, \log(t)}{T_j(t-1)} -2r\sqrt{\frac{2 \, \log(t)}{T_j(t-1)} }} \right\}  ,
\end{equation*}
which can be interpreted as the minimum number of times to pull arm $i$ such that \eqref{lemma_condition2} is satisfied if there were only two arms in the game (arm $j$ and arm $i$).\\ 
%Since there are $m-1$ arms different from $j$, the algorithm will have to pull each arm (different from $j$) $\phi(r,T_j(t-1))$ times to be sure that you will pull arm $j$ for the $(T_j(t-1)+1)$-th time. \\
We will show that the algorithm will have pulled arm $j$ for the $(T_j(t-1)+1)$-th time before or at round 
\begin{equation*}
	(m-1)\phi(r,T_j(t-1)).
\end{equation*}
In order to prove this, assume for the sake of contradiction that the algorithm has played the $j$-th arm $T_j(t-1)$ times at turn $t$, and will not play arm $j$ thereafter. Then, there will eventually be a time $s_1$ such that for some arm $i_1$ we have 
\[T_{i_1}(s_1) =  \phi(r,T_j(s_1)) .\] 
In other words, we will play arm $i_1$ at least $T_{i_1}(s_1)$ times. 
By definition of $\phi(r,T_j(s_1))$, this means that
\[
\widehat{X}_{i_1} + \sqrt{\frac{2 \, \log(s_1)}{T_{i_1}(s_1)} } < 
\widehat{X}_{j} + \sqrt{\frac{2\, \log(s_1)}{T_j(s_1)} }. %\,\,\,\,\,\,\forall t\geq s_1.
\]
Thus, between arm $i_1$ and $j$, the algorithm will always prefer arm $j$. Since by assumption the algorithm does not play arm $j$, it will not choose either arm $j$ nor arm $i_1$ for all the following turns $t\geq s_1$:
\[
\widehat{X}_{i_1} + \sqrt{\frac{2 \, \log(t)}{T_{i_1}(t)} } < 
\widehat{X}_{j} + \sqrt{\frac{2\, \log(t)}{T_j(t)} } \,\,\,\,\,\,\forall t\geq s_1.
\]
Therefore, there will be a time $s_2$ such that, for some arm $i_2\neq i_1, j$, we have 
\[T_{i_2}(s_2) =  \phi(r,T_j(s_2)).\] 
By the same argument, neither arm $i_2$ nor $i_1$ (nor $j$) can be pulled in the following turns $t\geq s_2$.
Since there are $m-1$ arms different from $j$, this argument repeats $m-1$ times, until we have that
\[T_{i}(s_i) =  \phi(r,T_j(s_i))  \,\,\,\,\,\,\forall i\neq j.\]
Since by assumption the number of times the algorithm pulls arm $j$ never changes, at turn $t$ we have $T_j(s_i) = T_j(t-1)$ $\forall i$. Thus, after $(m-1)\phi(r,T_j(t-1))$ pulls, the algorithm will prefer arm $j$ to all the other arms. This means arm $j$ will be pulled the next turn, leading to a contradiction. Thus after at most $(m-1)\phi(r,T_j(t-1))$ turns, the algorithm must have pulled arm $j$ once more. \\
%Note that once an arm $i$ has been pulled $\phi(r,T_j(t-1))$ times, its upper confidence bound will stay lower than the upper confidence bound of arm $j$, even when playing other arms different from both $j$ and $i$. This is because the upper confidence bound for arm $j$ increases at each step with $\sqrt{2\log(t) / T_j(t-1)}$ while the one of arm $i$ increases with $\sqrt{2\log(t) / T_i(t-1)}$, and $T_i(t-1) > T_j(t-1)$. \\
Now that we can compute the turns at which the algorithm will have pulled arm $j$ for at least the $[T_j(t-1)+1]$th time, we can define the minimum number of times $x_n$ that the algorithm will play arm $j$ during a game of $n$ rounds:
\begin{equation*}
	x_n = \max\left\{ T_j(t-1) \in \N : (m-1)\phi(r,T_j(t-1)) \leq n    \right\} + 1.
\end{equation*} 
Intuitively, the term $\max\left\{ T_j(t-1) \in \{ 1, \cdots, n\} : (m-1)\phi(r,T_j(t-1)) \leq n    \right\}$ is the maximum number of pulls to arm $j$ such that the next pull is still possible before the game ends. Therefore, by adding one more pull, $x_n$ counts the minimum number of pulls for arm $j$.
Note that this proof holds for each arm, so the lower bound on the number of times the algorithm plays an arm is the same for each arm and depends on the number of arms $m$, the range $r$, and the maximum number of turns in the game, $n$. 

Corollary \ref{Lemma::minimum_pulls_UCB_different_ranges} is generalization of this result.
%\newpage
\tcbset{colback=blue!2!white}
\begin{tcolorbox}
	\begin{corollary}\label{Lemma::minimum_pulls_UCB_different_ranges}
		Suppose the rewards of arm $i$ are bounded in $[a_i,b_i]$, and let us define $r_{ji} = b_i-a_j$. When using the UCB policy for a game consisting of $n$ turns, each arm $j$ will be pulled at least $x_n(j)$ times, where 
		\begin{eqnarray*}
			x_n(j) &=& \max\left\{ y\in \N \,: \sum_{i\neq j}\phi(r_{ji},y) \leq n    \right\} + 1,\\
			\phi(r_{ji},y)  &=&  \min\left\{t \geq \tau(r_{ji},y), \, t \in \N \,: t >  \frac{2\, \log(t)}{r_{ji}^2 + \frac{2 \, \log(t)}{y} -2r_{ji}\sqrt{\frac{2 \, \log(t)}{y} }}\right\} ,\\
			\tau(r_{ji},y) &=& \min\left\{t \in \N :  \sqrt{\frac{2\log(t)}{y} } \geq r_{ji}\right\}.
		\end{eqnarray*}
	\end{corollary}
\end{tcolorbox}
\noindent The proof of this corollary closely follows the proof of Lemma \ref{Lemma::minimum_pulls_UCB}.\\
During the initialization phase, each arm is pulled once, so by turn $t=m$ we have that $T_i(m) = 1$ $\forall i$. Let us consider arm $j$, and $t>m$. After arm $j$ has been played $T_j(t-1)$ times, the algorithm will play arm $j$ for the $(T_j(t-1)+1)$-th time when the following condition is met:
\begin{equation*}\label{eq1_lemma2}
	\widehat{X}_{j} + \sqrt{\frac{2 \, \log(t)}{T_j(t-1)} } > 
	\widehat{X}_{i} + \sqrt{\frac{2\, \log(t)}{T_i(t-1)} } \,\,\,\forall i\neq j.  
\end{equation*}
Since rewards are bounded in $[a_i,b_i]$ $\forall i$, we have that $\widehat{X}_{j}\geq a_j$ and $\widehat{X}_{i} \leq b_i$. Therefore,
\begin{equation*}\label{eq1.5_lemma2}
	a_j + \sqrt{\frac{2 \, \log(t)}{T_j(t-1)} } > 
	b_i + \sqrt{\frac{2\, \log(t)}{T_i(t-1)} } \,\,\,\forall i\neq j
	\,\,\Rightarrow\,\,
	\widehat{X}_{j} + \sqrt{\frac{2 \, \log(t)}{T_j(t-1)} } > 
	\widehat{X}_{i} + \sqrt{\frac{2\, \log(t)}{T_i(t-1)} } \,\,\,\forall i\neq j.  
\end{equation*}
The algorithm is forced to play arm $j$ at time $t$ when each of the $T_i(t-1)$'s are large enough so that the following holds:
\begin{equation}\label{eq2_lemma2}
	a_j + \sqrt{\frac{2 \, \log(t)}{T_j(t-1)} } > 
	b_i + \sqrt{\frac{2\, \log(t)}{T_i(t-1)} } \,\,\,\forall i\neq j.  
\end{equation}
We want to solve \eqref{eq2_lemma2} for $T_i(t-1)$, to see how many times the algorithm is required to pull each arm $i\neq j$ to satisfy \eqref{eq2_lemma2}, which can be rewritten as:
\begin{equation}\label{lemma2_condition2}
	\sqrt{\frac{2\, \log(t)}{T_i(t-1)} } < \sqrt{\frac{2 \, \log(t)}{T_j(t-1)} } - r_{ji}\,\,\,\forall i\neq j.
\end{equation}
Let us define
\begin{equation*}\label{lemma2_tau}
	\tau(r_{ji},T_j(t-1)) = \min\left\{t \in \N:  \sqrt{\frac{2\log(t)}{T_j(t-1)} } \geq r_{ji}\right\}.
\end{equation*}
For $t<\tau(r_{ji},T_j(t-1))$, there is no solution to inequality \eqref{lemma2_condition2} for $T_i(t-1)$ (because the right hand side of \eqref{lemma2_condition2} is negative), and therefore arm $j$ is not necessarily being pulled in turns $t<\tau(r_{ji},T_j(t-1))$. 

For $t\geq \tau(r_{ji},T_j(t-1))$, for \eqref{lemma2_condition2} to hold, we need that $\forall i \neq j$
\begin{eqnarray*}
	&& \frac{2\, \log(t)}{T_i(t-1)} < r_{ji}^2 + \frac{2 \, \log(t)}{T_j(t-1)} -2r_{ji}\sqrt{\frac{2 \, \log(t)}{T_j(t-1)} } \nonumber\\
	&\Leftrightarrow& T_i(t-1) > \frac{2\, \log(t)}{r_{ji}^2 + \frac{2 \, \log(t)}{T_j(t-1)} -2r_{ji}\sqrt{\frac{2 \, \log(t)}{T_j(t-1)} }}.%\label{eq3_lemma}
\end{eqnarray*} 
Let us define 
\begin{equation*}\label{phi2}
	\phi(r_{ji},T_j(t-1)) =  \min\left\{ t \geq \tau(r_{ji},T_j(t-1)), \, t\in \N\,: t >  \frac{2\, \log(t)}{r_{ji}^2 + \frac{2 \, \log(t)}{T_j(t-1)} -2r_{ji}\sqrt{\frac{2 \, \log(t)}{T_j(t-1)} }} \right\}  .
\end{equation*}
%which can be interpreted as the minimum number of times to pull arm $i$ such that \eqref{lemma2_condition2} is satisfied if there were only two arms in the game (arm $j$ and arm $i$).\\ 
%Since there are $m-1$ arms different from $j$, the algorithm will have to pull each arm (different from $j$) $\phi(r,T_j(t-1))$ times to be sure that you will pull arm $j$ for the $(T_j(t-1)+1)$-th time. \\
We will show that the algorithm will have pulled arm $j$ for the $(T_j(t-1)+1)$-th time before or at round 
\begin{equation*}
	\sum_{i\neq j}\phi(r_{ji},T_j(t-1)).
\end{equation*}
In order to prove this, assume for the sake of contradiction that the algorithm has played the $j$-th arm $T_j(t-1)$ times at turn $t$, and will not play arm $j$ thereafter. Then, there will eventually be a time $s_1$ such that for some arm $i_1$ we have 
\[T_{i_1}(s_1) =  \phi(r_{ji_1},T_j(s_1)) .\] 
In other words, we will play arm $i_1$ at least $T_{i_1}(s_1)$ times. 
By definition of $\phi(r_{ji_1},T_j(s_1))$, this means that
\[
\widehat{X}_{i_1} + \sqrt{\frac{2 \, \log(s_1)}{T_{i_1}(s_1)} } < 
\widehat{X}_{j} + \sqrt{\frac{2\, \log(s_1)}{T_j(s_1)} }.
\]
Thus, between arm $i_1$ and $j$, the algorithm will always prefer arm $j$. Since by assumption the algorithm does not play arm $j$, it will not choose either arm $j$ nor arm $i_1$ for all the following turns $t\geq s_1$:
\[
\widehat{X}_{i_1} + \sqrt{\frac{2 \, \log(t)}{T_{i_1}(t)} } < 
\widehat{X}_{j} + \sqrt{\frac{2\, \log(t)}{T_j(t)} } \,\,\,\,\,\,\forall t\geq s_1.
\]
Therefore, there will be a time $s_2$ such that, for some arm $i_2\neq i_1, j$, we have 
\[T_{i_2}(s_2) =  \phi(r_{ji_2},T_j(s_2)).\] 
By the same argument, neither arm $i_2$ nor $i_1$ (nor $j$) can be pulled in the following turns $t\geq s_2$.
This argument repeats for all the arms, until we have that
\[T_{i}(s_i) =  \phi(r_{ji},T_j(s_i))  \,\,\,\,\,\,\forall i\neq j.\]
Since by assumption the number of times the algorithm pulls arm $j$ never changes, at turn $t$ we have $T_j(s_i) = T_j(t-1)$ $\forall i$. Thus, after $\sum_{i\neq j}T_i(s_i) =\sum_{i\neq j}\phi(r_{ji},T_j(t-1))$ pulls, the algorithm will prefer arm $j$ to all the other arms. This means arm $j$ will be pulled the next turn, leading to a contradiction. Thus after at most $\sum_{i\neq j}\phi(r_{ji},T_j(t-1))$ turns, the algorithm must have pulled arm $j$ once more. 

%Note that once an arm $i$ has been pulled $\phi(r,T_j(t-1))$ times, its upper confidence bound will stay lower than the upper confidence bound of arm $j$, even when playing other arms different from both $j$ and $i$. This is because the upper confidence bound for arm $j$ increases at each step with $\sqrt{2\log(t) / T_j(t-1)}$ while the one of arm $i$ increases with $\sqrt{2\log(t) / T_i(t-1)}$, and $T_i(t-1) > T_j(t-1)$. \\
Now that we can compute the turns at which the algorithm will have pulled arm $j$ for at least the $[T_j(t-1)+1]$th time, we can define the minimum number of times $x_n(j)$ that the algorithm will play arm $j$ during a game of $n$ rounds:
\begin{equation*}
	x_n(j) = \max\left\{ T_j(t-1) \in \N : \sum_{i\neq j}\phi(r_{ji},T_j(t-1)) \leq n    \right\} + 1.
\end{equation*} 
Intuitively, the term $\max\left\{ T_j(t-1) \in \N : \sum_{i\neq j}\phi(r_{ji},T_j(t-1)) \leq n   \right\}$ is the maximum number of pulls to arm $j$ such that the next pull is still possible before the game ends. Therefore, by adding one more pull, $x_n(j)$ counts the minimum number of pulls for arm $j$.

%\newpage
%\section{Lower bound on $x_n$}
\tcbset{colback=blue!2!white}
\begin{tcolorbox}
	{\bf Theorem \ref{Theorem::lower_bound_x_n}}\textit{ 
		Suppose the rewards of the arms are bounded in $[a,b]$ and let us define $r = b-a$ the range of the possible rewards. Let $x_n$ be the  minimum number of times each arm will be pulled when using the UCB policy for a game consisting of $n$ turns ($n>m$). Then,
		\[x_n \in \Omega(\log(n)).\]
		In other words, the minimum number of times the algorithm pulls an arm in a game of $n$ turns grows at least logarithmically\footnotemark in $n$. %other UCB formulation would replace 2 with the UCB parameter
	}
\end{tcolorbox}

\footnotetext{$f(n) \in \Omega(g(n))$ if $\exists k> 0$ $\exists n_0$: $\forall n>n_0$ $f(n) \geq k g(n)$. In our case, $k=\frac{2 \beta }{c\,r^2}$ and $n_0 = m+1$.}
\noindent Recall that 
\begin{equation*}
	x_n = \max\left\{ y \geq 1, \, y \in \N : (m-1)\phi(r,y) \leq n    \right\} + 1,
\end{equation*}
\begin{equation*}
	\phi(r,y)  =  \min\left\{t \geq \tau(r,y),\, t \in \N: t >  \frac{2\, \log(t)}{r^2 + \frac{2 \, \log(t)}{y} -2r\sqrt{\frac{2 \, \log(t)}{y} }}\right\},
\end{equation*}
and
\begin{equation*}
	\tau(r,y) = \min\left\{t \in \N :  \sqrt{\frac{2\log(t)}{y} } \geq r\right\}.
\end{equation*}
Let us define the following:
\begin{eqnarray*}
	c &>& \left(\frac{\sqrt{2}}{r} + 1\right)^2, \;\;\; c \in \R^+,  \\
	q_n &=& \frac{1}{n}\left\lfloor\frac{n}{m-1}\right\rfloor, \;\;\; q_n \in \Q^+\\
	\alpha_n &=& \frac{c\,r^2}{\beta\log_n\left(q_n\right) +2\beta}, \;\;\; \alpha_n \in \R^+, \;\;\; 0<\beta<\frac{1}{2}. %note: log_n(q) is negative, but + 1 is positive
\end{eqnarray*}
To show that $x_n \in \Omega(\log(n))$, let us choose $$y=\left\lfloor\frac{2 \log(n)}{\alpha_n}\right\rfloor$$
and prove that $(m-1)\phi(r,y) \leq n$ (first, second, and third step).  This implies that $x_n \geq y + 1 \geq \frac{2 \log(n)}{\alpha_n}$. Then, by finding an upper bound on $\alpha_n$ that does not depend on $n$ (fourth step), we conclude that $x_n \in \Omega(\log(n))$. 
\tcbset{colback=white}
\begin{tcolorbox}
	\bf{First step:} $\tau(r,y) \leq   \left\lceil n^{\frac{1}{\alpha_n}r^2} \right\rceil$
\end{tcolorbox}
\noindent By definition,
$$\tau(r,y) = \tau\left(r, \left\lfloor\frac{2 \log(n)}{\alpha_n}\right\rfloor\right) =\min\left\{ t \in \N :  \sqrt{\frac{2\log(t)}{\left\lfloor\frac{2 \log(n)}{\alpha_n}\right\rfloor} } \geq r\right\} \leq \min\left\{ t \in \N :  \sqrt{\frac{\alpha_n \, \log(t)}{ \log(n)} } \geq r\right\}.$$
For $s\in\R^+$, we have that
\begin{eqnarray*}
	&&\sqrt{\frac{\alpha_n \, \log(s)}{ \log(n)} } \geq r \\
	&\Rightarrow& \log(s) \geq \frac{1}{\alpha_n} r^2 \, \log(n) \\
	&\Rightarrow& s \geq n^{\frac{1}{\alpha_n}r^2}\\
	&\Rightarrow& \min\left\{t \in \N :  \sqrt{\frac{\alpha_n \, \log(t)}{ \log(n)} } \geq r\right\} =   \left\lceil n^{\frac{1}{\alpha_n}r^2} \right\rceil\\
	&\Rightarrow& \tau(r,y) \leq   \left\lceil n^{\frac{1}{\alpha_n}r^2} \right\rceil.
\end{eqnarray*}
% 
%&\Rightarrow& \tau(r,y) \leq \left\lceil n^{\frac{1}{\alpha_n}r^2}\right\rceil.\\

\tcbset{colback=white}
\begin{tcolorbox}
	\bf{Second step:} $\phi(r,y) \leq \left\lceil n^{\frac{1}{\alpha_n}c\,r^2}
	\right\rceil$
\end{tcolorbox}
\noindent By definition,
$$\phi(r,y)  = \phi\left(r, \left\lfloor\frac{2 \log(n)}{\alpha_n}\right\rfloor\right) \leq  \min\left\{t \geq \left\lceil n^{\frac{1}{\alpha_n}r^2} \right\rceil,\, t \in \N: t >  \frac{2\, \log(t)}{r^2 + \frac{2 \, \log(t)}{\left\lfloor\frac{2 \log(n)}{\alpha_n}\right\rfloor} -2r\sqrt{\frac{2 \, \log(t)}{\left\lfloor\frac{2 \log(n)}{\alpha_n}\right\rfloor} }}\right\},$$
where the inequality follows from the fact that are using an upper bound on $\tau(r,y)$.\\
Let us prove that 
\begin{equation}\label{phi_upperbound_0}
	\min\left\{t \geq \left\lceil n^{\frac{1}{\alpha_n}r^2} \right\rceil,\, t \in \N: t >  \frac{2\, \log(t)}{r^2 + \frac{2 \, \log(t)}{\left\lfloor\frac{2 \log(n)}{\alpha_n}\right\rfloor} -2r\sqrt{\frac{2 \, \log(t)}{\left\lfloor\frac{2 \log(n)}{\alpha_n}\right\rfloor} }}\right\} \leq \left\lceil n^{\frac{1}{\alpha_n}c\,r^2} \right\rceil.
\end{equation}
Let us call 
\begin{equation}\label{inequality_constraint}
	t >  \frac{2\, \log(t)}{r^2 + \frac{2 \, \log(t)}{\left\lfloor\frac{2 \log(n)}{\alpha_n}\right\rfloor} -2r\sqrt{\frac{2 \, \log(t)}{\left\lfloor\frac{2 \log(n)}{\alpha_n}\right\rfloor} }}
\end{equation}
the ``inequality constraint''.
Let us set $t = \left\lceil n^{\frac{1}{\alpha_n}c\,r^2} \right\rceil$ and prove that the inequality constraint is satisfied. We have that 
\begin{equation}\label{phi_upperbound_1}
	\frac{2\, \log\left(\left\lceil n^{\frac{1}{\alpha_n}c\,r^2} \right\rceil\right)}{r^2 + \frac{2 \, \log\left(\left\lceil n^{\frac{1}{\alpha_n}c\,r^2} \right\rceil\right)}{\left\lfloor\frac{2 \log(n)}{\alpha_n}\right\rfloor} -2r\sqrt{\frac{2 \, \log\left(\left\lceil n^{\frac{1}{\alpha_n}c\,r^2} \right\rceil\right)}{\left\lfloor\frac{2 \log(n)}{\alpha_n}\right\rfloor} }} = 
	\frac{2\, \log\left(\left\lceil n^{\frac{1}{\alpha_n}c\,r^2} \right\rceil\right)}{\left(\sqrt{\frac{2 \, \log\left(\left\lceil n^{\frac{1}{\alpha_n}c\,r^2} \right\rceil\right)}{\left\lfloor\frac{2 \log(n)}{\alpha_n}\right\rfloor} } - r\right)^2}. 
\end{equation}
Since
$$
\sqrt{\frac{2 \, \log\left(\left\lceil n^{\frac{1}{\alpha_n}c\,r^2} \right\rceil\right)}{\left\lfloor\frac{2 \log(n)}{\alpha_n}\right\rfloor} } - r \geq \sqrt{\frac{2 \, \log\left( n^{\frac{1}{\alpha_n}c\,r^2} \right)}{\frac{2 \log(n)}{\alpha_n}} } - r = r\sqrt{c} - r > r\left(\frac{\sqrt{2}}{r} + 1\right) - r > \sqrt{2},
$$
then,
$$
\eqref{phi_upperbound_1} < \log\left(\left\lceil n^{\frac{1}{\alpha_n}c\,r^2} \right\rceil\right) \leq \left\lceil n^{\frac{1}{\alpha_n}c\,r^2} \right\rceil.
$$
Thus, $\left\lceil n^{\frac{1}{\alpha_n}c\,r^2} \right\rceil$ is an integer that satisfies the inequality constraint \eqref{inequality_constraint}. Therefore, the minimal integer $t \geq \left\lceil n^{\frac{1}{\alpha_n}r^2} \right\rceil$ satisfying the inequality constraint \eqref{inequality_constraint} must be less or equal to  $\left\lceil n^{\frac{1}{\alpha_n}c\,r^2} \right\rceil$, proving \eqref{phi_upperbound_0} and concluding the proof that $\phi(r,y) \leq \left\lceil n^{\frac{1}{\alpha_n}c\,r^2}
\right\rceil$, where $y = \left\lfloor\frac{2 \log(n)}{\alpha_n}\right\rfloor$.

\tcbset{colback=white}
\begin{tcolorbox}
	\bf{Third step:} $(m-1)\phi(r,y) < n$
\end{tcolorbox}
\noindent Using the upper bound on $\phi(r,y)$ from the previous step, we have that, for $y = \left\lfloor\frac{2 \log(n)}{\alpha_n}\right\rfloor$, 
\begin{eqnarray}
	(m-1)\phi(r,y) &\leq& (m-1)\left\lceil n^{\frac{1}{\alpha_n}c\,r^2} \right\rceil \nonumber\\ 
	&=& (m-1)\left\lceil n^{\frac{\beta\log_n\left(q_n\right)+2\beta}{c\,r^2}c\,r^2} \right\rceil \nonumber\\\label{phiLn1}
	&=& (m-1)\left\lceil n^{\frac{\beta\log_n\left( \frac{1}{n}\left\lfloor\frac{n}{m-1}\right\rfloor \right)+2\beta}{c\,r^2}c\,r^2} \right\rceil \nonumber\\ 
	&=& %(m-1)\left\lceil n^{\log_n\left( \frac{1}{n}\left\lfloor\frac{n}{m-1}\right\rfloor \right)+1} \right\rceil
	(m-1)\left\lceil  \left(\frac{1}{n}\left\lfloor\frac{n}{m-1}\right\rfloor \right)^{\beta} n^{2\beta} \right\rceil \label{phiLn2}\nonumber\\
	&=& (m-1)\left\lceil  n^{2\beta-1}\left\lfloor\frac{n}{m-1}\right\rfloor^{\beta}  \right\rceil \nonumber\\ 
	&<& (m-1)\left\lceil \left\lfloor\frac{n}{m-1}\right\rfloor  \right\rceil \label{phiLn3}\\
	&=& (m-1) \left\lfloor\frac{n}{m-1}\right\rfloor  \nonumber\\  
	&\leq& n,\nonumber
\end{eqnarray}
where in \eqref{phiLn3} we used the fact that $2\beta -1 < 0$ (therefore, $n^{2\beta-1} < 1$), and that $n/(m-1) > 1$ (because $n>m$, and therefore $\left\lfloor n/(m-1)\right\rfloor^{\beta} < \left\lfloor n/(m-1)\right\rfloor$). %This concludes the third step.\\
\tcbset{colback=white}
\begin{tcolorbox}
	\bf{Fourth step:}  $-1 \leq \log_n(q_n) < 0$ 
\end{tcolorbox}
\noindent Since $n>m$, 
\begin{eqnarray*}
	&& q_n = \frac{1}{n}\left\lfloor\frac{n}{m-1}\right\rfloor  \geq \frac{1}{n} \nonumber \\
	&\Rightarrow& \log_n(q_n) \geq \log_n\left( \frac{1}{n} \right) = -1.
\end{eqnarray*} 
We also have that
\begin{eqnarray*}
	&& q_n = \frac{1}{n}\left\lfloor\frac{n}{m-1}\right\rfloor  \leq  \frac{n}{n(m-1)} = \frac{1}{m-1} \nonumber \\
	&\Rightarrow& \log_n(q_n) < \log_n\left( \frac{1}{m-1}\right)<0.
\end{eqnarray*}
\begin{tcolorbox}
	\bf{Fifth step:} $ \frac{2 \log(n)}{\alpha_n} \geq k \log(n),\;\;\;\;k = \frac{2 \beta }{c\,r^2}$
\end{tcolorbox}
Using $-1 \leq \log_n(q_n) < 0$,
\begin{eqnarray*}
	&& \alpha_n = \frac{c\,r^2}{\beta\log_n\left(q_n\right) +2\beta} \leq \frac{c\,r^2}{-\beta +2\beta} = \frac{c\,r^2}{\beta}\\
	&\Rightarrow& \frac{2 \log(n)}{\alpha_n} \geq \frac{2 \beta }{c\,r^2}\log(n).
\end{eqnarray*}
In conclusion, we showed that $$x_n \geq y + 1=\left\lfloor\frac{2 \log(n)}{\alpha_n}\right\rfloor +1 \geq \frac{2 \log(n)}{\alpha_n}  \geq \frac{2 \beta }{c\,r^2}\log(n).$$ This means that $x_n$ is lower bounded by a quantity that increases logarithmically in $n$: $x_n \in \Omega(\log(n))$.

%\newpage
\tcbset{colback=blue!2!white}
\begin{tcolorbox}
	\begin{corollary}\label{Theorem::lower_bound_x_n_different_range}
		Suppose the rewards of arm $i$ are bounded in $[a_i,b_i]$ and let us define $r_{ji} = b_i-a_j$ the range of the possible rewards. Let $x_n(j)$ be the  minimum number of times each arm will be pulled when using the UCB policy for a game consisting of $n$ turns ($n>m$). Then,
		\[x_n(j) \in \Omega(\log(n)).\]
		In other words, the minimum number of times the algorithm pulls arm $j$ in a game of $n$ turns grows at least logarithmically\footnotemark in $n$. %other UCB formulation would replace 2 with the UCB parameter
	\end{corollary}
\end{tcolorbox}
\footnotetext{$f(n) \in \Omega(g(n))$ if $\exists k> 0$ $\exists n_0$: $\forall n>n_0$ $f(n) \geq k g(n)$. In our case, $k=(2 \beta)/(c\,r^2)$ and $n_0 = m+1$.}
\noindent The proof closely follows the proof of Theorem \ref{Theorem::lower_bound_x_n}, with some adjustments to account for the different $r_{ji}$. 

Recall that 
\begin{align*}
	x_n(j) &= \max\left\{ y \geq 1, \, y \in \N : \sum_{i\neq j}\phi(r_{ji},y) \leq n    \right\} + 1, \\
	\phi(r_{ji},y)  &=  \min\left\{t \geq \tau(r_{ji},y),\, t \in \N: t >  \frac{2\, \log(t)}{r_{ji}^2 + \frac{2 \, \log(t)}{y} -2r_{ji}\sqrt{\frac{2 \, \log(t)}{y} }}\right\},\\
	\tau(r_{ji},y) &= \min\left\{t \in \N :  \sqrt{\frac{2\log(t)}{y} } \geq r_{ji}\right\}.
\end{align*}
Let us define the following quantities:
\begin{align*}
	r &= \max_{i\neq j} r_{ji}\\
	c_i &= \left(\frac{\sqrt{2}}{r_{ji}} + 1\right)^2, && c_i \in \R^+ \;\;\forall i\neq j \\
	c &= \max_{i\neq j}\; c_i\\
	q_n &= \frac{1}{n}\left\lfloor\frac{n}{m-1}\right\rfloor, && q_n \in \Q^+\\
	\alpha_{n,i} &= \frac{c\,r_{ji}^2}{\beta\log_n\left(q_n\right) +2\beta}, && \alpha_{n,i} \in \R^+, \;\;\forall i\neq j, \;\;\; 0<\beta<\frac{1}{2}\\ %note: log_n(q) is negative, but + 1 is positive
	\alpha_n &= \min_{i\neq j}\; \alpha_{n,i}.
\end{align*}
To show that $x_n(j) \in \Omega(\log(n))$, let us choose $$y=\left\lfloor\frac{2 \log(n)}{\alpha_{n}}\right\rfloor$$
and prove that $\sum_{i\neq j}\phi(r_{ji},y) \leq n$ (first, second, and third step).  This implies that $x_n(j) \geq y + 1 \geq \frac{2 \log(n)}{\alpha_n}$. Then, by finding an upper bound on $\alpha_n$ that does not depend on $n$ (fourth and fifth step), we conclude that $x_n \in \Omega(\log(n))$. \\
\tcbset{colback=white}
\begin{tcolorbox}
	\noindent {\bf{First step:}}  $\tau(r_{ji},y) \leq   \left\lceil n^{\frac{1}{\alpha_n}r_{ji}^2} \right\rceil$ $\;\;\forall i\neq j$\\
\end{tcolorbox}
\noindent Same proof as in Theorem \ref{Theorem::lower_bound_x_n} for each $i\neq j$, with $r_{ji}$.\\
\tcbset{colback=white}
\begin{tcolorbox}
	\noindent{\bf{Second step:}} $\phi(r_{ji},y) \leq \left\lceil n^{\frac{1}{\alpha_n}c\,r_{ji}^2}
	\right\rceil$ $\;\;\forall i\neq j$\\
\end{tcolorbox}
\noindent Same proof as in Theorem \ref{Theorem::lower_bound_x_n} for each $i\neq j$, with $r_{ji}$, $c_i$, and $\alpha_{n,i}$ to prove that \[\phi(r_{ji},y) \leq \left\lceil n^{\frac{1}{\alpha_{n,i}}c_i\,r_{ji}^2}
\right\rceil.\] 
Then, since $c\geq c_i$ , $r \geq r_{ji}$ , and $\alpha_n \leq 
\alpha_{n,i}$ $\forall i\neq j$, it follows that
$$\phi(r_{ji},y) \leq \left\lceil n^{\frac{1}{\alpha_n}c\,r^2}
\right\rceil,  \;\;\forall i\neq j.  $$
\tcbset{colback=white}
\begin{tcolorbox}
	\noindent{\bf{Third step:}} $\sum_{i\neq j}\phi(r_{ji},y) < n$\\
\end{tcolorbox}
\noindent Same proof as in Theorem \ref{Theorem::lower_bound_x_n} since 
$$\phi(r_{ji},y) \leq \left\lceil n^{\frac{1}{\alpha_n}c\,r^2} \right\rceil \;\;\;\forall i \neq j \;\;\;\Rightarrow \sum_{i\neq j}\phi(r_{ji},y) \leq (m-1)\left\lceil n^{\frac{1}{\alpha_n}c\,r^2} \right\rceil < n.$$  \\
\tcbset{colback=white}
\begin{tcolorbox}
	\noindent {\bf{Fourth step:}}  $-1 \leq \log_n(q_n) < 0$.\\
\end{tcolorbox}
\noindent Same proof as in Theorem \ref{Theorem::lower_bound_x_n}.\\
\tcbset{colback=white}
\begin{tcolorbox}
	\noindent{\bf{Fifth step:}} $ \frac{2 \log(n)}{\alpha_n} \geq k \log(n),\;\;\;\;k = \frac{2 \beta }{c\,r^2}$\\
\end{tcolorbox}
\noindent Using $-1 \leq \log_n(q_n) < 0$, $r \geq r_{ji}$ $\forall j\neq i$
\begin{eqnarray*}
	&& \alpha_n = \min_{i\neq j}\{\alpha_{n,i}\} = \min_{i\neq j}\left\{ \frac{c\,r_{ji}^2}{\beta\log_n\left(q_n\right) +2\beta}\right\} \leq \frac{c\,r^2}{-\beta +2\beta} = \frac{c\,r^2}{\beta}\\
	&\Rightarrow& \frac{2 \log(n)}{\alpha_n} \geq \frac{2 \beta }{c\,r^2}\log(n).
\end{eqnarray*}
In conclusion, we showed that $$x_n \geq y + 1=\left\lfloor\frac{2 \log(n)}{\alpha_n}\right\rfloor +1 \geq \frac{2 \log(n)}{\alpha_n}  \geq \frac{2 \beta }{c\,r^2}\log(n).$$ This means that $x_n$ is lower bounded by a quantity that increases logarithmically in $n$: $x_n \in \Omega(\log(n))$.

\tcbset{colback=blue!2!white}
\begin{tcolorbox}
	\begin{corollary}\label{Corollary::P-UCB_x_n}
		Suppose we used a $P$-UCB policy that plays at turn $t$ arm $j$ with highest upper confidence bound $\widehat{X}_j + \sqrt{\frac{P \, \log(t)}{T_j(t-1)}}$ (the classic UCB is a $P$-UCB policy with $P=2$). Then, $x_n \geq \frac{P \beta }{c\,r^2}\log(n)$.
	\end{corollary}
\end{tcolorbox}
\noindent To prove this define
\begin{align*}
	\phi(r_{ji},y)  &=  \min\left\{t \geq \tau(r_{ji},y),\, t \in \N: t >  \frac{P\, \log(t)}{r_{ji}^2 + \frac{P \, \log(t)}{y} -2r_{ji}\sqrt{\frac{P \, \log(t)}{y} }}\right\},\\
	\tau(r_{ji},y) &= \min\left\{t \in \N :  \sqrt{\frac{P\log(t)}{y} } \geq r_{ji}\right\},\\
	c_i &= \left(\frac{\sqrt{P}}{r_{ji}} + 1\right)^2, \;\;\;\;c_i \in \R^+ \;\;\forall i\neq j, \\
	y &=\left\lfloor\frac{P \log(n)}{\alpha_{n}}\right\rfloor.
\end{align*}

\newpage

\section{The regret bound of the Soft UCB algorithm}\label{proof_UCB2}

\tcbset{colback=blue!2!white}
\begin{tcolorbox}
	\begin{proposition}\label{Proposition::at_least_one_holds}
		When
		\begin{equation}\label{condition}
			\ONE\left\{ \widehat{X}_{j} + \sqrt{\frac{2 \, \log\xi(t)}{T_j(t-1)} } \geq 
			\widehat{X}_{*} + \sqrt{\frac{2\, \log\xi(t)}{T_*(t-1)} } \right\}
		\end{equation}
		is equal to one, at least one of the following has to be true:
		\begin{eqnarray}
			\widehat{X}_{*} &\leq& \mu_* - \sqrt{\frac{2\, \log\xi(t)}{T_*(t-1)} }, \label{first_inequality}\\
			\widehat{X}_{j} &\geq& \mu_j + \sqrt{\frac{2\, \log\xi(t)}{T_j(t-1)} }, \label{second_inequality}\\
			\mu_* &<& \mu_j + 2\sqrt{\frac{2 \, \log\xi(t)}{T_j(t-1)} }. \label{third_inequality}
		\end{eqnarray}
	\end{proposition}
\end{tcolorbox}
\noindent Assume for the sake of contradiction that none of them hold simultaneously. Then from \eqref{first_inequality} we would have that $$\widehat{X}_{*} > \mu_* - \sqrt{\frac{2\log\xi(t)}{T_*(t-1)} };$$ then, by applying \eqref{third_inequality} (with opposite verse since we are assuming it does not hold) we have that  $$\widehat{X}_{*} > \mu_j + 2\sqrt{\frac{2\log\xi(t)}{T_j(t-1)} } - \sqrt{\frac{2\log\xi(t)}{T_*(t-1)} }$$ and then from \eqref{second_inequality} (again, with opposite verse) follows that $$\widehat{X}_{*} > \widehat{X}_{j}  + \sqrt{\frac{2\log\xi(t)}{T_j(t-1)} } - \sqrt{\frac{2\log\xi(t)}{T_*(t-1)} }$$ which is in contradiction with \eqref{condition}.

\tcbset{colback=blue!2!white}
\begin{tcolorbox}
	\begin{proposition}\label{Proposition::one_does_not_hold}
		%If $u= \left\lceil \frac{8}{\Delta_j^2}\log \left( \displaystyle \max_{t\in\{m+1,\cdots,n \}} \xi(t)  \right) \right\rceil$, then for $T_j(t-1)\geq u$ \footnotesize
		%	\begin{equation}
		%		\ONE\left\{ \widehat{X}_{j} + \sqrt{\frac{2 \, \log\xi(t)}{T_j(t-1)} } \geq \widehat{X}_{*} + \sqrt{\frac{2\, \log\xi(t)}{T_*(t-1)} } \right\} 
		%		\leq 
		%		\ONE\left\{ \widehat{X}_{*} \leq \mu_* - \sqrt{\frac{2\, \log\xi(t)}{T_*(t-1)} } \right\} + 
		%		\ONE\left\{ \widehat{X}_{j} \geq \mu_j + \sqrt{\frac{2\, \log\xi(t)}{T_j(t-1)} } \right\}
		%	\end{equation}
		%	\end{proposition}
		%\end{tcolorbox}\normalsize
		\footnotesize
		\begin{eqnarray}
			\ONE\left\{ \widehat{X}_{j} + \sqrt{\frac{2 \, \log\xi(t)}{T_j(t-1)} } \geq \widehat{X}_{*} + \sqrt{\frac{2\, \log\xi(t)}{T_*(t-1)} } \right\} 
			&\leq& 
			\ONE\left\{ \widehat{X}_{*} \leq \mu_* - \sqrt{\frac{2\, \log\xi(t)}{T_*(t-1)} } \right\} \nonumber\\
			&+& 
			\ONE\left\{ \widehat{X}_{j} \geq \mu_j + \sqrt{\frac{2\, \log\xi(t)}{T_j(t-1)} } \right\}
		\end{eqnarray}
	\end{proposition}
\end{tcolorbox}\normalsize
We have that
\begin{eqnarray*}
	&&\mu_* - \mu_j - 2\sqrt{\frac{2\log\xi(t)}{T_j(t-1)} } \\
	&\geq& \mu_* - \mu_j - 2\sqrt{\frac{2\log\xi(t)}{u } } \\
	&=&\mu_* - \mu_j-\Delta_j \sqrt{\frac{\log\xi(t)}{\left\lceil \log \left( \displaystyle \max_{t\in\{m+1,\cdots,n \}} \xi(t)  \right) \right\rceil } } \\
	&\geq& \mu_* - \mu_j - \Delta_j =0,
\end{eqnarray*}
therefore, with this choice of $u$, \eqref{third_inequality} can not hold. Using Preposition \ref{Proposition::at_least_one_holds} we have the result.
%When
%\begin{equation}\label{condition}
%\ONE\left\{ \widehat{X}_{j} + \sqrt{\frac{2 \, \log\xi(t)}{T_j(t-1)} } \geq 
%\widehat{X}_{*} + \sqrt{\frac{2\, \log\xi(t)}{T_*(t-1)} } \right\}
%\end{equation}
%is equal to one, at least one of the following has to be true:
%\begin{eqnarray}
%\widehat{X}_{*} &\leq& \mu_* - \sqrt{\frac{2\, \log\xi(t)}{T_*(t-1)} }; \label{first_inequality}\\
%\widehat{X}_{j} &\geq& \mu_j + \sqrt{\frac{2\, \log\xi(t)}{T_j(t-1)} }; \label{second_inequality}\\
%\mu_* &<& \mu_j + 2\sqrt{\frac{2 \, \log\xi(t)}{T_j(t-1)} }. \label{third_inequality}
%\end{eqnarray}
%In fact, assume for the sake of contradiction that none of them hold simultaneously. Then from \eqref{first_inequality} we would have that $\widehat{X}_{*} > \mu_* - \sqrt{\frac{2\log\xi(t)}{T_*(t-1)} }$; then, by applying \eqref{third_inequality} (with opposite verse since we are assuming it does not hold) we get $\widehat{X}_{*} > \mu_j + 2\sqrt{\frac{2\log\xi(t)}{T_j(t-1)} } - \sqrt{\frac{2\log\xi(t)}{T_*(t-1)} }$ and then from \eqref{second_inequality} (again, with opposite verse) follows that $\widehat{X}_{*} > \widehat{X}_{j}  + \sqrt{\frac{2\log\xi(t)}{T_j(t-1)} } - \sqrt{\frac{2\log\xi(t)}{T_*(t-1)} }$ which is in contradiction with \eqref{condition}.

\tcbset{colback=blue!2!white}
\begin{tcolorbox}
	{\bf Theorem \ref{Theorem::soft_UCB}}\textit{ 
		Let $S=\{m+1, \dots, n\}$. The bound on the mean regret $\E[R_n]$ at time $n$ is given by
		\footnotesize
		\begin{eqnarray*}
			\E[R_n]\displaystyle & \leq & \displaystyle\sum_{j=1}^m G(j) \Delta_j  \label{initialization_Algorithm::UCB_soft} \\
			& + &\displaystyle \max_{t\in S}G(t) \left[   \sum_{j : \mu_j<\mu_*}  \frac{8}{\Delta_j}\log \left( \displaystyle \max_{t\in S} \xi(t) \right)  + \sum_{j=1}^m \Delta_j \left( 1 + \sum_{t=m+1}^n \frac{2(t-1-m)^2}{\xi(t)^{4}} \right)\right]. \label{convSeries}
		\end{eqnarray*}	
	}
\end{tcolorbox}
\tcbset{colback=white}
\begin{tcolorbox}
	{\bf First step:} Derivation of $\E[R_n]$.
\end{tcolorbox}
The total mean regret of a game $\mathcal{I} = \{I_t\}_{t=1}^n$ at round $n$ is given by
\begin{eqnarray*}
	R_n &=&\displaystyle\sum_{j=1}^m G(j) \Delta_j + \sum_{t=m+1}^n \sum_{j=1}^m \Delta_j G(t) \ONE_{\{I_t=j\}} \\ &\leq& \displaystyle\sum_{j=1}^m G(j) \Delta_j + \displaystyle \left( \max_{t\in\{m+1, \cdots, n\}}G(t) \right)\sum_{j=1}^m \Delta_j \sum_{t=m+1}^n \ONE_{\{I_t=j\}}.
\end{eqnarray*}
The expected regret $\E[R_n]$ (expectation taken over the policy) at round $n$ is bounded by
\begin{equation}\label{expectedRegret}
	\E[R_n] \leq \displaystyle\sum_{j=1}^m G(j) \Delta_j + \left( \max_{t\in\{m+1, \cdots, n\}}G(t) \right)\sum_{j=1}^m \Delta_j \E[T_{j}(n)].
\end{equation}
where $T_{j}(n) = \sum_{t=1}^n \ONE_{\{I_t=j\}}$ is the number of times the sub-optimal arm $j$ has been chosen up to round $n$.
\tcbset{colback=white}
\begin{tcolorbox}
	{\bf Second step:} Upper bound on the probability of overestimating or underestimating the expected reward of arm $j$ by more than $\sqrt{2 \log\xi(t)/s_j}$ when $T_j(t-1) = s_j$.
\end{tcolorbox}
Recall from \eqref{mean_estimator} that
\begin{equation*}
	\widehat{X}_{j} = \frac{1}{T_j(t-1)}\sum_{s=1}^{T_j(t-1)} X_j(s).
\end{equation*}
Suppose rewards are bounded\footnote{If rewards are bounded in $[a,b]$, with $r=b-a$, in the following choose $\varepsilon = \sqrt{\frac{2 r\,  \log\xi(t)}{s_j}}$} in $[0,1]$. From Hoeffding's inequality we have that 
\begin{equation*}\label{CH2}
	\P\left( \frac{1}{s_j} \sum_{i=1}^{s_j} X_{j,i} - \mu_j  \leq -\varepsilon \,\bigg|\, T_j(t-1) = s_j \right) \leq \exp\{{-2s_j\varepsilon^2}\},
\end{equation*}
and
\begin{equation*}\label{CH1}
	\P\left( \frac{1}{s_j} \sum_{i=1}^{s_j} X_{j,i} - \mu_j \geq \varepsilon \,\bigg|\, T_j(t-1) = s_j \right) \leq \exp\{{-2s_j\varepsilon^2}\}.
\end{equation*}
Let us define the following function:
\begin{equation*}
	\xi(t) = \left(1 + \frac{t}{G(t)}\right),
\end{equation*}
by selecting $\varepsilon = \sqrt{\frac{2\,  \log\xi(t)}{s_j}}$ we have 
\begin{equation}\label{CH2app}
	\P\left( \widehat{X}_{j} + \sqrt{\frac{2\, \log\xi(t)}{s_j} } \leq \mu_j \,\bigg|\, T_j(t-1) = s_j \right) \leq \xi(t)^{-4},
\end{equation}
and
\begin{equation}\label{CH1app}
	\P\left( \widehat{X}_{j} - \sqrt{\frac{2\, \log\xi(t)}{s_j} } \geq \mu_j \,\bigg|\, T_j(t-1) = s_j \right) \leq \xi(t)^{-4}.
\end{equation}
%Equivalently, we may write for every $j$
%
%\begin{equation}\label{CH2app2}
%\mu_j -\sqrt{\frac{2\,\log\xi(t)}{T_j(t-1)} } \leq \widehat{X}_{j} \;\;\;\text{with probability at least}\;\;1-\xi(t)^{-4},
%\end{equation}
%
%\begin{equation}\label{CH1app1}
%\mu_j + \sqrt{\frac{2\, \log\xi(t)}{T_j(t-1)} } \geq \widehat{X}_{j} \;\;\;\text{with probability at least}\;\;1-\xi(t)^{-4}.
%\end{equation}
%If we choose arm $j$ at round $t$ (i.e., the event $\{I_t=j\}$ occurs) we have that 
%\begin{equation}\label{choosej}
%\widehat{X}_{j} + \sqrt{\frac{2\, \log\xi(t)}{T_j(t-1)} } \geq 
%\widehat{X}_{*} + \sqrt{\frac{2\,  \log\xi(t)}{T_*(t-1)}}.
%\end{equation}
%Let us use \eqref{CH1app1} to upper bound the LHS and \eqref{CH2app2} to lower bound the RHS of \eqref{choosej}, then we get with probability at least $1-2\xi(t)^{-4}$
%
%\begin{equation*}
%\mu_j+2\sqrt{\frac{2 \, \log\xi(t)}{T_j(t-1)} } \geq \mu_*,
%\end{equation*}
%from which we get with probability at least $1-2\xi(t)^{-4}$
%
%\begin{equation}\label{TjBound}
%T_j(t-1) \leq \frac{8}{\Delta_j^2}\log\xi(t) .
%\end{equation}
\tcbset{colback=white}
\begin{tcolorbox}
	{\bf Third step:} Upper bound on  $T_j(n)$ (the number of times that arm $j$ is played by the end of the game).
\end{tcolorbox}
In order to emphasize the dependence of $\widehat{X}_j$ from $T_j(t-1)$ we will sometimes write $\widehat{X}_{j,T_j(t-1)}$.
In the following, notice that in \eqref{step_one_UCB_soft} the summation starts from $m+1$ because in the first $m$ initialization rounds each arm is played once. Moreover, step \eqref{step_two_UCB_soft} follows from \eqref{step_one_UCB_soft} by assuming that arm $j$ has already been played $u$ times. Then, for each $t$, 
\begin{align}
	&\left\{ \widehat{X}_{j,T_j(t-1)} + \sqrt{\frac{2\log\xi(t)}{T_j(t-1)} } \geq 
	\widehat{X}_{*,T_*(t-1)} + \sqrt{\frac{2\log\xi(t)}{T_*(t-1)} }, T_j(t-1)\geq u\right\} \subset  \nonumber \\  
	& \left\{ \displaystyle\max_{s_j \in \{u,  \ldots , T_j(t-1)\}} \widehat{X}_{j,s_j} + \sqrt{\frac{2\log\xi(t)}{s_j} } \geq 
	\displaystyle\min_{s_* \in \{1,  \ldots,  T_*(t-1)\}}\widehat{X}_{*,s_*} + \sqrt{\frac{2\log\xi(t)}{s_*} } \right\}\label{rhsSETS}
\end{align}
which justifies \eqref{step_four_UCB_soft}. We also have that \eqref{rhsSETS} is included in
\normalsize
\begin{equation*}
	%\left\{ \displaystyle\max_{ s_j \in \{u,  \ldots , T_j(t-1)\}} \widehat{X}_{j,s_j} + \sqrt{\frac{2\log\xi(t)}{s_j} } \geq 
	%\displaystyle\min_{s_* \in \{1,  \ldots,  T_*(t-1)\}}\widehat{X}_{*,s_*} + \sqrt{\frac{2\log\xi(t)}{s_*} } \right\}  \subset 
	\bigcup_{s_*=1}^{T_*(t-1)}\bigcup_{s_j=u}^{T_j(t-1)}\left\{ \widehat{X}_{j,s_j} + \sqrt{\frac{2\log\xi(t)}{s_j} } \geq 
	\widehat{X}_{*,s_*} + \sqrt{\frac{2\log\xi(t)}{s_*} } \right\}.
\end{equation*} 
\normalsize Thus, for any integer $u$, we may write
\footnotesize
\begin{eqnarray}
	T_j(n) & = & 1 + \displaystyle\sum_{t=m+1}^n \ONE\{I_t=j\} \label{step_one_UCB_soft}\\
	&=& u + \displaystyle\sum_{t=m+1}^n \ONE\{I_t=j, T_j(t-1)\geq u\}  \label{step_two_UCB_soft}\\
	&= & u + \displaystyle\sum_{t=m+1}^n \ONE\left\{ \widehat{X}_{j,T_j(t-1)} + \sqrt{\frac{2\log\xi(t)}{T_j(t-1)} } \geq 
	\widehat{X}_{*,T_*(t-1)} + \sqrt{\frac{2\log\xi(t)}{T_*(t-1)} }, T_j(t-1)\geq u\right\}  \label{step_three_UCB_soft}\\
	&\leq & u + \displaystyle\sum_{t=m+1}^n \ONE\left\{ \displaystyle\max_{s_j \in \{u,  \ldots , T_j(t-1)\}} \widehat{X}_{j,s_j} + \sqrt{\frac{2\log\xi(t)}{s_j} } \geq 
	\displaystyle\min_{s_* \in \{1,  \ldots,  T_*(t-1)\}}\widehat{X}_{*,s_*} + \sqrt{\frac{2\log\xi(t)}{s_*} } \right\}  \label{step_four_UCB_soft}\\
	&\leq & u + \displaystyle\sum_{t=m+1}^n \;\; \displaystyle\sum_{s_*=1}^{T_*(t-1)}\;\; \sum_{s_j=u}^{T_j(t-1)} \ONE\left\{ \widehat{X}_{j,s_j} + \sqrt{\frac{2\log\xi(t)}{s_j} } \geq 
	\widehat{X}_{*,s_*} + \sqrt{\frac{2\log\xi(t)}{s_*} } \right\}.\label{step_five_UCB_soft}
\end{eqnarray}
\normalsize
Now, by setting $u= \left\lceil \frac{8}{\Delta_j^2}\log \left( \displaystyle \max_{t\in\{m+1,\cdots,n \}} \xi(t)  \right) \right\rceil$ and using Proposition \ref{Proposition::one_does_not_hold}, for $T_j(t-1)\geq u$ we have that
%\begin{eqnarray*}
%&&\mu_* - \mu_j - 2\sqrt{\frac{2\log\xi(t)}{T_j(t-1)} } \\
%&\geq& \mu_* - \mu_j - 2\sqrt{\frac{2\log\xi(t)}{u } } \\
%&=&\mu_* - \mu_j-\Delta_j \sqrt{\frac{\log\xi(t)}{\left\lceil \log \left( \displaystyle \max_{t\in\{m+1,\cdots,n \}} \xi(t)  \right) \right\rceil } } \\
%&\geq& \mu_* - \mu_j - \Delta_j =0,
%\end{eqnarray*}
%therefore, with this choice of $u$, \eqref{third_inequality} can not hold. 
Thus, using this choice of $u$ and \eqref{step_five_UCB_soft}, we have that 
\begin{eqnarray*}
	T_j(n) & \leq & \displaystyle \left\lceil \frac{8}{\Delta_j^2}\log \left( \displaystyle \max_{t\in\{m+1,\cdots,n \}} \xi(t)  \right) \right\rceil \\
	& + & \displaystyle\sum_{t=m+1}^n \;\; \displaystyle\sum_{s_*=1}^{T_*(t-1)}\;\; \sum_{s_j=u}^{T_j(t-1)} \ONE\left\{ \widehat{X}_{*,s_*} \leq \mu_* - \sqrt{\frac{2\log\xi(t)}{s_*} }\right\} \\ 
	& + & \displaystyle\sum_{t=m+1}^n \;\; \displaystyle\sum_{s_*=1}^{T_*(t-1)}\;\; \sum_{s_j=u}^{T_j(t-1)}  \ONE\left\{ \widehat{X}_{j,s_j} \geq \mu_j + \sqrt{\frac{2\log\xi(t)}{s_j} }\right\} 
\end{eqnarray*}
\tcbset{colback=white}
\begin{tcolorbox}
	{\bf Fourth step:} Upper bound on  $\E[T_j(n)]$ (the expected number of times that arm $j$ is played by the end of the game).
\end{tcolorbox}
By taking the expected value of $T_j(n)$ and using the result from the {\bf Second step}, we have that
\begin{eqnarray*}
	\E[T_j(n)] & \leq & \displaystyle \left\lceil \frac{8}{\Delta_j^2}\log \left( \displaystyle \max_{t\in\{m+1,\cdots,n \}} \xi(t)  \right) \right\rceil \\
	& + & \displaystyle\sum_{t=m+1}^n \;\; \displaystyle\sum_{s_*=1}^{T_*(t-1)}\;\; \sum_{s_j=u}^{T_j(t-1)} \P\left\{ \widehat{X}_{*,s_*} \leq \mu_* - \sqrt{\frac{2\log\xi(t)}{s_*} }\right\} \\ 
	& + & \displaystyle\sum_{t=m+1}^n \;\; \displaystyle\sum_{s_*=1}^{T_*(t-1)}\;\; \sum_{s_j=u}^{T_j(t-1)}  \P\left\{ \widehat{X}_{j,s_j} \geq \mu_j + \sqrt{\frac{2\log\xi(t)}{s_j} }\right\} \\
	&\leq&  \frac{8}{\Delta_j^2}\log \left( \displaystyle \max_{t\in\{m+1,\cdots,n \}} \xi(t) \right) + 1  + 2\sum_{t=m+1}^n \xi(t)^{-4}(t-1-m)^2 .
\end{eqnarray*}
where in the last step we use $(t-1-m)$ as upper bound  for $T_*(t-1)$ and $T_j(t-1)$ (cases where we have only played the best arm or arm $j$).
\tcbset{colback=white}
\begin{tcolorbox}
	{\bf Fifth step:} Determine the upper bound on $\E[R_n]$.
\end{tcolorbox}
Using \eqref{expectedRegret} and the result from the previous step, we have that \footnotesize
\begin{eqnarray*}
	\E[R_n]\displaystyle & \leq & \displaystyle\sum_{j=1}^m G(j) \Delta_j  \\
	& + &\displaystyle \max_{t\in\{m+1, \cdots, n\}}G(t) \left(   \sum_{j : \mu_j<\mu_*}  \frac{8}{\Delta_j}\log \left( \displaystyle \max_{t\in\{m+1,\cdots,n \}} \xi(t) \right)  + \sum_{j=1}^m \Delta_j \left[ 1 + \sum_{t=m+1}^n \frac{2(t-1-m)^2}{\xi(t)^{4}} \right]\right).
\end{eqnarray*}

\normalsize

%\newpage

%\newpage
\tcbset{colback=blue!2!white}
\begin{tcolorbox}
	\begin{lemma}\label{Lemma::minimum_pulls_UCB_soft}
		Suppose the rewards of the arms are bounded in $[a,b]$ and let us define $r = b-a$ the range of the possible rewards. When using the UCB soft policy for a game consisting of $n$ turns and a bounded multiplier function $G(t)$, each arm will be pulled at least $x_n$ times, where 
		\begin{eqnarray*}
			x_n &=& \max\left\{ y\in \N \, : (m-1)\phi(r,y,G) \leq n    \right\} + 1,\\
			\phi(r,y, G)  &=&  \min\left\{ t \geq \tau(r,y), \, t \in \N \,: t >  \frac{2\, \log(\overline{\xi(t)})}{r^2 + \frac{2 \, \log(\underline{\xi(t)})}{y} -2r\sqrt{\frac{2 \, \log(\underline{\xi(t)})}{y} }}\right\} ,\\
			\tau(r,y,G) &=& \min\left\{t \in \N :  \sqrt{\frac{2\log(\underline{\xi(t)})}{y} } \geq r\right\},\\
			\overline{\xi(t)} &=& \log\left(1+ \frac{t}{\min_{s>m} G(s)}\right) \;\;\;\;\text{and}\;\;\;\;  \underline{\xi(t)} = \log\left(1+ \frac{t}{\max_{s>m} G(s)}\right).
		\end{eqnarray*}
	\end{lemma}
\end{tcolorbox}
During the initialization phase, each arm is pulled once, so by turn $t=m$ we have that $T_i(m) = 1$ $\forall i$. Let us consider arm $j$, and $t>m$. After arm $j$ has been played $T_j(t-1)$ times, the algorithm will play arm $j$ for the $(T_j(t-1)+1)$-th time when the following condition is met:
\begin{equation*}\label{eq1_lemma_soft}
	\widehat{X}_{j} + \sqrt{\frac{2 \, \log(\xi(t))}{T_j(t-1)} } > 
	\widehat{X}_{i} + \sqrt{\frac{2\, \log(\xi(t))}{T_i(t-1)} } \,\,\,\forall i\neq j.  
\end{equation*}
Since rewards are bounded in $[a,b]$, we have that $\widehat{X}_{j}\geq a$ and $\widehat{X}_{i} \leq b$. 
Let us define 
\begin{equation*}
	\displaystyle \overline{\xi(t)} = \log\left(1+ \frac{t}{\min_{s>m} G(s)}\right) \;\;\;\;\text{and}\;\;\;\;  \underline{\xi(t)} = \log\left(1+ \frac{t}{\max_{s>m} G(s)}\right).
\end{equation*}
Then,
\begin{equation*}\label{eq1.5_lemma_soft}
	a + \sqrt{\frac{2 \, \log(\underline{\xi(t)})}{T_j(t-1)} } > 
	b + \sqrt{\frac{2\, \log(\overline{\xi(t)})}{T_i(t-1)} } \,\,\,\forall i\neq j
	\,\,\Rightarrow\,\,
	\widehat{X}_{j} + \sqrt{\frac{2 \, \log(\xi(t))}{T_j(t-1)} } > 
	\widehat{X}_{i} + \sqrt{\frac{2\, \log(\xi(t))}{T_i(t-1)} } \,\,\,\forall i\neq j.  
\end{equation*}
The algorithm is forced to play arm $j$ at time $t$ when each of the $T_i(t-1)$'s are large enough so that the following holds:
\begin{equation}\label{eq2_lemma_soft}
	a + \sqrt{\frac{2 \, \log(\underline{\xi(t)})}{T_j(t-1)} } > 
	b + \sqrt{\frac{2\, \log(\overline{\xi(t)})}{T_i(t-1)} } \,\,\,\forall i\neq j.  
\end{equation}
We want to solve \eqref{eq2_lemma_soft} for $T_i(t-1)$, to see how many times the algorithm is required to pull each arm $i\neq j$ to satisfy \eqref{eq2_lemma_soft}, which can be rewritten as:
\begin{equation}\label{lemma_condition2_soft}
	\sqrt{\frac{2\, \log(\overline{\xi(t)})}{T_i(t-1)} } < \sqrt{\frac{2 \, \log(\underline{\xi(t)})}{T_j(t-1)} } - r\,\,\,\forall i\neq j.
\end{equation}
Let us define
\begin{equation*}\label{tau_soft}
	\tau(r,T_j(t-1)) = \min\left\{t >m :  \sqrt{\frac{2\log(\underline{\xi(t)})}{T_j(t-1)} } \geq r\right\}.
\end{equation*}
For $t<\tau(r,T_j(t-1))$, there is no solution to inequality \eqref{lemma_condition2_soft} for $T_i(t-1)$ (because the right hand side of \eqref{lemma_condition2_soft} is negative), and therefore arm $j$ is not necessarily being pulled in turns $t<\tau(r,T_j(t-1))$. \\
For $t\geq \tau(r,T_j(t-1))$, for \eqref{lemma_condition2_soft} to hold, we need that $\forall i \neq j$
\begin{eqnarray*}
	&& \frac{2\, \log(\overline{\xi(t)})}{T_i(t-1)} < r^2 + \frac{2 \, \log(\underline{\xi(t)})}{T_j(t-1)} -2r\sqrt{\frac{2 \, \log(\underline{\xi(t)})}{T_j(t-1)} } \nonumber\\
	&\Leftrightarrow& T_i(t-1) > \frac{2\, \log(\overline{\xi(t)})}{r^2 + \frac{2 \, \log(\underline{\xi(t)})}{T_j(t-1)} -2r\sqrt{\frac{2 \, \log(\underline{\xi(t)})}{T_j(t-1)} }}.\label{eq3_lemma_soft}
\end{eqnarray*} 
Let us define 
\begin{equation*}\label{phi_soft}
	\phi(r,T_j(t-1), G) =  \min\left\{ t \in\{ \tau(r,T_j(t-1)), \cdots, n \}: t >  \frac{2\, \log(\overline{\xi(t)})}{r^2 + \frac{2 \, \log(\underline{\xi(t)})}{T_j(t-1)} -2r\sqrt{\frac{2 \, \log(\underline{\xi(t)})}{T_j(t-1)} }} \right\}  ,
\end{equation*}
which can be interpreted as the minimum number of times to pull arm $i$ such that \eqref{lemma_condition2_soft} is satisfied if there were only two arms in the game (arm $j$ and arm $i$).\\ 
%Since there are $m-1$ arms different from $j$, the algorithm will have to pull each arm (different from $j$) $\phi(r,T_j(t-1))$ times to be sure that you will pull arm $j$ for the $(T_j(t-1)+1)$-th time. \\
We will show that the algorithm will have pulled arm $j$ for the $(T_j(t-1)+1)$-th time before or at round 
\begin{equation*}
	(m-1)\phi(r,T_j(t-1),G).
\end{equation*}
In order to prove this, assume for the sake of contradiction that the algorithm has played the $j$-th arm $T_j(t-1)$ times at turn $t$, and will not play arm $j$ thereafter. Then, there will eventually be a time $s_1$ such that for some arm $i_1$ we have 
\[T_{i_1}(s_1) =  \phi(r,T_j(s_1),G) .\] 
In other words, we will play arm $i_1$ at least $T_{i_1}(s_1)$ times. 
By definition of $\phi(r,T_j(s_1),G)$, this means that
\[
\widehat{X}_{i_1} + \sqrt{\frac{2 \, \log(\overline{\xi(s_1)})}{T_{i_1}(s_1)} } < 
\widehat{X}_{j} + \sqrt{\frac{2\, \log(\underline{\xi(s_1)})}{T_j(s_1)} }. %\,\,\,\,\,\,\forall t\geq s_1.
\]
Since by assumption the algorithm does not play arm $j$, it will not choose either arm $j$ nor arm $i_1$ for all the following turns $t\geq s_1$:
\[
\widehat{X}_{i_1} + \sqrt{\frac{2 \, \log(\overline{\xi(t)})}{T_{i_1}(t)} } < 
\widehat{X}_{j} + \sqrt{\frac{2\, \log(\underline{\xi(t)})}{T_j(t)} } \,\,\,\,\,\,\forall t\geq s_1.
\]
Thus, between arm $i_1$ and $j$, the algorithm will always prefer arm $j$.
Therefore, there will be a time $s_2$ such that, for some arm $i_2\neq i_1, j$, we have 
\[T_{i_2}(s_2) =  \phi(r,T_j(s_2),G).\] 
By the same argument, neither arm $i_2$ nor $i_1$ (nor $j$) can be pulled in the following turns $t\geq s_2$.
Since there are $m-1$ arms different from $j$, this argument repeats $m-1$ times, until we have that
\[T_{i}(s_i) =  \phi(r,T_j(s_i),G)  \,\,\,\,\,\,\forall i\neq j.\]
Since by assumption the number of times the algorithm pulls arm $j$ never changes, at turn $t$ we have $T_j(s_i) = T_j(t-1)$ $\forall i$. Thus, after $(m-1)\phi(r,T_j(t-1),G)$ pulls, the algorithm will prefer arm $j$ to all the other arms. This means arm $j$ will be pulled the next turn, leading to a contradiction. Thus after at most $(m-1)\phi(r,T_j(t-1),G)$ turns, the algorithm must have pulled arm $j$ once more. \\
%Note that once an arm $i$ has been pulled $\phi(r,T_j(t-1))$ times, its upper confidence bound will stay lower than the upper confidence bound of arm $j$, even when playing other arms different from both $j$ and $i$. This is because the upper confidence bound for arm $j$ increases at each step with $\sqrt{2\log(\xi(t)) / T_j(t-1)}$ while the one of arm $i$ increases with $\sqrt{2\log(\xi(t)) / T_i(t-1)}$, and $T_i(t-1) > T_j(t-1)$. \\
Now that we can compute the turns at which the algorithm will have pulled arm $j$ for at least the $[T_j(t-1)+1]$th time, we can define the minimum number of times $x_n$ that the algorithm will play arm $j$ during a game of $n$ rounds:
\begin{equation*}
	x_n = \max\left\{ T_j(t-1) \in \N : (m-1)\phi(r,T_j(t-1),G) \leq n    \right\} + 1.
\end{equation*} 
Intuitively, the term $\max\left\{ T_j(t-1) \in \N : (m-1)\phi(r,T_j(t-1),G) \leq n    \right\}$ is the maximum number of pulls to arm $j$ such that the next pull is still possible before the game ends. Therefore, by adding one more pull, $x_n$ counts the minimum number of pulls for arm $j$.
Note that this proof holds for each arm, so the lower bound on the number of times the algorithm plays an arm is the same for each arm and depends on the number of arms $m$, the range $r$, and the maximum number of turns in the game, $n$. \\
Lemma \ref{Lemma::minimum_pulls_UCB_soft_different_ranges} is generalization of this result.

%\newpage
\begin{tcolorbox}
	\begin{lemma}\label{Lemma::minimum_pulls_UCB_soft_different_ranges}
		Suppose the rewards of arm $i$ are bounded in $[a_i,b_i]$, and let us define $r_{ji} = b_i-a_j$. When using the UCB soft policy for a game consisting of $n$ turns and a bounded multiplier function $G(t)$, each arm will be pulled at least $x_n(j)$ times, where 
		\begin{eqnarray*}
			x_n(j) &=& \max\left\{ y \geq 1,\, y\in \N \, : \sum_{i\neq j}\phi(r_{ji},y,G) \leq n    \right\} + 1,\\
			\phi(r_{ji},y, G)  &=&  \min\left\{ t \geq \tau(r_{ji},y), \, t \in \N \,: t >  \frac{2\, \log(\overline{\xi(t)})}{r_{ji}^2 + \frac{2 \, \log(\underline{\xi(t)})}{y} -2r_{ji}\sqrt{\frac{2 \, \log(\underline{\xi(t)})}{y} }}\right\} ,\\
			\tau(r_{ji},y,G) &=& \min\left\{t \in \N :  \sqrt{\frac{2\log(\underline{\xi(t)})}{y} } \geq r_{ji}\right\},\\
			\overline{\xi(t)} &=& \log\left(1+ \frac{t}{\min_{s>m} G(s)}\right) \;\;\;\;\text{and}\;\;\;\;  \underline{\xi(t)} = \log\left(1+ \frac{t}{\max_{s>m} G(s)}\right).
		\end{eqnarray*}
	\end{lemma}
\end{tcolorbox}
\noindent The proof of this Lemma closely follows the proof of Lemma \ref{Lemma::minimum_pulls_UCB_soft}.\\
During the initialization phase, each arm is pulled once, so by turn $t=m$ we have that $T_i(m) = 1$ $\forall i$. Let us consider arm $j$, and $t>m$. After arm $j$ has been played $T_j(t-1)$ times, the algorithm will play arm $j$ for the $(T_j(t-1)+1)$-th time when the following condition is met:
\begin{equation*}\label{eq1_lemma2_j_soft}
	\widehat{X}_{j} + \sqrt{\frac{2 \, \log(t)}{T_j(t-1)} } > 
	\widehat{X}_{i} + \sqrt{\frac{2\, \log(t)}{T_i(t-1)} } \,\,\,\forall i\neq j.  
\end{equation*}
Since rewards are bounded in $[a_i,b_i]$ $\forall i$, we have that $\widehat{X}_{j}\geq a_j$ and $\widehat{X}_{i} \leq b_i$. Let us define 
\begin{equation*}
	\displaystyle \overline{\xi(t)} = \log\left(1+ \frac{1}{\min_{t>m} G(t)}\right) \;\;\;\;\text{and}\;\;\;\;  \underline{\xi(t)} = \log\left(1+ \frac{1}{\max_{t>m} G(t)}\right).
\end{equation*}
Then,
\begin{equation*}\label{eq1.5_lemma2_j_soft}
	a_j + \sqrt{\frac{2 \, \log(\underline{\xi(t)})}{T_j(t-1)} } > 
	b_i + \sqrt{\frac{2\, \log(\overline{\xi(t)})}{T_i(t-1)} } \,\,\,\forall i\neq j
	\,\,\Rightarrow\,\,
	\widehat{X}_{j} + \sqrt{\frac{2 \, \log(\xi(t))}{T_j(t-1)} } > 
	\widehat{X}_{i} + \sqrt{\frac{2\, \log(\xi(t))}{T_i(t-1)} } \,\,\,\forall i\neq j.  
\end{equation*}
The algorithm is forced to play arm $j$ at time $t$ when each of the $T_i(t-1)$'s are large enough so that the following holds:
\begin{equation}\label{eq2_lemma2_j_soft}
	a_j + \sqrt{\frac{2 \, \log(\underline{\xi(t)})}{T_j(t-1)} } > 
	b_i + \sqrt{\frac{2\, \log(\overline{\xi(t)})}{T_i(t-1)} } \,\,\,\forall i\neq j.  
\end{equation}
We want to solve \eqref{eq2_lemma2_j_soft} for $T_i(t-1)$, to see how many times the algorithm is required to pull each arm $i\neq j$ to satisfy \eqref{eq2_lemma2_j_soft}, which can be rewritten as:
\begin{equation}\label{lemma2_condition2_j_soft}
	\sqrt{\frac{2\, \log(\overline{\xi(t)})}{T_i(t-1)} } < \sqrt{\frac{2 \, \log(\underline{\xi(t)})}{T_j(t-1)} } - r_{ji}\,\,\,\forall i\neq j.
\end{equation}
Let us define
\begin{equation*}\label{lemma2_tau_j_soft}
	\tau(r_{ji},T_j(t-1)) = \min\left\{t >m :  \sqrt{\frac{2\log(\underline{\xi(t)})}{T_j(t-1)} } \geq r_{ji}\right\}.
\end{equation*}
For $t<\tau(r_{ji},T_j(t-1))$, there is no solution to inequality \eqref{lemma2_condition2_j_soft} for $T_i(t-1)$ (because the right hand side of \eqref{lemma2_condition2_j_soft} is negative), and therefore arm $j$ is not necessarily being pulled in turns $t<\tau(r_{ji},T_j(t-1))$. \\
For $t\geq \tau(r_{ji},T_j(t-1))$, for \eqref{lemma2_condition2_j_soft} to hold, we need that $\forall i \neq j$
\begin{eqnarray*}
	&& \frac{2\, \log(\overline{\xi(t)})}{T_i(t-1)} < r_{ji}^2 + \frac{2 \, \log(\underline{\xi(t)})}{T_j(t-1)} -2r_{ji}\sqrt{\frac{2 \, \log(\underline{\xi(t)})}{T_j(t-1)} } \nonumber\\
	&\Leftrightarrow& T_i(t-1) > \frac{2\, \log(\overline{\xi(t)})}{r_{ji}^2 + \frac{2 \, \log(\underline{\xi(t)})}{T_j(t-1)} -2r_{ji}\sqrt{\frac{2 \, \log(\underline{\xi(t)})}{T_j(t-1)} }}.\label{eq3_lemma_j_soft}
\end{eqnarray*} 
Let us define 
\begin{equation*}\label{phi_j_soft}
	\phi(r_{ji},T_j(t-1)) =  \min\left\{ t \in\{ \tau(r_{ji},T_j(t-1)), \cdots, n \}: t >  \frac{2\, \log(\overline{\xi(t)})}{r_{ji}^2 + \frac{2 \, \log(\underline{\xi(t)})}{T_j(t-1)} -2r_{ji}\sqrt{\frac{2 \, \log(\underline{\xi(t)})}{T_j(t-1)} }} \right\}  .
\end{equation*}
%which can be interpreted as the minimum number of times to pull arm $i$ such that \eqref{lemma2_condition2} is satisfied if there were only two arms in the game (arm $j$ and arm $i$).\\ 
%Since there are $m-1$ arms different from $j$, the algorithm will have to pull each arm (different from $j$) $\phi(r,T_j(t-1))$ times to be sure that you will pull arm $j$ for the $(T_j(t-1)+1)$-th time. \\
We will show that the algorithm will have pulled arm $j$ for the $(T_j(t-1)+1)$-th time before or at round 
\begin{equation*}
	\sum_{i\neq j}\phi(r_{ji},T_j(t-1)).
\end{equation*}
In order to prove this, assume for the sake of contradiction that the algorithm has played the $j$-th arm $T_j(t-1)$ times at turn $t$, and will not play arm $j$ thereafter. Then, there will eventually be a time $s_1$ such that for some arm $i_1$ we have 
\[T_{i_1}(s_1) =  \phi(r_{ji_1},T_j(s_1)) .\] 
In other words, we will play arm $i_1$ at least $T_{i_1}(s_1)$ times. 
By definition of $\phi(r_{ji_1},T_j(s_1))$, this means that
\[
\widehat{X}_{i_1} + \sqrt{\frac{2 \, \log(\overline{\xi(s_1)})}{T_{i_1}(s_1)} } < 
\widehat{X}_{j} + \sqrt{\frac{2\, \log(\underline{\xi(s_1)})}{T_j(s_1)} }.
\]
Thus, between arm $i_1$ and $j$, the algorithm will always prefer arm $j$. Since by assumption the algorithm does not play arm $j$, it will not choose either arm $j$ nor arm $i_1$ for all the following turns $t\geq s_1$:
\[
\widehat{X}_{i_1} + \sqrt{\frac{2 \, \log(t)}{T_{i_1}(t)} } < 
\widehat{X}_{j} + \sqrt{\frac{2\, \log(t)}{T_j(t)} } \,\,\,\,\,\,\forall t\geq s_1.
\]
Therefore, there will be a time $s_2$ such that, for some arm $i_2\neq i_1, j$, we have 
\[T_{i_2}(s_2) =  \phi(r_{ji_2},T_j(s_2)).\] 
By the same argument, neither arm $i_2$ nor $i_1$ (nor $j$) can be pulled in the following turns $t\geq s_2$.
This argument repeats for all the arms, until we have that
\[T_{i}(s_i) =  \phi(r_{ji},T_j(s_i))  \,\,\,\,\,\,\forall i\neq j.\]
Since by assumption the number of times the algorithm pulls arm $j$ never changes, at turn $t$ we have $T_j(s_i) = T_j(t-1)$ $\forall i$. Thus, after $\sum_{i\neq j}T_i(s_i) =\sum_{i\neq j}\phi(r_{ji},T_j(t-1))$ pulls, the algorithm will prefer arm $j$ to all the other arms. This means arm $j$ will be pulled the next turn, leading to a contradiction. Thus after at most $\sum_{i\neq j}\phi(r_{ji},T_j(t-1))$ turns, the algorithm must have pulled arm $j$ once more. \\
%Note that once an arm $i$ has been pulled $\phi(r,T_j(t-1))$ times, its upper confidence bound will stay lower than the upper confidence bound of arm $j$, even when playing other arms different from both $j$ and $i$. This is because the upper confidence bound for arm $j$ increases at each step with $\sqrt{2\log(t) / T_j(t-1)}$ while the one of arm $i$ increases with $\sqrt{2\log(t) / T_i(t-1)}$, and $T_i(t-1) > T_j(t-1)$. \\
Now that we can compute the turns at which the algorithm will have pulled arm $j$ for at least the $[T_j(t-1)+1]$th time, we can define the minimum number of times $x_n(j)$ that the algorithm will play arm $j$ during a game of $n$ rounds:
\begin{equation*}
	x_n(j) = \max\left\{ T_j(t-1) \in \N : \sum_{i\neq j}\phi(r_{ji},T_j(t-1)) \leq n    \right\} + 1.
\end{equation*} 
Intuitively, the term $\max\left\{ T_j(t-1) \in \N : \sum_{i\neq j}\phi(r_{ji},T_j(t-1)) \leq n   \right\}$ is the maximum number of pulls to arm $j$ such that the next pull is still possible before the game ends. Therefore, by adding one more pull, $x_n(j)$ counts the minimum number of pulls for arm $j$.

\newpage

\section{Regret bound proof for regulating greed with variable arm pool size}\label{Appendix::variable_pool}

\tcbset{colback=blue!2!white}
\begin{tcolorbox}
	\begin{proposition}\label{Proposition::inclusion_variable_pool}
		Let us call $\Delta_j^{*-m_t}=\mu^{*-m_t}-\mu_j$ the difference between the mean reward of the $m_t$th-best arm and the mean reward of arm $j$ (following definition, $\Delta_j=\Delta_j^{*-1}$ and $\mu^*=\mu^{*-1}$). Let us define the following events:\small
		\begin{eqnarray*}
			A &=& \left\{ \widehat{X}_{j,T_j(t-1)} > \widehat{X}_{*-m_t,T_{*-m_t}(t-1)} \right\},\\
			B &=& \left\{ \widehat{X}_{j,T_j(t-1)} > \mu_j + \frac{\Delta^{*-m_t}_j}{2}  \right\},\\
			C &=& \left\{\widehat{X}_{*,T_*(t-1)} <  \mu^{*-m_t} - \frac{\Delta^{*-m_t}_j}{2}  \right\}.
		\end{eqnarray*}
		\normalsize Then,
		\begin{equation}\label{_____inclusion1}
			A \subset \left(  B \cup  C  \right).
		\end{equation}
	\end{proposition}
\end{tcolorbox}
%\begin{itemize}
%	\item $\widehat{X}_{*,T_*(t-1)} \leq \mu_* - \frac{\Delta_j}{2}$;
%	\item $\widehat{X}_{j,T_j(t-1)} \geq \mu_j + \frac{\Delta_j}{2}$.
%\end{itemize}
Intuitively, inclusion \eqref{_____inclusion1} means that we play arm $j$ when we underestimate the mean reward of the $m_t$th best arm, or when we overestimate that of arm $j$.
Assume for the sake of contradiction that there exists an element $\omega \in A$ that does not belong to $B \cup C$. Then, we have that $\omega \in \left(B \cup C\right)^{\mathcal{C}}$\small
\begin{eqnarray}
	\Rightarrow \;\;\omega & \in & \left(  \left\{ \widehat{X}_{j,T_j(t-1)} > \mu_j + \frac{\Delta^{*-m_t}_j}{2}  \right\} \cup 
	\left\{\widehat{X}_{*-m_t,T_{*-m_t}(t-1)} <  \mu^{*-m_t} - \frac{\Delta^{*-m_t}_j}{2}  \right\}  \right)^{\mathcal{C}} \label{_____toContradict1} \\
	\Rightarrow \;\;\omega &\in&   \left\{ \widehat{X}_{j,T_j(t-1)} \leq \mu_j + \frac{\Delta^{*-m_t}_j}{2}  \right\} \cap 
	\left\{\widehat{X}_{*-m_t,T_{*-m_t}(t-1)} \geq  \mu^{*-m_t} - \frac{\Delta^{*-m_t}_j}{2}  \right\}. \label{_____NONinclusion1}
\end{eqnarray}\normalsize
By definition we have $\mu^{*-m_t} - \frac{\Delta^{*-m_t}_j}{2} = \mu_* - \frac{\mu^{*-m_t}-\mu_j}{2} = \frac{\mu^{*-m_t}+\mu_j}{2} =  \mu_j + \frac{\Delta^{*-m_t}_j}{2}$. From the inequalities given in \eqref{_____NONinclusion1} it follows that  
\begin{eqnarray*}
	\widehat{X}_{*-m_t,T_{*-m_t}(t-1)} \geq \mu^{*-m_t} - \frac{\Delta^{*-m_t}_j}{2}  = \mu_j + \frac{\Delta^{*-m_t}_j}{2} 
	\geq \widehat{X}_{j,T_j(t-1)},
\end{eqnarray*}
but this contradicts our assumption that $ \omega \in A = \left\{ \widehat{X}_{j,T_j(t-1)} > \widehat{X}_{*-m_t,T_{*-m_t}(t-1)} \right\}$.\\ 
Therefore, all elements of $A$ belong to $B \cup C$.

\tcbset{colback=blue!2!white}
\begin{tcolorbox}
	{\bf Theorem \ref{Theorem::regret_variable_pool}}\textit{
		The bound on the mean regret $\E[R_n]$ at time $n$ is given by
		\begin{equation*}
			\E[R_n]\displaystyle  \leq  \sum_{t=1}^n \sum_{j=1}^m \Delta_j G(t) \frac{2}{m_t}\beta_j ,
		\end{equation*}	
		where
		\begin{equation*}\label{beta::z_pool}
			\beta_j(t)=\gamma\log(t)(t)^{-\gamma/5} + \frac{2}{(\Delta_j^{*-m_t})^2} t^{-\frac{\gamma(\Delta_j^{*-m_t})^2}{2}}.
		\end{equation*}	
	}
\end{tcolorbox}
\tcbset{colback=white}
\begin{tcolorbox}
	{\bf First step:} Derivation of $\E[R_n]$.
\end{tcolorbox}
The total mean regret of a game $\mathcal{I} = \{I_t\}_{t=1}^n$ at round $n$ is given by
\begin{equation}
	R_n=\sum_{t=1}^n \sum_{j=1}^m \Delta_j G(t) \ONE_{\{I_t=j\}},\label{regret::z_pool}
\end{equation}
where $G(t)$ is the greed function evaluated at time $t$, $\ONE_{\{I_t=j\}}$ is an indicator function equal to $1$ if arm $j$ is played at time $t$ (otherwise its value is $0$) and $\Delta_j=\mu^*-\mu_j$ is the difference between the mean of the best arm reward distribution and the mean of the $j$'s arm reward distribution. Similarly, $\Delta_j^{*-m_t}=\mu^{*-m_t}-\mu_j$ is the difference between the mean reward of the $m_t$th-best arm and the mean reward of arm $j$ (with this definition we can also write $\Delta_j=\Delta_j^{*-1}$ and $\mu^*=\mu^{*-1}$). 
By taking the expectation with respect to the policy of \eqref{regret::z_pool} we get 
\begin{equation}
	\E[R_n]=\sum_{t=1}^n  \frac{G(t)}{m_t}  \sum_{j=1}^m \Delta_j \P\left( \widehat{X}_{j,T_j(t-1)} >  \widehat{X}_{i,T_i(t-1)} \; \;\;\text{for at least}\;\;m-m_t\;\;\text{indexes}\;\;i \right),\label{mean_regret::z_pool}
\end{equation}
where $m_t$ is the size of the pool of arms at time $t$, defined by $m_t = \min\left( m, \max\left( 1,  \left\lfloor\frac{c m}{t G(t)}\right\rfloor  \right) \right)$.
\tcbset{colback=white}
\begin{tcolorbox}
	{\bf Second step:} Upper bound on $ \P\left( \widehat{X}_{j,T_j(t-1)} >  \widehat{X}_{i,T_i(t-1)} \; \;\;\text{for at least}\;\;m-m_t\;\;\text{indexes}\;\;i \right)$.
\end{tcolorbox}
We have that
\begin{eqnarray}
	&&\P\left( \widehat{X}_{j,T_j(t-1)} > \widehat{X}_{i,T_i(t-1)} \;\; \;\;\text{for at least}\;\;m-m_t\;\;\text{indexes}\;\;i \right)\nonumber\\
	&& \leq \P\left(\widehat{X}_{j,T_j(t-1)}\nonumber > \widehat{X}_{*-m_t,T_{*-m_t}(t-1)}\right)\\
	&& \leq \P\left(\widehat{X}_{j,T_j(t-1)} > \mu_j + \frac{\Delta^{*-m_t}_j}{2} \right) + \P\left(\widehat{X}_{*,T_*(t-1)} <  \mu^{*-m_t} - \frac{\Delta^{*-m_t}_j}{2} \right)  \label{probs::z_pool}
\end{eqnarray} \normalsize
the last inequality follows from Proposition \ref{Proposition::inclusion_variable_pool}.
%the fact that 
%\begin{equation}\label{inclusion::z_pool}
%\left\{ \widehat{X}_{j,T_j(t-1)} \geq \widehat{X}_{*-m_t,T_{*-m_t}(t-1)} \right\} \subset \left(  \left\{ \widehat{X}_{j,T_j(t-1)} \geq \mu_j + \frac{\Delta^{*-m_t}_j}{2}  \right\} \cup \left\{\widehat{X}_{*,T_*(t-1)} \leq  \mu^{*-m_t} - \frac{\Delta^{*-m_t}_j}{2}  \right\}  \right).
%\end{equation}
%In fact, suppose that the inclusion \eqref{inclusion::z_pool} does not hold. Then we would have that \footnotesize
%\begin{eqnarray}
%\left\{ \widehat{X}_{j,T_j(t-1)} \geq \widehat{X}_{*-m_t,T_{*-m_t}(t-1)} \right\} & \subset & \left(  \left\{ \widehat{X}_{*,T_*(t-1)}  \leq  \mu^{*-m_t} - \frac{\Delta^{*-m_t}_j}{2} \right\} \cup \left\{ \widehat{X}_{j,T_j(t-1)} \geq \mu_j + \frac{\Delta^{*-m_t}_j}{2}  \right\}  \right)^C \label{toContradict::z_pool} \\
%&=&   \left\{ \widehat{X}_{*,T_*(t-1)}  >  \mu^{*-m_t} - \frac{\Delta^{*-m_t}_j}{2} \right\} \cap \left\{ \widehat{X}_{j,T_j(t-1)} < \mu_j + \frac{\Delta^{*-m_t}_j}{2} \right\} , \label{NONinclusion::z_pool}
%\end{eqnarray} \normalsize
%but from the intersection of events given in \eqref{NONinclusion::z_pool} it follows that  $\widehat{X}_{*-m_t,T_{*-m_t}(t-1)} > \mu^{*-m_t} - \frac{\Delta^{*-m_t}_jj}{2} > \mu_j - \frac{\Delta^{*-m_t}_j}{2} > \widehat{X}_{j,T_j(t-1)}$  which contradicts \eqref{toContradict::z_pool}.
\tcbset{colback=white}
\begin{tcolorbox}
	{\bf Third step:} Upper bound for \eqref{probs::z_pool}.
\end{tcolorbox}
Let us consider the first term of \eqref{probs::z_pool} (the computations for the second term are similar),\footnotesize
\begin{eqnarray}
	\P\left(\widehat{X}_{j,T_j(t-1)} > \mu_j + \frac{\Delta^{*-m_t}_j}{2} \right) 
	&= & \sum_{s=1}^{t-1} \P\left(T_j(t-1)=s , \widehat{X}_{j,s} > \mu_j + \frac{\Delta_j^{*-m_t}}{2}\right) \nonumber \\
	&= & \sum_{s=1}^{t-1} \P\left(T_j(t-1)=s \,\bigg|\, \widehat{X}_{j,s} > \mu_j + \frac{\Delta_j^{*-m_t}}{2}\right)\P\left(\widehat{X}_{j,s} > \mu_j + \frac{\Delta_j^{*-m_t}}{2}\right) \nonumber\\
	&\leq & \sum_{s=1}^{t-1} \P\left(T_j(t-1)=s \,\bigg|\, \widehat{X}_{j,s} > \mu_j + \frac{\Delta_j^{*-m_t}}{2}\right)e^{-\frac{s(\Delta_j^{*-m_t})^2}{2}}, \label{ToCont1::z_pool}
\end{eqnarray}\normalsize
where in the last inequality we used the Hoeffding's bound\footnote{{\bf Hoeffding's bound:} Let $X_1, \cdots, X_n$ be r.v. bounded in $[a_i,b_i]$ $\forall i$. Let $\widehat{X} = \frac{1}{n}\sum_{i=1}^n X_i$ and $\mu = \E[\widehat{X}]$. \\Then, $\P\left(\widehat{X} - \mu \geq \varepsilon\right) \leq \exp\left\{ -\frac{2n^2\varepsilon^2}{\sum_{i=1}^n (b_i-a_i)^2 }  \right\}$.}. 
\tcbset{colback=white}
\begin{tcolorbox}
	{\bf Fourth step:} Upper bound for \eqref{ToCont1::z_pool}.
\end{tcolorbox}
Let us define $T_j^R(t-1)$ as the number of times arm $j$ is played at random when the pool size is full before round $t$ starts, and let us define
$$
\lambda_t=\frac{1}{2m}\sum_{s=1}^{t} \ONE\left\{\min\left( m, \max\left( 1,\left\lfloor \frac{c m}{s G(s)} \right\rfloor \right) \right) = m\right\}. 
$$
Then,
\begin{eqnarray}
	\eqref{ToCont1::z_pool} &\leq& \sum_{s=1}^{ \lfloor \lambda_t \rfloor } \P\left(T_j(t-1)=s \bigg| \widehat{X}_{j,s} \geq \mu_j + \frac{\Delta_j^{*-m_t}}{2}\right) + \sum_{s=\lfloor \lambda_t \rfloor + 1}^{t-1}e^{-\frac{(\Delta_j^{*-m_t})^2}{2}s}  \nonumber\\
	&\leq&  \sum_{s=1}^{ \lfloor \lambda_t \rfloor } \P\left(T_j(t-1)=s \bigg| \widehat{X}_{j,s} \geq \mu_j + \frac{\Delta_j^{*-m_t}}{2}\right) + \frac{2}{\Delta_j^2}e^{-\frac{(\Delta_j^{*-m_t})^2}{2} \lfloor \lambda_t \rfloor} \nonumber\\
	&\leq & \sum_{s=1}^{ \lfloor \lambda_t \rfloor } \P\left(T^R_j(t-1) \leq s \bigg| \widehat{X}_{j,s} \geq \mu_j + \frac{\Delta_j^{*-m_t}}{2}\right) + \frac{2}{\Delta_j^2}e^{-\frac{(\Delta_j^{*-m_t})^2}{2} \lfloor \lambda_t \rfloor}\nonumber\\
	&\leq & \lfloor \lambda_t \rfloor \P\left(T^R_j(t-1) \leq  \lambda_t \right) + \frac{2}{(\Delta_j^{*-m_t})^2}e^{-\frac{(\Delta_j^{*-m_t})^2}{2} \lfloor \lambda_t \rfloor}\label{B::z_pool}
\end{eqnarray}
where for the first $\lfloor \lambda_t \rfloor$ terms of the sum we upper-bounded $e^{-\frac{\Delta_j^2}{2}s}$ by $1$, and for the remaining terms we used the fact that $ \sum_{\lfloor \lambda_t \rfloor+1}^{\infty} e^{-ks} \leq \frac{1}{k}e^{-kx}$, where in our case $k=\frac{\Delta_j^2}{2}$.
\begin{tcolorbox}
	{\bf Fifth step:} Upper bound for $\P\left(T^R_j(t-1) \leq  \lambda_t \right)$.
\end{tcolorbox}
Since $T_j^R(t-1)$ is a sum of $\lambda = \sum_{s=1}^{t} \ONE\left\{ m_s=m\right\}$ independent Bernoulli r.v. with parameter $1/m_s$, we have that \small
\begin{eqnarray*} 
	\E[T_j^R(t-1)] &=& \frac{1}{m} \sum_{s=1}^{t}  \ONE\left\{ \min\left( m, \max\left( 1, \left\lfloor \frac{c m}{s G(s)} \right\rfloor \right) \right)=m \right\} = 2\lambda_t, \\ 
	Var(T_j^R(t-1)) &=& 
	\frac{1}{m}\left(1-\frac{1}{m}\right) \sum_{s=1}^{t}  \ONE\left\{\min\left( m, \max\left( 1, \left\lfloor \frac{c m}{s G(s)}\right\rfloor \right) \right)=m \right\} \leq \E[T_j^R(t-1)] ,
\end{eqnarray*}\normalsize
and, using the Bernstein inequality $\P(S_n \leq \E[S_n] - a) \leq \exp\{-\frac{a^2/2}{\sigma^2  +  a/2}\}$ with $S_n = T_j^R(t-1)$ and $a=\frac{1}{2}\E[T_j^R(t-1)]$,
\begin{eqnarray}
	\P(T_j^R(t-1) \leq \lambda_t ) &=& \P\left( T_j^R(t-1) \leq  \E[T_j^R(t-1)] - \frac{1}{2}\E[T_j^R(t-1)] \right) \nonumber\\
	&\leq & \exp \left\{ -\frac{  \frac{1}{8}(\E[T_j^R(t-1)])^2   }{\E[T_j^R(t-1)] + \frac{1}{4}\E[T_j^R(t-1)]   } \right\} \nonumber\\
	&= & \exp  \left\{ -\frac{4}{5}\frac{1}{8}\E[T_j^R(t-1)] \right\}  = 
	\exp \left\{ -\frac{1}{5}  \lambda_t  \right\}.\label{BoundPx0::z_pool}
\end{eqnarray}
In order to bound \eqref{BoundPx0::z_pool} we need $\lambda_t \geq \gamma \log(t)$ with $\gamma > 5$ so that $\P(T_j^R(t-1) \leq \lambda_t ) < t^{-\frac{\gamma}{5}}$.
\tcbset{colback=white}
\begin{tcolorbox}
	%\begin{remark}\label{Remark::G_prime}
	\textit{Remark \ref{Remark::G_prime}.}
	If $G(t)$ does not satisfy the requirement that $\lambda_t \geq \gamma \log(t)$ it is easy to construct a new multiplier function $G'(t)$ by first finding the set $S=\{ t: \lambda_t > \lceil \gamma \log(t-1) \rceil \}$  and then by defining 
	\begin{equation*}
		G'(s) = \begin{cases}
			(c-1)/s & \mbox{for } s \in \{t, t+1, \cdots, t+2m\} \mbox{ if }  s\in S  \\
			G(s)    & \mbox{otherwise}.
		\end{cases}
	\end{equation*}
	If we do use $G'(t)$ instead of $G(t)$ in the algorithm, the bound holds for $G'(t)$ instead of $G(t)$.
	%\end{remark}
\end{tcolorbox}
\tcbset{colback=white}
\begin{tcolorbox}
	{\bf Sixth step:} Bringing together all bounding quantities.
\end{tcolorbox}
We have that \eqref{B::z_pool} is bounded by
\begin{equation*}
	\beta_j^{VP}(t)=\gamma\log(t)(t)^{-\gamma/5} + \frac{2}{(\Delta_j^{*-m_t})^2} t^{-\frac{\gamma(\Delta_j^{*-m_t})^2}{2}}.
\end{equation*}
%We have now an upper bound for $\P( \widehat{X}_{j,T_j(t-1)} \geq \widehat{X}_{i,T_i(t-1)} \;\; \forall i )$. We can use this to easily bound $\P( \widehat{X}_{j,T_j(t-1)} \geq \widehat{X}_{i,T_i(t-1)} \;\; \forall i )$ in \eqref{ExpectedRegretA} which yields the following bound on the mean regret at time $n$:
The computations for the second term in \eqref{ToCont1::z_pool} are similar, therefore
\begin{equation*}
	\E[R_n]\displaystyle  \leq  \sum_{t=1}^n \sum_{j=1}^m \Delta_j G(t) \frac{2}{m_t}\beta_j^{VP}(t) .
\end{equation*}
\normalsize

\newpage 

\section{Regret bound proof for the UCB soft mortal algorithm}\label{Appendix::UCB_mortal_regret_bound}

\tcbset{colback=blue!2!white}
\begin{tcolorbox}
	\begin{proposition}\label{Proposition::at_least_one_of_three}
		The event
		\begin{equation*}
			A = \left\{ \widehat{X}_{j} + \psi_{\text{future}}(j,t)\sqrt{\frac{2 \log\xi_{\text{present}}(t-s_j)}{T_j(t-1)}} > \widehat{X}_{i^*_{t}} + \psi_{\text{future}}(i^*_{t},t)\sqrt{\frac{2 \log\xi_{\text{present}}(t-s_{i^*_{t}})}{T_{i^*_{t}}(t-1)}}\right\}
		\end{equation*}
		is included in $B \cup C \cup D$, where
		\begin{eqnarray*}
			B &=& \left\{ \widehat{X}_{i^*_{t}} < \mu_{i^*_{t}} - \psi_{\text{future}}(i^*_{t},t)\sqrt{\frac{2 \log\xi_{\text{present}}(t-s_{i^*_{t}})}{T_{i^*_{t}}(t-1)}} \right\}\\
			C &=& \left\{ \widehat{X}_{j} >       \mu_j         + \psi_{\text{future}}(j,t)     \sqrt{\frac{2 \log\xi_{\text{present}}(t-s_j)}{T_j(t-1)}}  \right\}\\
			D &=& \left\{ \mu_{i^*_{t}} - \mu_j < 2\psi_{\text{future}}(j,t)     \sqrt{\frac{2 \log\xi_{\text{present}}(t-s_j)}{T_j(t-1)}} \right\}.
		\end{eqnarray*}
		The inclusion $A \subset (B \cup C \cup D)$ intuitively means that if the algorithm is choosing to play suboptimal arm $j$ at turn $t$, then it is underestimating the best arm available (event $B$), or it is overestimating arm $j$ (event $C$), or it has not pulled enough times arm $j$ to distinguish its performance from the one of arm $i^*_{t}$ (event $D$).
	\end{proposition}
\end{tcolorbox}
For the sake of contradiction let us assume there exists $\omega \in A$ such that $\omega \in (B \cup C \cup D)^{\mathcal{C}}$. Then, for that $\omega$, none of the inequalities that define the events $B$, $C$, and $D$ would hold, i.e. (using, in order, the inequality in $B$, then the one in $D$, then the one in $C$):
\begin{eqnarray*}
	\widehat{X}_{i^*_{t}} &\geq& \mu_{i^*_{t}} - \psi_{\text{future}}(i^*_{t},t)\sqrt{\frac{2 \log\xi_{\text{present}}(t-s_{i^*_{t}})}{T_{i^*_{t}}(t-1)}} \\
	&\geq& \mu_j + 2\psi_{\text{future}}(j,t)     \sqrt{\frac{2 \log\xi_{\text{present}}(t-s_j)}{T_j(t-1)}} - \psi_{\text{future}}(i^*_{t},t)\sqrt{\frac{2 \log\xi_{\text{present}}(t-s_{i^*_{t}})}{T_{i^*_{t}}(t-1)}} \\
	&\geq& \widehat{X}_{j} + \psi_{\text{future}}(j,t)     \sqrt{\frac{2 \log\xi_{\text{present}}(t-s_j)}{T_j(t-1)}} - \psi_{\text{future}}(i^*_{t},t)\sqrt{\frac{2 \log\xi_{\text{present}} (t-s_{i^*_{t}})}{T_{i^*_{t}}(t-1)}} ,
\end{eqnarray*}
which contradicts $\omega \in A$.

\tcbset{colback=blue!2!white}
\begin{tcolorbox}
	\begin{proposition}\label{Proposition::the_third_cannot_hold}
		When
		\begin{equation*}
			T_j(t-1) \geq \left\lceil \frac{8 \psi_{\text{future}}^2(j,t) \log\xi_{\text{present}}(t-s_j)}{\Delta_{j,i^*_{t}}^2} \right\rceil
		\end{equation*}
		event D in Preposition \ref{Proposition::at_least_one_of_three} can not happen.
	\end{proposition}
\end{tcolorbox}
In fact, 
\begin{eqnarray*}
	&&\mu_{i^*_{t}} - \mu_j - 2\psi_{\text{future}}(j,t)     \sqrt{\frac{2 \log\xi_{\text{present}}(t-s_j)}{T_j(t-1)}} \\
	&\geq& \mu_{i^*_{t}} - \mu_j - 2\psi_{\text{future}}(j,t)     \sqrt{\frac{2 \log\xi_{\text{present}}(t-s_j)}{\left\lceil \frac{8 \psi_{\text{future}}^2(j,t) \log\xi_{\text{present}}(t-s_j)}{\Delta_{j,i^*_{t}}^2} \right\rceil}} \\
	&\geq& \mu_{i^*_{t}} - \mu_j - 2\psi_{\text{future}}(j,t)     \sqrt{\frac{ \log\xi_{\text{present}}(t-s_j) \Delta_{j,i^*_{t}}^2}{ 4 \psi_{\text{future}}^2(j,t) \log\xi_{\text{present}}(t-s_j) }} \\
	&=& \mu_{i^*_{t}} - \mu_j - \Delta_{j,i^*_{t}} = 0.
\end{eqnarray*}

\normalsize

The proof of the following theorem follows steps similar to the proof of the UCB-L algorithm presented in \cite{TracaRuYa2019}.

\tcbset{colback=blue!2!white}
\begin{tcolorbox}
	%\begin{theorem}
	{\bf Theorem \ref{Theorem::G_mortal}}\textit{ 
		Let $\bigcup_{z=1}^{E_j}L_j^z$ be a partition of $L_j$ into epochs with different best available arm, $s_j^z$ and $l_j^z$ be the first and last step of epoch $L_j^z$, and for each epoch let $u_{j,z}$ be defined as
		\begin{equation}
			u_{j,z} = \max_{t\in\{s_j^z,\cdots,l_j^z\}}\left\lceil \frac{8 \psi_{\text{future}}(j,t) \log\xi_{\text{present}}(t-s_j)}{\Delta_{j,z}^2}  \right\rceil,
		\end{equation}
		where
		\begin{equation}
			\Delta_{j,i^*_{t}} = \Delta_{j,z} \;\;\text{for}\;t \in L_j^z.
		\end{equation}
		Then, the bound on the mean regret $\E[R_n]$ at time $n$ is given by
		\footnotesize
		\begin{eqnarray*}
			&&\E[R_n] \leq \sum_{j \in M_I}G(j)\Delta_{j,i^*_{t}} +  \sum_{j \in M}\; \sum_{z =1}^{E_j}  \left(\max_{t \in E_j} G(t)\right)\Delta_{j,z}     \min\left(l_j^z-s_j^z \;,\; \beta_j^{M}\right) ,
		\end{eqnarray*}	\normalsize
		where \footnotesize
		\begin{equation}
			\beta_j^{M} =  u_{j,z} + \displaystyle\sum_{\substack{t \in L_j^z \\ t>m_I}}\; (t-s_{i^*_{t}}) (t-s_j-u_{j,z}+1)\left[  \xi_{\text{present}}(t-s_j)^{-\frac{4}{r^2}\psi_{\text{future}}(j,t)} +  \xi_{\text{present}}(t-s_{i^*_{t}})^{-\frac{4}{r^2} \psi_{\text{future}}(i^*_{t},t)}  \right]. 
		\end{equation} \normalsize
	}
	%\end{theorem}
\end{tcolorbox}
\tcbset{colback=white}

\tcbset{colback=white}
\begin{tcolorbox}
	{\bf First step:} Decomposition of $\E[R_n]$.
\end{tcolorbox}
Let us partition the set of steps $L_j$ during which arm $j$ is available into $E_j$ epochs $L_j^z$, such that 
\begin{itemize}
	\item $\bigcup_{z=1}^{E_j}L_j^z = L_j$,
	\item $L_j^{z_1} \cap L_j^{z_2} = \emptyset$ if $z_1 \neq z_2$,
	\item $i_t^* \neq i_s^*$ if $t \in L_j^{z_1}$ and $s\in L_j^{z_2}$ (i.e., if different epochs have different best arm available).
\end{itemize}
Since during the same epoch the best arm available does not change, let us define 
\begin{equation}
	\Delta_{j,i^*_{t}} = \Delta_{j,z} \;\;\text{for}\;t \in L_j^z,
\end{equation}
and $s_j^z = \min{L_j^z}$, $l_j^z = \max{L_j^z}$ the first and last step of epoch $L_j^z$.\\
We have that 
\begin{eqnarray}
	R_n &=& \sum_{j \in M_I}G(j)\Delta_{j,i^*_{t}} +   \sum_{j \in M}\; \sum_{\substack{t \in L_j \\ t>m_I}} G(t) \Delta_{j,i^*_{t}} \ONE{\{t \in I(j)\}} \label{Equation::R_n_to_take_expectation_0}\\
	&\leq& \sum_{j \in M_I}\Delta_{j,i^*_{t}} +   \sum_{j \in M}\; \sum_{z =1}^{E_j}  \left(\max_{t \in E_j} G(t)\right) \Delta_{j,z} \sum_{\substack{t \in L_j^z \\ t>m_I}}  \ONE{\{t \in I(j)\}} \label{Equation::R_n_to_take_expectation}
\end{eqnarray}
Let us call
\begin{equation*}\label{Equation::T_j_total}
	T_j^z(l_j^z) = \sum_{\substack{t \in L_j^z \\ t>m_I}}   \ONE{\{t \in I(j)\}}
\end{equation*}
the total number of times we choose arm $j$ in epoch $z$ during the game (after initialization). Then, by taking the expectation with respect to the policy of \eqref{Equation::R_n_to_take_expectation} we get 
\begin{equation}\label{Equation::ER_n_UCB_mortal}
	\E[R_n] = \sum_{j \in M_I}G(j)\Delta_{j,i^*_{t}} +   \sum_{j \in M}\; \sum_{z =1}^{E_j} \left(\max_{t \in E_j} G(t)\right) \Delta_{j,z}\, \E\left[ T_j^z(l_j^z) \right]. 
\end{equation}
Therefore, finding an upper bound for the expected value of \eqref{Equation::R_n_to_take_expectation_0} can be accomplished by bounding the expected value of $T_j^z(l_j^z)$.
%\tcbset{colback=white}
%\begin{tcolorbox}
%	{\bf First step:} Decomposition of $\E[R_n]$.
%\end{tcolorbox}
%Let us call 
%\[\Delta_{j}^{\max} = \max_{t \in L_j}\Delta_{j,i^*_{t}} \]
%the maximum gap in mean reward between arm $j$ and the best arm appeared during the life of arm $j$. 
%Using the second formulation of the cumulative regret given in \eqref{Equation::cumulative_regret_second_form} we have that 
%\begin{eqnarray}
%	R_n &=& \sum_{j \in M_I}\Delta_{j,i^*_{t}} +   \sum_{j \in M}\; \sum_{\substack{t \in L_j \\ t>m_I}}  \Delta_{j,i^*_{t}} \ONE{\{t \in I(j)\}} \label{Equation::R_n_to_take_expectation_0} \\
%	&\leq& \sum_{j \in M_I}\Delta_{j,i^*_{t}} +   \sum_{j \in M}\; \Delta_{j}^{\max} \sum_{\substack{t \in L_j \\ t>m_I}}   \ONE{\{t \in I(j)\}}. \label{Equation::R_n_to_take_expectation}
%\end{eqnarray}
%Let us call
%\begin{equation*}\label{Equation::T_j_total}
%T_j(s_j+l_j) = \sum_{\substack{t \in L_j \\ t>m_I}}   \ONE{\{t \in I(j)\}}
%\end{equation*}
%the total number of times we choose arm $j$ during the game (after initialization). Then, by taking the expectation of \eqref{Equation::R_n_to_take_expectation} we get 
%\begin{equation}\label{Equation::ER_n_UCB_mortal}
%	\E[R_n] \leq \sum_{j \in M_I}\Delta_{j,i^*_{t}} +   \sum_{j \in M}\; \Delta_{j}^{\max} \E\left[ T_j(s_j+l_j) \right]. 
%\end{equation}
%Therefore, finding an upper bound for \eqref{Equation::R_n_to_take_expectation_0} can be accomplished by bounding the expected value of $T_j(s_j+l_j)$.
\tcbset{colback=white}
\begin{tcolorbox}
	{\bf Second step:} Decomposition of $T_j^z(l_j^z)$.
\end{tcolorbox}
Recall that with $T_j(t-1)$ we indicate the number of times we played arm $j$ before turn $t$ starts.  For any integer $u_{j,z}$, we can write
\footnotesize
\begin{eqnarray*}
	&&T_j^z(l_j^z)  =   u_{j,z} + \displaystyle\sum_{\substack{t \in L_j^z \\ t>m_I}}   \ONE{\{t \in I(j), T_j(t-1) \geq u_{j,z}\}}  \label{step_two}\\
	&&= u_{j,z} \\&&+ \displaystyle\sum_{\substack{t \in L_j^z \\ t>m_I}}  \ONE\left\{ \widehat{X}_{j} + \psi_{\text{fut.}}(j,t)\sqrt{\frac{2 \log\xi_{\text{pres.}}(t-s_j)}{T_{j}(t-1)}} > 
	\widehat{X}_{i^*_{t}} + \psi_{\text{fut.}}(i^*_{t},t)\sqrt{\frac{2 \log\xi_{\text{pres.}}(t-s_{i^*_{t}})}{T_{i^*_{t}}(t-1)}}, T_j(t-1)\geq u_{j,z}\right\}  \label{step_three}\\
	&&\leq  u_{j,z} \\&&+ \displaystyle\sum_{\substack{t \in L_j^z \\ t>m_I}}\; \sum_{k_j = u_{j,z}}^{t-s_j}\; \sum_{k_{i^*_{t}} = 1}^{t-s_{i^*_{t}}} \ONE\left\{ \widehat{X}_{j} + \psi_{\text{fut.}}(j,t)\sqrt{\frac{2 \log\xi_{\text{pres.}}(t-s_j)}{k_j}} > \widehat{X}_{i^*_{t}} + \psi_{\text{fut.}}(i^*_{t},t)\sqrt{\frac{2 \log\xi_{\text{pres.}}(t-s_{i^*_{t}})}{k_{i^*_{t}}}}\right\}.  \label{step_four}
\end{eqnarray*}
\normalsize
Therefore we can find an upper bound for the expectation of $T_j^z(l_j^z)$ by finding an upper bound for the probability of the event $$A = \left\{ \widehat{X}_{j} + \psi_{\text{future}}(j,t)\sqrt{\frac{2 \log\xi_{\text{present}}(t-s_j)}{k_j}} > \widehat{X}_{i^*_{t}} + \psi_{\text{future}}(i^*_{t},t)\sqrt{\frac{2 \log\xi_{\text{present}}(t-s_{i^*_{t}})}{k_{i^*_{t}}}}\right\}.$$
%\begin{tcolorbox}
%	{\bf Second step:} Decomposition of $T_j(s_j+l_j)$.
%\end{tcolorbox}
%Recall that with $T_j(t-1)$ we indicate the number of times we played arm $j$ before turn $t$ starts.  For any integer $u_{j,z}$, we can write
%\footnotesize
%\begin{eqnarray*}
%	T_j(s_j+l_j) & = &  u_{j,z} + \displaystyle\sum_{\substack{t \in L_j \\ t>m_I}}   \ONE{\{t \in I(j), T_j(t-1) \geq u_{j,z}\}}  \label{step_two}\\
%	&= & u_{j,z} + \displaystyle\sum_{\substack{t \in L_j \\ t>m_I}}  \ONE\left\{ \widehat{X}_{j} + \psi_{\text{future}}(j,t)\sqrt{\frac{2 \log\xi_{\text{present}}(t-s_j)}{T_{j}(t-1)}} > 
%	\widehat{X}_{i^*_{t}} + \psi_{\text{future}}(i^*_{t},t)\sqrt{\frac{2 \log\xi_{\text{present}}(t-s_{i^*_{t}})}{T_{i^*_{t}}(t-1)}}, T_j(t-1)\geq u_{j,z}\right\}  \label{step_three}\\
%	&\leq & u_{j,z} + \displaystyle\sum_{\substack{t \in L_j \\ t>m_I}}\; \sum_{k_j = u_{j,z}}^{t-s_j}\; \sum_{k_{i^*_{t}} = 1}^{t-s_{i^*_{t}}} \ONE\left\{ \widehat{X}_{j} + \psi_{\text{future}}(j,t)\sqrt{\frac{2 \log\xi_{\text{present}}(t-s_j)}{k_j}} > \widehat{X}_{i^*_{t}} + \psi_{\text{future}}(i^*_{t},t)\sqrt{\frac{2 \log\xi_{\text{present}}(t-s_{i^*_{t}})}{k_{i^*_{t}}}}\right\}  \label{step_four}
%\end{eqnarray*}\normalsize
%Therefore we can find an upper bound the expectation of $T_j(s_j+l_j)$ by finding an upper bound for the probability of the event $$A = \left\{ \widehat{X}_{j} + \psi_{\text{future}}(j,t)\sqrt{\frac{2 \log\xi_{\text{present}}(t-s_j)}{k_j}} > \widehat{X}_{i^*_{t}} + \psi_{\text{future}}(i^*_{t},t)\sqrt{\frac{2 \log\xi_{\text{present}}(t-s_{i^*_{t}})}{k_{i^*_{t}}}}\right\}$$.
\tcbset{colback=white}
\begin{tcolorbox}
	{\bf Third step:} Upper bound for $\E[T_j^z(l_j^z)]$.
\end{tcolorbox}
By choosing $u_{j,z} = \max_{t\in\{s_j^z,\cdots,l_j^z\}}\left\lceil \frac{8 \psi_{\text{future}}^2(j,t) \log (t-s_j^z)}{\Delta_{j,z}^2} \right\rceil$ and using Proposition \ref{Proposition::at_least_one_of_three} and Proposition \ref{Proposition::the_third_cannot_hold}, we have that, 
\footnotesize
\begin{equation}\label{Equation::inclusion_0}
	A \subset \left(\left\{ \widehat{X}_{i^*_{t}} < \mu_{i^*_{t}} - \psi_{\text{future}}(i^*_{t},t)\sqrt{\frac{2 \log\xi_{\text{present}}(t-s_{i^*_{t}})}{k_{i^*_{t}}}} \right\} \cup \left\{ \widehat{X}_{j} >       \mu_j         + \psi_{\text{future}}(j,t)     \sqrt{\frac{2 \log\xi_{\text{present}}(t-s_j)}{k_j}}  \right\} \right).
\end{equation}
\normalsize
Using Hoeffding's\footnote{{\bf Hoeffding's bound:} Let $X_1, \cdots, X_n$ be r.v. bounded in $[a_i,b_i]$ $\forall i$. Let $\widehat{X} = \frac{1}{n}\sum_{i=1}^n X_i$ and $\mu = \E[\widehat{X}]$. \\Then, $\P\left(\widehat{X} - \mu \geq \varepsilon\right) \leq \exp\left\{ -\frac{2n^2\varepsilon^2}{\sum_{i=1}^n (b_i-a_i)^2 }  \right\}$. \\In our case, $n$ is $k_j$ or $k_{i^*_{t}}$, $b_i - a_i$ is $r$, $\mu$ is $\mu_j$ or $\mu_{i^*_{t}}$, and $\varepsilon$ is  $\psi_{\text{future}}(j,t)     \sqrt{\frac{2 \log\xi_{\text{present}}(t-s_j)}{T_j(t-1)}}$ or $\psi_{\text{future}}(i^*_{t},t)\sqrt{\frac{2 \log\xi_{\text{present}}(t-s_{i^*_{t}})}{T_{i^*_{t}}(t-1)}}$.} bound we have that 
%\footnotesize
%\begin{eqnarray*}
%	&&\P\left(\widehat{X}_{i^*_{t}} < \mu_{i^*_{t}} - \psi_{\text{fut.}}(i^*_{t},t)\sqrt{\frac{2 \log\xi_{\text{pres.}}(t-s_{i^*_{t}})}{T_{i^*_{t}}(t-1)}}\right) \leq 
%	\exp\left\{-\frac{2 k_{i^*_{t}}^2 \psi_{\text{fut.}}^2(i^*_{t},t) \frac{2 \log\xi_{\text{pres.}}(t-s_{i^*_{t}}) }{k_{i^*_{t}}} }{k_{i^*_{t}} r^2} \right\} =  \xi_{\text{pres.}}(t-s_{i^*_{t}})^{-\frac{4}{r^2} \psi_{\text{fut.}}(i^*_{t},t)}\\
%	&&\P\left(\widehat{X}_{j} >       \mu_j         + \psi_{\text{fut.}}(j,t)     \sqrt{\frac{2 \log\xi_{\text{pres.}}(t-s_j)}{T_j(t-1)}}\right) \leq 
%	\exp\left\{-\frac{2 k_j^2 \psi_{\text{fut.}}^2(j,t) \frac{2 \log\xi_{\text{pres.}}(t-s_j) }{k_j} }{k_j r^2} \right\} =  \xi_{\text{pres.}}(t-s_j)^{-\frac{4}{r^2}\psi_{\text{fut.}}(j,t)}.
%\end{eqnarray*}
%\normalsize
\small
\begin{eqnarray*}
	\P\left(\widehat{X}_{i^*_{t}} < \mu_{i^*_{t}} - \psi_{\text{fut.}}(i^*_{t},t)\sqrt{\frac{2 \log\xi_{\text{pres.}}(t-s_{i^*_{t}})}{T_{i^*_{t}}(t-1)}}\right) &\leq& 
	\exp\left\{-\frac{2 k_{i^*_{t}}^2 \psi_{\text{fut.}}^2(i^*_{t},t) \frac{2 \log\xi_{\text{pres.}}(t-s_{i^*_{t}}) }{k_{i^*_{t}}} }{k_{i^*_{t}} r^2} \right\} 
	\\
	&=&  \xi_{\text{pres.}}(t-s_{i^*_{t}})^{-\frac{4}{r^2} \psi_{\text{fut.}}(i^*_{t},t)}
\end{eqnarray*}
\begin{eqnarray*}
	\P\left(\widehat{X}_{j} >       \mu_j         + \psi_{\text{fut.}}(j,t)     \sqrt{\frac{2 \log\xi_{\text{pres.}}(t-s_j)}{T_j(t-1)}}\right) 
	&\leq& 
	\exp\left\{-\frac{2 k_j^2 \psi_{\text{fut.}}^2(j,t) \frac{2 \log\xi_{\text{pres.}}(t-s_j) }{k_j} }{k_j r^2} \right\}\\
	&=&  \xi_{\text{pres.}}(t-s_j)^{-\frac{4}{r^2}\psi_{\text{fut.}}(j,t)}.
\end{eqnarray*}
\normalsize
Using the inclusion in \eqref{Equation::inclusion_0} in combination with Hoeffding's bounds, we have that 
\footnotesize
\begin{eqnarray}
	&&\E\left[T_j^z(l_j^z)\right] \leq u_{j,z} \nonumber\\
	&&+ \displaystyle\sum_{\substack{t \in L_j^z \\ t>m_I}}\; \sum_{k_j = u_{j,z}}^{l_j}\; \sum_{k_{i^*_{t}} = 1}^{t-s_{i^*_{t}}} \P\left\{ \widehat{X}_{j} + \psi_{\text{fut.}}(j,t)\sqrt{\frac{2 \log\xi_{\text{pres.}}(t-s_j)}{k_j}} > \widehat{X}_{i^*_{t}} + \psi_{\text{fut.}}(i^*_{t},t)\sqrt{\frac{2 \log\xi_{\text{pres.}}(t-s_{i^*_{t}})}{k_{i^*_{t}}}}\right\}\nonumber\\
	&&\leq u_{j,z} + \displaystyle\sum_{\substack{t \in L_j^z \\ t>m_I}}\; \sum_{k_j = u_{j,z}}^{t-s_j}\; \sum_{k_{i^*_{t}} = 1}^{t-s_{i^*_{t}}} \left[\xi_{\text{pres.}}(t-s_{i^*_{t}})^{-\frac{4}{r^2} \psi_{\text{fut.}}(i^*_{t},t)} + \xi_{\text{pres.}}(t-s_j)^{-\frac{4}{r^2}\psi_{\text{fut.}}(j,t)}\right]\nonumber\\
	&&=u_{j,z} + \displaystyle\sum_{\substack{t \in L_j^z \\ t>m_I}}\; (t-s_{i^*_{t}}) (t-s_j-u_{j,z}+1)\left[  \xi_{\text{pres.}}(t-s_j)^{-\frac{4}{r^2}\psi_{\text{fut.}}(j,t)} +  \xi_{\text{pres.}}(t-s_{i^*_{t}})^{-\frac{4}{r^2} \psi_{\text{fut.}}(i^*_{t},t)}  \right]. \label{Equation::bound_on_ET_j_UCB_mortal_0}
\end{eqnarray}
\normalsize
\tcbset{colback=white}
\begin{tcolorbox}
	{\bf Fourth step:} Upper bound for $\E[R_n]$.
\end{tcolorbox}
Combining \eqref{Equation::bound_on_ET_j_UCB_mortal_0} with \eqref{Equation::ER_n_UCB_mortal} we get that the bound on the cumulative regret is given by\footnotesize
\begin{eqnarray*}
	\E[R_n] &\leq& \sum_{j \in M_I}G(j)\Delta_{j,i^*_{t}} +  \sum_{j \in M}\; \sum_{z =1}^{E_j}  \left(\max_{t \in E_j} G(t)\right)\Delta_{j,z}  \times   \min\big(l_j^z-s_j^z \;,\; \\&&u_{j,z} + \displaystyle\sum_{\substack{t \in L_j^z \\ t>m_I}}\; (t-s_{i^*_{t}}) (t-s_j-u_{j,z}+1)\left[  \xi_{\text{pres.}}(t-s_j)^{-\frac{4}{r^2}\psi_{\text{fut.}}(j,t)} +  \xi_{\text{pres.}}(t-s_{i^*_{t}})^{-\frac{4}{r^2} \psi_{\text{fut.}}(i^*_{t},t)}  \right] \big) . 
\end{eqnarray*}\normalsize

Notice that if $\psi_{\text{future}}(j,t) = 1$ , $G(t) =1$,  $s_j = 0$ and $l_j > n$ $\forall j, t$, you can recover the bound of the standard UCB algorithm used in the stochastic case.
(Note that you should use $P>2$ instead of $2$ when $r$ is not $1$ to create the UCB.)

\newpage 

%\newpage
\section{Experiments in a simulated environment}\label{Appendix::Experiments}

% description - table - [H]

\begin{figure}[]%
	\makebox[\textwidth][c]{ %to center figures!
		\begin{subfloatrow}
			\subfloat[\small{The \emph{Wave Greed}}]{{\includegraphics[width=4.0cm,height=4.5cm]{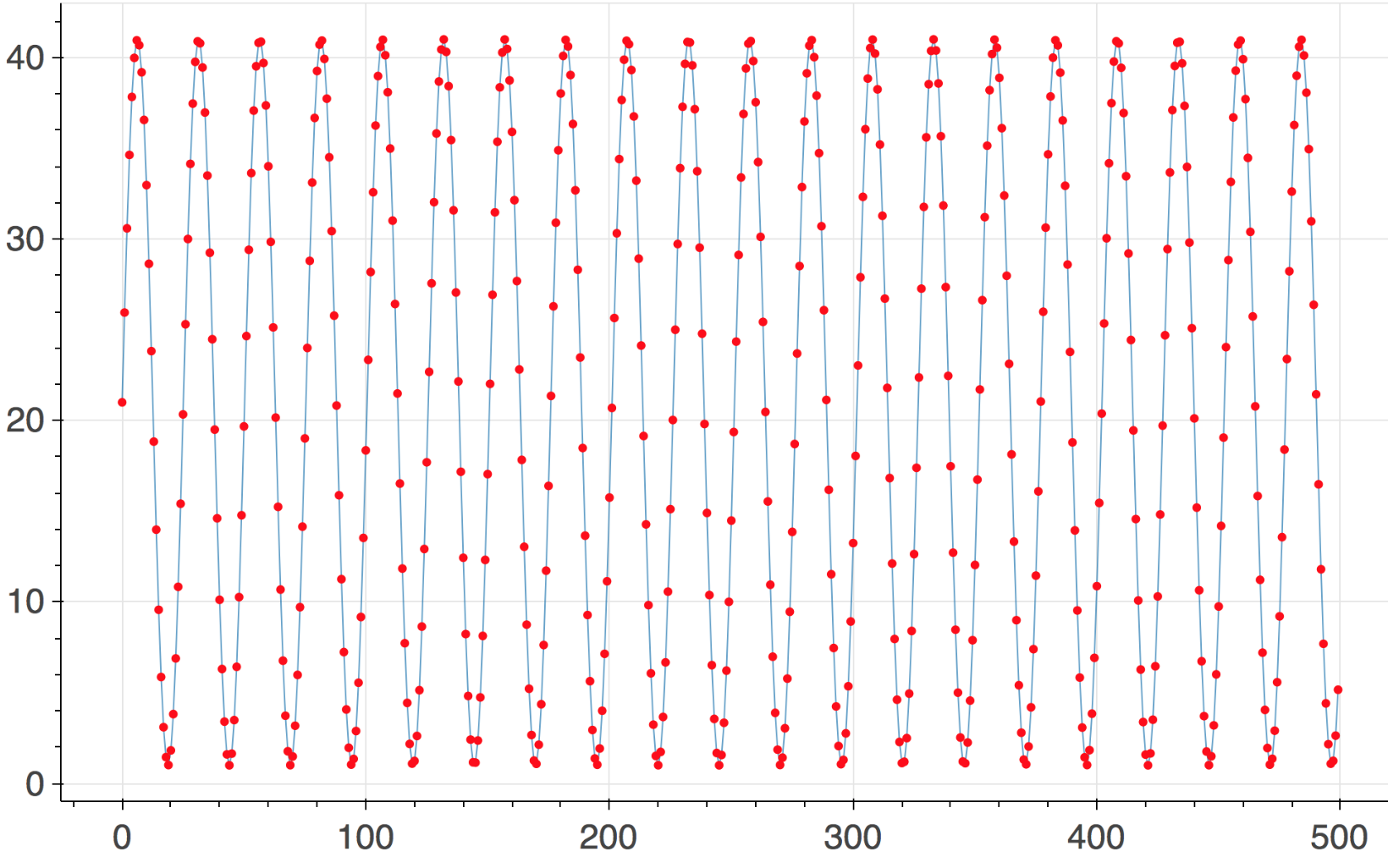} }\label{Figure::Wave_greed}}%
			%\;\;
			\qquad
			\subfloat[\small{The \emph{Christmas Greed}}]{{\includegraphics[width=4.0cm,height=4.5cm]{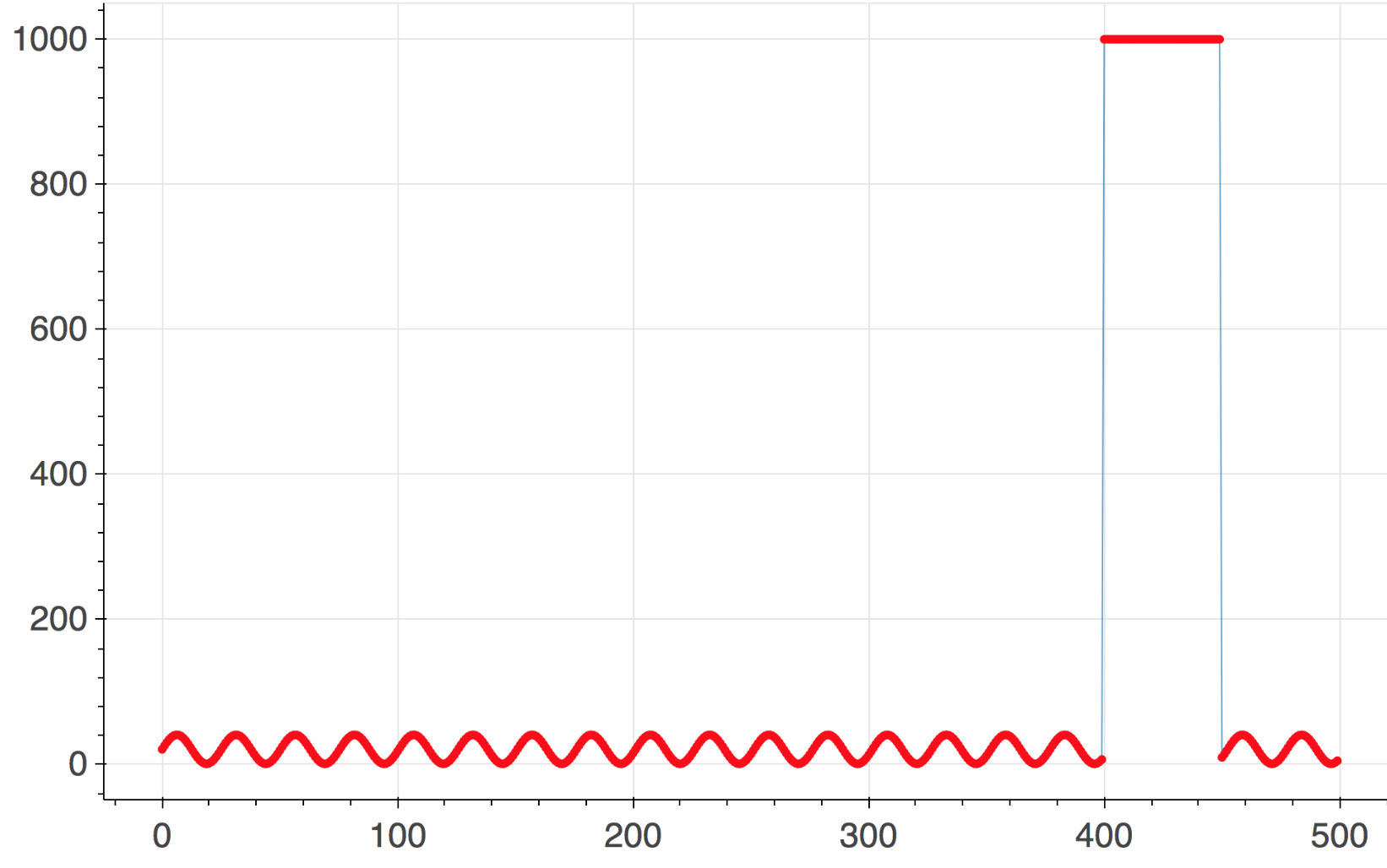} }\label{Figure::Christmas_greed}}%\;\;
			\qquad
			\subfloat[\small{The \emph{Step Greed}}]{{\includegraphics[width=4.0cm,height=4.5cm]{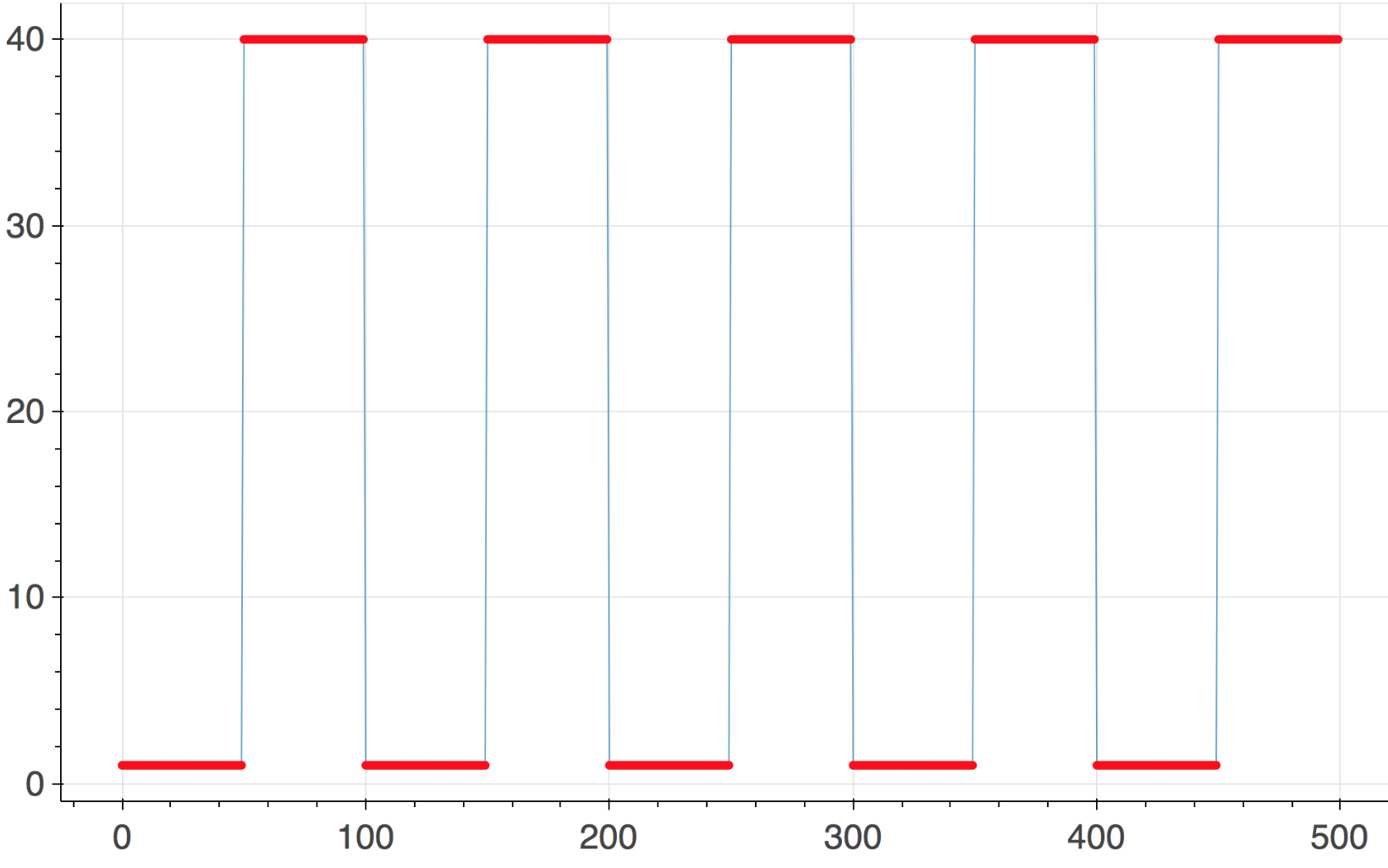} }\label{Figure::Step_greed}}%\;\;
		\end{subfloatrow}
	}
	\caption{Shapes of the multiplier functions used in the experiments.}\label{Figure::appendix_greed_functions}
\end{figure}

We present plots on the average final rewards in different settings %(over 50 games each consisting of 1500 turns)
for each type of the greed functions introduced in Section \ref{Section::Experimental_Results} (Wave, Step, and Christmas) and shown again in Figure \ref{Figure::appendix_greed_functions}. 

The red part of the bars indicate the portion of the final cumulative rewards coming from ``pure exploitation''. The definition of ``pure exploitation'' depends on the algorithm used:
\begin{itemize}
	\item pure exploitation in $\varepsilon$-greedy algorithms: when the algorithms decide to exploit. This is forced in algorithms with threshold when the greed function is above that threshold;
	\item pure exploitation in UCB algorithms: when the arm played has the highest estimated mean reward. This is forced in algorithms with threshold when the greed function is above that threshold;
	\item pure exploitation in the variable-pool algorithm: when the pool size is 1.
\end{itemize}

The rewards are generated from Bernoulli distributions (each arm has probability of success drawn from a Uniform distribution in $[0,1]$) or from Truncated-Normal distributions (each arm has mean drawn from a Uniform distribution in $[0,1]$ and standard deviation equal to $1$, and the rewards are bounded in $[0,1]$).
The list of figures regarding experiments in various settings that used a Wave-type greed function is reported in Table \ref{Table::experiments_Wave}, the list of the ones using a Step-type greed function is reported in Table \ref{Table::experiments_Step}, and the list of the ones using a Christmas-type greed function is reported in Table \ref{Table::experiments_Christmas}.

\begin{figure}[]
	\caption{Average increase in rewards (coming from Bernoulli distributions) compared to the (smarter) version of the $\varepsilon$-greedy algorithm (Algorithm \ref{Algorithm::UCB_slightly_smarter}) with a Wave-type greed function.}
	\centering
	\includegraphics[width=0.95\textwidth]{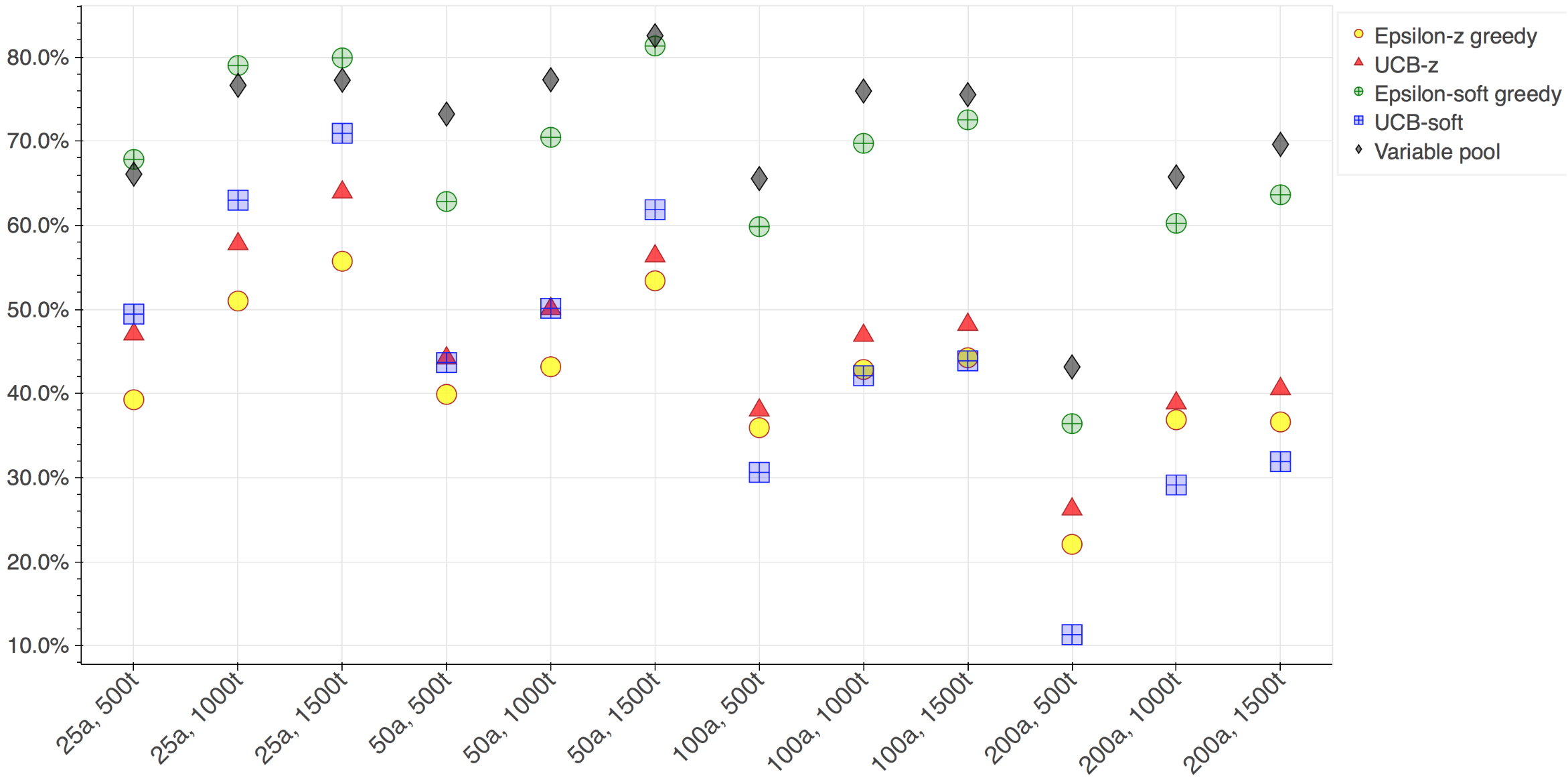}
	\label{BEW}
\end{figure}

We also report the average increase in rewards with respect to the (smarter) versions of the standard $\varepsilon$-greedy algorithm (Algorithm \ref{Algorithm::epsilon_slightly_smarter}) and the standard UCB algorithm (Algorithm \ref{Algorithm::UCB_slightly_smarter}). The list of figures that show the average percentage increase in rewards is reported in Table \ref{Table::increase_percentage}. In these plots, on the $x$-axis there are different combinations of number of arms and number of turns played in each game (for example, $100a, 1500t$ means that there were $100$ arms and the game had $1500$ turns.) On the $y$-axis is reported the average increase in rewards for each of the algorithms that regulate greed over time. For example, each dot in Figure \ref{BEW} represents the percentage increase of rewards for a regulating greed over time algorithm compared with the standard $\varepsilon$-greedy algorithm. The reward multiplier function is Wave-type greed function. For instance, the green circle in the upper left indicates that there was a 68\% increase in rewards of Soft $\varepsilon$-greedy as compared with smarter $\varepsilon$-greedy for a simulation with 25 arms and 500 rounds. The plot shows that across different numbers of arms, across different rounds, the algorithms that regulate greed over time generally increase rewards by anywhere between 10 and 80\%.

\begin{table}[]
	\centering
	\caption{List of the figures showing the final cumulative rewards when using a Wave-type greed function.}
	\label{Table::experiments_Wave}
	\begin{tabular}{ | c | c | c | c | c | } 
		\hline
		{\bf Figure} & {\bf Number of arms} & {\bf Number of turns} & {\bf Reward distribution} \\ 
		\hline
		\ref{Bernoulli_Wave_25a_500t} & 25 & 500 & Bernoulli \\ \hline 
		\ref{Truncated_Normal_Wave_25a_500t} & 25 & 500 & Truncated Normal \\ \hline 
		\ref{Bernoulli_Wave_50a_500t} & 50 & 500 & Bernoulli \\ \hline 
		\ref{Truncated_Normal_Wave_50a_500t} & 50 & 500 & Truncated Normal \\ \hline 
		\ref{Bernoulli_Wave_100a_500t} & 100 & 500 & Bernoulli \\ \hline 
		\ref{Truncated_Normal_Wave_100a_500t} & 100 & 500 & Truncated Normal \\ \hline 
		\ref{Bernoulli_Wave_200a_500t} & 200 & 500 & Bernoulli \\ \hline 
		\ref{Truncated_Normal_Wave_200a_500t} & 200 & 500 & Truncated Normal \\ \hline 
		\ref{Bernoulli_Wave_25a_1000t} & 25 & 1000 & Bernoulli \\ \hline 
		\ref{Truncated_Normal_Wave_25a_1000t} & 25 & 1000 & Truncated Normal \\ \hline 
		\ref{Bernoulli_Wave_50a_1000t} & 50 & 1000 & Bernoulli \\ \hline 
		\ref{Truncated_Normal_Wave_50a_1000t} & 50 & 1000 & Truncated Normal \\ \hline 
		\ref{Bernoulli_Wave_100a_1000t} & 100 & 1000 & Bernoulli \\ \hline 
		\ref{Truncated_Normal_Wave_100a_1000t} & 100 & 1000 & Truncated Normal \\ \hline 
		\ref{Bernoulli_Wave_200a_1000t} & 200 & 1000 & Bernoulli \\ \hline 
		\ref{Truncated_Normal_Wave_200a_1000t} & 200 & 1000 & Truncated Normal \\ \hline 
		\ref{Bernoulli_Wave_25a_1500t} & 25 & 1500 & Bernoulli \\ \hline 
		\ref{Truncated_Normal_Wave_25a_1500t} & 25 & 1500 & Truncated Normal \\ \hline 
		\ref{Bernoulli_Wave_50a_1500t} & 50 & 1500 & Bernoulli \\ \hline 
		\ref{Truncated_Normal_Wave_50a_1500t} & 50 & 1500 & Truncated Normal \\ \hline 
		\ref{Bernoulli_Wave_100a_1500t} & 100 & 1500 & Bernoulli \\ \hline 
		\ref{Truncated_Normal_Wave_100a_1500t} & 100 & 1500 & Truncated Normal \\ \hline 
		\ref{Bernoulli_Wave_200a_1500t} & 200 & 1500 & Bernoulli \\ \hline 
		\ref{Truncated_Normal_Wave_200a_1500t} & 200 & 1500 & Truncated Normal \\ 
		\hline
	\end{tabular}
\end{table}

\begin{table}[]
	\centering
	\caption{List of the figures showing the final cumulative rewards when using a Step-type greed function.}
	\label{Table::experiments_Step}
	\begin{tabular}{ | c | c | c | c | c | } 
		\hline
		{\bf Figure} & {\bf Number of arms} & {\bf Number of turns} & {\bf Reward distribution} \\ 
		\hline
		\ref{Bernoulli_Step_25a_500t} & 25 & 500 & Bernoulli \\ \hline 
		\ref{Truncated_Normal_Step_25a_500t} & 25 & 500 & Truncated Normal \\ \hline 
		\ref{Bernoulli_Step_50a_500t} & 50 & 500 & Bernoulli \\ \hline 
		\ref{Truncated_Normal_Step_50a_500t} & 50 & 500 & Truncated Normal \\ \hline 
		\ref{Bernoulli_Step_100a_500t} & 100 & 500 & Bernoulli \\ \hline 
		\ref{Truncated_Normal_Step_100a_500t} & 100 & 500 & Truncated Normal \\ \hline 
		\ref{Bernoulli_Step_200a_500t} & 200 & 500 & Bernoulli \\ \hline 
		\ref{Truncated_Normal_Step_200a_500t} & 200 & 500 & Truncated Normal \\ \hline 
		\ref{Bernoulli_Step_25a_1000t} & 25 & 1000 & Bernoulli \\ \hline 
		\ref{Truncated_Normal_Step_25a_1000t} & 25 & 1000 & Truncated Normal \\ \hline 
		\ref{Bernoulli_Step_50a_1000t} & 50 & 1000 & Bernoulli \\ \hline 
		\ref{Truncated_Normal_Step_50a_1000t} & 50 & 1000 & Truncated Normal \\ \hline 
		\ref{Bernoulli_Step_100a_1000t} & 100 & 1000 & Bernoulli \\ \hline 
		\ref{Truncated_Normal_Step_100a_1000t} & 100 & 1000 & Truncated Normal \\ \hline 
		\ref{Bernoulli_Step_200a_1000t} & 200 & 1000 & Bernoulli \\ \hline 
		\ref{Truncated_Normal_Step_200a_1000t} & 200 & 1000 & Truncated Normal \\ \hline 
		\ref{Bernoulli_Step_25a_1500t} & 25 & 1500 & Bernoulli \\ \hline 
		\ref{Truncated_Normal_Step_25a_1500t} & 25 & 1500 & Truncated Normal \\ \hline 
		\ref{Bernoulli_Step_50a_1500t} & 50 & 1500 & Bernoulli \\ \hline 
		\ref{Truncated_Normal_Step_50a_1500t} & 50 & 1500 & Truncated Normal \\ \hline 
		\ref{Bernoulli_Step_100a_1500t} & 100 & 1500 & Bernoulli \\ \hline 
		\ref{Truncated_Normal_Step_100a_1500t} & 100 & 1500 & Truncated Normal \\ \hline 
		\ref{Bernoulli_Step_200a_1500t} & 200 & 1500 & Bernoulli \\ \hline 
		\ref{Truncated_Normal_Step_200a_1500t} & 200 & 1500 & Truncated Normal \\ 
		\hline
	\end{tabular}
\end{table}

\begin{table}[]
	\centering
	\caption{List of the figures showing the final cumulative rewards when using a Christmas-type greed function.}
	\label{Table::experiments_Christmas}
	\begin{tabular}{ | c | c | c | c | c | } 
		\hline
		{\bf Figure} & {\bf Number of arms} & {\bf Number of turns} & {\bf Reward distribution} \\ 
		\hline
		\ref{Bernoulli_Christmas_25a_500t} & 25 & 500 & Bernoulli \\ \hline 
		\ref{Truncated_Normal_Christmas_25a_500t} & 25 & 500 & Truncated Normal \\ \hline 
		\ref{Bernoulli_Christmas_50a_500t} & 50 & 500 & Bernoulli \\ \hline 
		\ref{Truncated_Normal_Christmas_50a_500t} & 50 & 500 & Truncated Normal \\ \hline 
		\ref{Bernoulli_Christmas_100a_500t} & 100 & 500 & Bernoulli \\ \hline 
		\ref{Truncated_Normal_Christmas_100a_500t} & 100 & 500 & Truncated Normal \\ \hline 
		\ref{Bernoulli_Christmas_200a_500t} & 200 & 500 & Bernoulli \\ \hline 
		\ref{Truncated_Normal_Christmas_200a_500t} & 200 & 500 & Truncated Normal \\ \hline 
		\ref{Bernoulli_Christmas_25a_1000t} & 25 & 1000 & Bernoulli \\ \hline 
		\ref{Truncated_Normal_Christmas_25a_1000t} & 25 & 1000 & Truncated Normal \\ \hline 
		\ref{Bernoulli_Christmas_50a_1000t} & 50 & 1000 & Bernoulli \\ \hline 
		\ref{Truncated_Normal_Christmas_50a_1000t} & 50 & 1000 & Truncated Normal \\ \hline 
		\ref{Bernoulli_Christmas_100a_1000t} & 100 & 1000 & Bernoulli \\ \hline 
		\ref{Truncated_Normal_Christmas_100a_1000t} & 100 & 1000 & Truncated Normal \\ \hline 
		\ref{Bernoulli_Christmas_200a_1000t} & 200 & 1000 & Bernoulli \\ \hline 
		\ref{Truncated_Normal_Christmas_200a_1000t} & 200 & 1000 & Truncated Normal \\ \hline 
		\ref{Bernoulli_Christmas_25a_1500t} & 25 & 1500 & Bernoulli \\ \hline 
		\ref{Truncated_Normal_Christmas_25a_1500t} & 25 & 1500 & Truncated Normal \\ \hline 
		\ref{Bernoulli_Christmas_50a_1500t} & 50 & 1500 & Bernoulli \\ \hline 
		\ref{Truncated_Normal_Christmas_50a_1500t} & 50 & 1500 & Truncated Normal \\ \hline 
		\ref{Bernoulli_Christmas_100a_1500t} & 100 & 1500 & Bernoulli \\ \hline 
		\ref{Truncated_Normal_Christmas_100a_1500t} & 100 & 1500 & Truncated Normal \\ \hline 
		\ref{Bernoulli_Christmas_200a_1500t} & 200 & 1500 & Bernoulli \\ \hline 
		\ref{Truncated_Normal_Christmas_200a_1500t} & 200 & 1500 & Truncated Normal \\ 
		\hline
	\end{tabular}
\end{table}

\begin{table}[]
	\centering
	\caption{List of the figures showing the average increase in rewards with respect to the (smarter) versions of the standard $\varepsilon$-greedy algorithm (Algorithm \ref{Algorithm::epsilon_slightly_smarter}) and the standard UCB algorithm (Algorithm \ref{Algorithm::UCB_slightly_smarter}).}
	\label{Table::increase_percentage}
	\begin{tabular}{ | c | c | c | c | } 
		\hline
		{\bf Figure} & {\bf Greed function} & {\bf Baseline} & {\bf Reward distribution} \\ 
		\hline
		\ref{BEW} & Wave & $\varepsilon$-greedy & Bernoulli \\ \hline
		\ref{BUW} & Wave & UCB & Bernoulli \\ \hline  
		\ref{NEW} & Wave & $\varepsilon$-greedy & Truncated Normal \\ \hline
		\ref{NUW} & Wave & UCB & Truncated Normal \\ \hline  
		\ref{BES} & Step & $\varepsilon$-greedy & Bernoulli \\ \hline
		\ref{BUS} & Step & UCB & Bernoulli \\ \hline  
		\ref{NES} & Step & $\varepsilon$-greedy & Truncated Normal \\ \hline
		\ref{NUS} & Step & UCB & Truncated Normal \\ \hline  
		\ref{BEC} & Christmas & $\varepsilon$-greedy & Bernoulli \\ \hline
		\ref{BUC} & Christmas & UCB & Bernoulli \\ \hline  
		\ref{NEC} & Christmas & $\varepsilon$-greedy & Truncated Normal \\ \hline
		\ref{NUC} & Christmas & UCB & Truncated Normal \\ 
		\hline
	\end{tabular}
\end{table}

%The average is computed over 50 games for each type of greed functions (Christmas, Step, and Wave) and for rewards coming from Bernoulli distributions (each arm has probability of success drawn from a Uniform distribution in $[0,1]$) or from Truncated-Normal distributions (each arm has mean drawn from a Uniform distribution in $[0,1]$ and standard deviation equal to $1$, and the rewards are bounded in $[0,1]$). On the $x$-axis there are different combinations of number of arms and number of turns played in each game (for example, $100a, 1500t$ means that there were $100$ arms and the game had $1500$ turns.) On the $y$-axis is reported the average increase in rewards for each of the algorithms that regulate greed over time.
\clearpage
%%\subsection{Wave-type greed function with 500 turns per game}

\begin{figure}[]%
	\makebox[\textwidth][c]{ %to center figures!
		\begin{subfloatrow}
			\subfloat[\small{Rewards from Bernoulli distributions. }]{{\includegraphics[width=6.7cm,height=4.0cm]{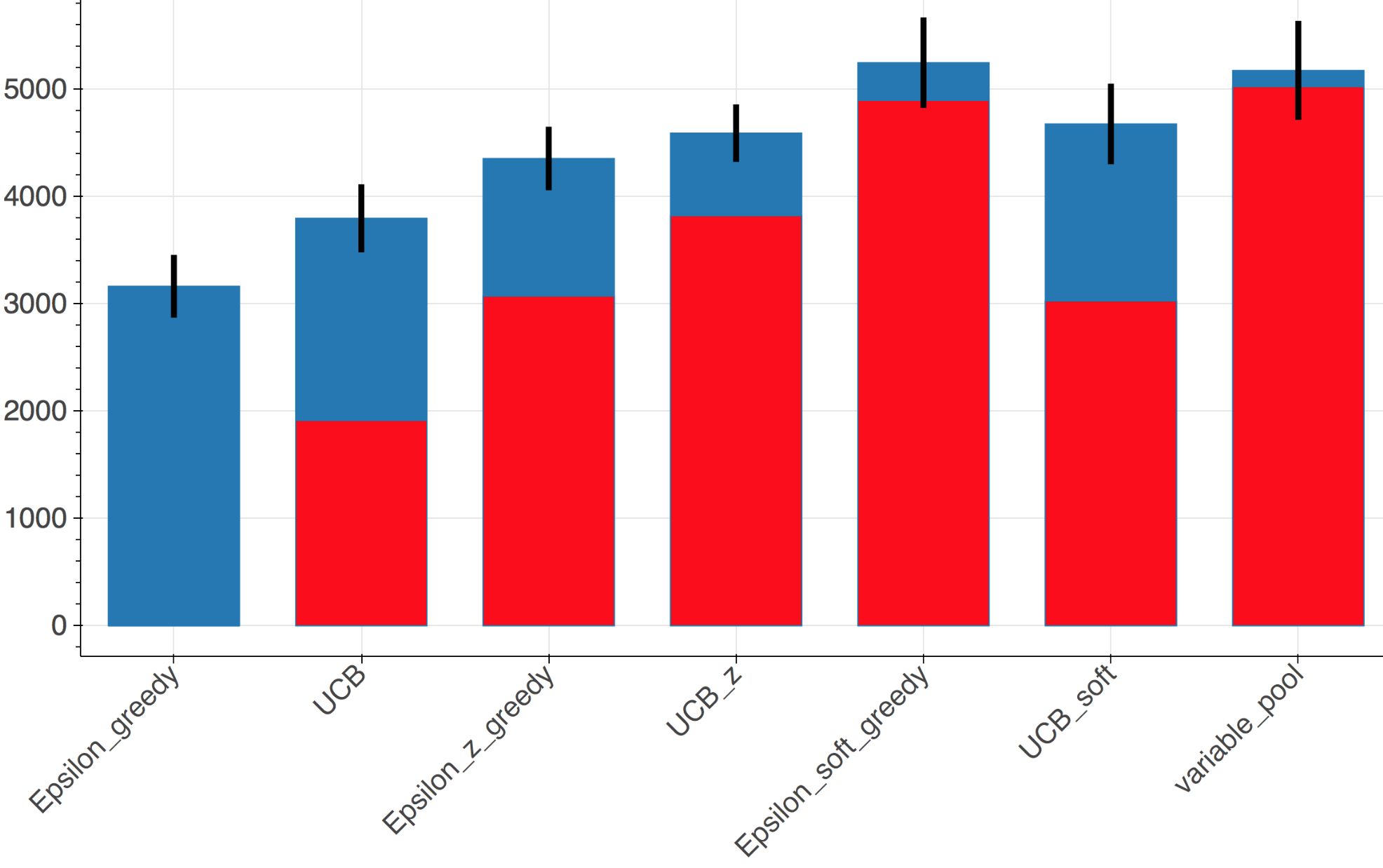} }\label{Bernoulli_Wave_25a_500t}}%
			\;\;
			\qquad
			\subfloat[\small{Rewards from truncated Normal distributions.
			}]{{\includegraphics[width=6.7cm,height=4.0cm]{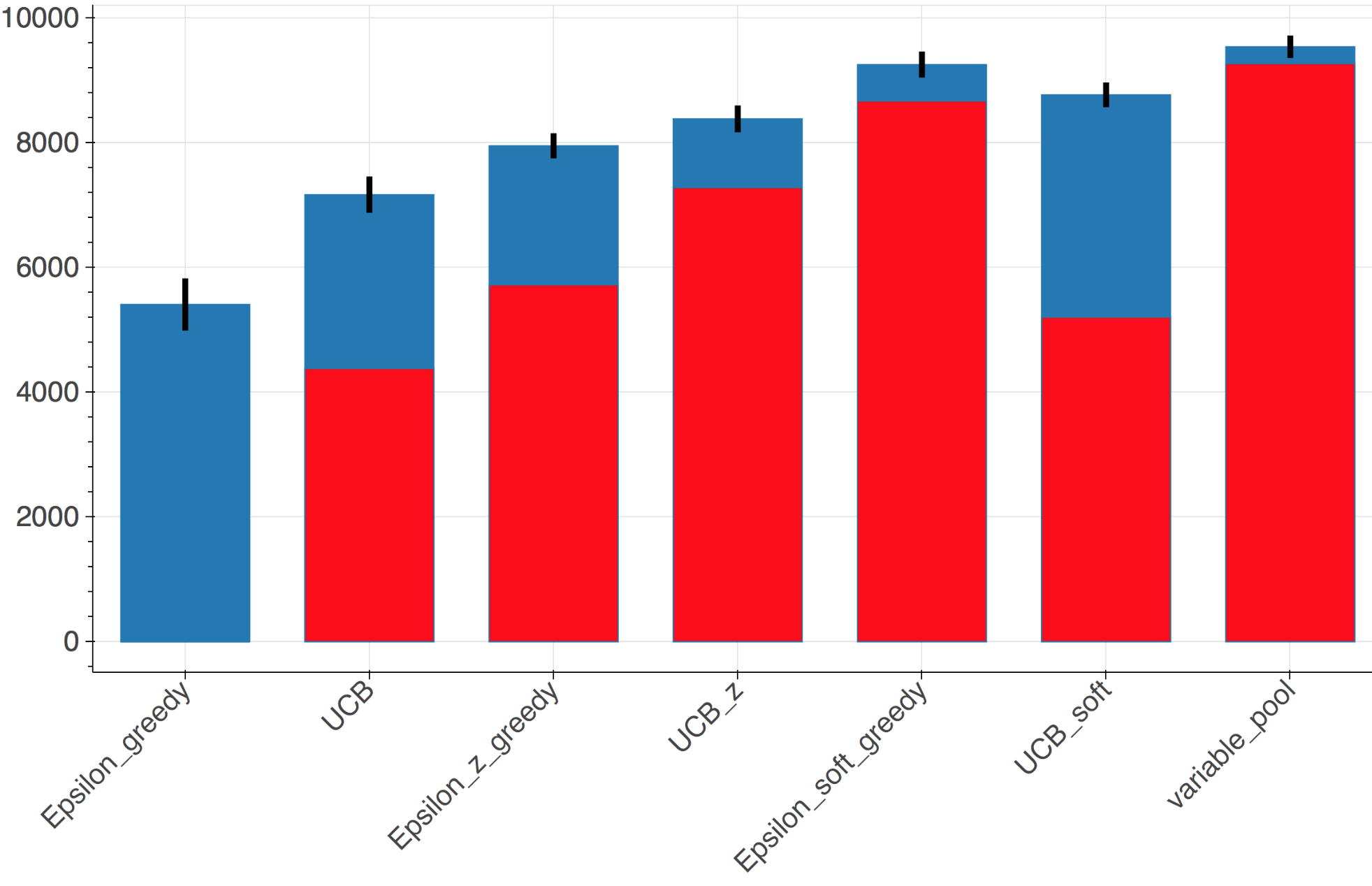} }\label{Truncated_Normal_Wave_25a_500t}}\;\;
		\end{subfloatrow}
	}
	\caption{Comparison of average final rewards in games with 25 arms, 500 turns, and a Wave-type greed function.}\label{Figure::25a_500t_Wave}
\end{figure}

\begin{figure}[]%
	\makebox[\textwidth][c]{ %to center figures!
		\begin{subfloatrow}
			\subfloat[\small{Rewards from Bernoulli distributions. }]{{\includegraphics[width=6.7cm,height=4.0cm]{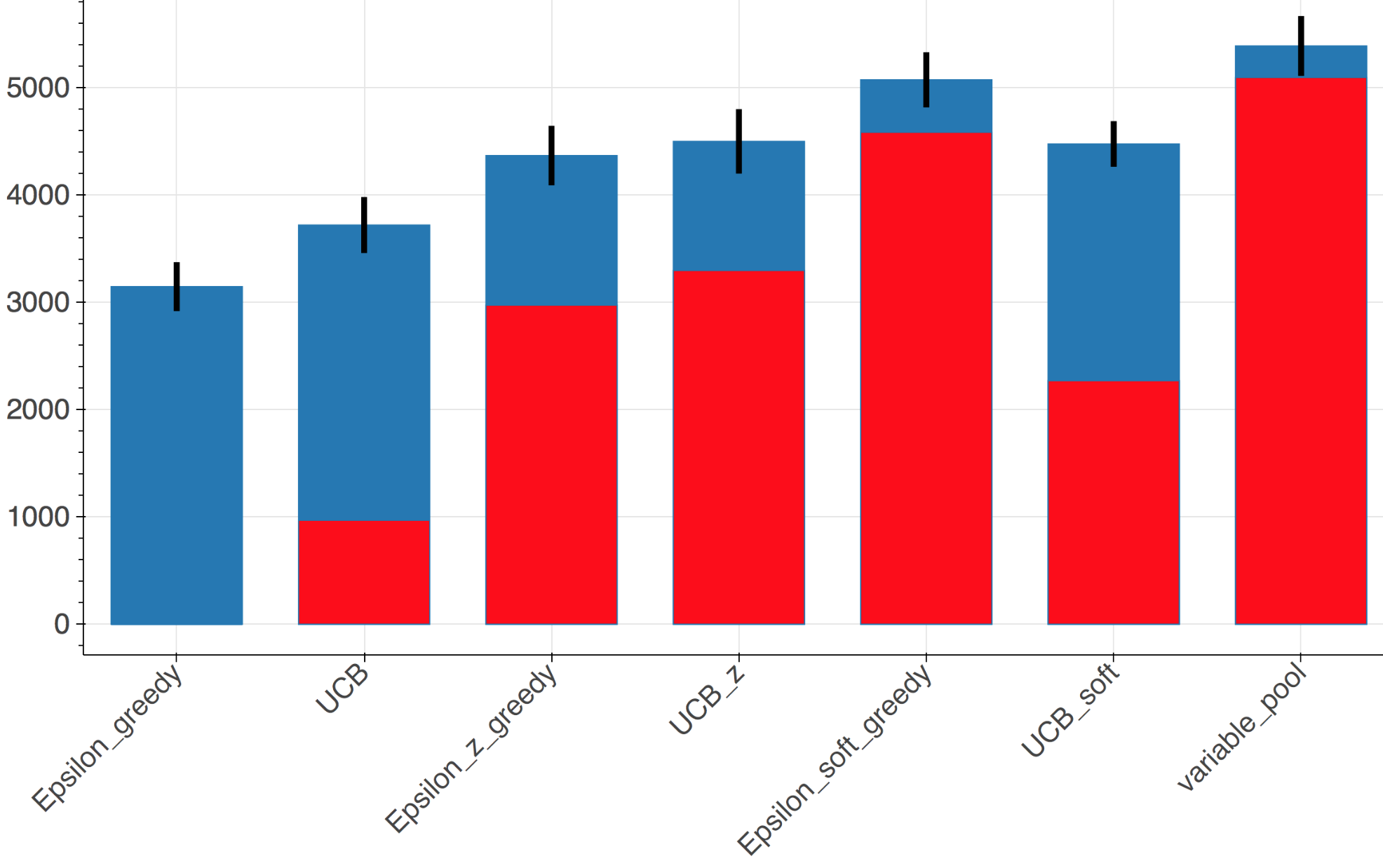} }\label{Bernoulli_Wave_50a_500t}}%
			\;\;
			\qquad
			\subfloat[\small{Rewards from truncated Normal distributions.
			}]{{\includegraphics[width=6.7cm,height=4.0cm]{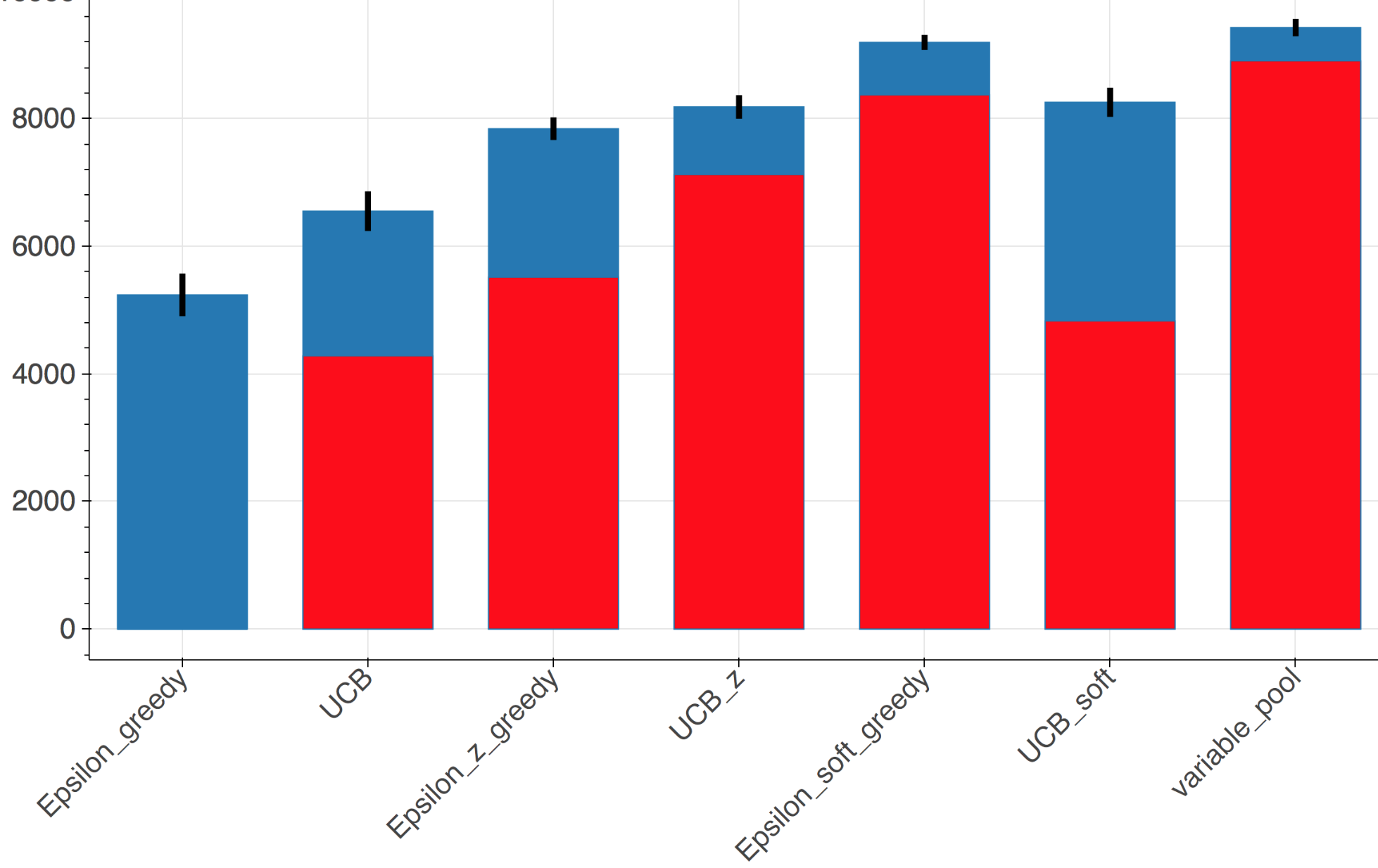} }\label{Truncated_Normal_Wave_50a_500t}}\;\;
		\end{subfloatrow}
	}
	\caption{Comparison of average final rewards in games with 50 arms, 500 turns, and a Wave-type greed function.}\label{Figure::50a_500t_Wave}
\end{figure}

\begin{figure}[]%
	\makebox[\textwidth][c]{ %to center figures!
		\begin{subfloatrow}
			\subfloat[\small{Rewards from Bernoulli distributions. }]{{\includegraphics[width=6.7cm,height=4.0cm]{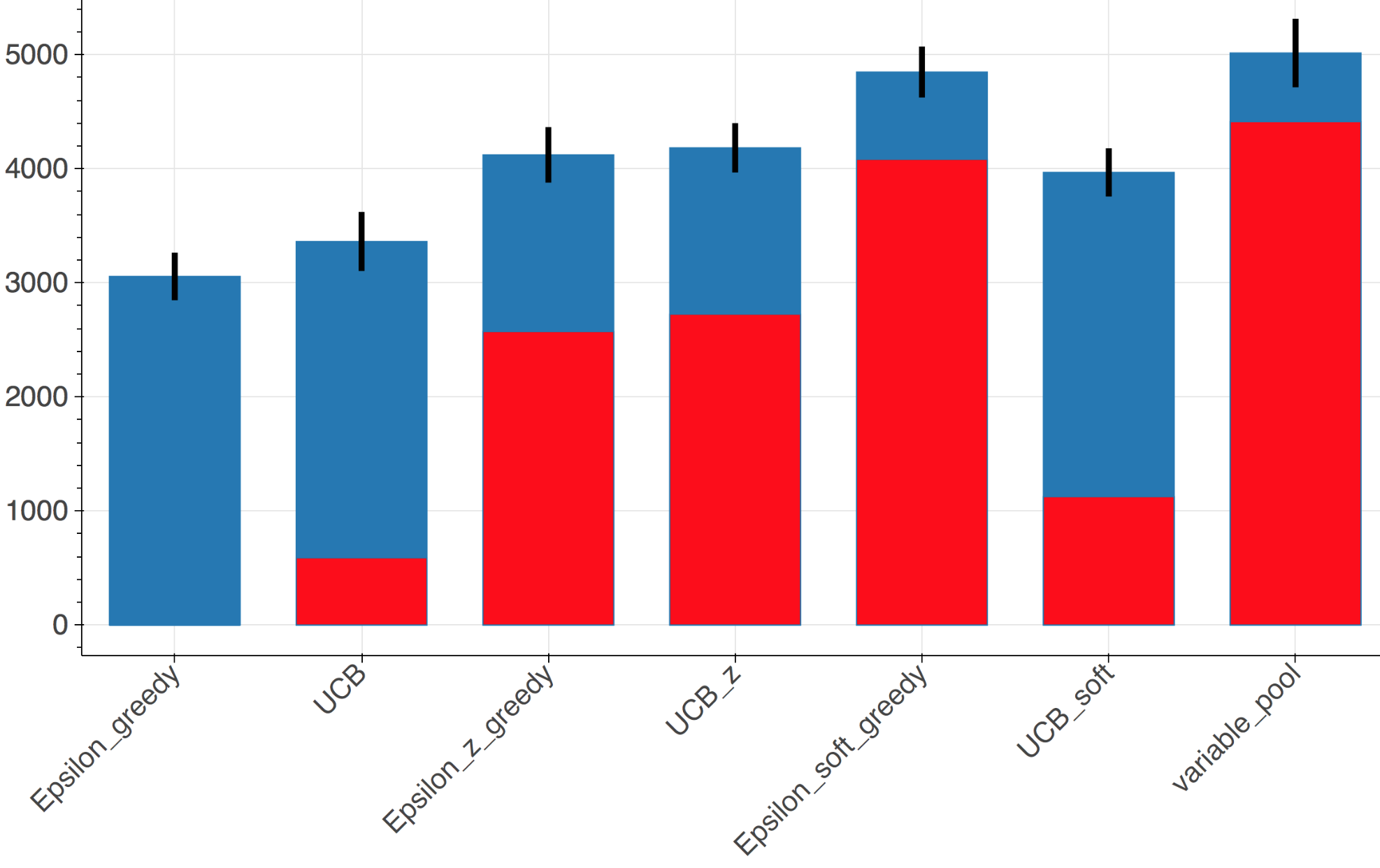} }\label{Bernoulli_Wave_100a_500t}}%
			\;\;
			\qquad
			\subfloat[\small{Rewards from truncated Normal distributions.
			}]{{\includegraphics[width=6.7cm,height=4.0cm]{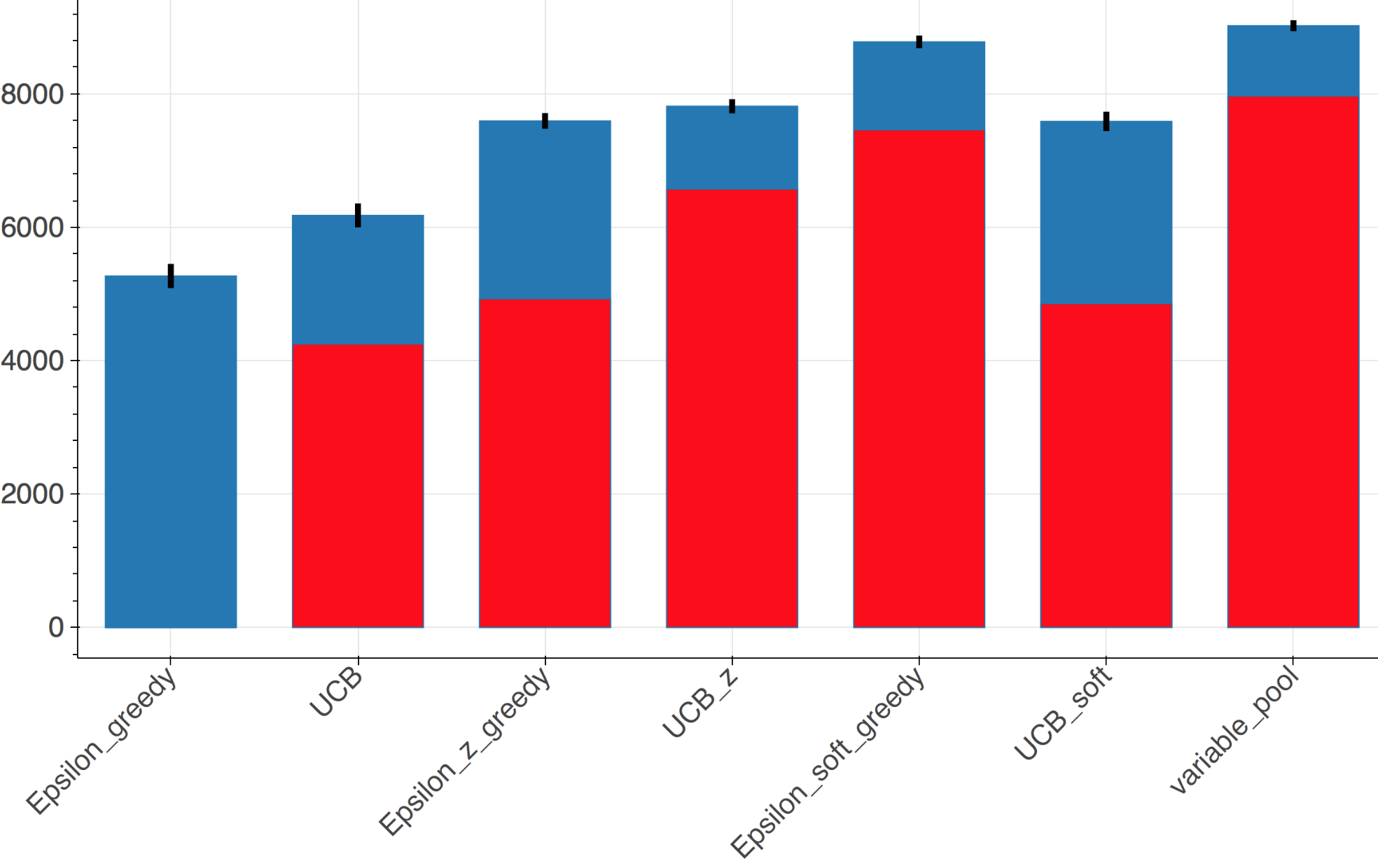} }\label{Truncated_Normal_Wave_100a_500t}}\;\;
		\end{subfloatrow}
	}
	\caption{Comparison of average final rewards in games with 100 arms, 500 turns, and a Wave-type greed function.}\label{Figure::100a_500t_Wave}
\end{figure}

\begin{figure}[]%
	\makebox[\textwidth][c]{ %to center figures!
		\begin{subfloatrow}
			\subfloat[\small{Rewards from Bernoulli distributions. }]{{\includegraphics[width=6.7cm,height=4.0cm]{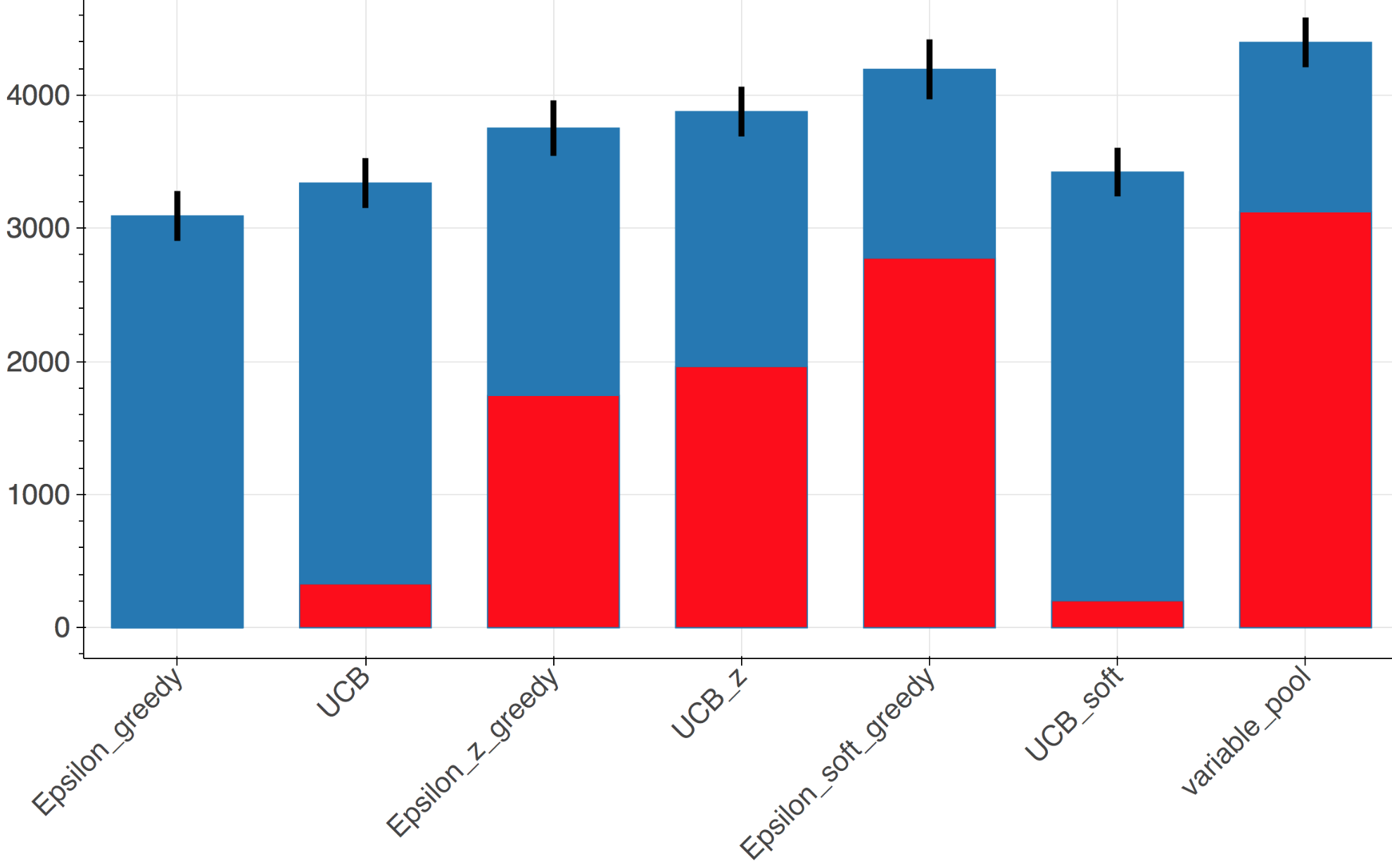} }\label{Bernoulli_Wave_200a_500t}}%
			\;\;
			\qquad
			\subfloat[\small{Rewards from truncated Normal distributions.
			}]{{\includegraphics[width=6.7cm,height=4.0cm]{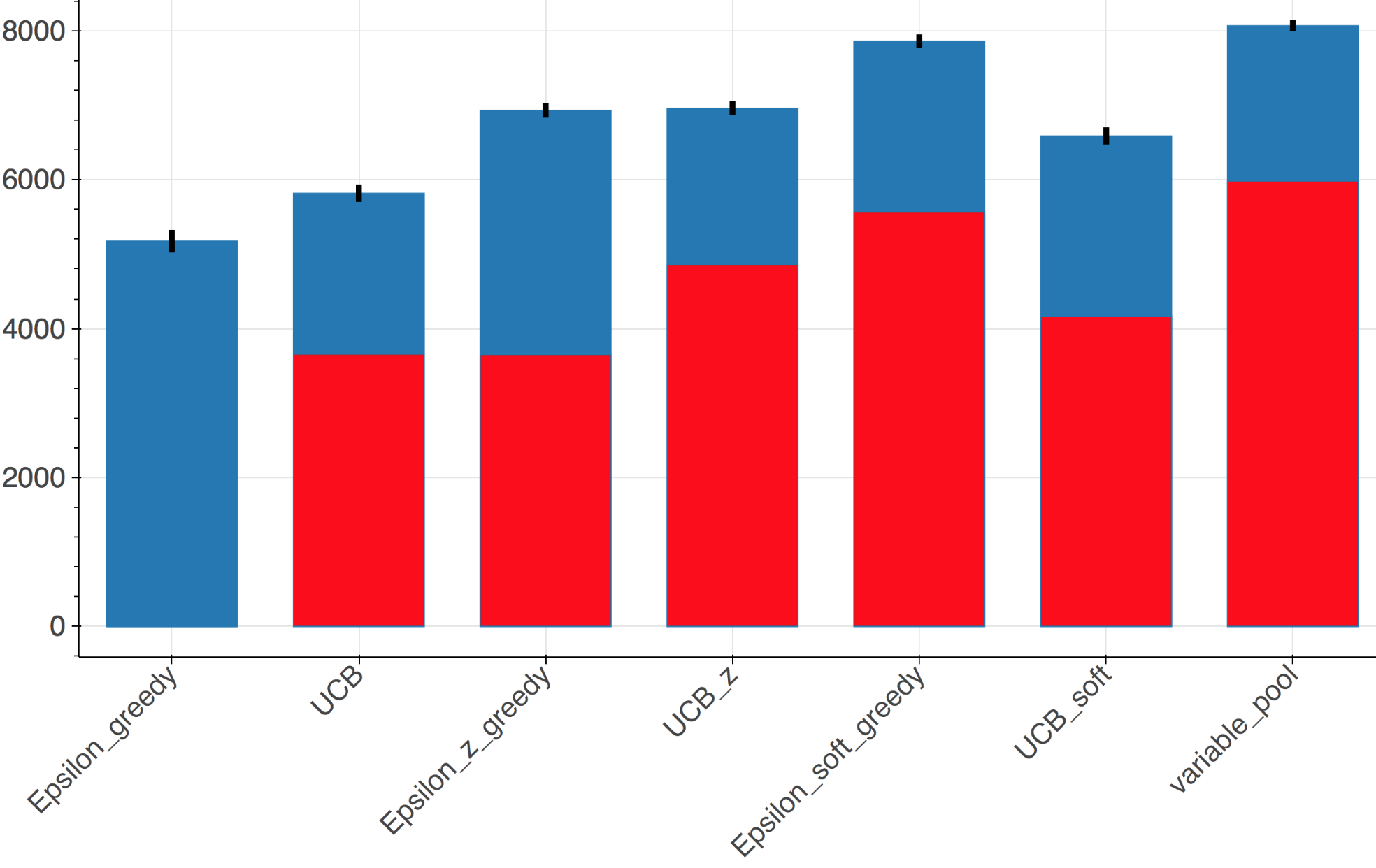} }\label{Truncated_Normal_Wave_200a_500t}}\;\;
		\end{subfloatrow}
	}
	\caption{Comparison of average final rewards in games with 200 arms, 500 turns, and a Wave-type greed function.}\label{Figure::200a_500t_Wave}
\end{figure}

%\clearpage
%%\subsection{Wave-type greed function with 1000 turns per game}

\begin{figure}[]%
	\makebox[\textwidth][c]{ %to center figures!
		\begin{subfloatrow}
			\subfloat[\small{Rewards from Bernoulli distributions. }]{{\includegraphics[width=6.7cm,height=4.0cm]{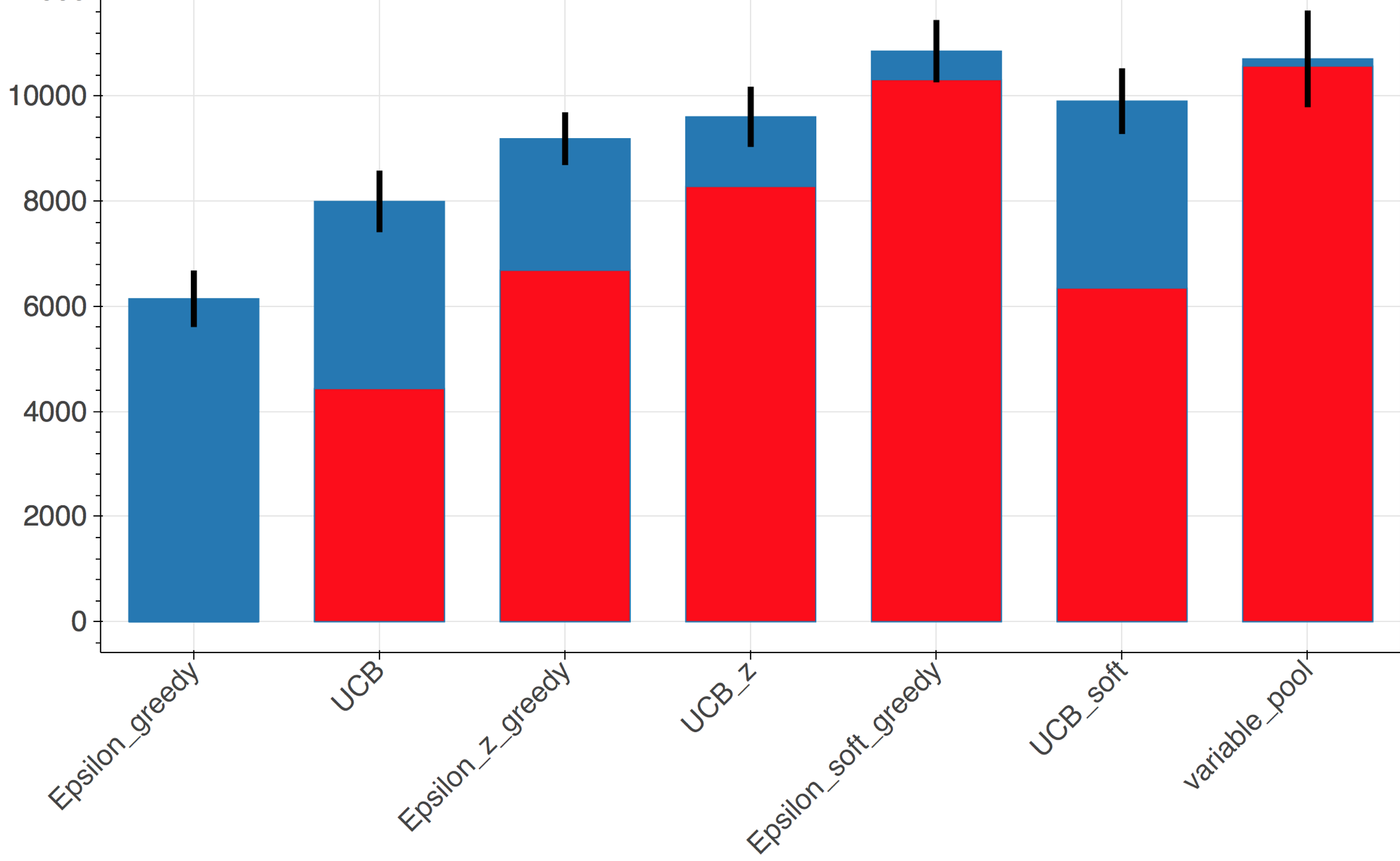} }\label{Bernoulli_Wave_25a_1000t}}%
			\;\;
			\qquad
			\subfloat[\small{Rewards from truncated Normal distributions.
			}]{{\includegraphics[width=6.7cm,height=4.0cm]{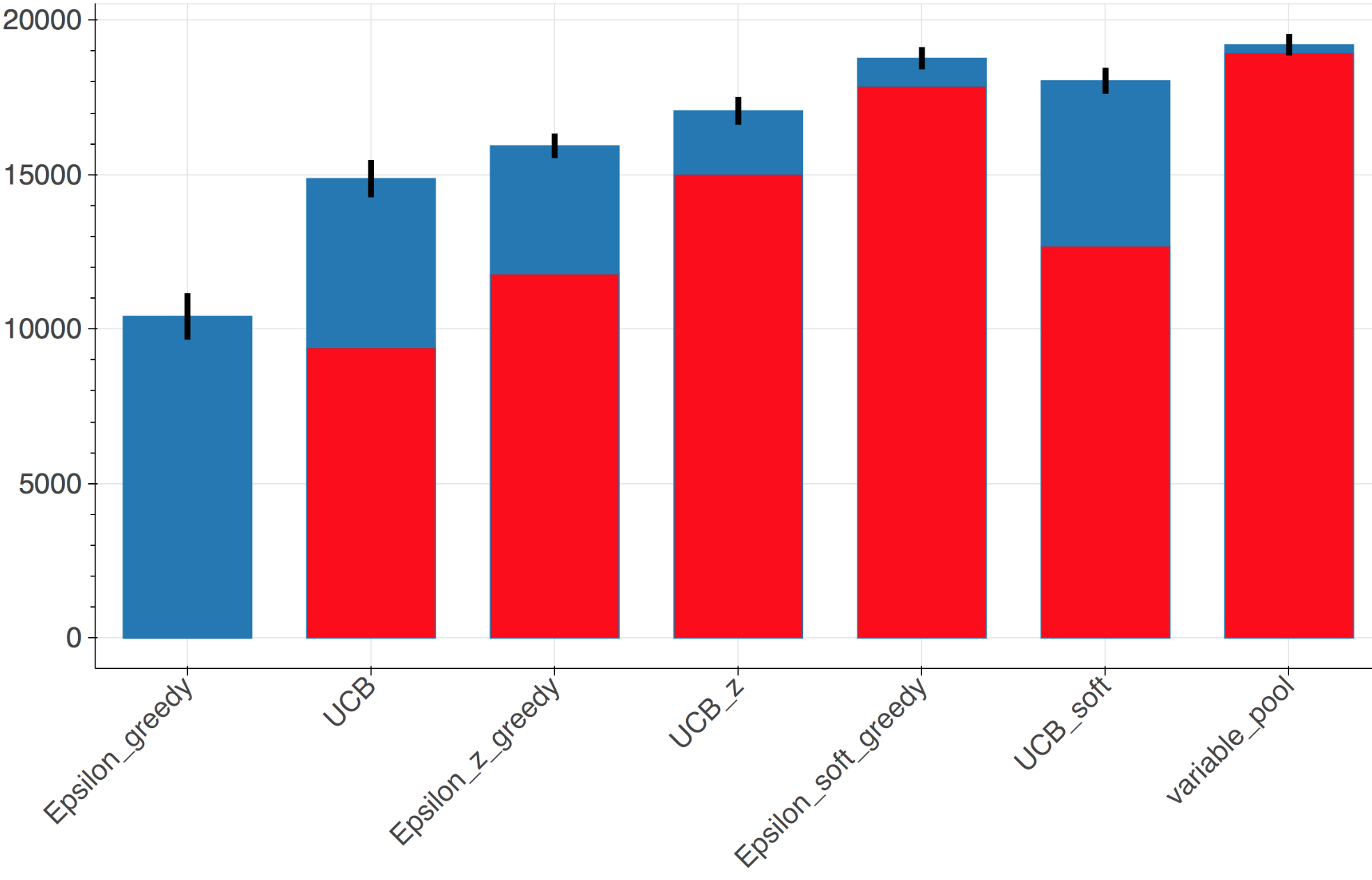} }\label{Truncated_Normal_Wave_25a_1000t}}\;\;
		\end{subfloatrow}
	}
	\caption{Comparison of average final rewards in games with 25 arms, 1000 turns, and a Wave-type greed function.}\label{Figure::25a_1000t_Wave}
\end{figure}

\begin{figure}[]%
	\makebox[\textwidth][c]{ %to center figures!
		\begin{subfloatrow}
			\subfloat[\small{Rewards from Bernoulli distributions. }]{{\includegraphics[width=6.7cm,height=4.0cm]{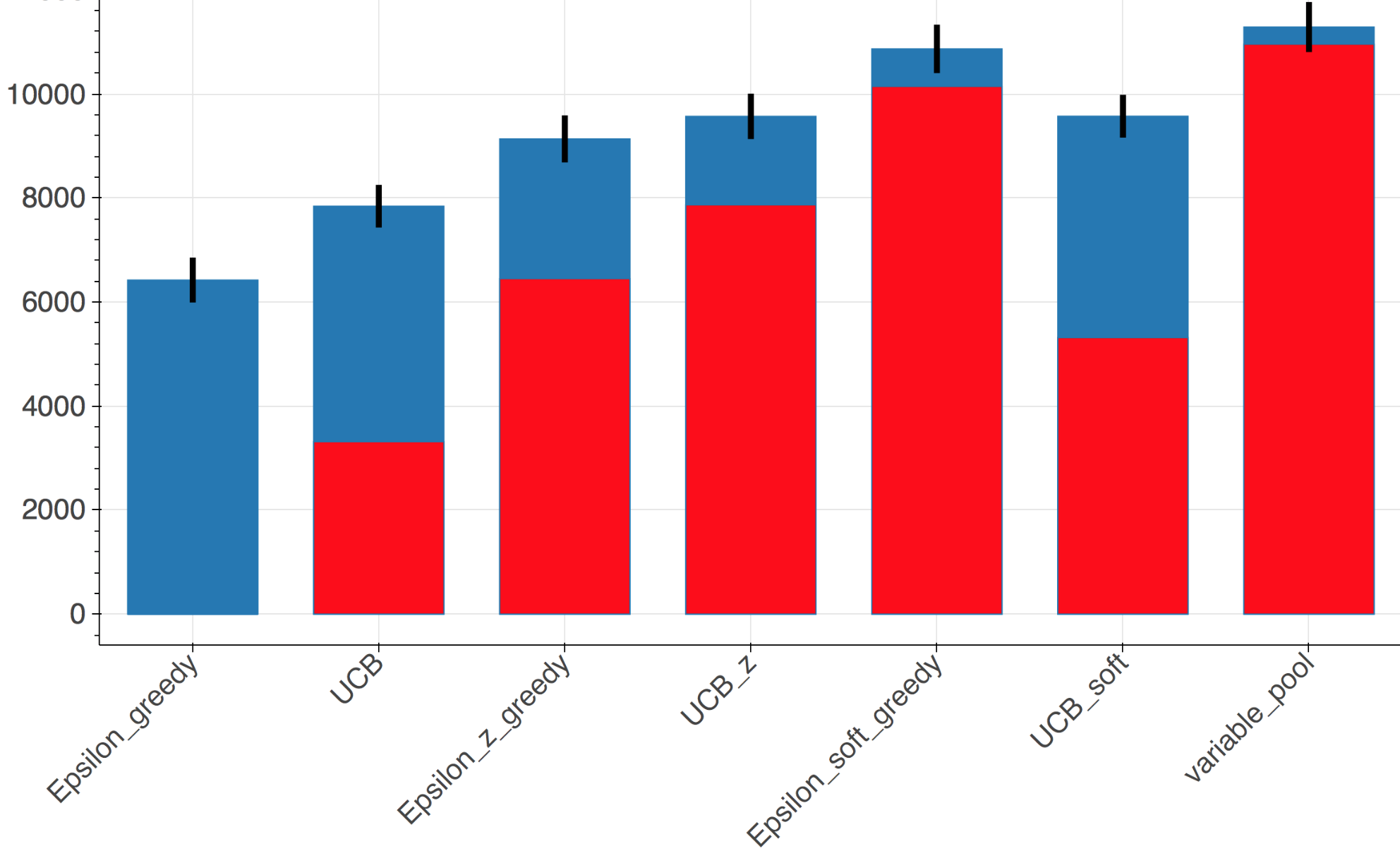} }\label{Bernoulli_Wave_50a_1000t}}%
			\;\;
			\qquad
			\subfloat[\small{Rewards from truncated Normal distributions.
			}]{{\includegraphics[width=6.7cm,height=4.0cm]{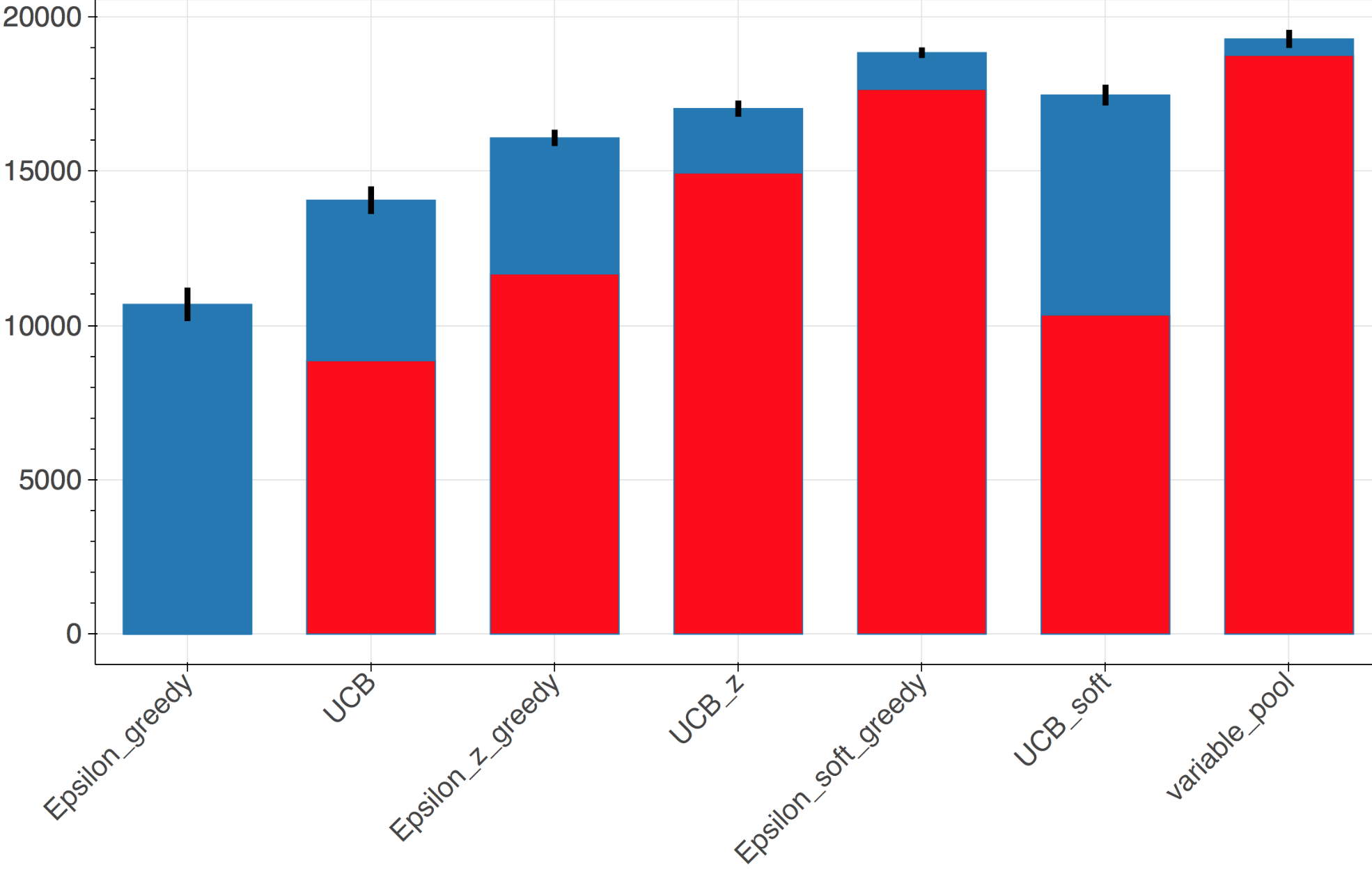} }\label{Truncated_Normal_Wave_50a_1000t}}\;\;
		\end{subfloatrow}
	}
	\caption{Comparison of average final rewards in games with 50 arms, 1000 turns, and a Wave-type greed function.}\label{Figure::50a_1000t_Wave}
\end{figure}

\begin{figure}[]%
	\makebox[\textwidth][c]{ %to center figures!
		\begin{subfloatrow}
			\subfloat[\small{Rewards from Bernoulli distributions. }]{{\includegraphics[width=6.7cm,height=4.0cm]{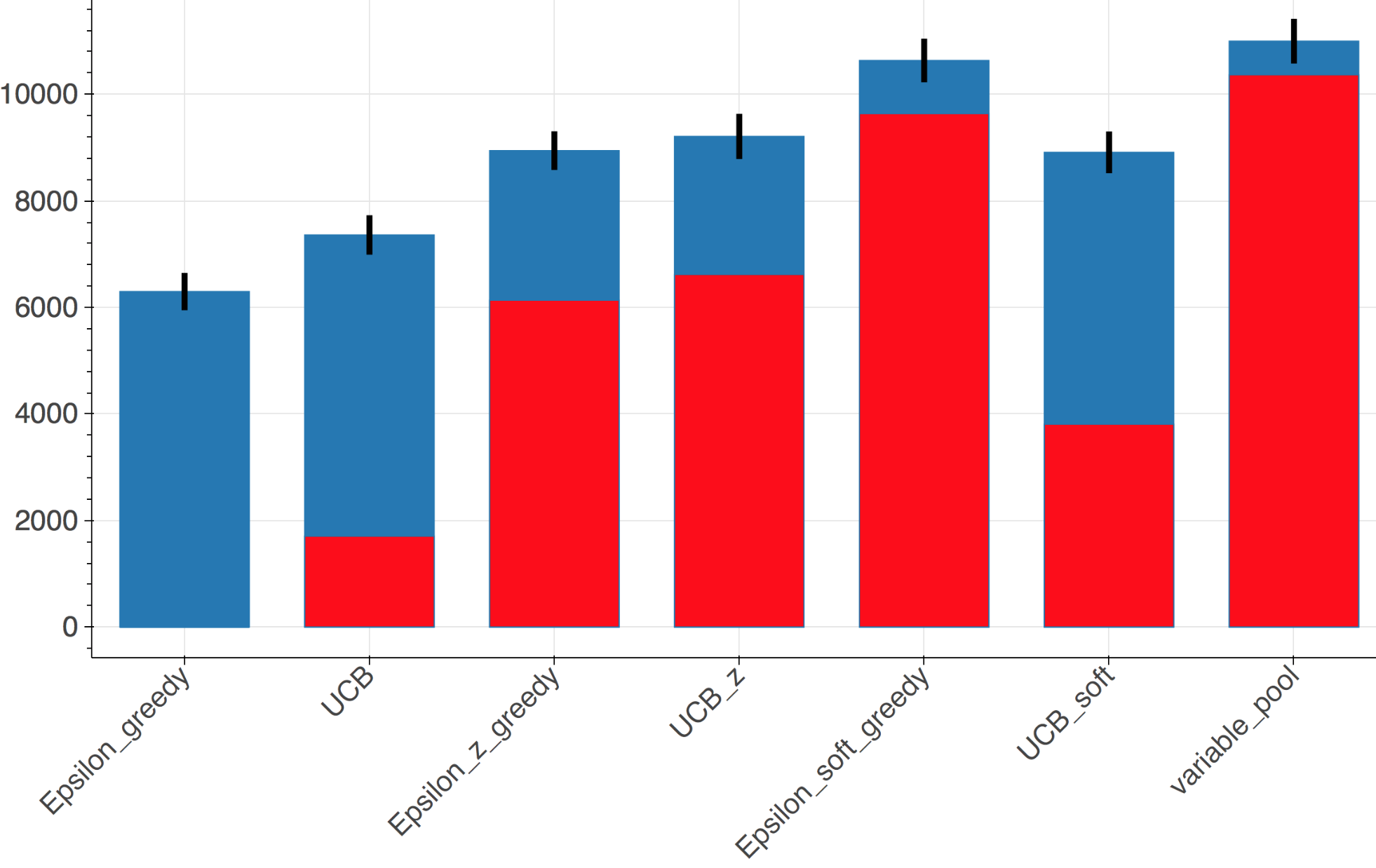} }\label{Bernoulli_Wave_100a_1000t}}%
			\;\;
			\qquad
			\subfloat[\small{Rewards from truncated Normal distributions.
			}]{{\includegraphics[width=6.7cm,height=4.0cm]{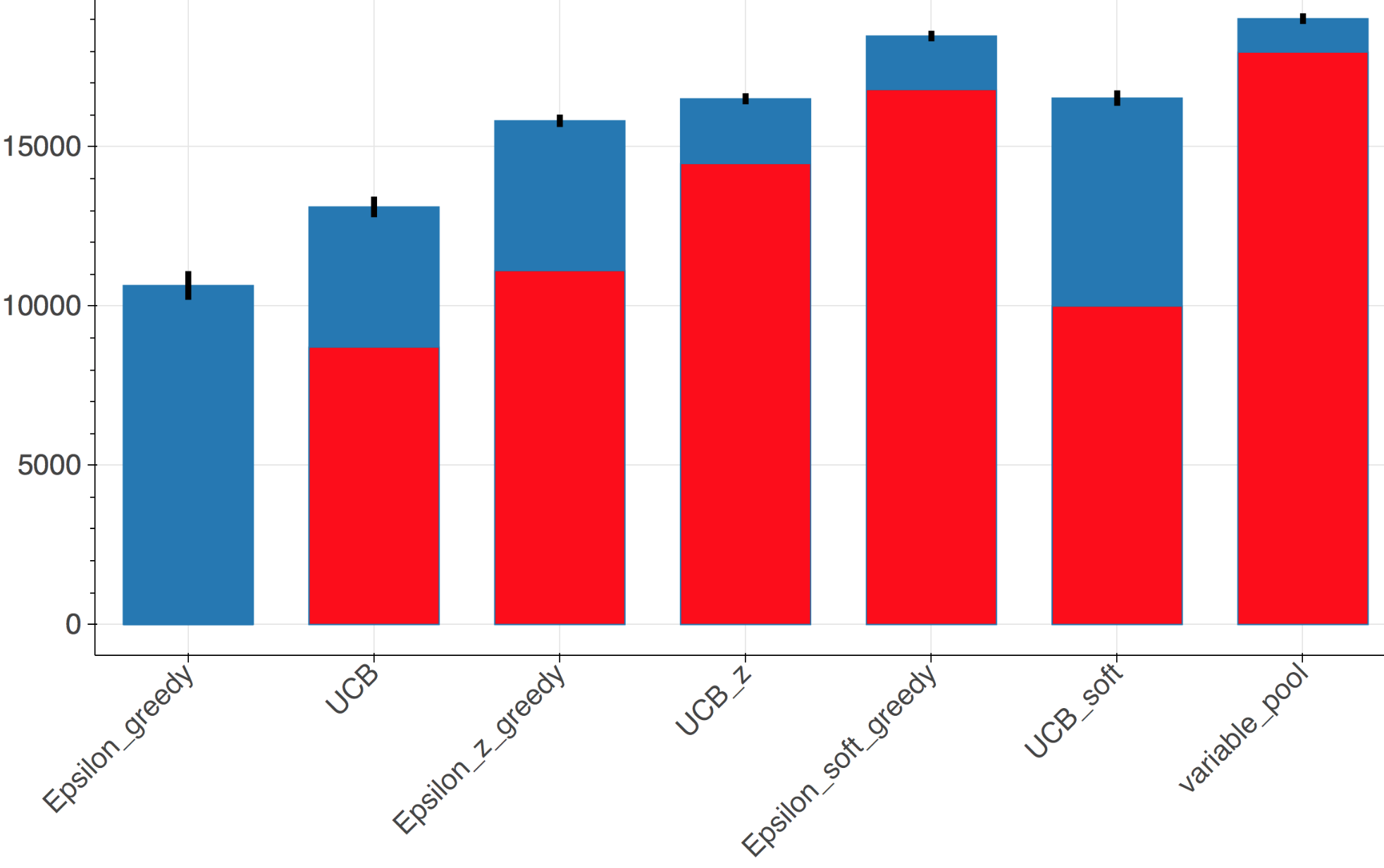} }\label{Truncated_Normal_Wave_100a_1000t}}\;\;
		\end{subfloatrow}
	}
	\caption{Comparison of average final rewards in games with 100 arms, 1000 turns, and a Wave-type greed function.}\label{Figure::100a_1000t_Wave}
\end{figure}

\begin{figure}[]%
	\makebox[\textwidth][c]{ %to center figures!
		\begin{subfloatrow}
			\subfloat[\small{Rewards from Bernoulli distributions. }]{{\includegraphics[width=6.7cm,height=4.0cm]{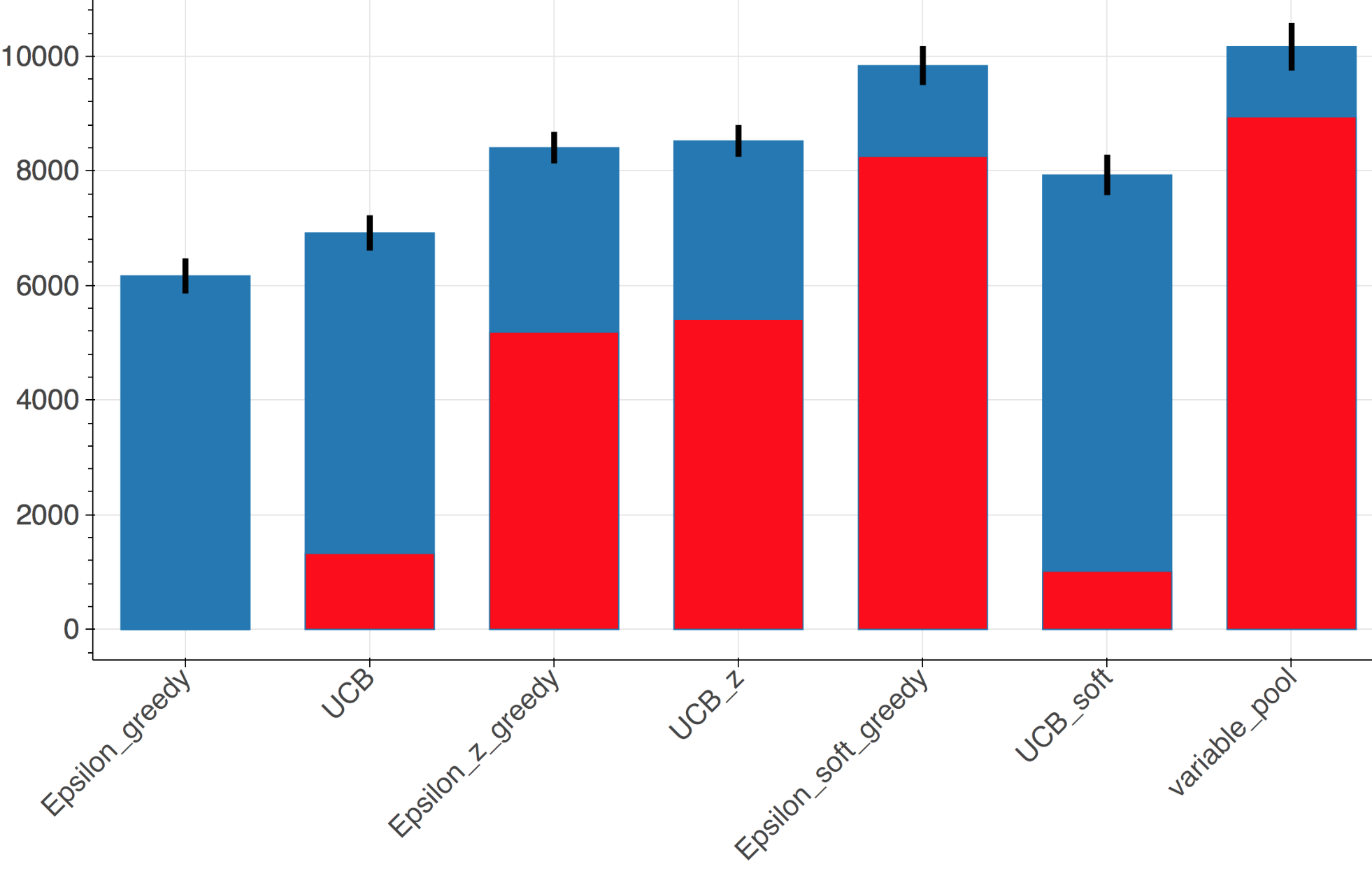} }\label{Bernoulli_Wave_200a_1000t}}%
			\;\;
			\qquad
			\subfloat[\small{Rewards from truncated Normal distributions.
			}]{{\includegraphics[width=6.7cm,height=4.0cm]{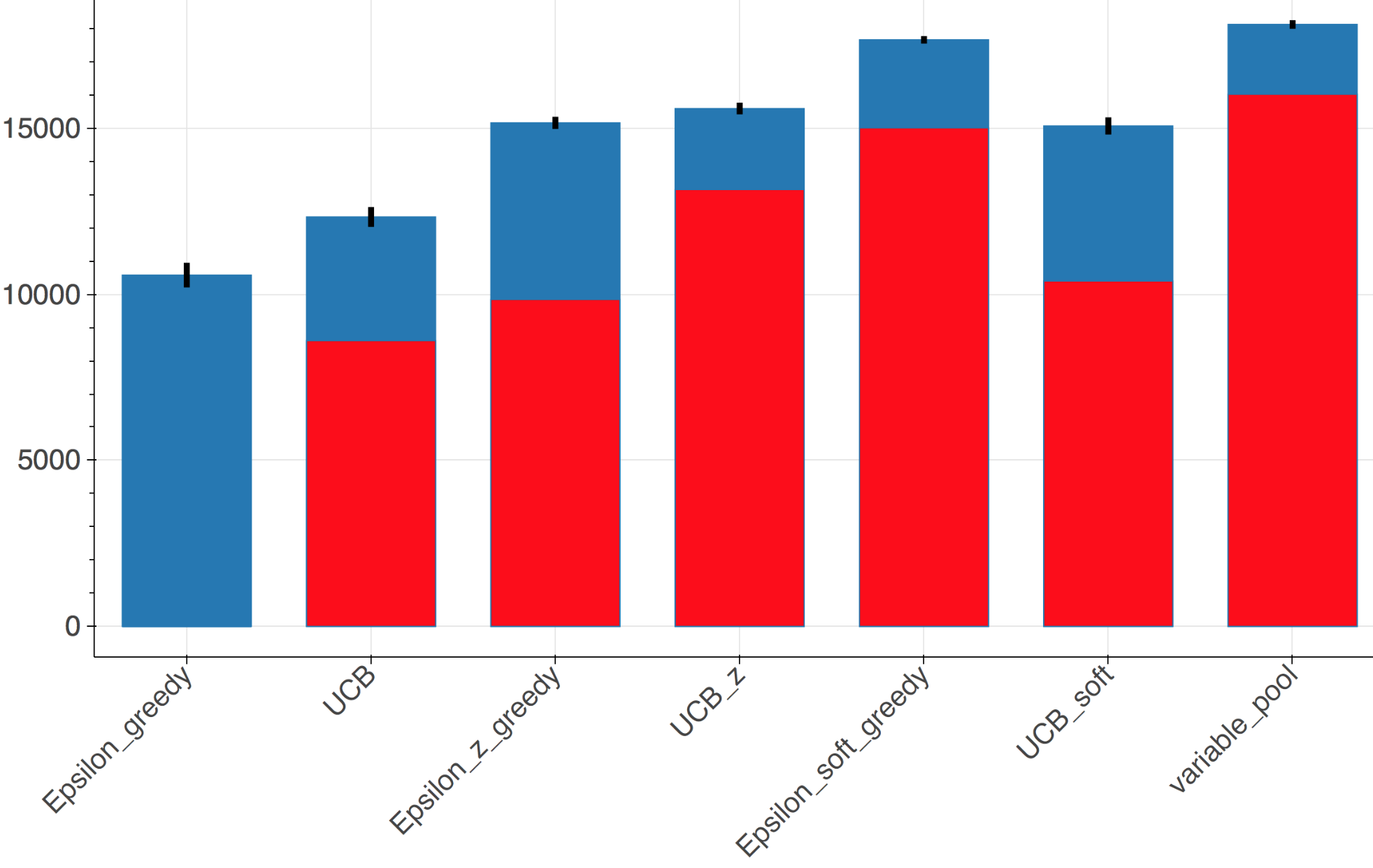} }\label{Truncated_Normal_Wave_200a_1000t}}\;\;
		\end{subfloatrow}
	}
	\caption{Comparison of average final rewards in games with 200 arms, 1000 turns, and a Wave-type greed function.}\label{Figure::200a_1000t_Wave}
\end{figure}

%\clearpage
%\subsection{Wave-type greed function with 1500 turns per game}

\begin{figure}[]%
	\makebox[\textwidth][c]{ %to center figures!
		\begin{subfloatrow}
			\subfloat[\small{Rewards from Bernoulli distributions. }]{{\includegraphics[width=6.7cm,height=4.0cm]{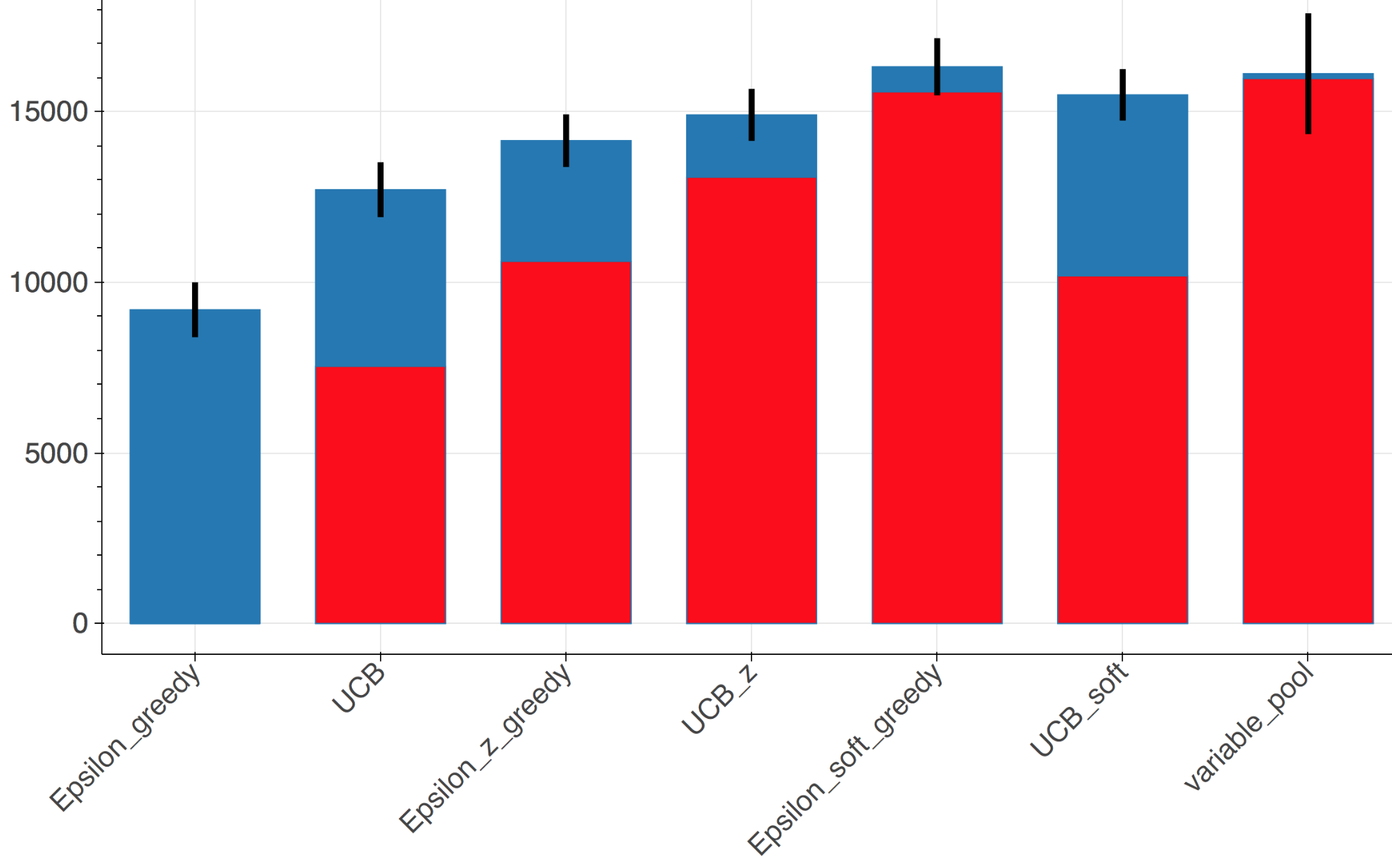} }\label{Bernoulli_Wave_25a_1500t}}%
			\;\;
			\qquad
			\subfloat[\small{Rewards from truncated Normal distributions.
			}]{{\includegraphics[width=6.7cm,height=4.0cm]{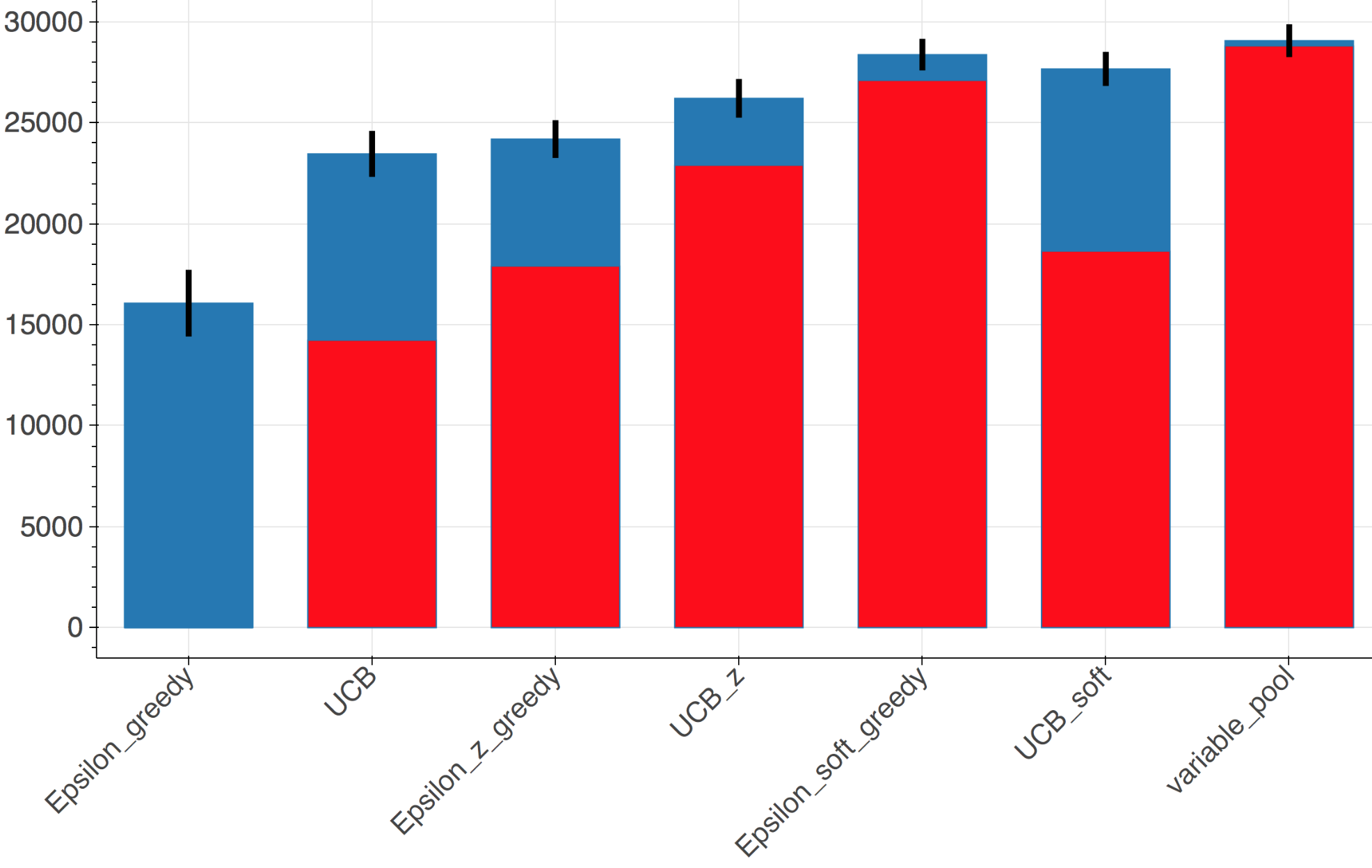} }\label{Truncated_Normal_Wave_25a_1500t}}\;\;
		\end{subfloatrow}
	}
	\caption{Comparison of average final rewards in games with 25 arms, 1500 turns, and a Wave-type greed function.}\label{Figure::25a_1500t_Wave}
\end{figure}

\begin{figure}[]%
	\makebox[\textwidth][c]{ %to center figures!
		\begin{subfloatrow}
			\subfloat[\small{Rewards from Bernoulli distributions. }]{{\includegraphics[width=6.7cm,height=4.0cm]{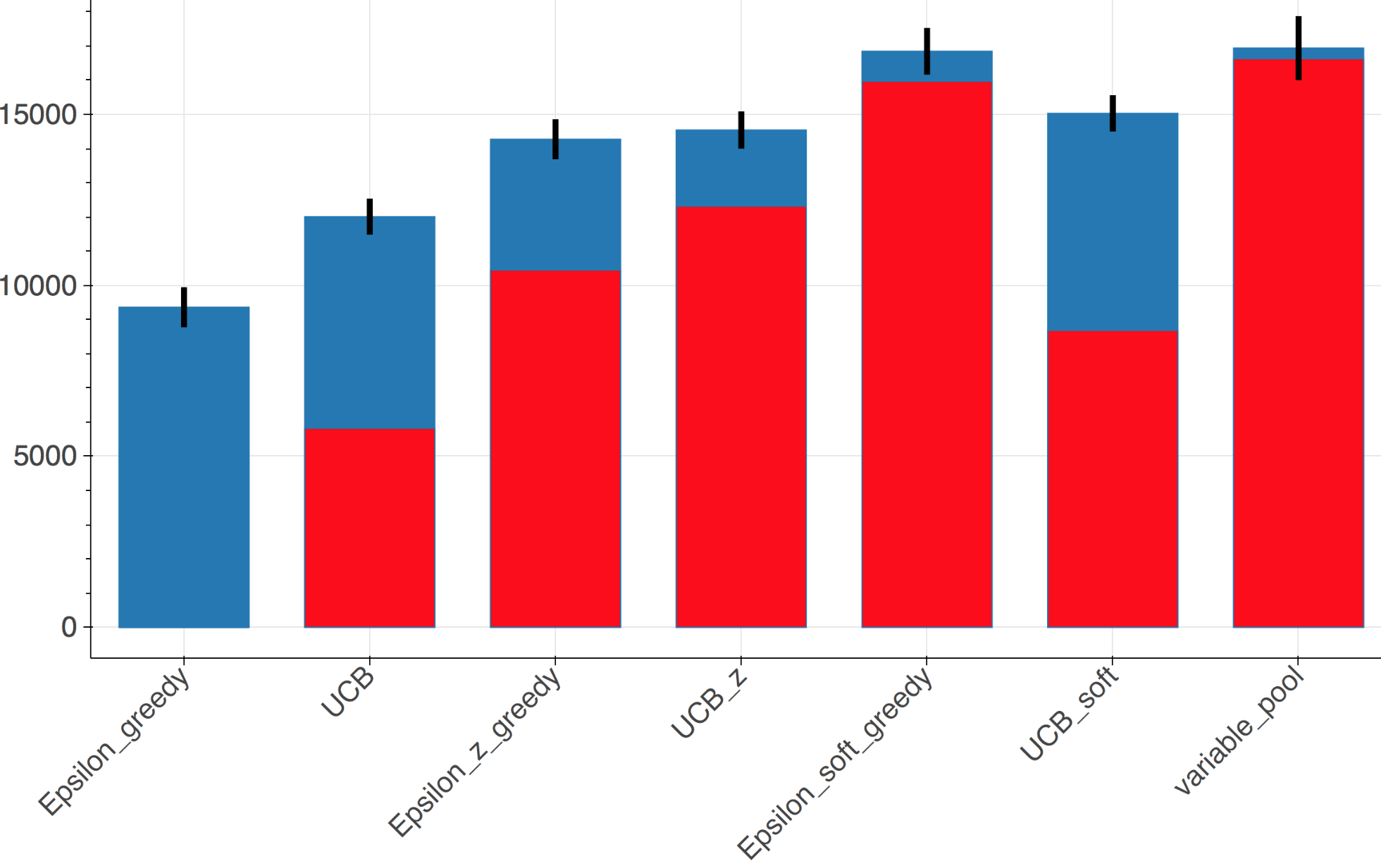} }\label{Bernoulli_Wave_50a_1500t}}%
			\;\;
			\qquad
			\subfloat[\small{Rewards from truncated Normal distributions.
			}]{{\includegraphics[width=6.7cm,height=4.0cm]{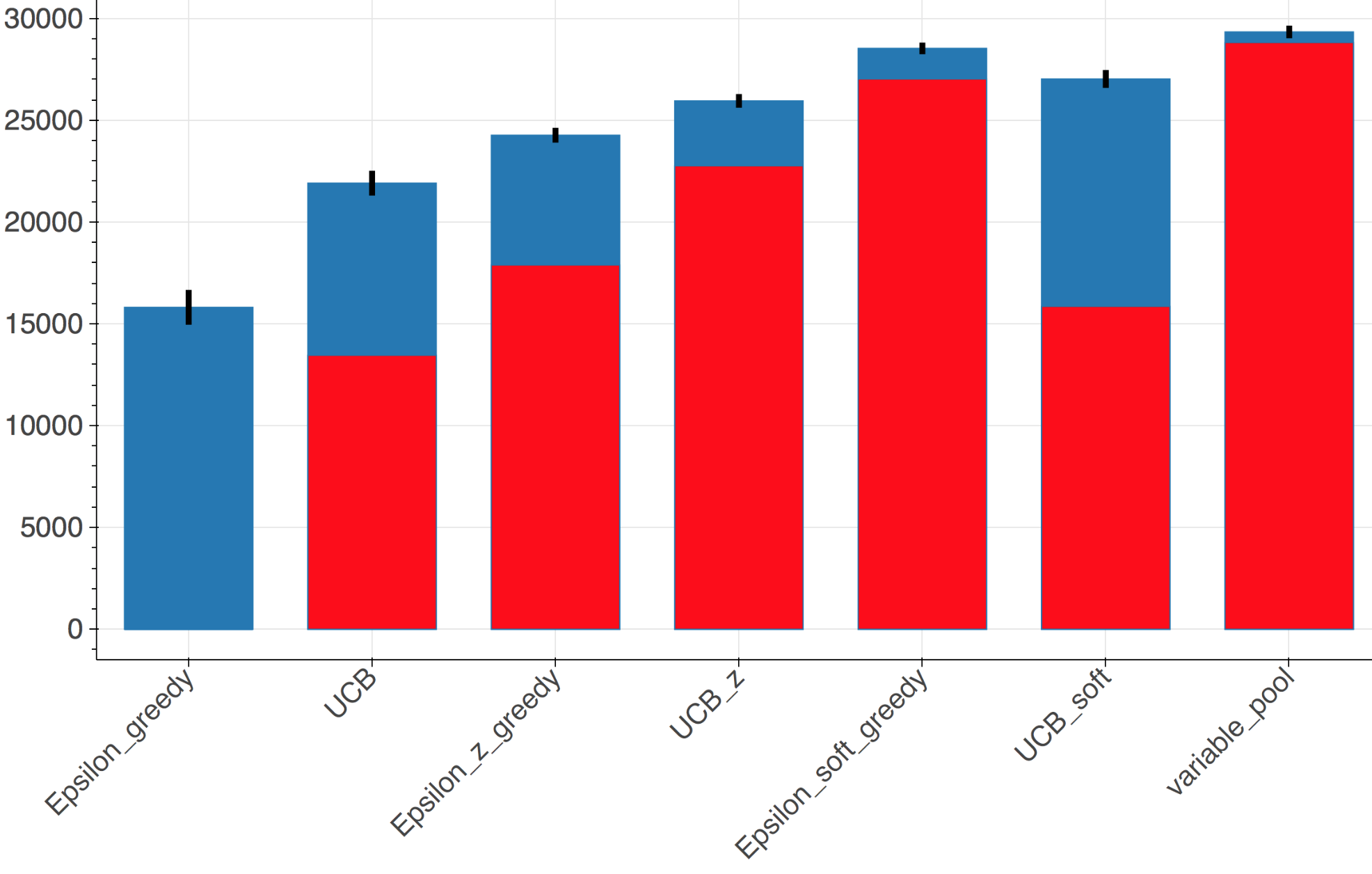} }\label{Truncated_Normal_Wave_50a_1500t}}\;\;
		\end{subfloatrow}
	}
	\caption{Comparison of average final rewards in games with 50 arms, 1500 turns, and a Wave-type greed function.}\label{Figure::50a_1500t_Wave}
\end{figure}

\begin{figure}[]%
	\makebox[\textwidth][c]{ %to center figures!
		\begin{subfloatrow}
			\subfloat[\small{Rewards from Bernoulli distributions. }]{{\includegraphics[width=6.7cm,height=4.0cm]{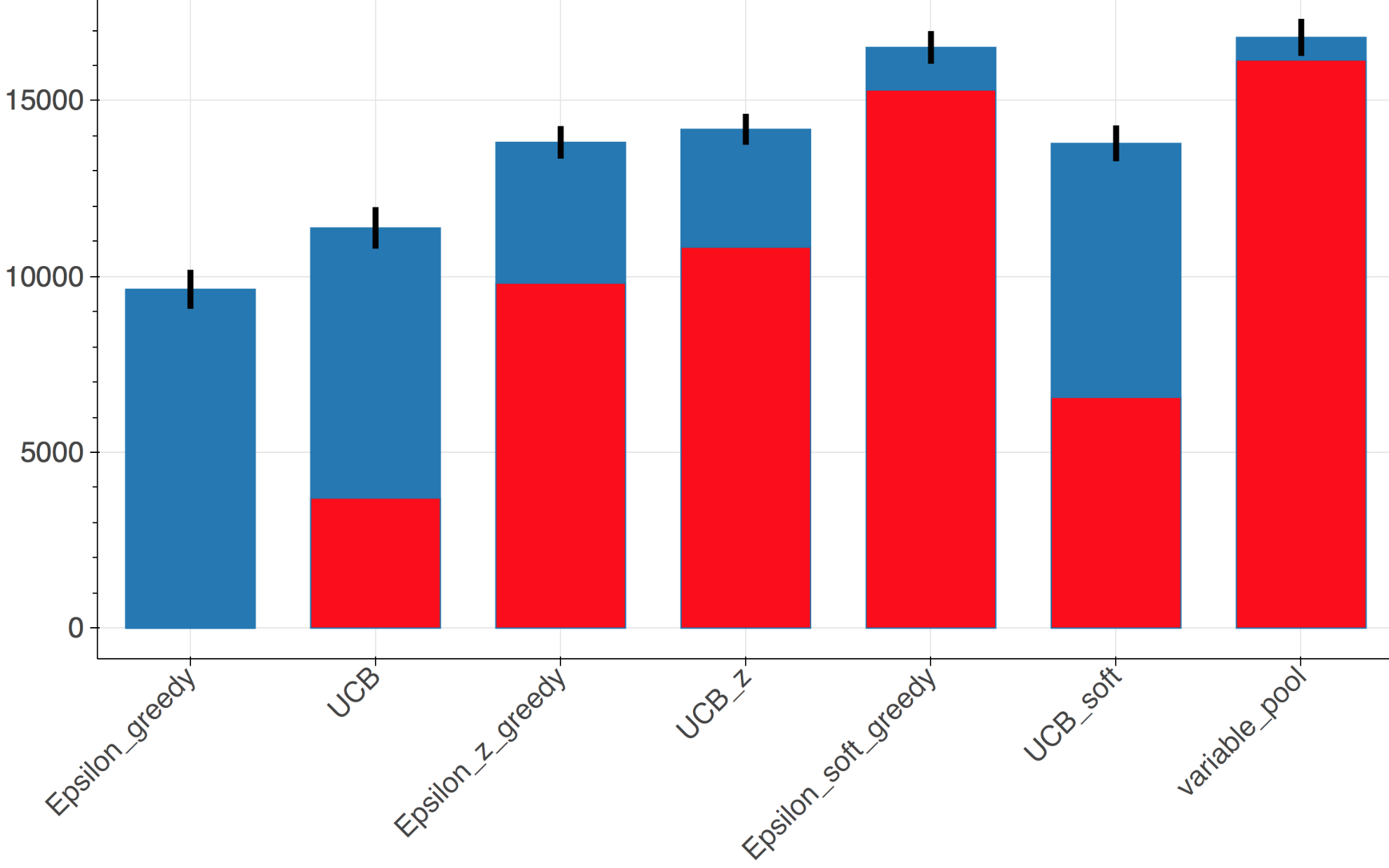} }\label{Bernoulli_Wave_100a_1500t}}%
			\;\;
			\qquad
			\subfloat[\small{Rewards from truncated Normal distributions.
			}]{{\includegraphics[width=6.7cm,height=4.0cm]{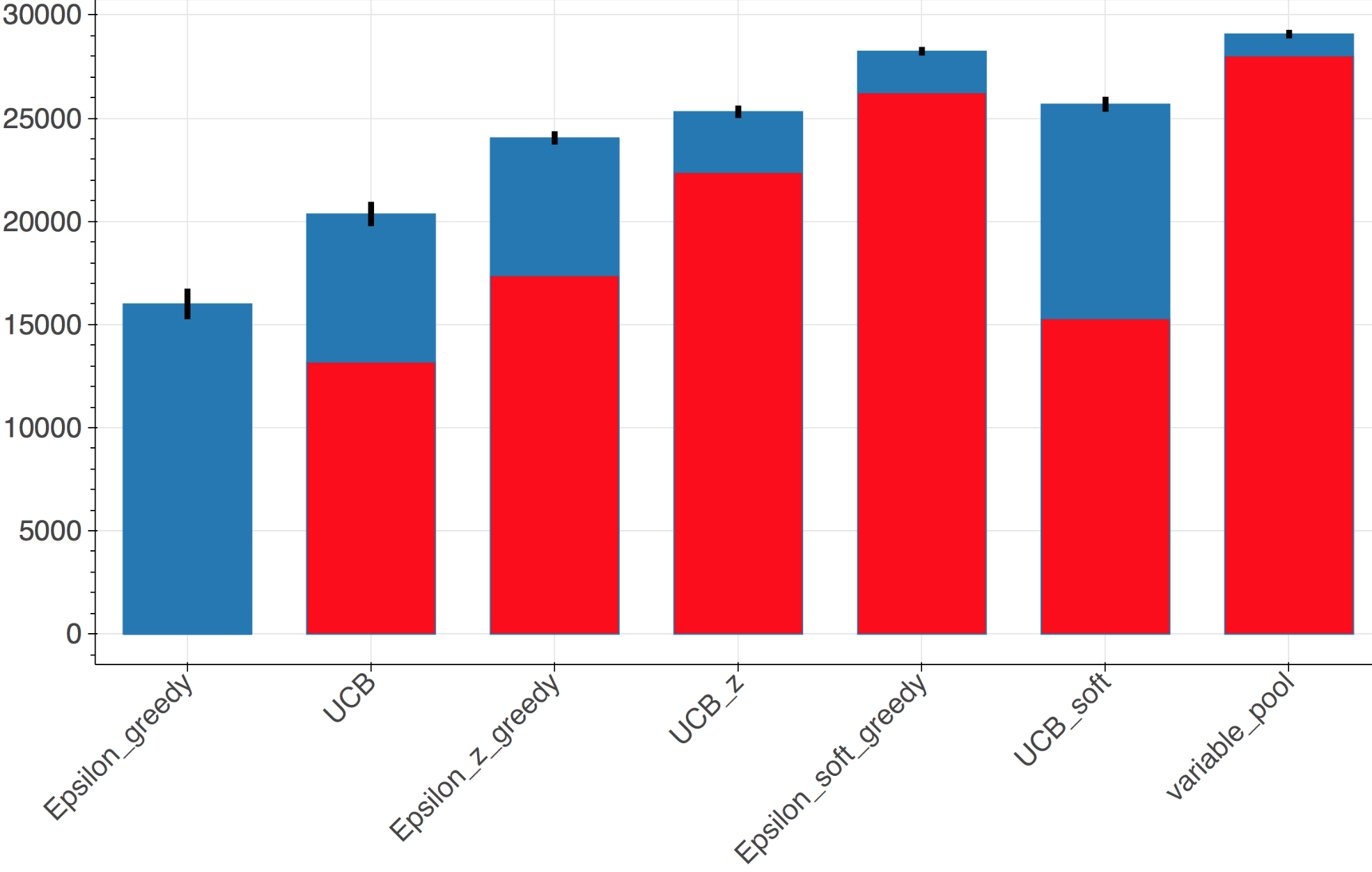} }\label{Truncated_Normal_Wave_100a_1500t}}\;\;
		\end{subfloatrow}
	}
	\caption{Comparison of average final rewards in games with 100 arms, 1500 turns, and a Wave-type greed function.}\label{Figure::100a_1500t_Wave}
\end{figure}

\begin{figure}[]%
	\makebox[\textwidth][c]{ %to center figures!
		\begin{subfloatrow}
			\subfloat[\small{Rewards from Bernoulli distributions. }]{{\includegraphics[width=6.7cm,height=4.0cm]{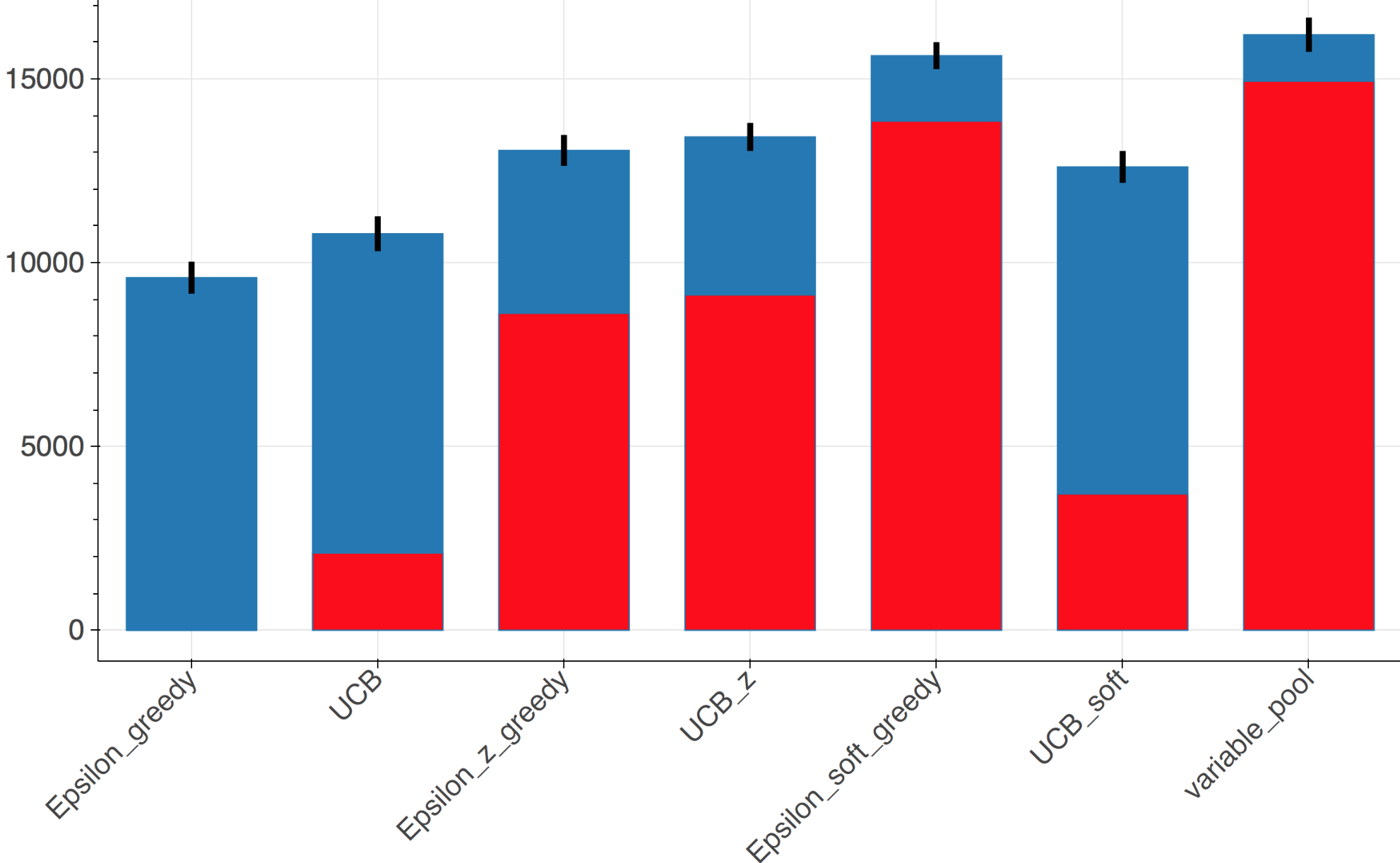} }\label{Bernoulli_Wave_200a_1500t}}%
			\;\;
			\qquad
			\subfloat[\small{Rewards from truncated Normal distributions.
			}]{{\includegraphics[width=6.7cm,height=4.0cm]{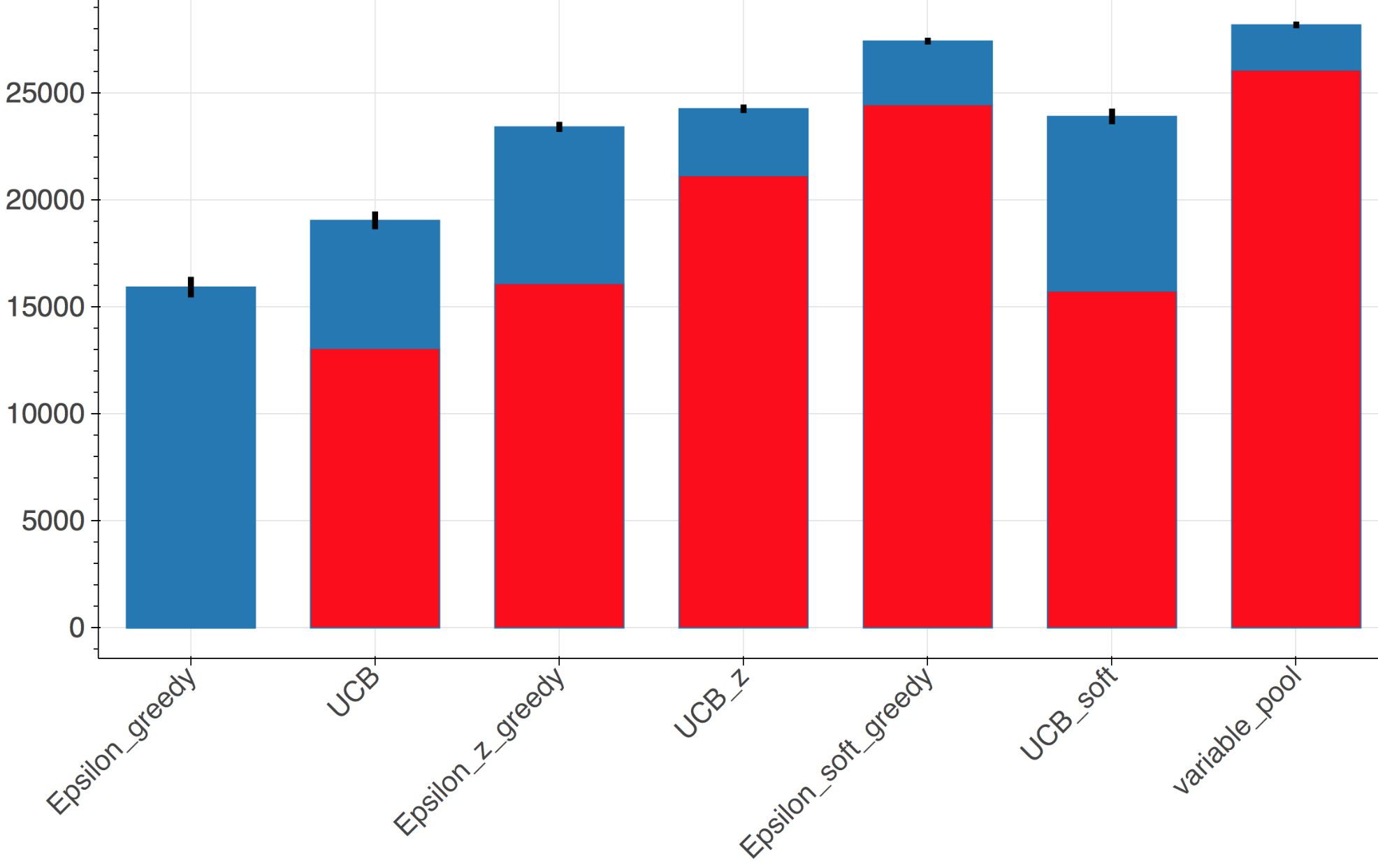} }\label{Truncated_Normal_Wave_200a_1500t}}\;\;
		\end{subfloatrow}
	}
	\caption{Comparison of average final rewards in games with 200 arms, 1500 turns, and a Wave-type greed function.}\label{Figure::200a_1500t_Wave}
\end{figure}

%%\subsection{Average increase in rewards with respect to the (smarter) versions of the $\varepsilon$-greedy algorithm and UCB algorithm with a Wave-type greed function}

%%\subsection{Cumulative Reward increase when regulating greed over time compared to the (smarter) $\varepsilon$-greedy algorithm and the (smarter) UCB algorithm with a Wave-type greed function.}

%\subsection{Percentage reward increase of regulating greed algorithms over a baseline algorithm that does not regulate greed. The reward multiplier is of the Wave-type.}

\begin{figure}[]
	\caption{Average increase in rewards (coming from Bernoulli distributions) compared to the (smarter) version of the UCB algorithm (Algorithm \ref{Algorithm::UCB_slightly_smarter}) with a Wave-type greed function. }
	\centering
	\includegraphics[width=0.95\textwidth]{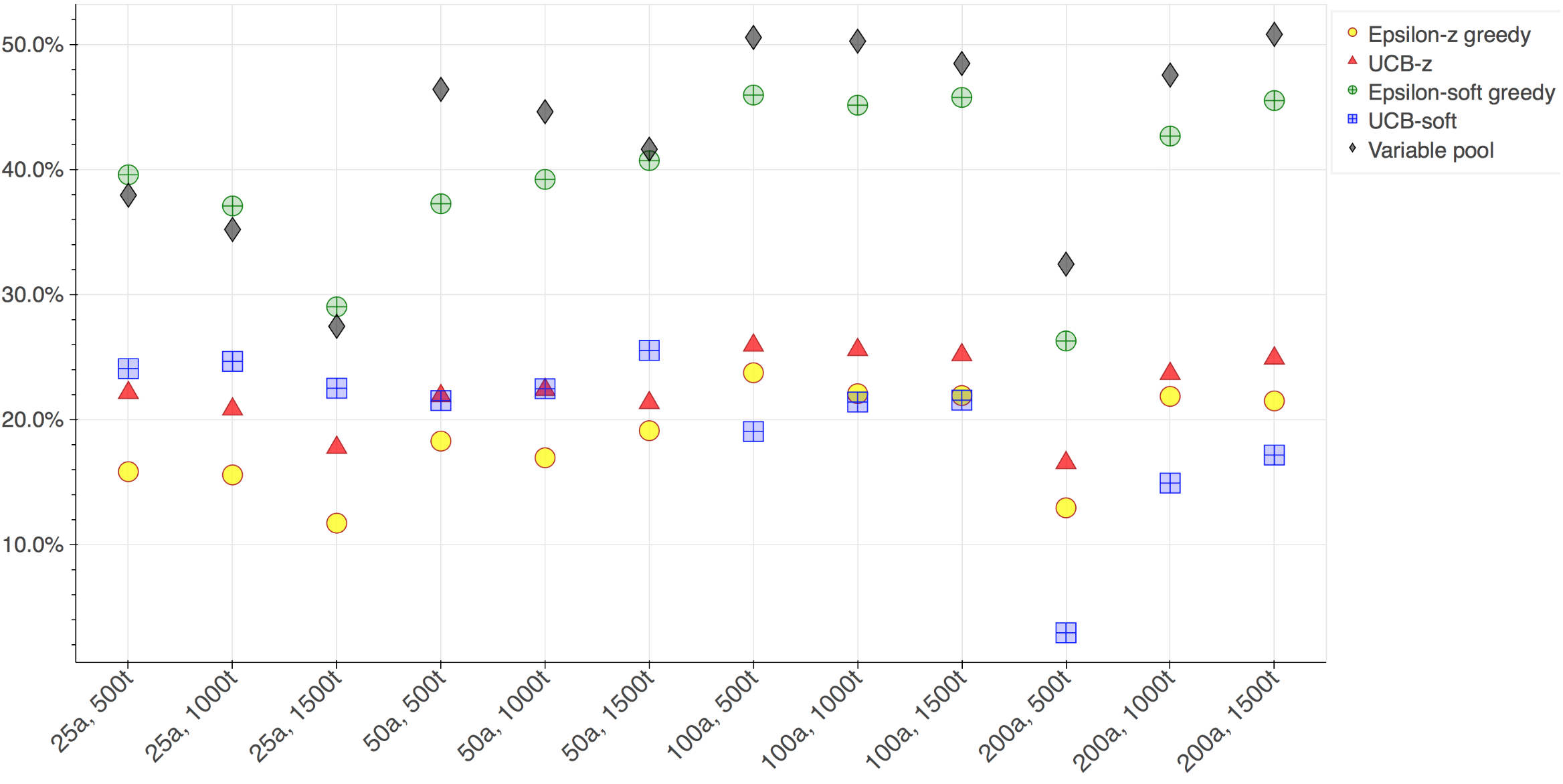}
	\label{BUW}
\end{figure}

\begin{figure}[]
	\caption{Average increase in rewards (coming from Truncated-Normal distributions) compared to the (smarter) version of the $\varepsilon$-greedy algorithm (Algorithm \ref{Algorithm::epsilon_slightly_smarter}) with a Wave-type greed function. }
	\centering
	\includegraphics[width=0.95\textwidth]{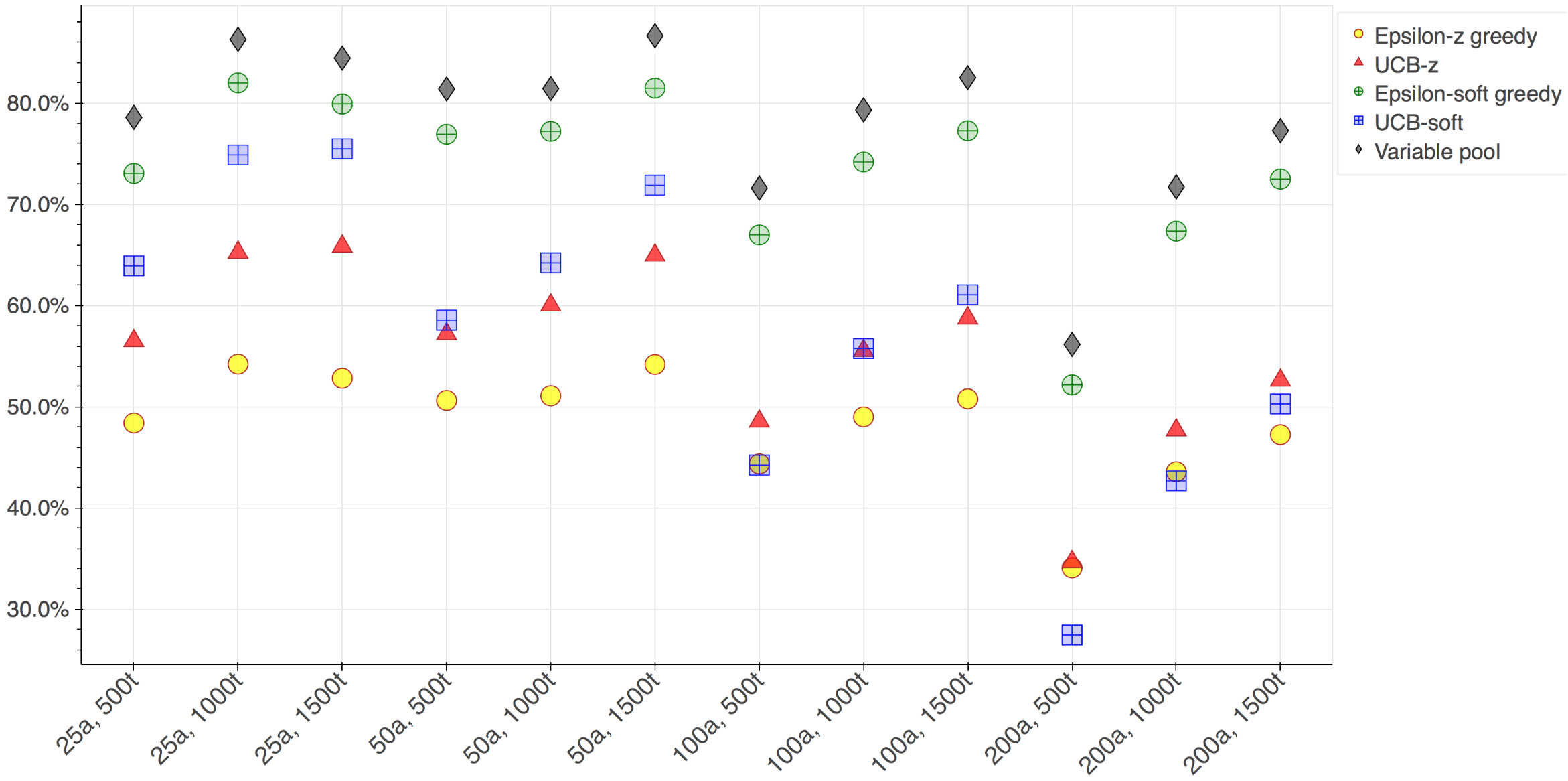}
	\label{NEW}
\end{figure}

\begin{figure}[]
	\caption{Average increase in rewards (coming from Truncated-Normal distributions) compared to the (smarter) version of the UCB algorithm (Algorithm \ref{Algorithm::UCB_slightly_smarter}) with a Wave-type greed function. }
	\centering
	\includegraphics[width=0.95\textwidth]{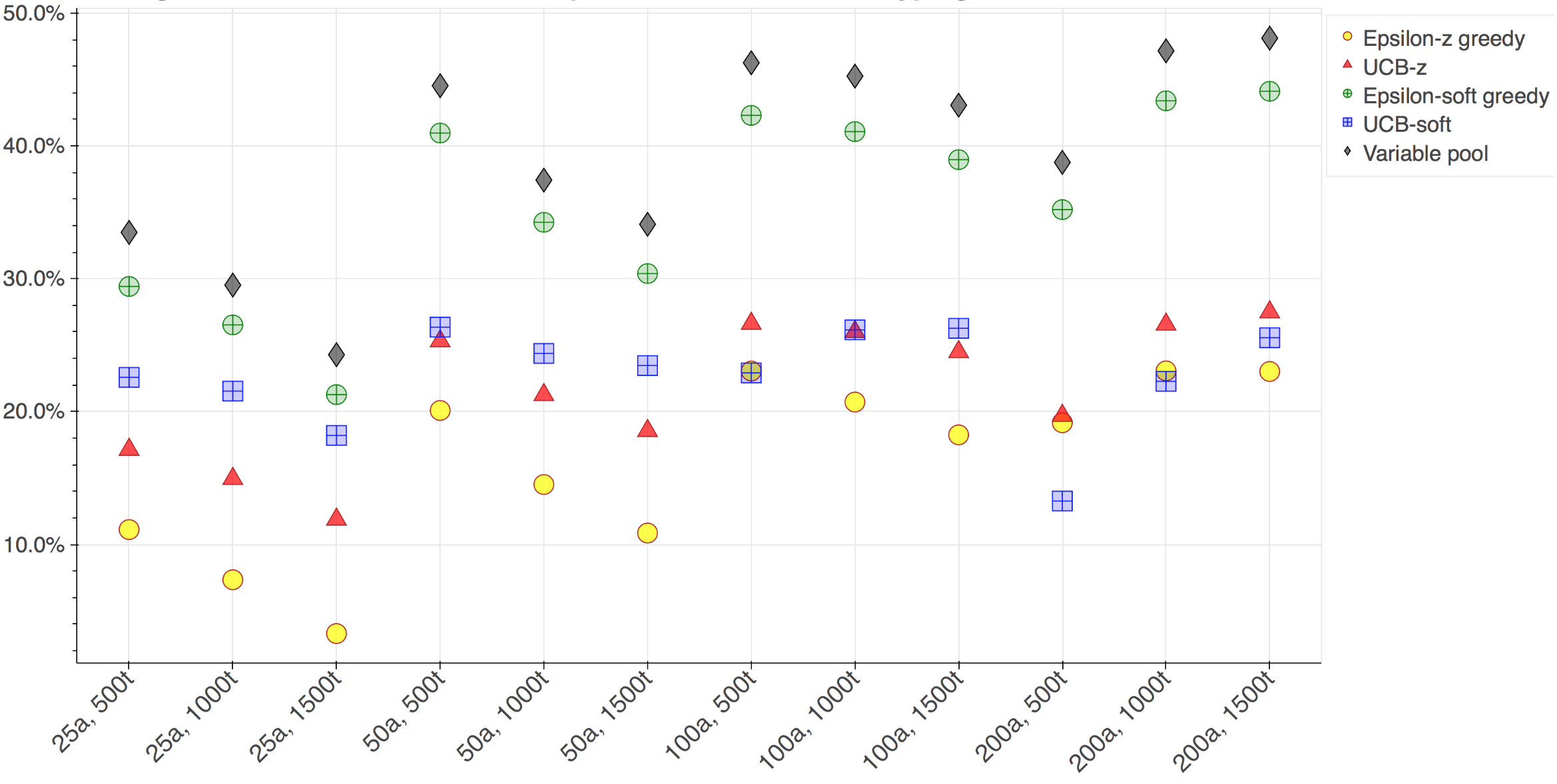}
	\label{NUW}
\end{figure}

%% ------------------
\clearpage
%\subsection{Step-type greed function with 500 turns per game}

\begin{figure}[]%
	\makebox[\textwidth][c]{ %to center figures!
		\begin{subfloatrow}
			\subfloat[\small{Rewards from Bernoulli distributions. }]{{\includegraphics[width=6.7cm,height=4.0cm]{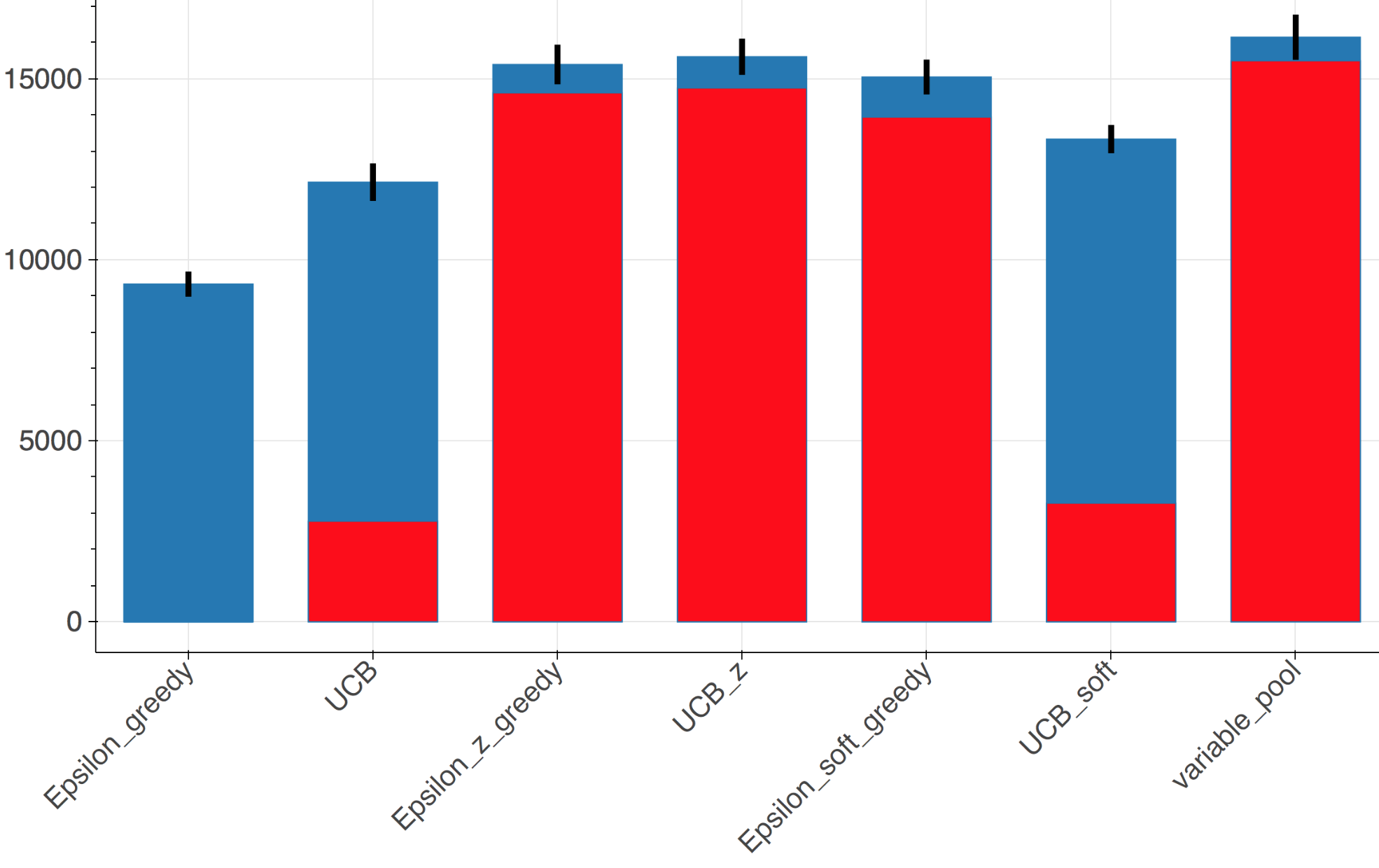} }\label{Bernoulli_Step_25a_500t}}%
			\;\;
			\qquad
			\subfloat[\small{Rewards from truncated Normal distributions.
			}]{{\includegraphics[width=6.7cm,height=4.0cm]{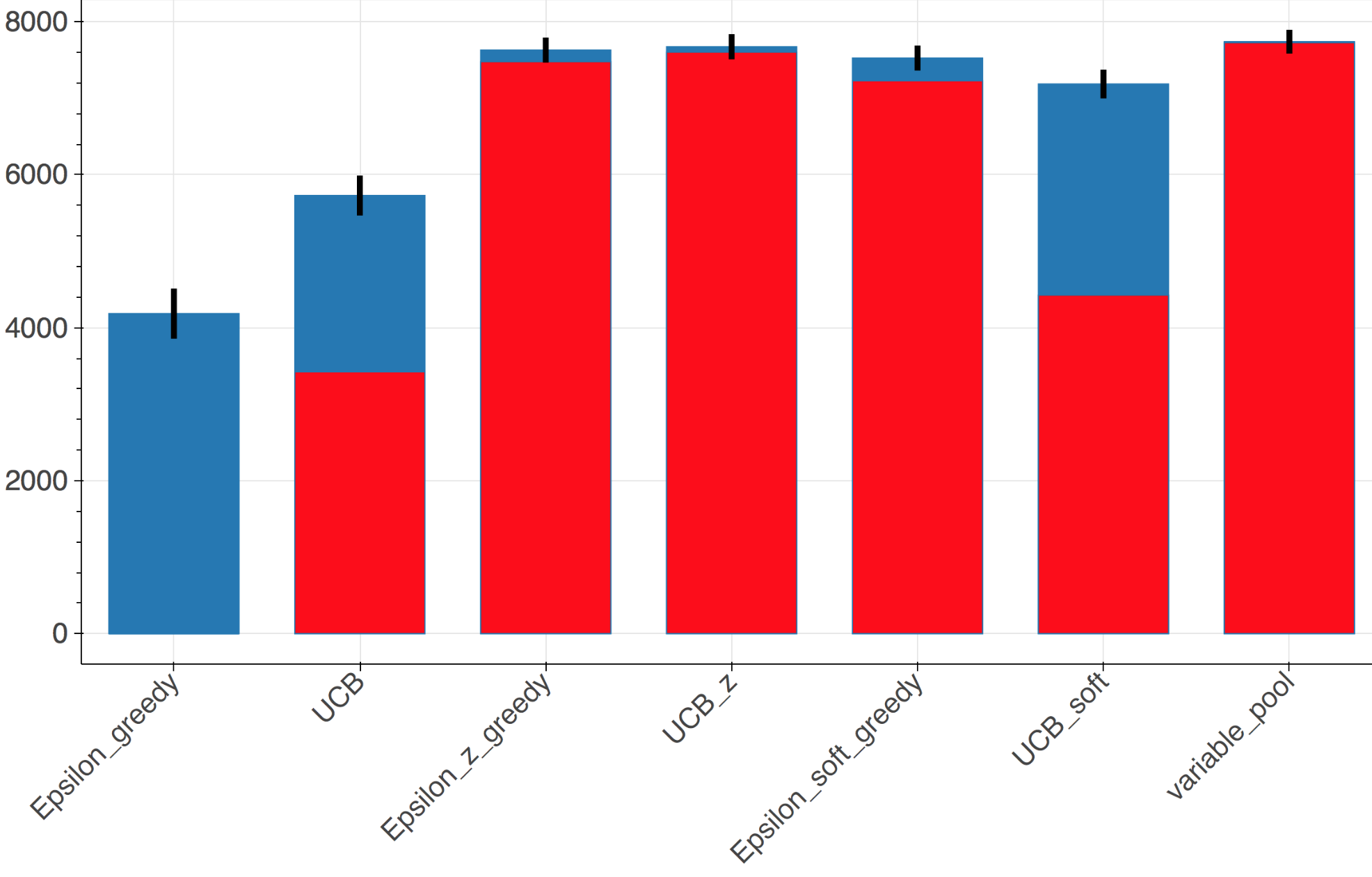} }\label{Truncated_Normal_Step_25a_500t}}\;\;
		\end{subfloatrow}
	}
	\caption{Comparison of average final rewards in games with 25 arms, 500 turns, and a Step-type greed function.}\label{Figure::25a_500t_Step}
\end{figure}

\begin{figure}[]%
	\makebox[\textwidth][c]{ %to center figures!
		\begin{subfloatrow}
			\subfloat[\small{Rewards from Bernoulli distributions. }]{{\includegraphics[width=6.7cm,height=4.0cm]{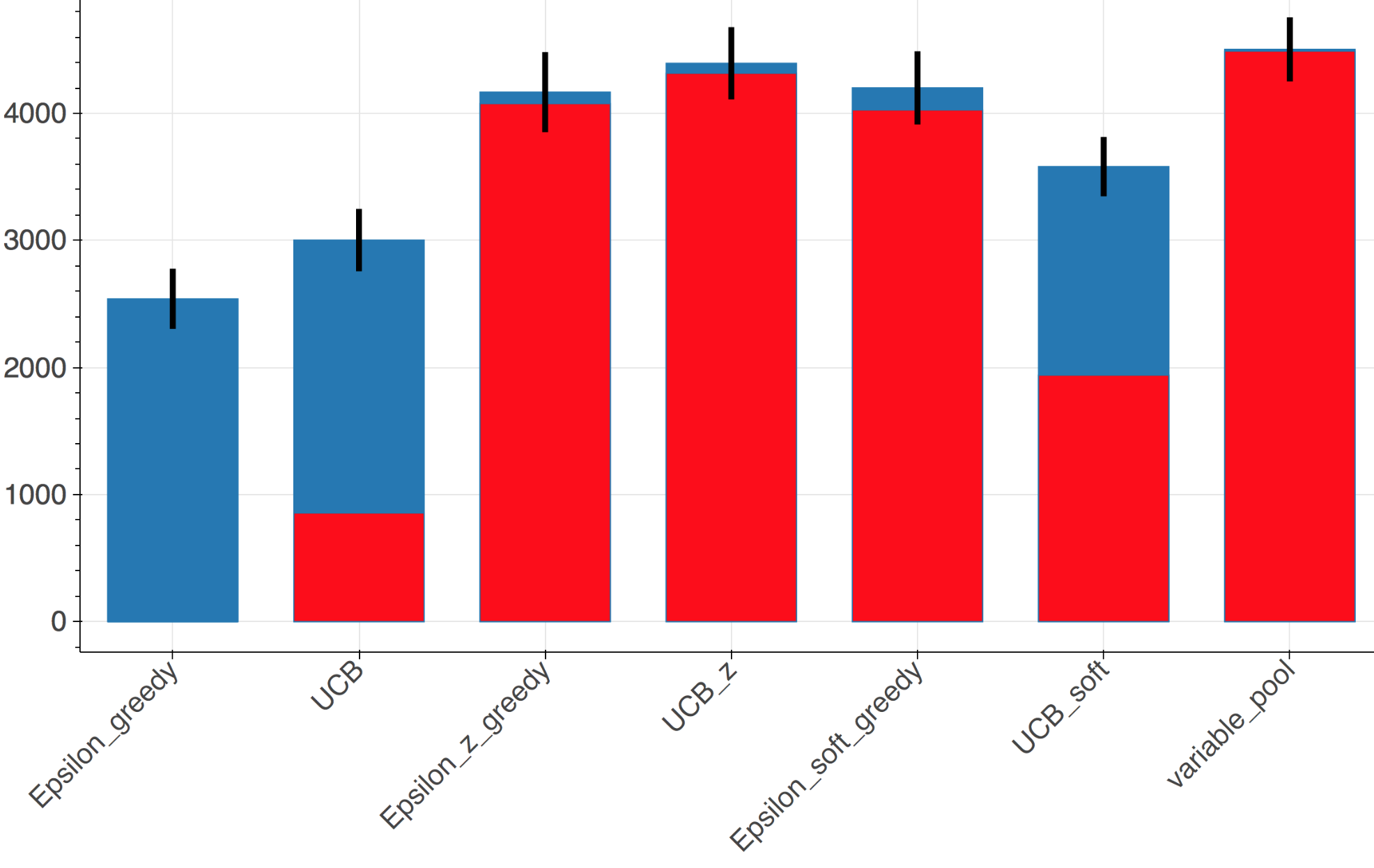} }\label{Bernoulli_Step_50a_500t}}%
			\;\;
			\qquad
			\subfloat[\small{Rewards from truncated Normal distributions.
			}]{{\includegraphics[width=6.7cm,height=4.0cm]{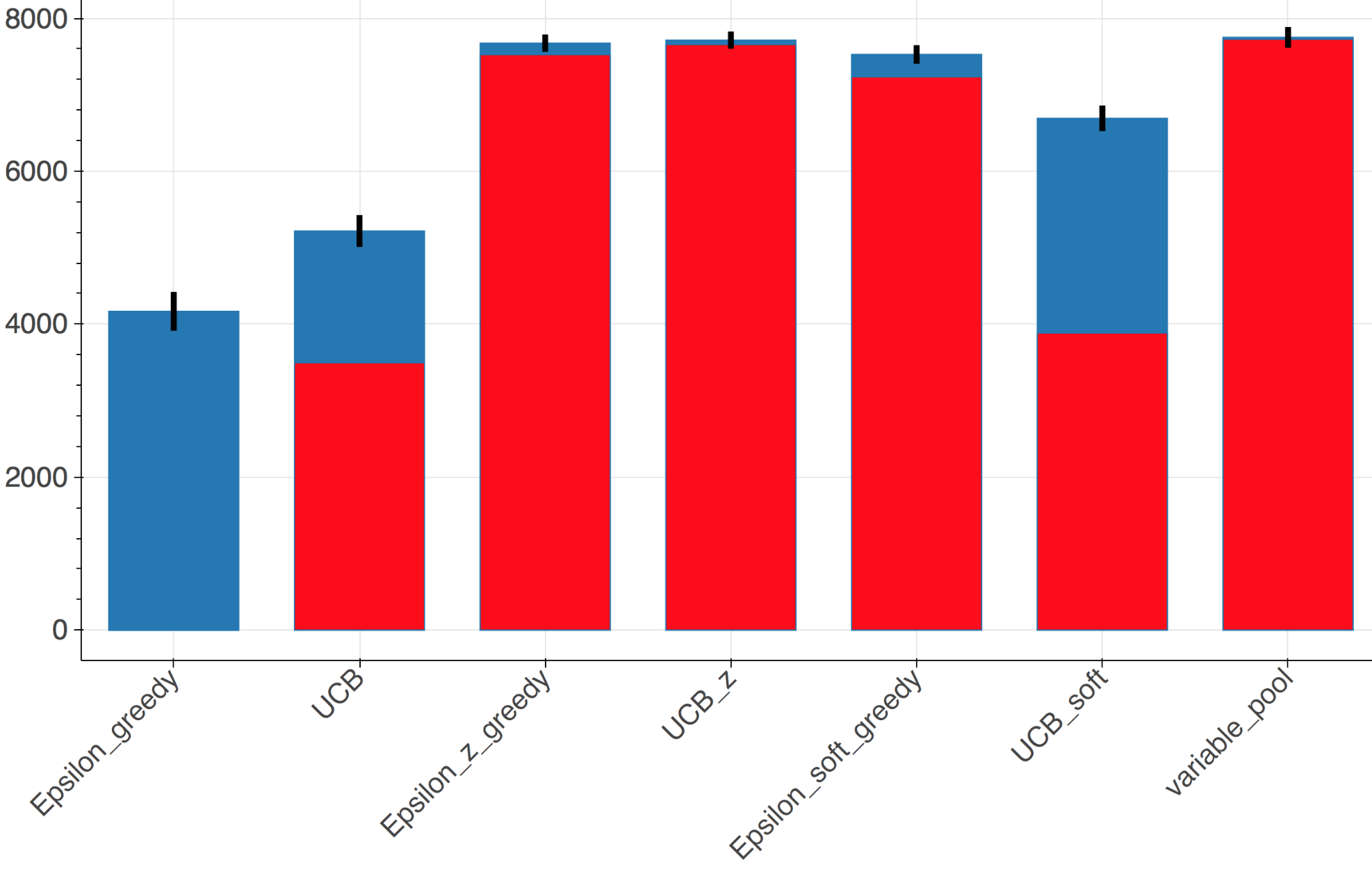} }\label{Truncated_Normal_Step_50a_500t}}\;\;
		\end{subfloatrow}
	}
	\caption{Comparison of average final rewards in games with 50 arms, 500 turns, and a Step-type greed function.}\label{Figure::50a_500t_Step}
\end{figure}

\begin{figure}[]%
	\makebox[\textwidth][c]{ %to center figures!
		\begin{subfloatrow}
			\subfloat[\small{Rewards from Bernoulli distributions. }]{{\includegraphics[width=6.7cm,height=4.0cm]{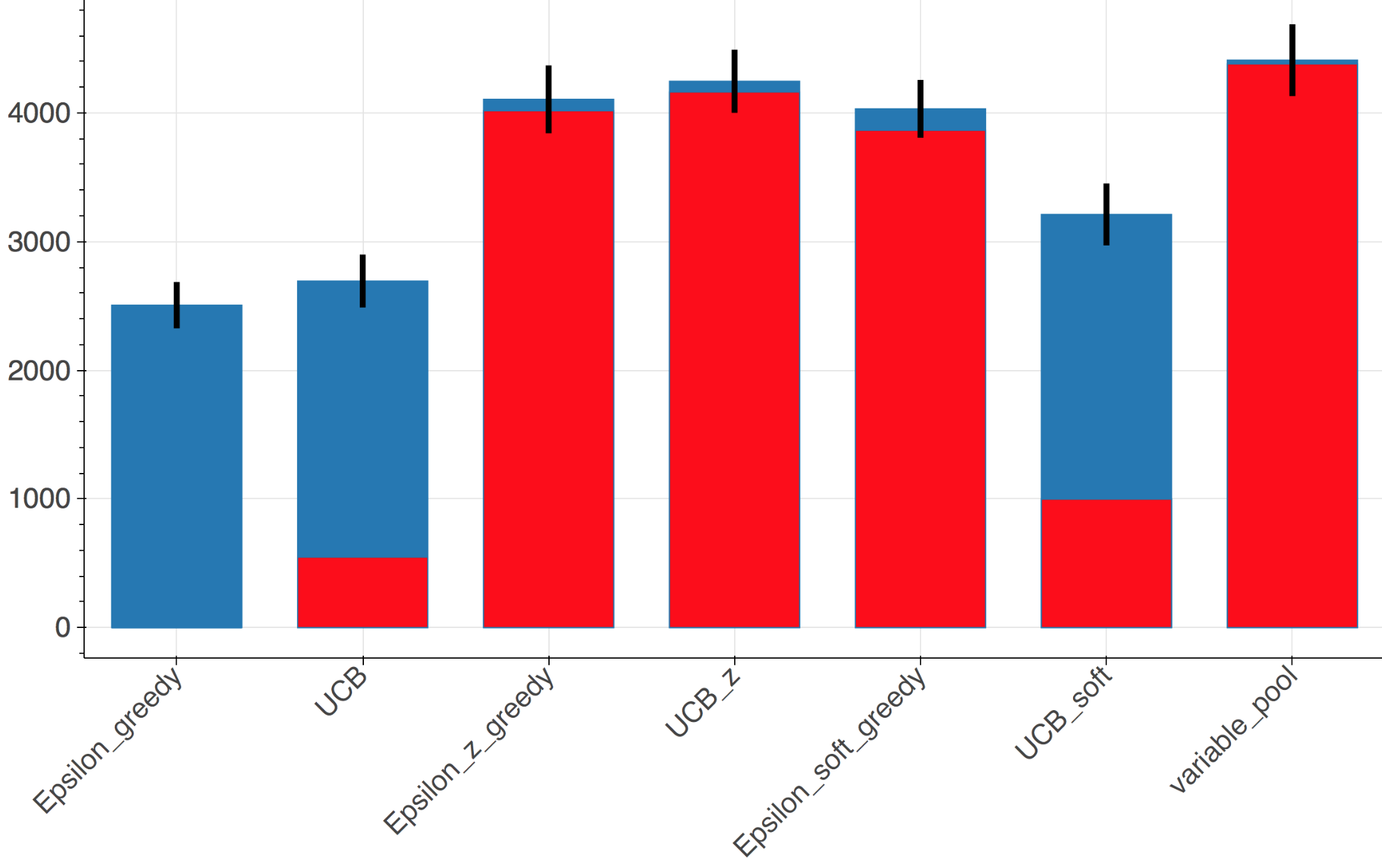} }\label{Bernoulli_Step_100a_500t}}%
			\;\;
			\qquad
			\subfloat[\small{Rewards from truncated Normal distributions.
			}]{{\includegraphics[width=6.7cm,height=4.0cm]{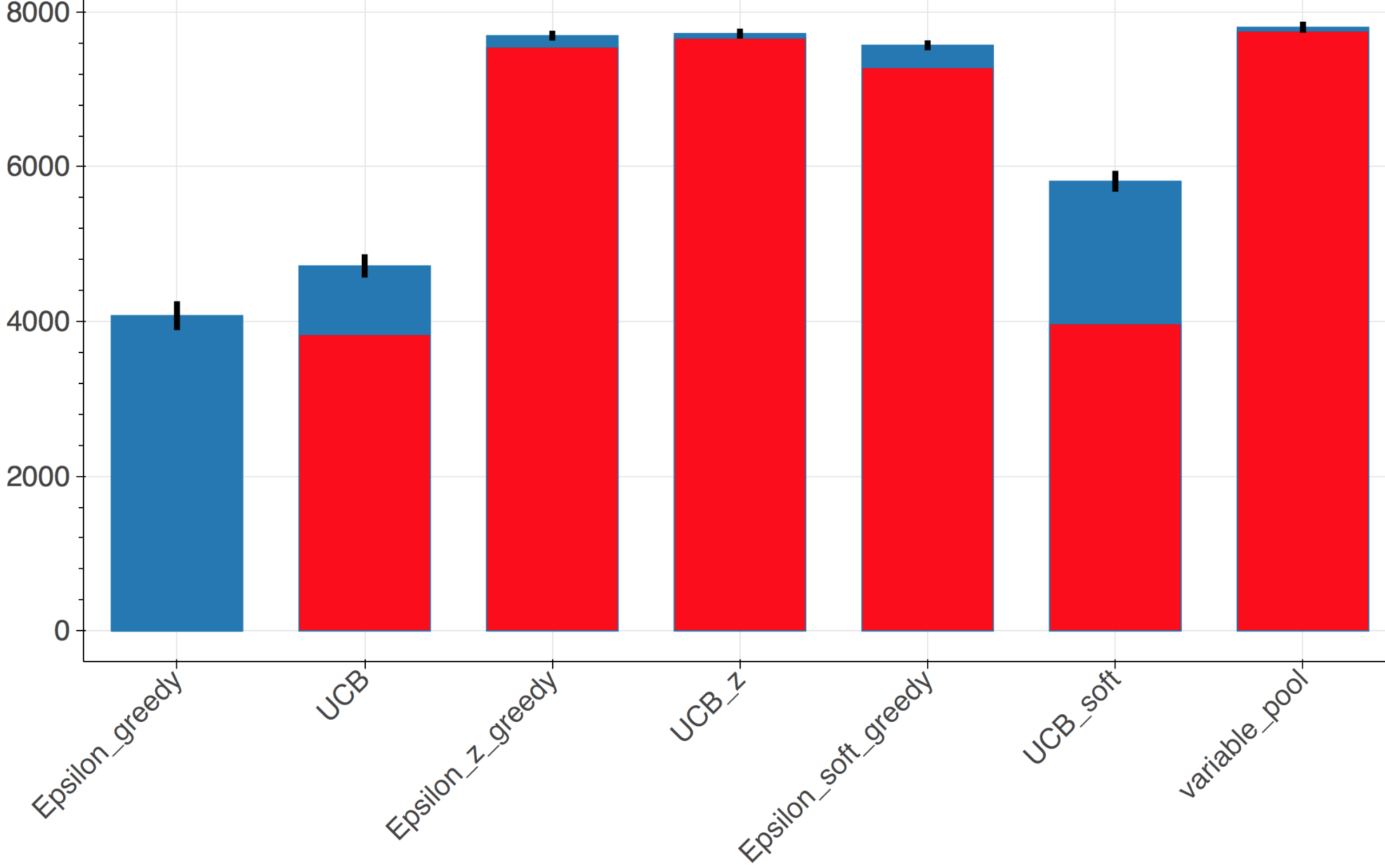} }\label{Truncated_Normal_Step_100a_500t}}\;\;
		\end{subfloatrow}
	}
	\caption{Comparison of average final rewards in games with 100 arms, 500 turns, and a Step-type greed function.}\label{Figure::100a_500t_Step}
\end{figure}

\begin{figure}[]%
	\makebox[\textwidth][c]{ %to center figures!
		\begin{subfloatrow}
			\subfloat[\small{Rewards from Bernoulli distributions. }]{{\includegraphics[width=6.7cm,height=4.0cm]{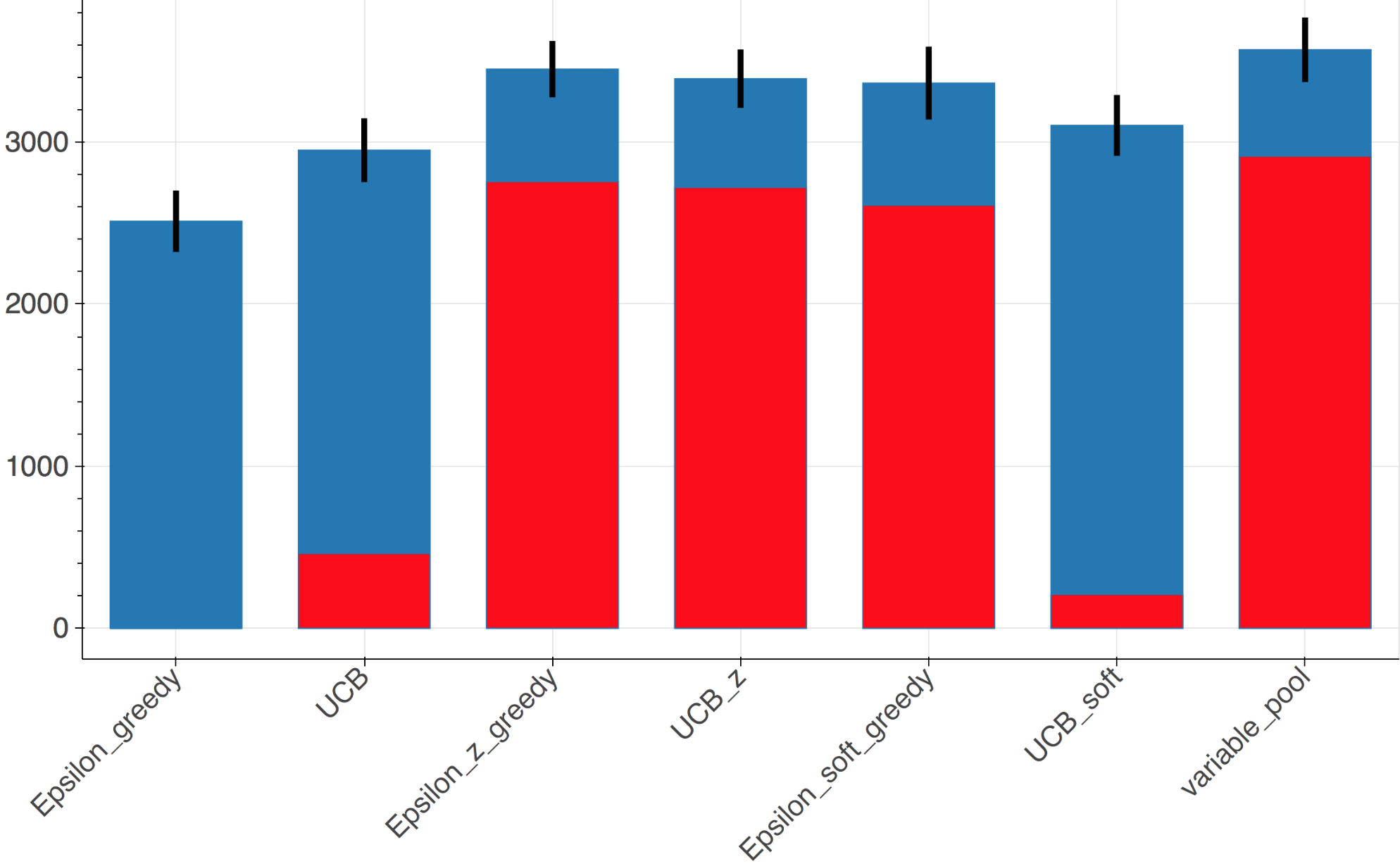} }\label{Bernoulli_Step_200a_500t}}%
			\;\;
			\qquad
			\subfloat[\small{Rewards from truncated Normal distributions.
			}]{{\includegraphics[width=6.7cm,height=4.0cm]{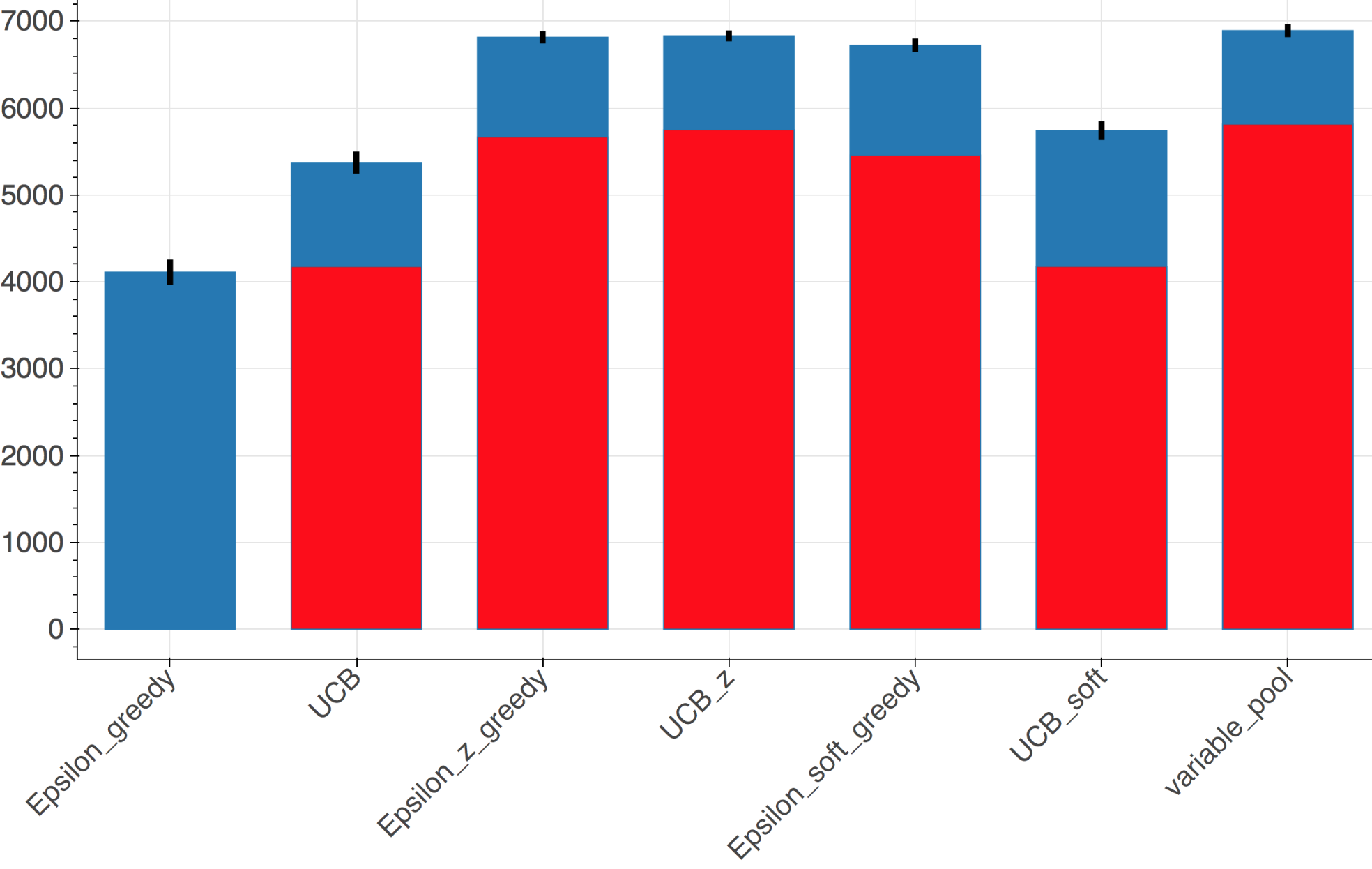} }\label{Truncated_Normal_Step_200a_500t}}\;\;
		\end{subfloatrow}
	}
	\caption{Comparison of average final rewards in games with 200 arms, 500 turns, and a Step-type greed function.}\label{Figure::200a_500t_Step}
\end{figure}

%\subsection{Step-type greed function with 1000 turns per game}

\begin{figure}[H]%
	\makebox[\textwidth][c]{ %to center figures!
		\begin{subfloatrow}
			\subfloat[\small{Rewards from Bernoulli distributions. }]{{\includegraphics[width=6.7cm,height=4.0cm]{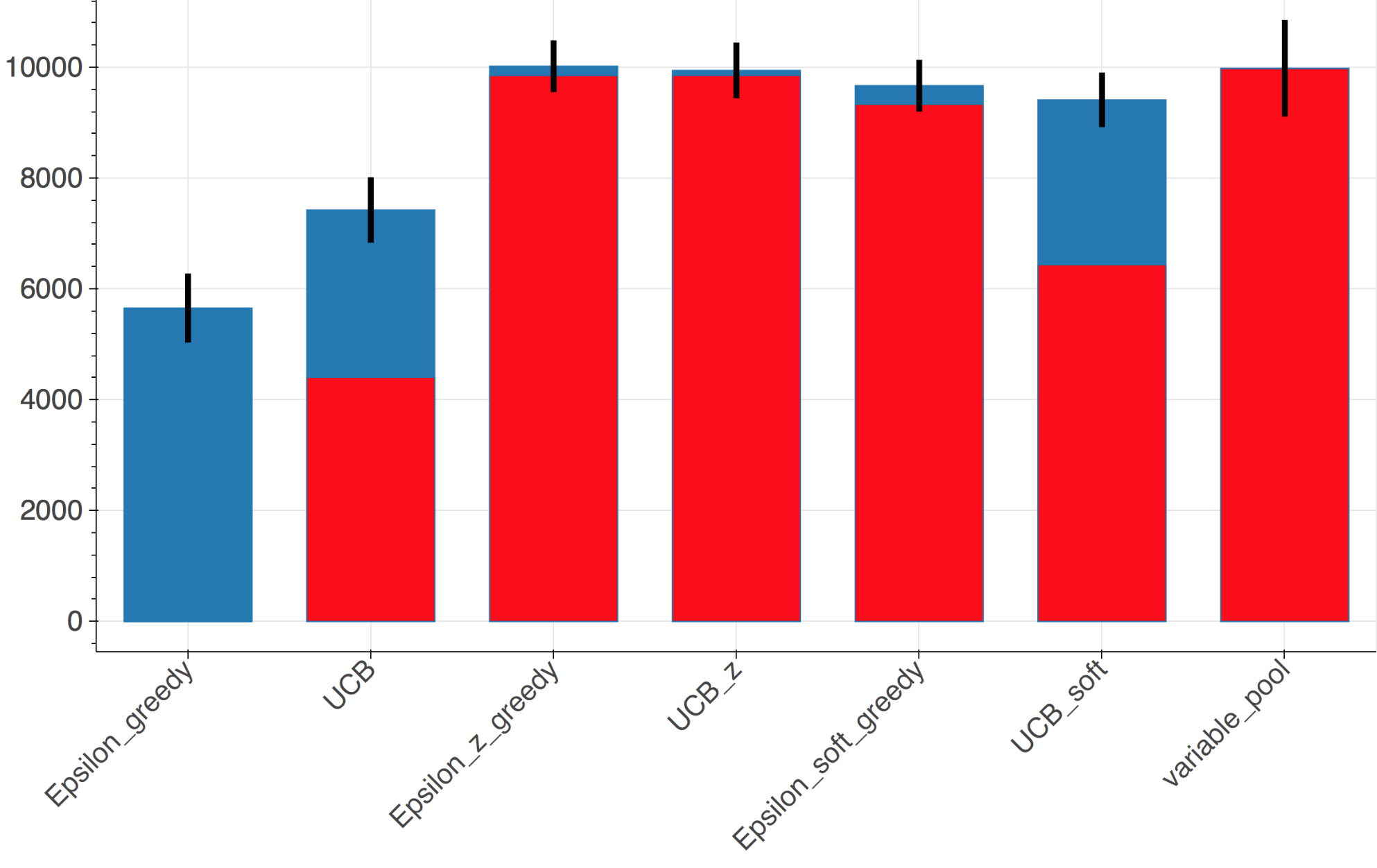} }\label{Bernoulli_Step_25a_1000t}}%
			\;\;
			\qquad
			\subfloat[\small{Rewards from truncated Normal distributions.
			}]{{\includegraphics[width=6.7cm,height=4.0cm]{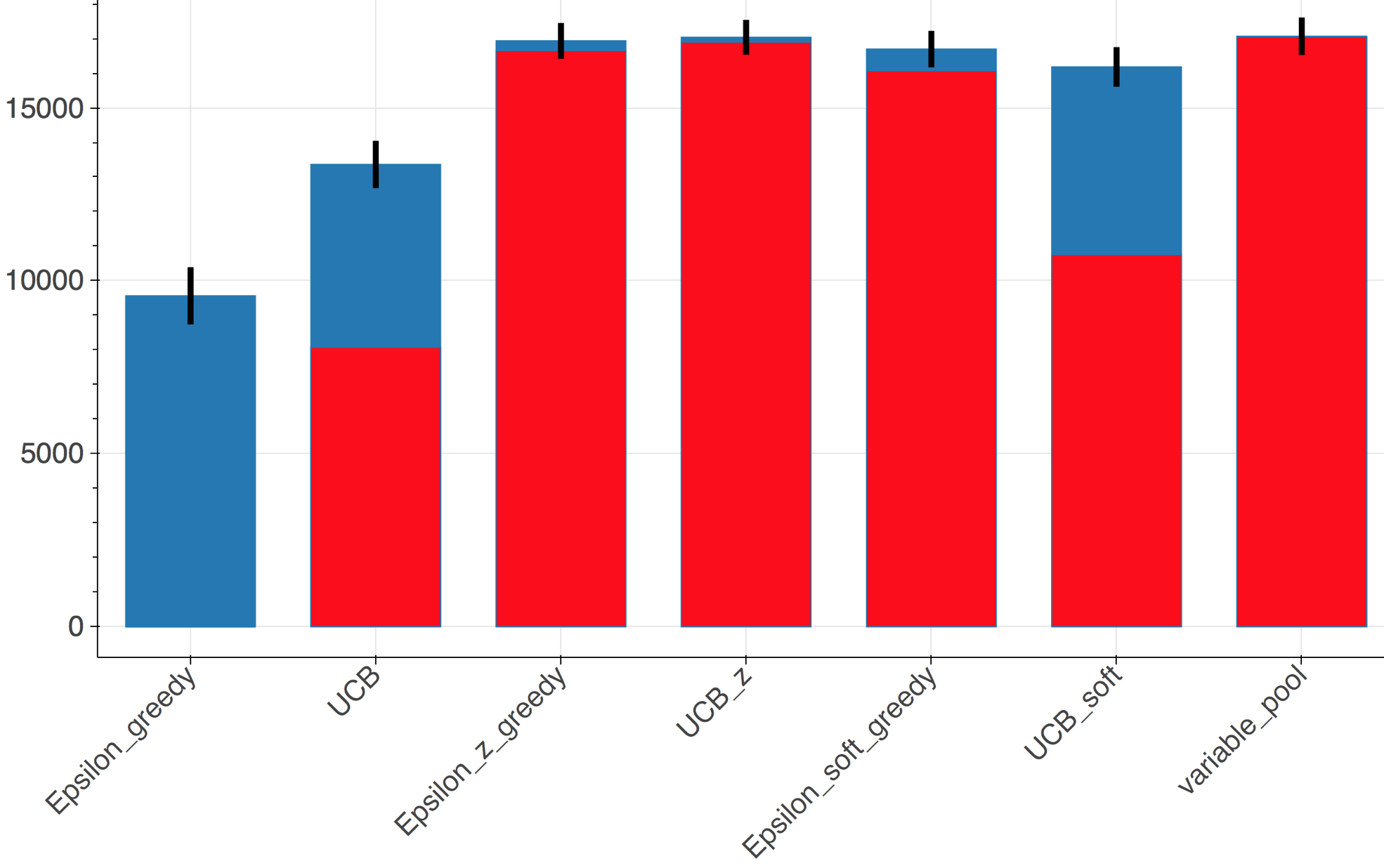} }\label{Truncated_Normal_Step_25a_1000t}}\;\;
		\end{subfloatrow}
	}
	\caption{Comparison of average final rewards in games with 25 arms, 1000 turns, and a Step-type greed function.}\label{Figure::25a_1000t_Step}
\end{figure}

\begin{figure}[H]%
	\makebox[\textwidth][c]{ %to center figures!
		\begin{subfloatrow}
			\subfloat[\small{Rewards from Bernoulli distributions. }]{{\includegraphics[width=6.7cm,height=4.0cm]{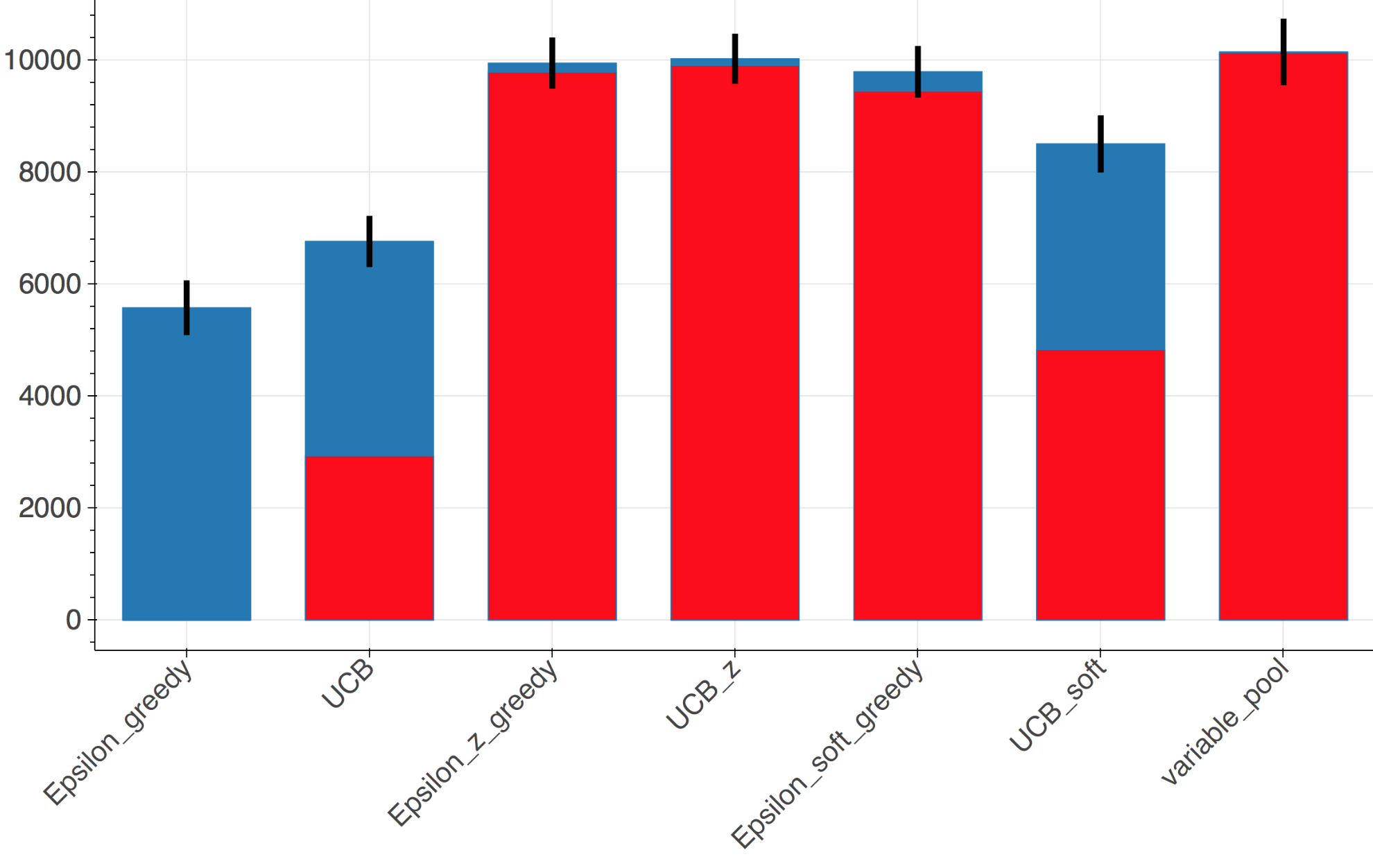} }\label{Bernoulli_Step_50a_1000t}}%
			\;\;
			\qquad
			\subfloat[\small{Rewards from truncated Normal distributions.
			}]{{\includegraphics[width=6.7cm,height=4.0cm]{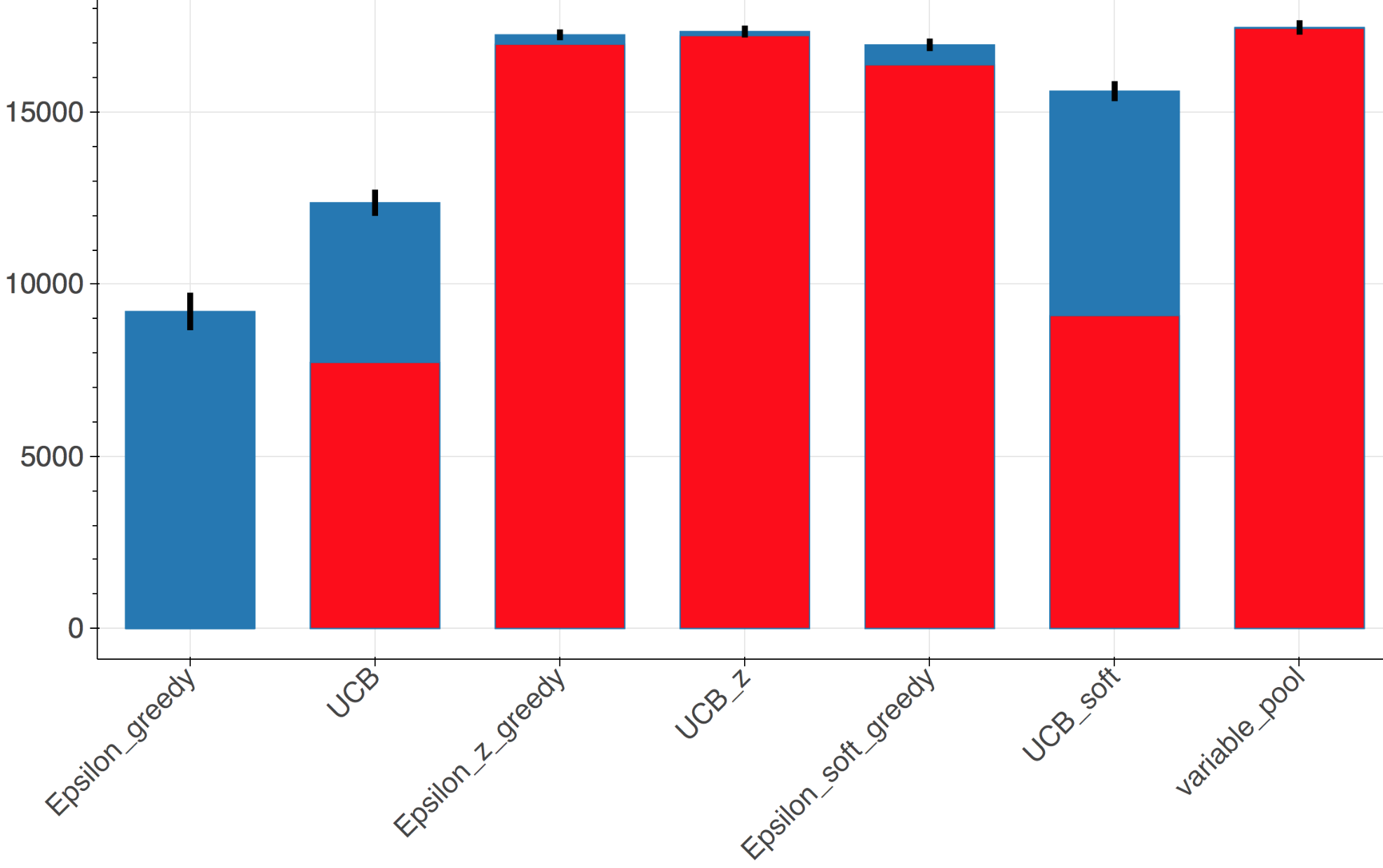} }\label{Truncated_Normal_Step_50a_1000t}}\;\;
		\end{subfloatrow}
	}
	\caption{Comparison of average final rewards in games with 50 arms, 1000 turns, and a Step-type greed function.}\label{Figure::50a_1000t_Step}
\end{figure}

\begin{figure}[H]%
	\makebox[\textwidth][c]{ %to center figures!
		\begin{subfloatrow}
			\subfloat[\small{Rewards from Bernoulli distributions. }]{{\includegraphics[width=6.7cm,height=4.0cm]{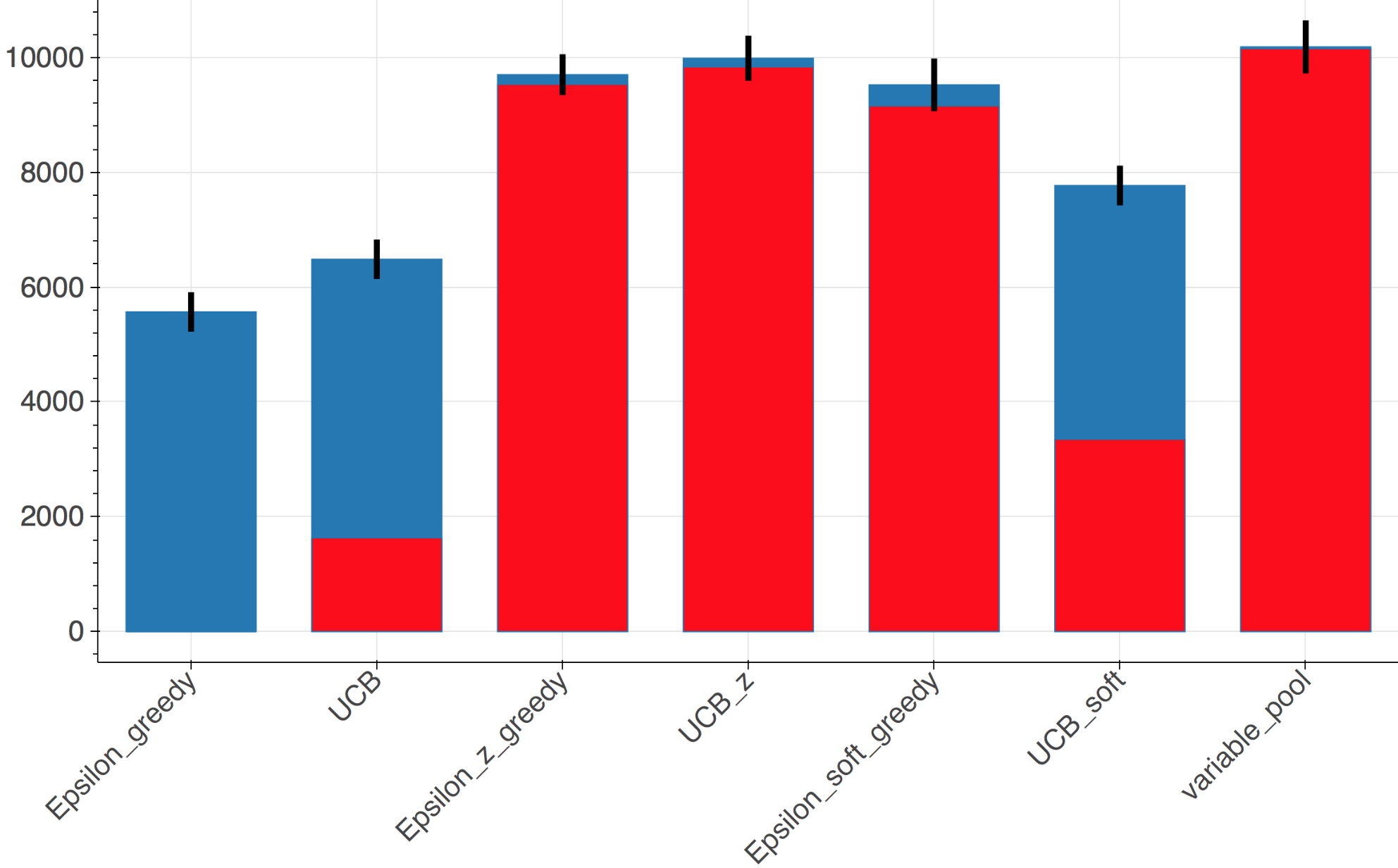} }\label{Bernoulli_Step_100a_1000t}}%
			\;\;
			\qquad
			\subfloat[\small{Rewards from truncated Normal distributions.
			}]{{\includegraphics[width=6.7cm,height=4.0cm]{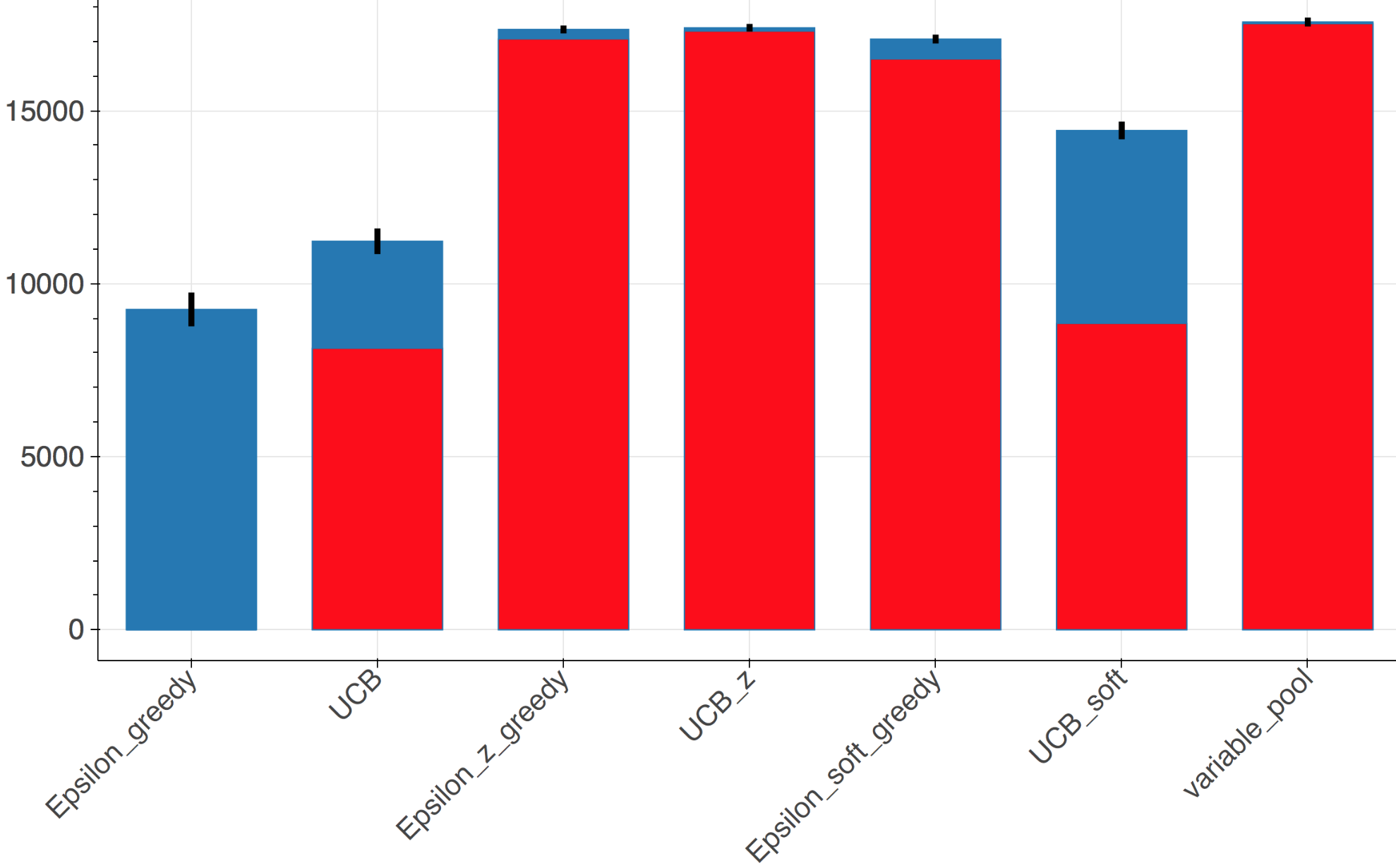} }\label{Truncated_Normal_Step_100a_1000t}}\;\;
		\end{subfloatrow}
	}
	\caption{Comparison of average final rewards in games with 100 arms, 1000 turns, and a Step-type greed function.}\label{Figure::100a_1000t_Step}
\end{figure}

\begin{figure}[H]%
	\makebox[\textwidth][c]{ %to center figures!
		\begin{subfloatrow}
			\subfloat[\small{Rewards from Bernoulli distributions. }]{{\includegraphics[width=6.7cm,height=4.0cm]{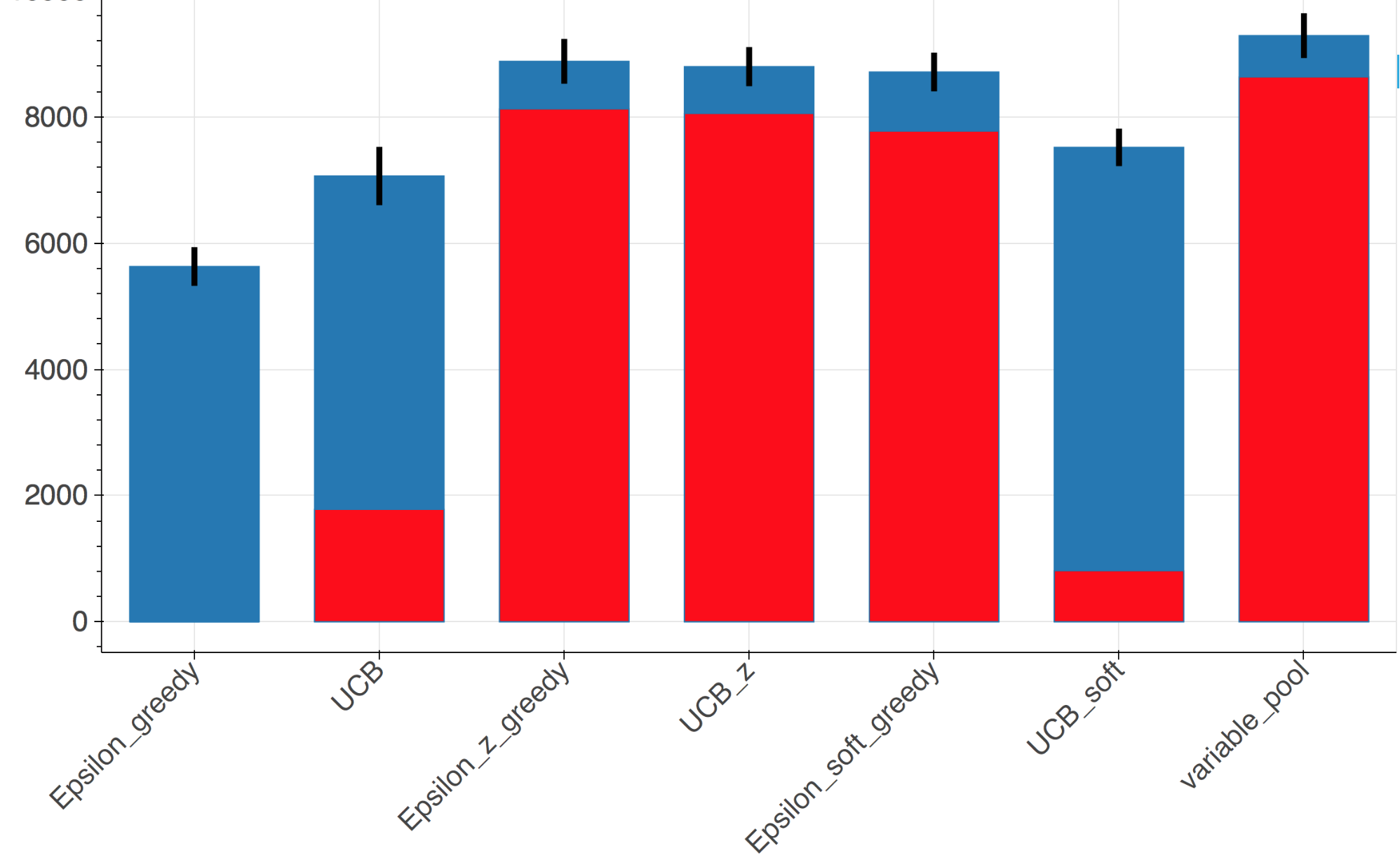} }\label{Bernoulli_Step_200a_1000t}}%
			\;\;
			\qquad
			\subfloat[\small{Rewards from truncated Normal distributions.
			}]{{\includegraphics[width=6.7cm,height=4.0cm]{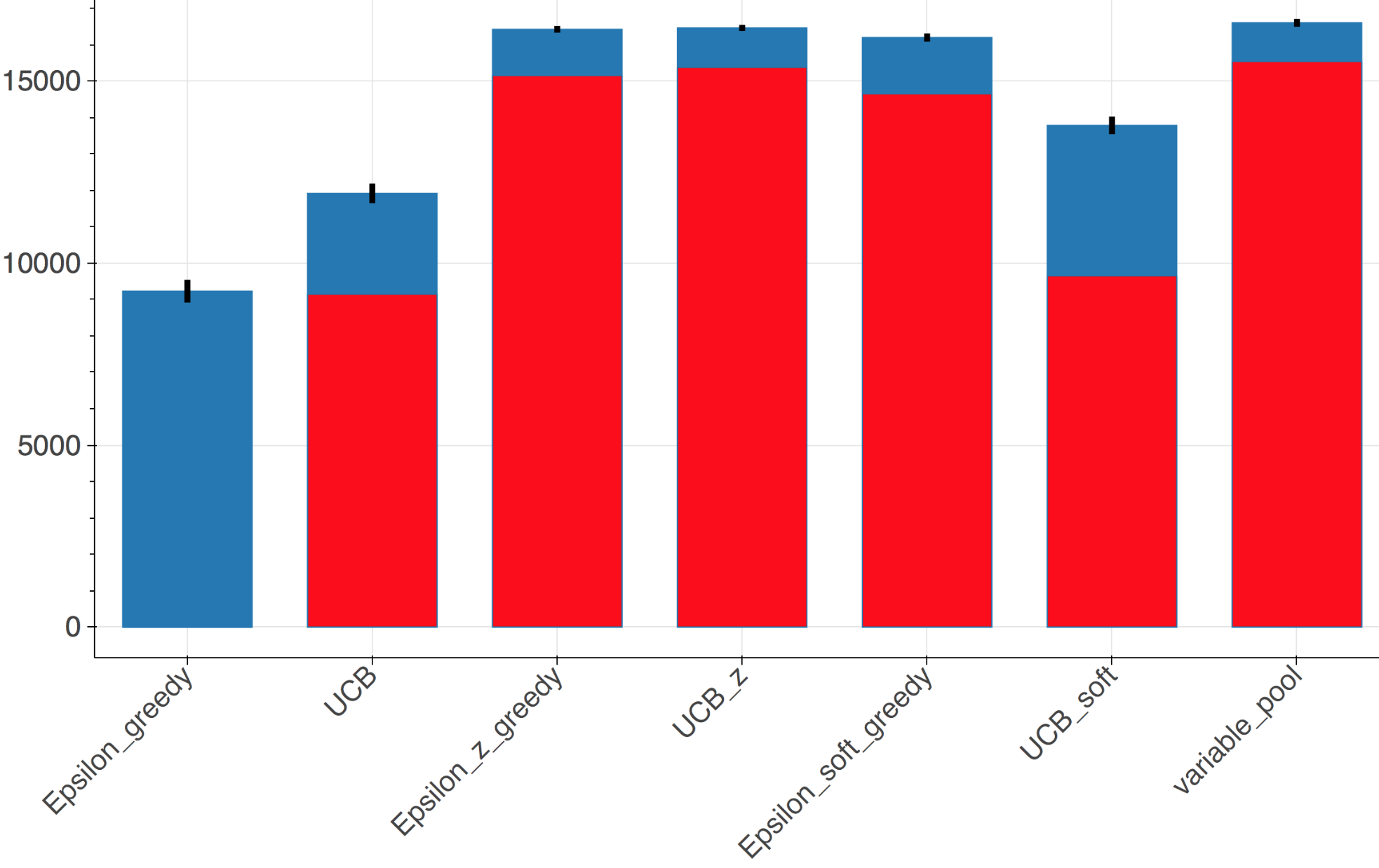} }\label{Truncated_Normal_Step_200a_1000t}}\;\;
		\end{subfloatrow}
	}
	\caption{Comparison of average final rewards in games with 200 arms, 1000 turns, and a Step-type greed function.}\label{Figure::200a_1000t_Step}
\end{figure}

%\subsection{Step-type greed function with 1500 turns per game}

\begin{figure}[H]%
	\makebox[\textwidth][c]{ %to center figures!
		\begin{subfloatrow}
			\subfloat[\small{Rewards from Bernoulli distributions. }]{{\includegraphics[width=6.7cm,height=4.0cm]{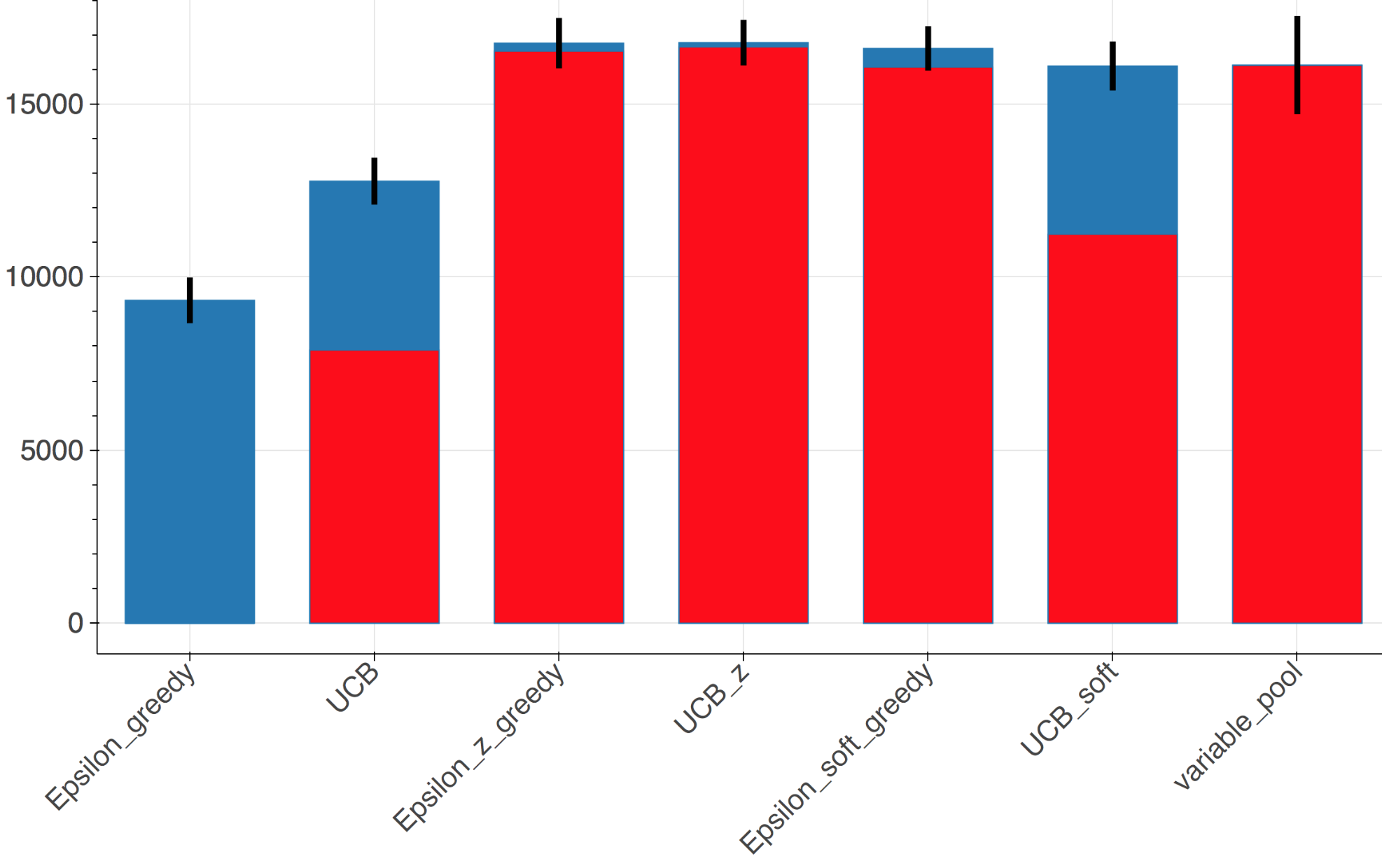} }\label{Bernoulli_Step_25a_1500t}}%
			\;\;
			\qquad
			\subfloat[\small{Rewards from truncated Normal distributions.
			}]{{\includegraphics[width=6.7cm,height=4.0cm]{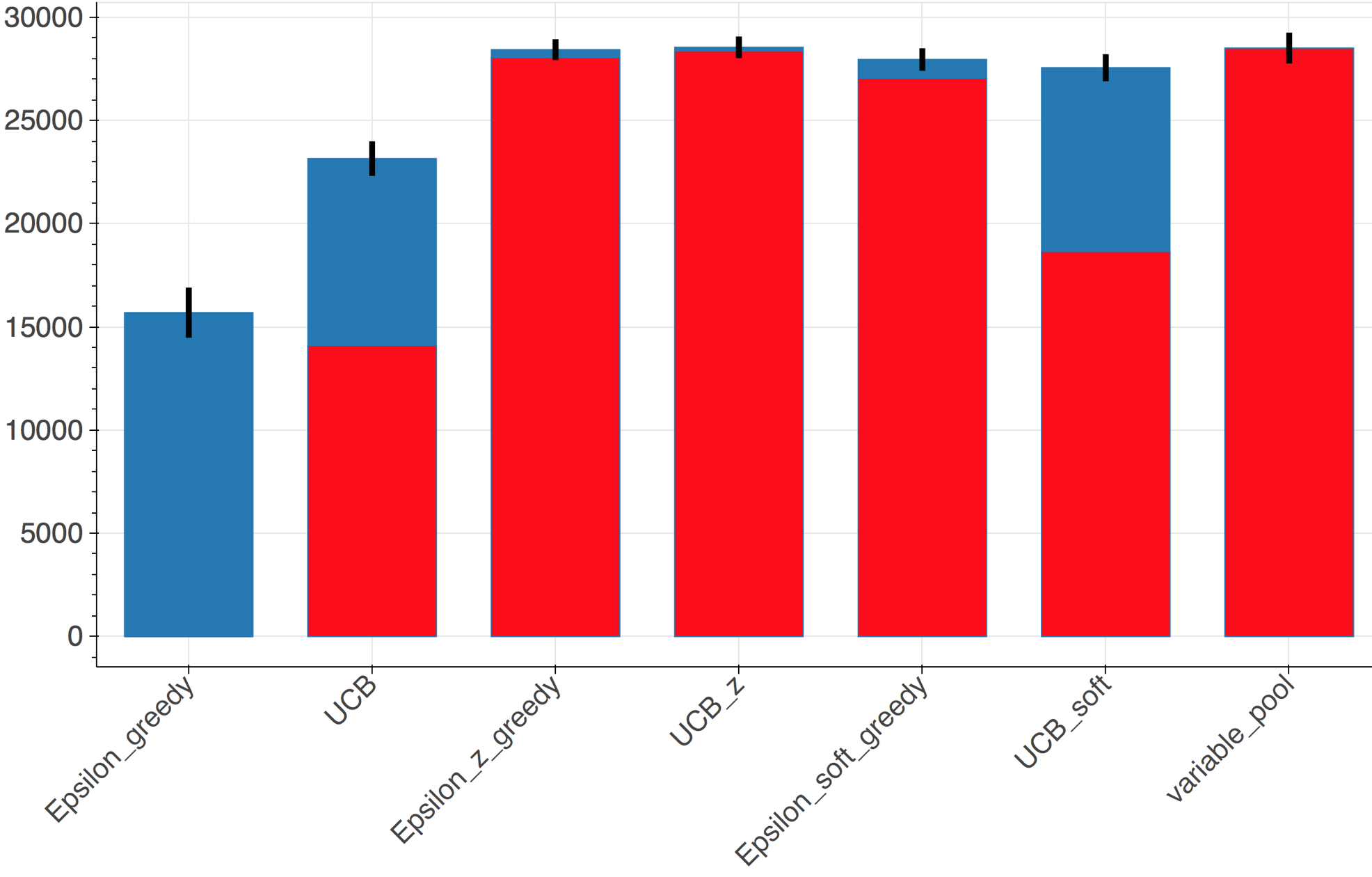} }\label{Truncated_Normal_Step_25a_1500t}}\;\;
		\end{subfloatrow}
	}
	\caption{Comparison of average final rewards in games with 25 arms, 1500 turns, and a Step-type greed function.}\label{Figure::25a_1500t_Step}
\end{figure}

\begin{figure}[H]%
	\makebox[\textwidth][c]{ %to center figures!
		\begin{subfloatrow}
			\subfloat[\small{Rewards from Bernoulli distributions. }]{{\includegraphics[width=6.7cm,height=4.0cm]{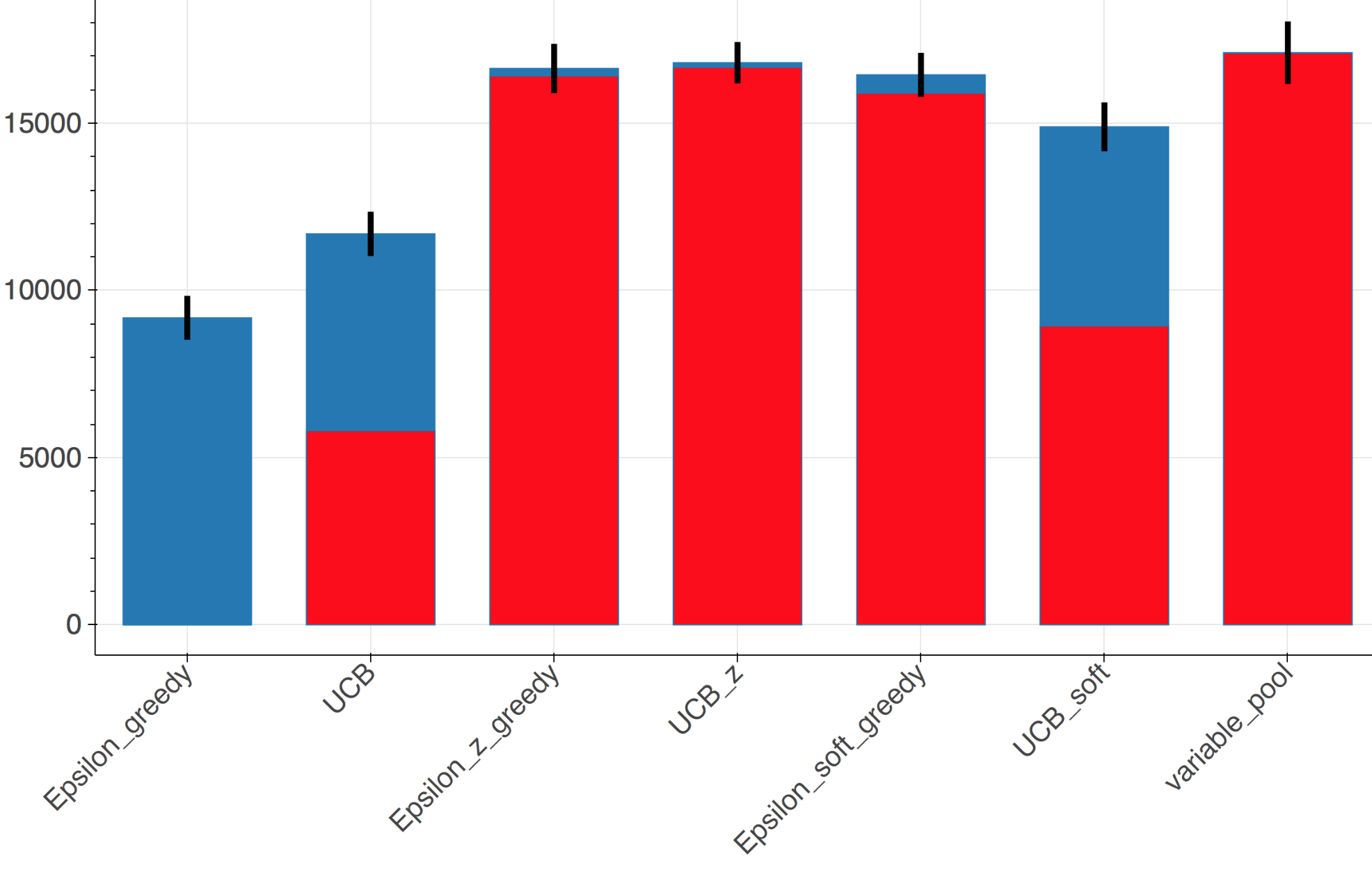} }\label{Bernoulli_Step_50a_1500t}}%
			\;\;
			\qquad
			\subfloat[\small{Rewards from truncated Normal distributions.
			}]{{\includegraphics[width=6.7cm,height=4.0cm]{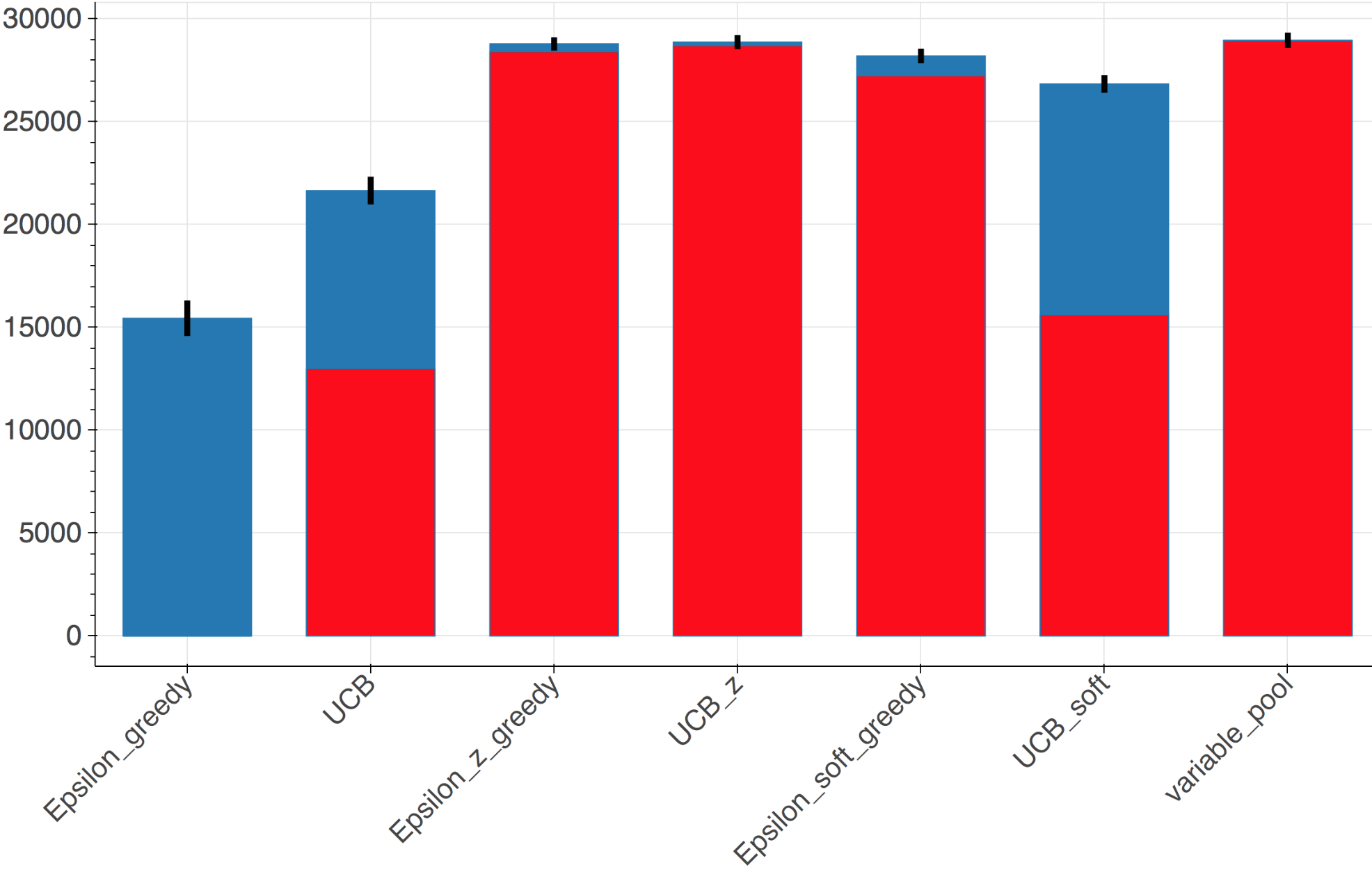} }\label{Truncated_Normal_Step_50a_1500t}}\;\;
		\end{subfloatrow}
	}
	\caption{Comparison of average final rewards in games with 50 arms, 1500 turns, and a Step-type greed function.}\label{Figure::50a_1500t_Step}
\end{figure}

\begin{figure}[H]%
	\makebox[\textwidth][c]{ %to center figures!
		\begin{subfloatrow}
			\subfloat[\small{Rewards from Bernoulli distributions. }]{{\includegraphics[width=6.7cm,height=4.0cm]{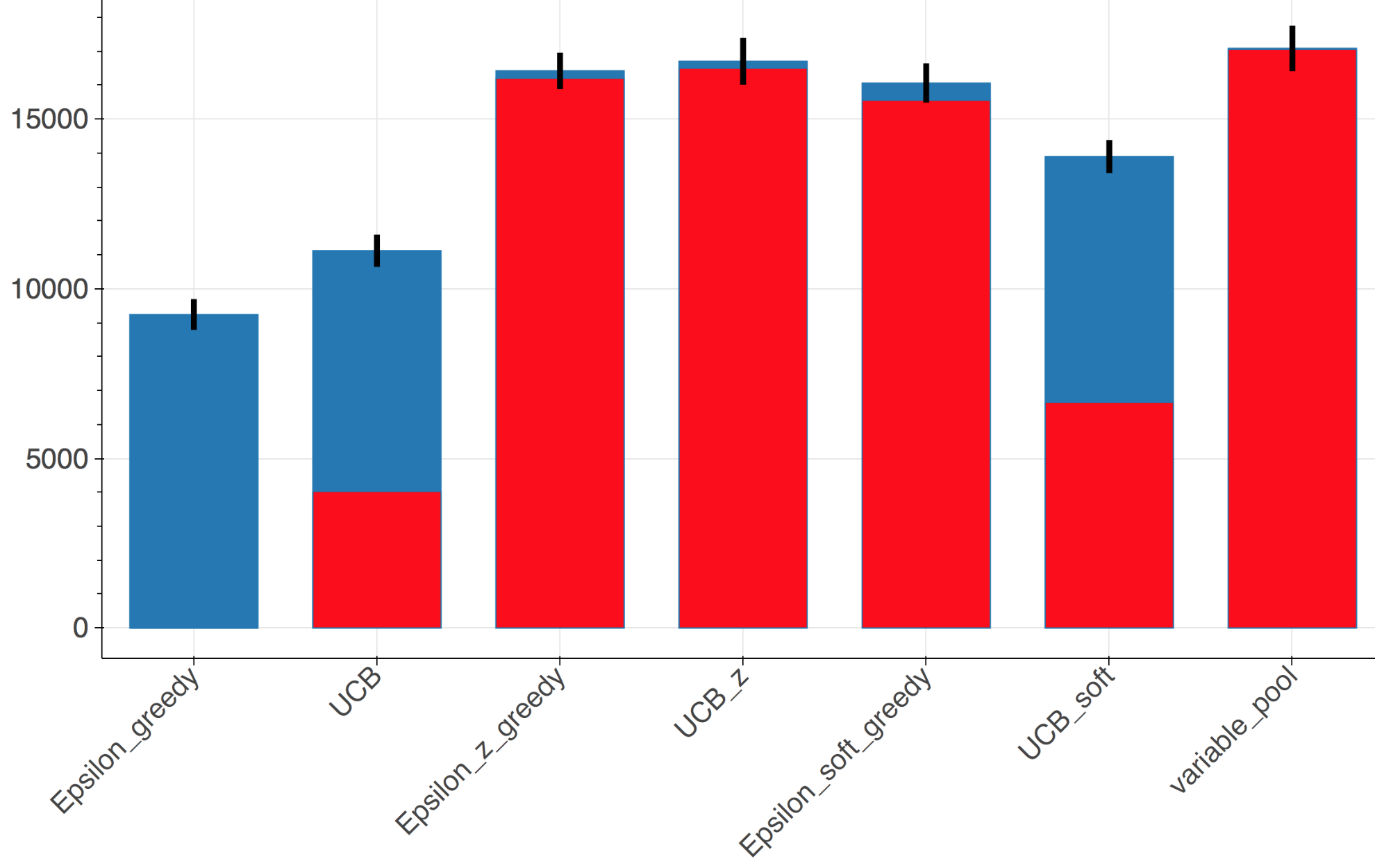} }\label{Bernoulli_Step_100a_1500t}}%
			\;\;
			\qquad
			\subfloat[\small{Rewards from truncated Normal distributions.
			}]{{\includegraphics[width=6.7cm,height=4.0cm]{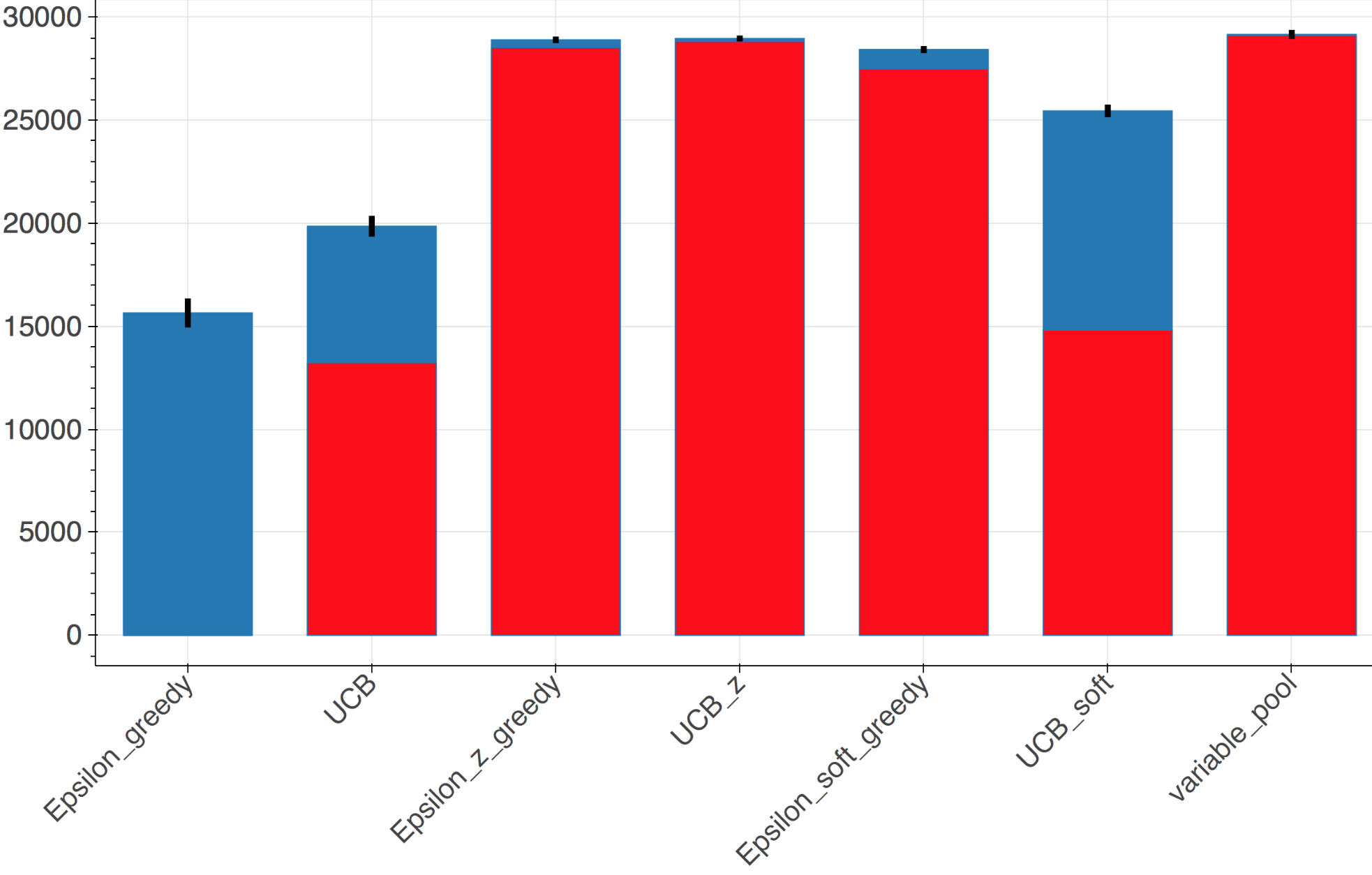} }\label{Truncated_Normal_Step_100a_1500t}}\;\;
		\end{subfloatrow}
	}
	\caption{Comparison of average final rewards in games with 100 arms, 1500 turns, and a Step-type greed function.}\label{Figure::100a_1500t_Step}
\end{figure}

\begin{figure}[H]%
	\makebox[\textwidth][c]{ %to center figures!
		\begin{subfloatrow}
			\subfloat[\small{Rewards from Bernoulli distributions. }]{{\includegraphics[width=6.7cm,height=4.0cm]{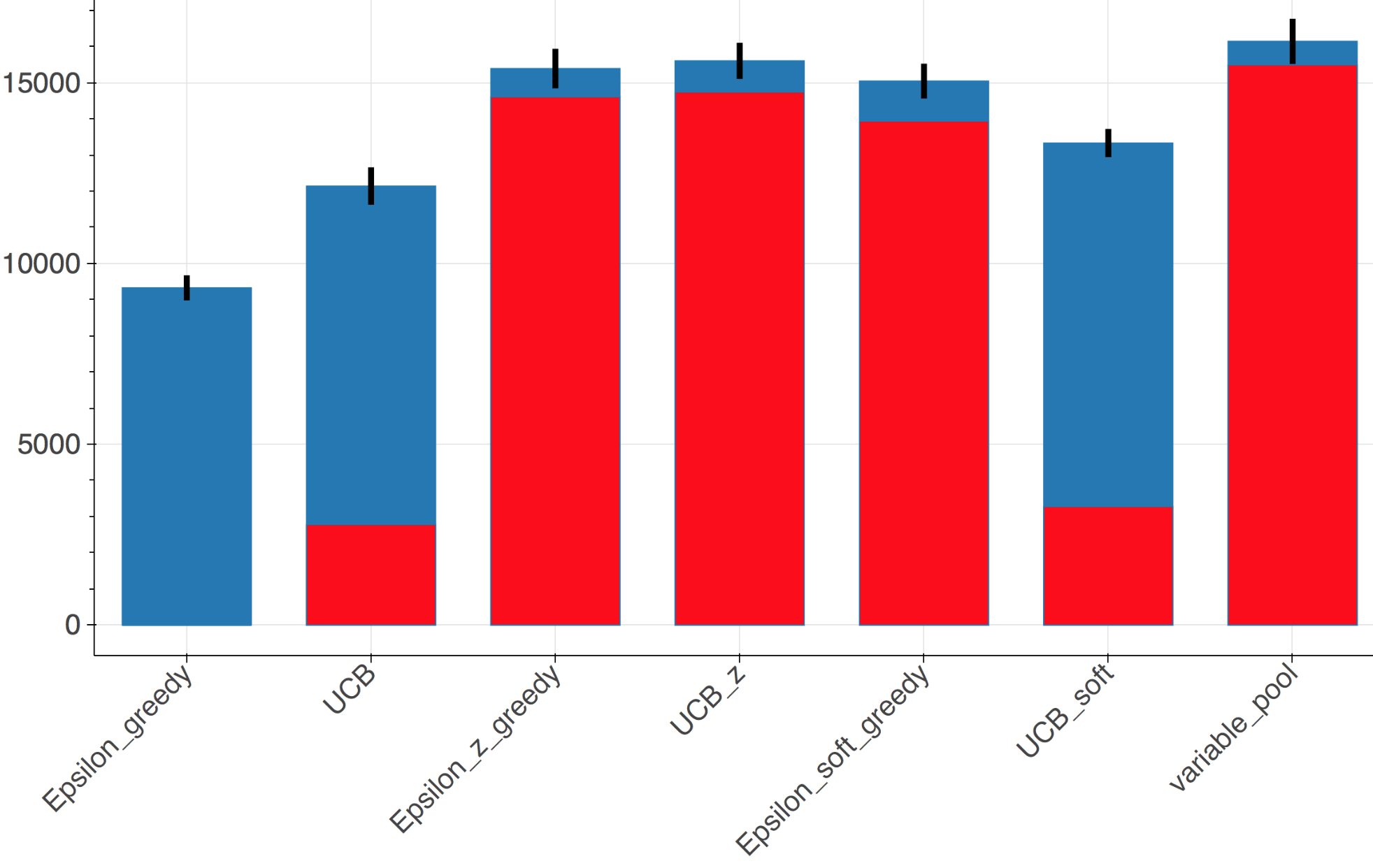} }\label{Bernoulli_Step_200a_1500t}}%
			\;\;
			\qquad
			\subfloat[\small{Rewards from truncated Normal distributions.
			}]{{\includegraphics[width=6.7cm,height=4.0cm]{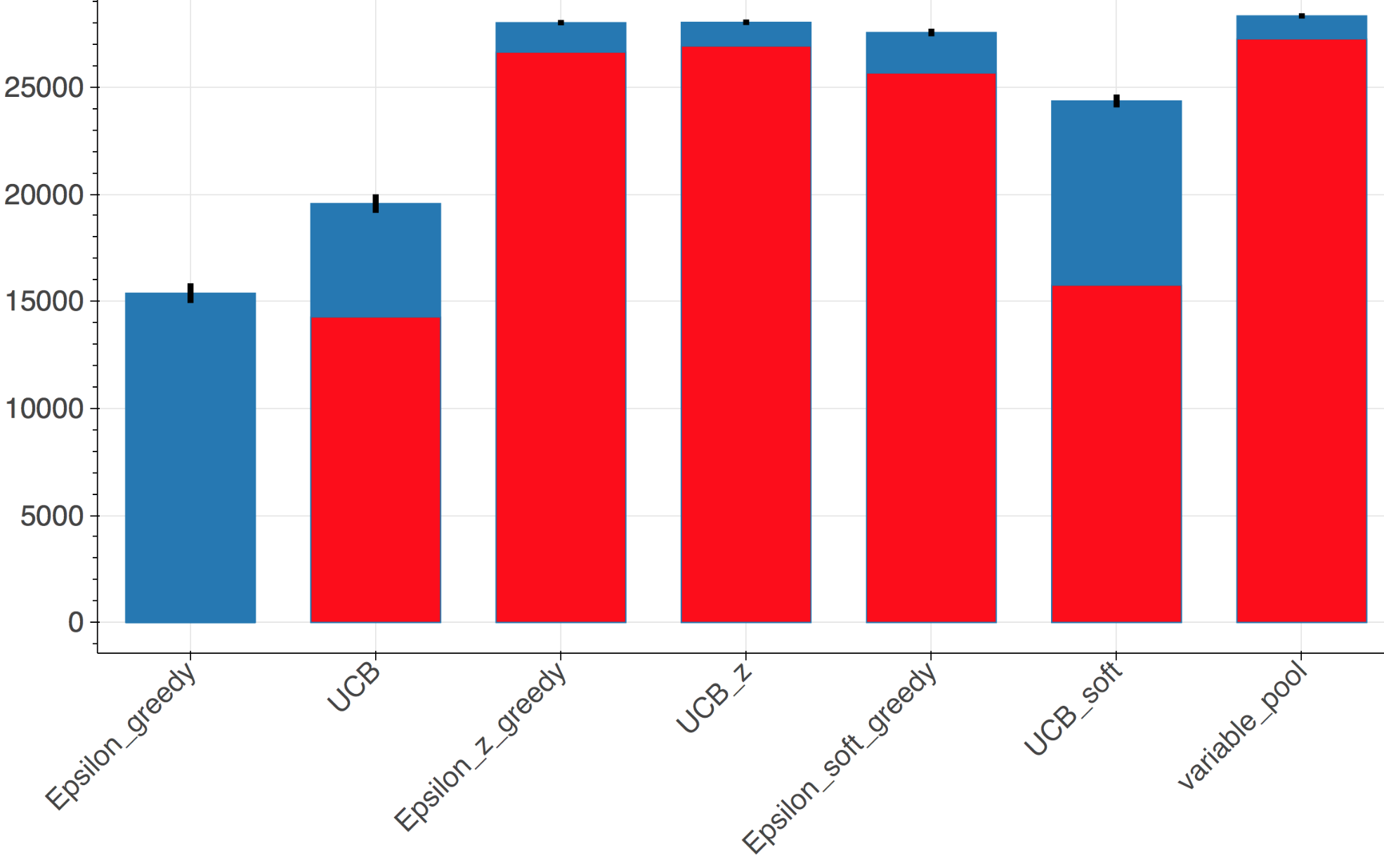} }\label{Truncated_Normal_Step_200a_1500t}}\;\;
		\end{subfloatrow}
	}
	\caption{Comparison of average final rewards in games with 200 arms, 1500 turns, and a Step-type greed function.}\label{Figure::200a_1500t_Step}
\end{figure}

%\subsection{Cumulative Reward increase when regulating greed over time compared to the (smarter) $\varepsilon$-greedy algorithm and the (smarter) UCB algorithm with a Step-type greed function.}

\begin{figure}[H]
	\caption{Average increase in rewards (coming from Bernoulli distributions) compared to the (smarter) version of the $\varepsilon$-greedy algorithm (Algorithm \ref{Algorithm::epsilon_slightly_smarter}) with a Step-type greed function. }
	\centering
	\includegraphics[width=0.95\textwidth]{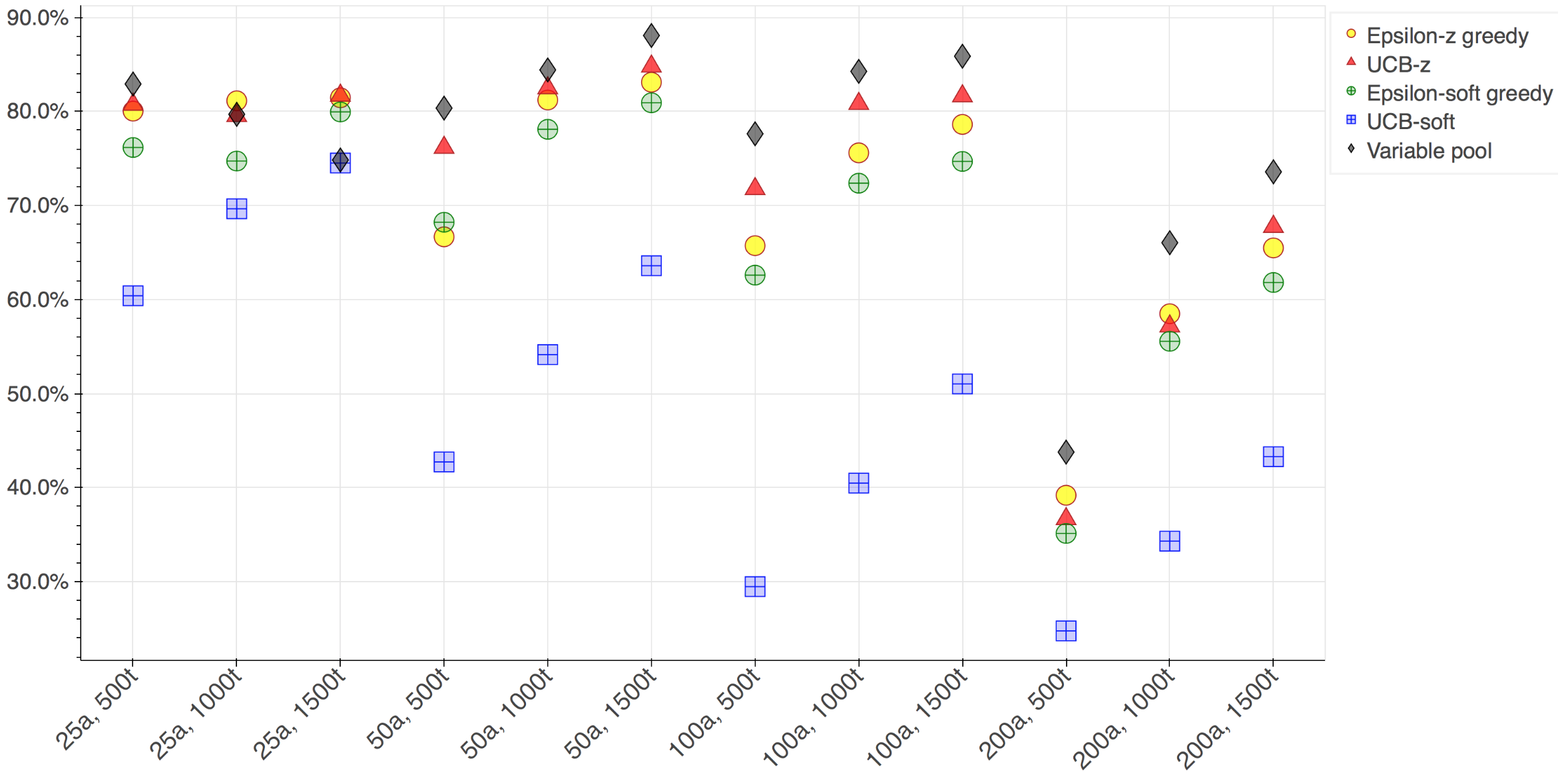}
	\label{BES}
\end{figure}

\begin{figure}[H]
	\caption{Average increase in rewards (coming from Bernoulli distributions) compared to the (smarter) version of the UCB algorithm (Algorithm \ref{Algorithm::UCB_slightly_smarter}) with a Step-type greed function. }
	\centering
	\includegraphics[width=0.95\textwidth]{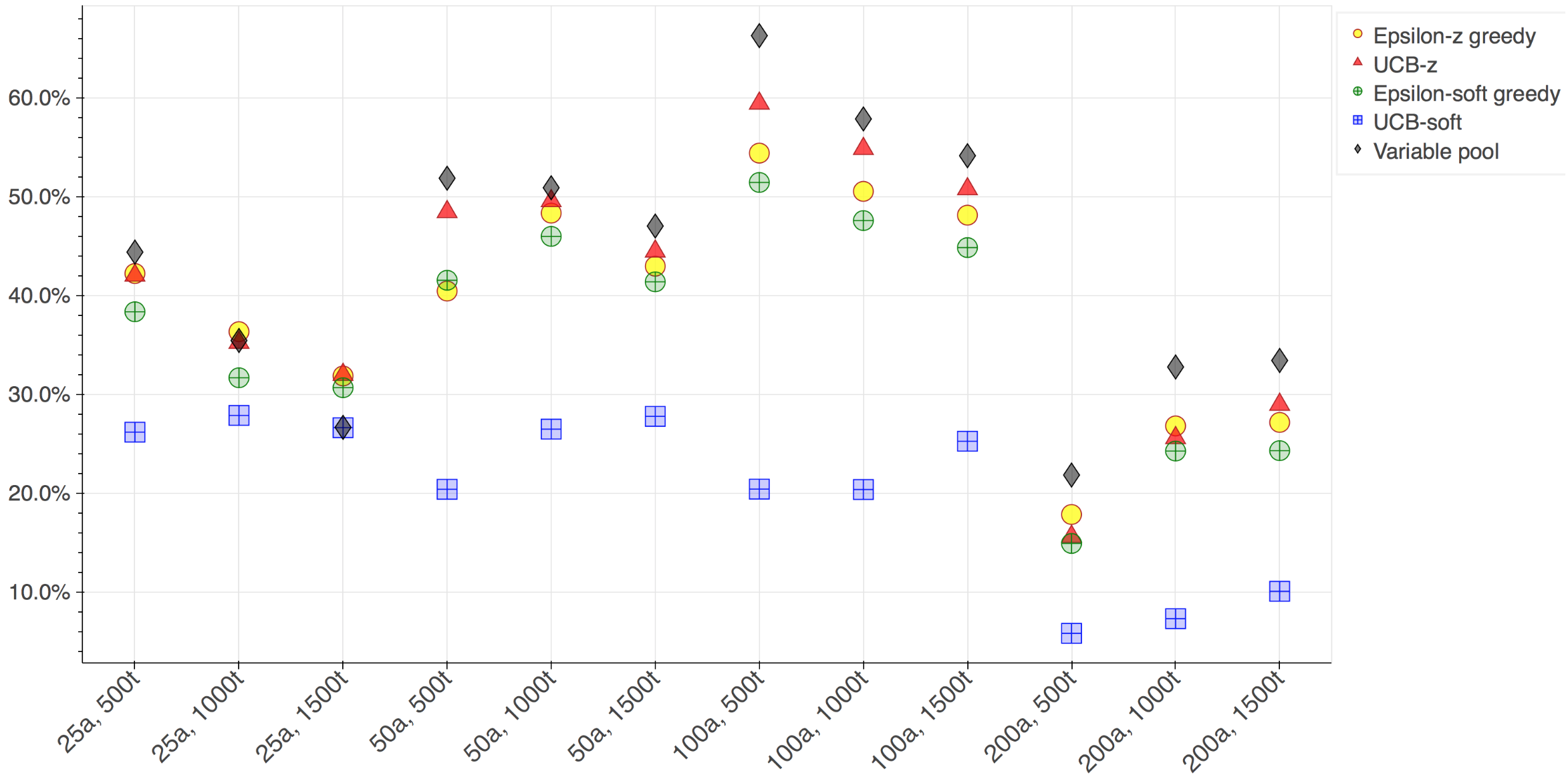}
	\label{BUS}
\end{figure}

\begin{figure}[H]
	\caption{Average increase in rewards (coming from Truncated-Normal distributions) compared to the (smarter) version of the $\varepsilon$-greedy algorithm (Algorithm \ref{Algorithm::epsilon_slightly_smarter}) with a Step-type greed function. }
	\centering
	\includegraphics[width=0.95\textwidth]{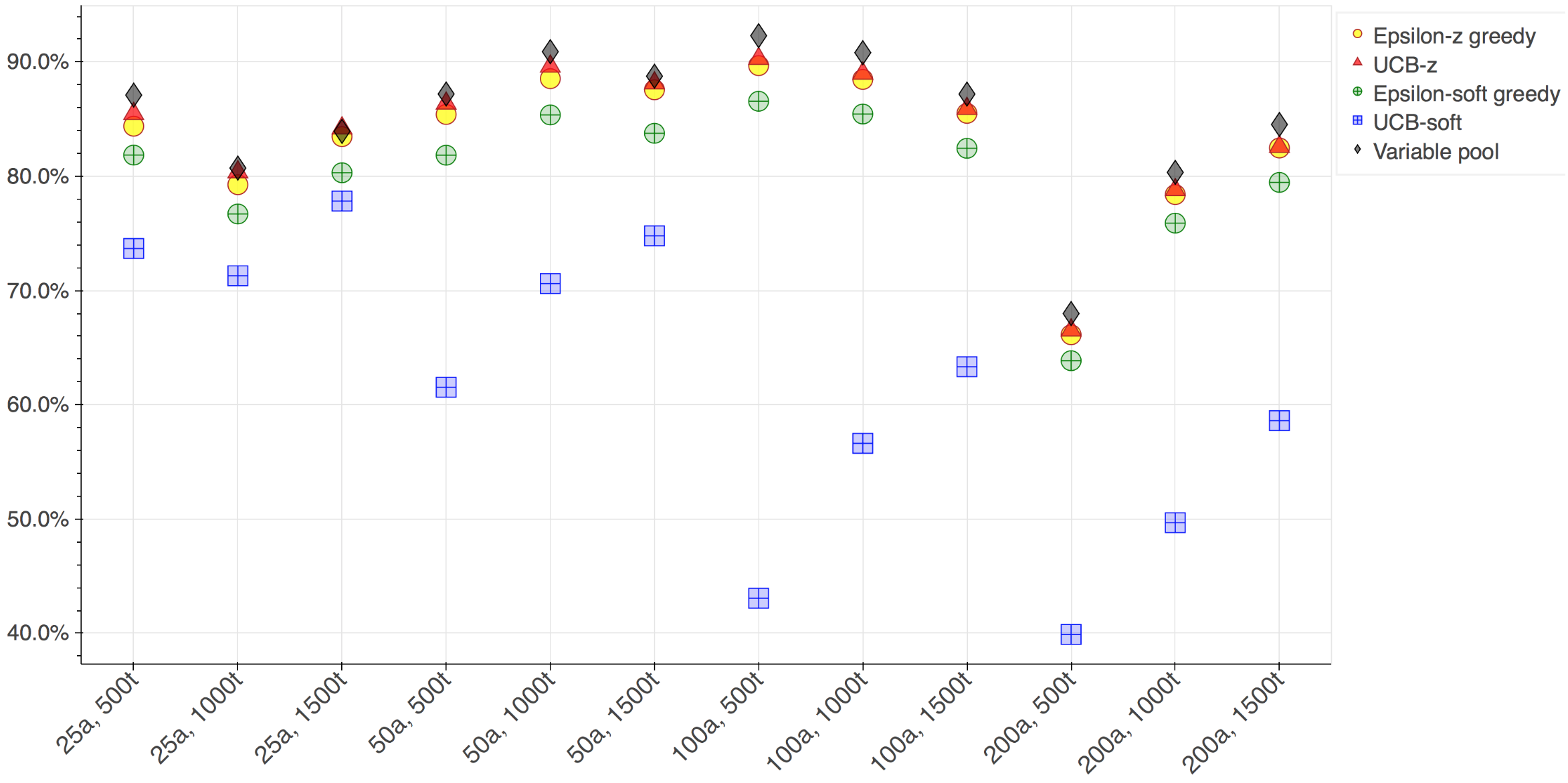}
	\label{NES}
\end{figure}

\begin{figure}[H]
	\caption{Average increase in rewards (coming from Truncated-Normal distributions) compared to the (smarter) version of the UCB algorithm (Algorithm \ref{Algorithm::UCB_slightly_smarter}) with a Step-type greed function. }
	\centering
	\includegraphics[width=0.95\textwidth]{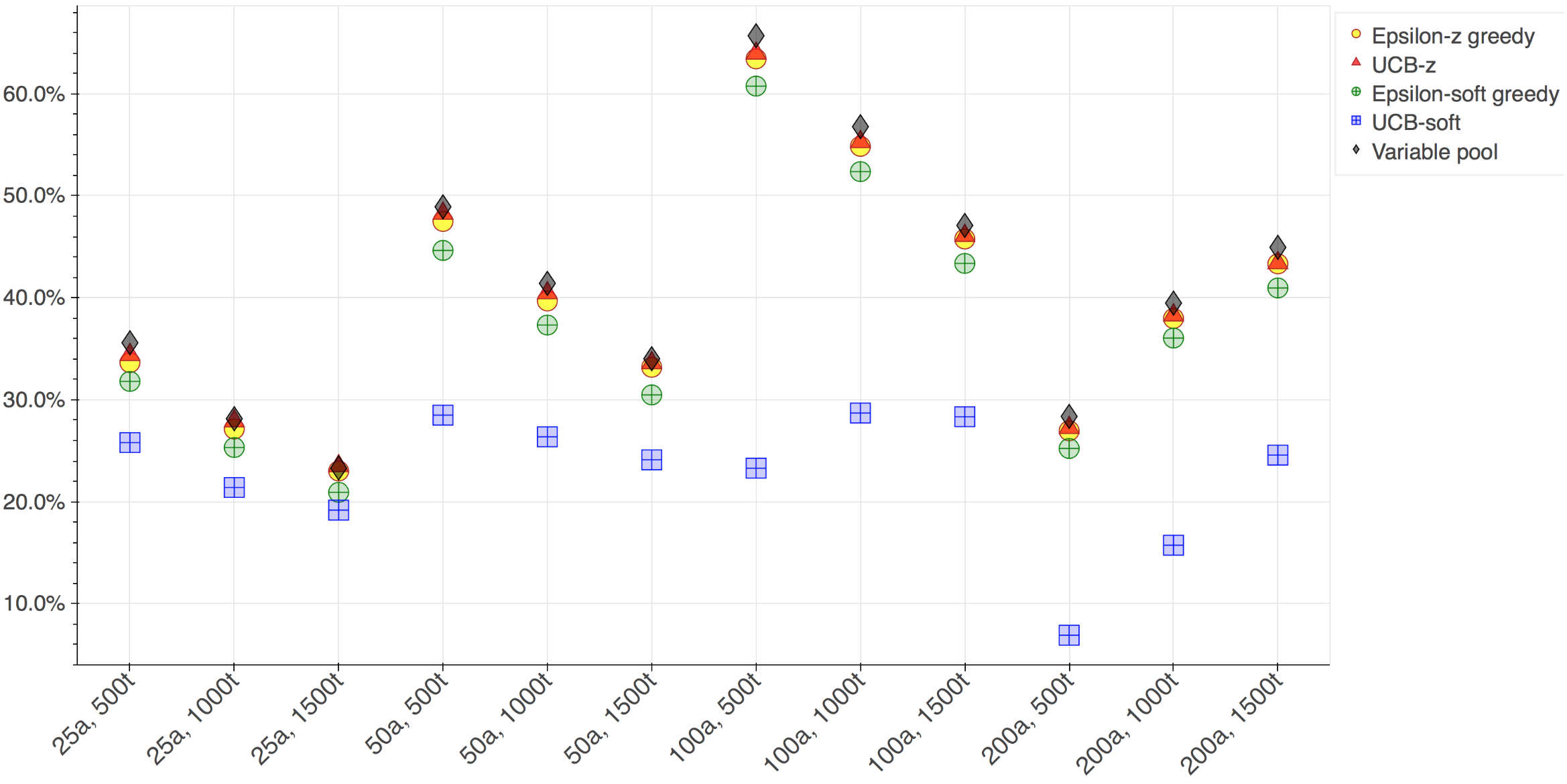}
	\label{NUS}
\end{figure}

\clearpage
%\subsection{Christmas-type greed function with 500 turns per game}

\begin{figure}[H]%
	\makebox[\textwidth][c]{ %to center figures!
		\begin{subfloatrow}
			\subfloat[\small{Rewards from Bernoulli distributions. }]{{\includegraphics[width=6.7cm,height=4.0cm]{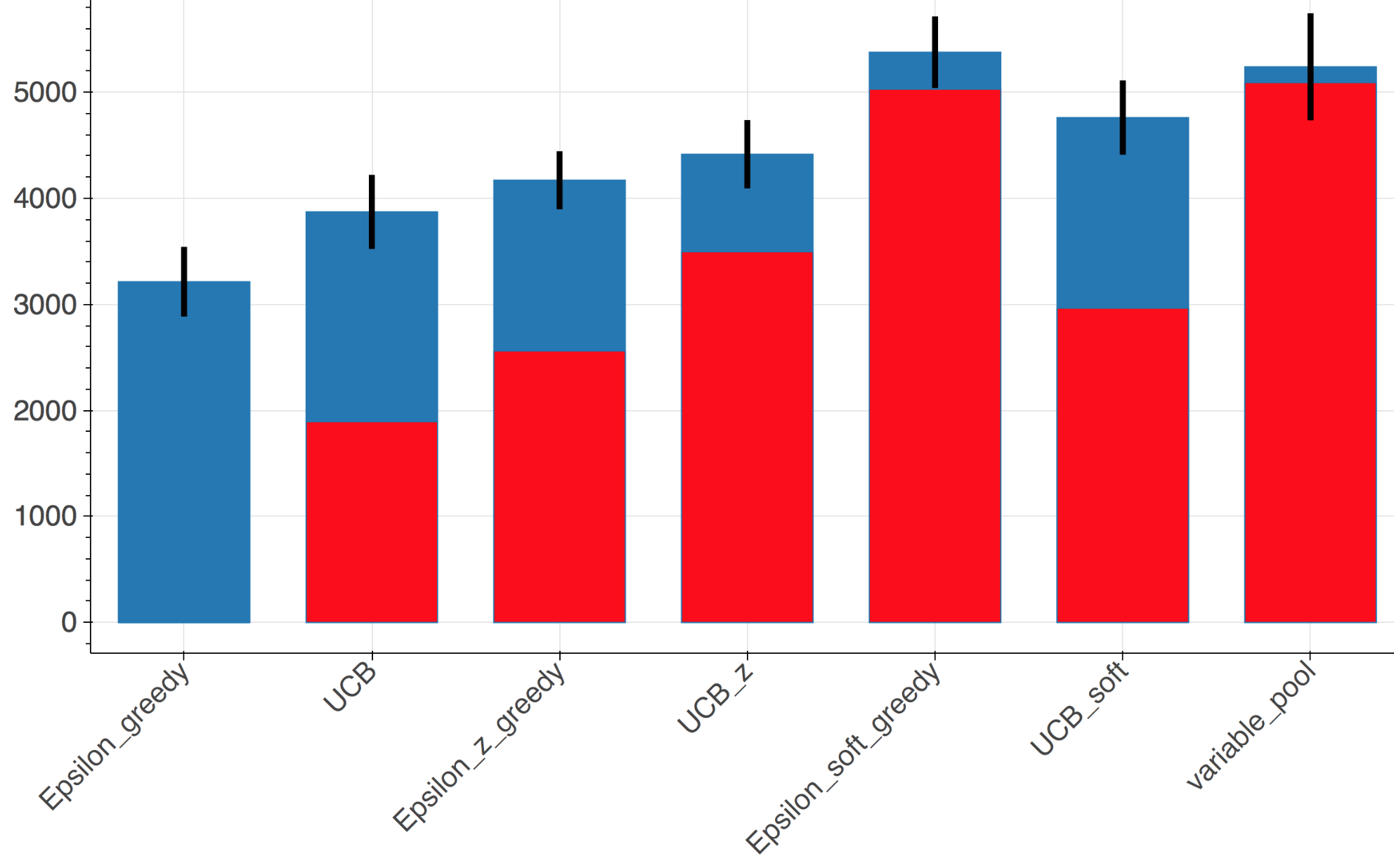} }\label{Bernoulli_Christmas_25a_500t}}%
			\;\;
			\qquad
			\subfloat[\small{Rewards from truncated Normal distributions.
			}]{{\includegraphics[width=6.7cm,height=4.0cm]{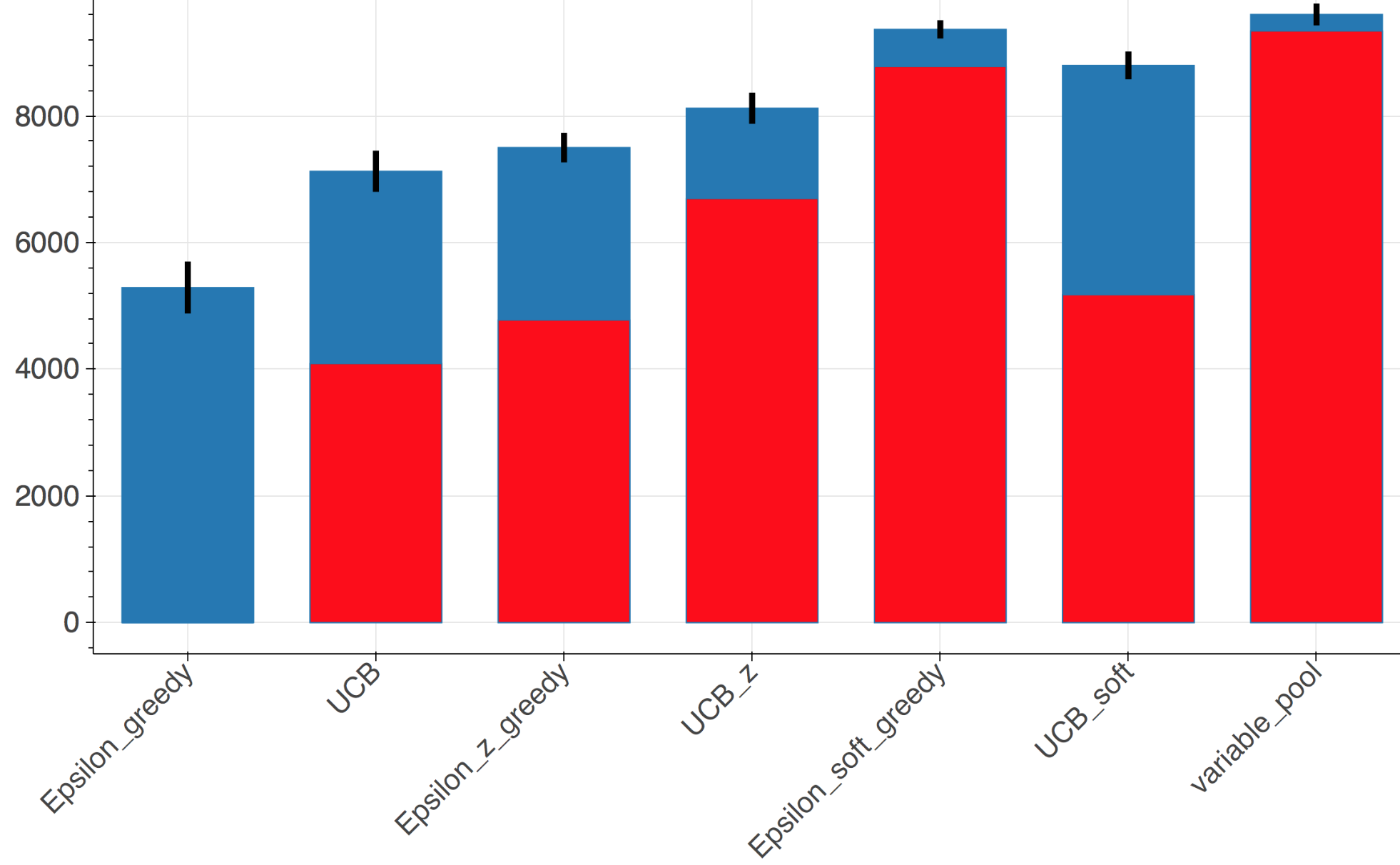} }\label{Truncated_Normal_Christmas_25a_500t}}\;\;
		\end{subfloatrow}
	}
	\caption{Comparison of average final rewards in games with 25 arms, 500 turns, and a Christmas-type greed function.}\label{Figure::25a_500t_Christmas}
\end{figure}

\begin{figure}[H]%
	\makebox[\textwidth][c]{ %to center figures!
		\begin{subfloatrow}
			\subfloat[\small{Rewards from Bernoulli distributions. }]{{\includegraphics[width=6.7cm,height=4.0cm]{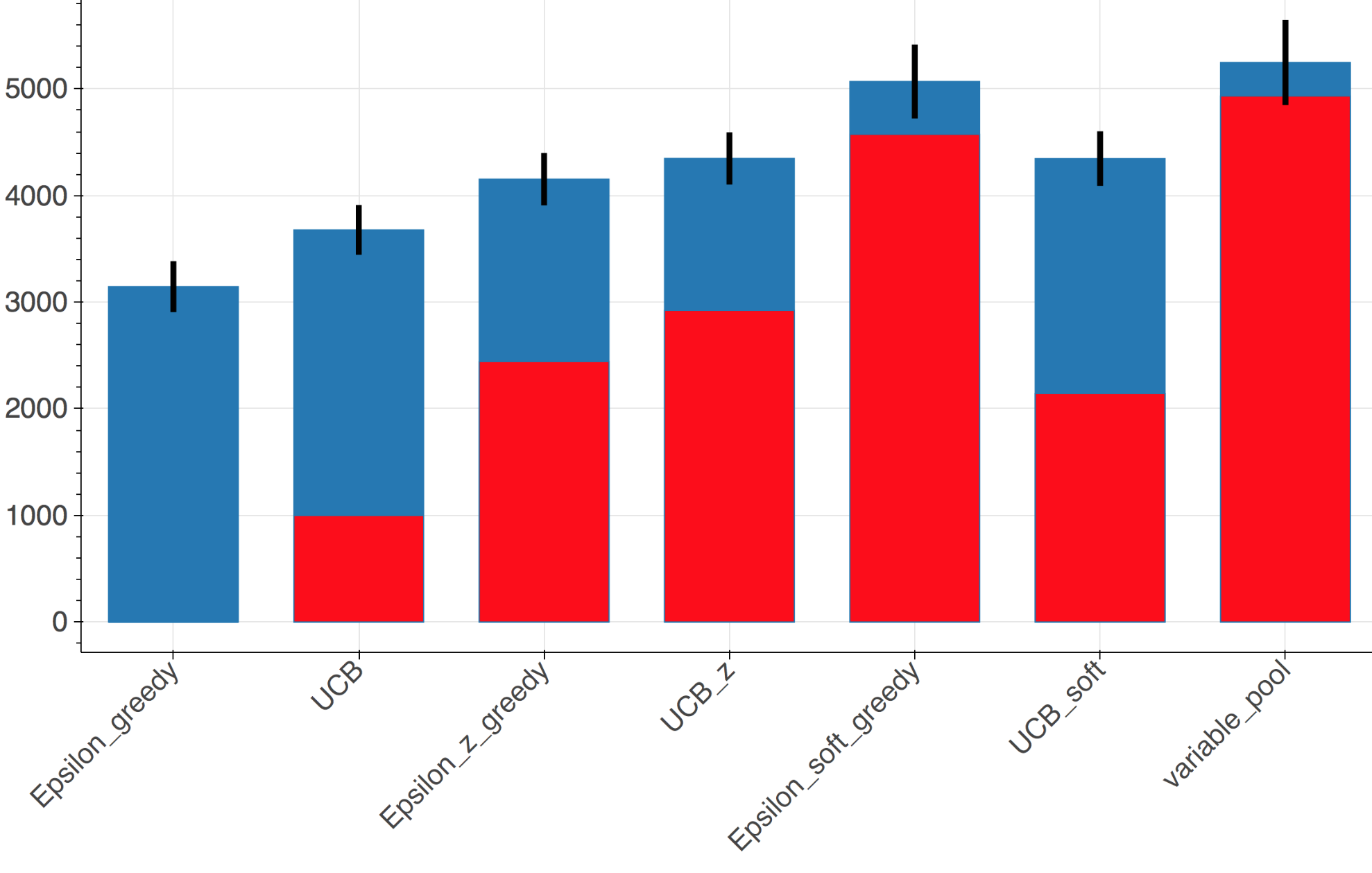} }\label{Bernoulli_Christmas_50a_500t}}%
			\;\;
			\qquad
			\subfloat[\small{Rewards from truncated Normal distributions.
			}]{{\includegraphics[width=6.7cm,height=4.0cm]{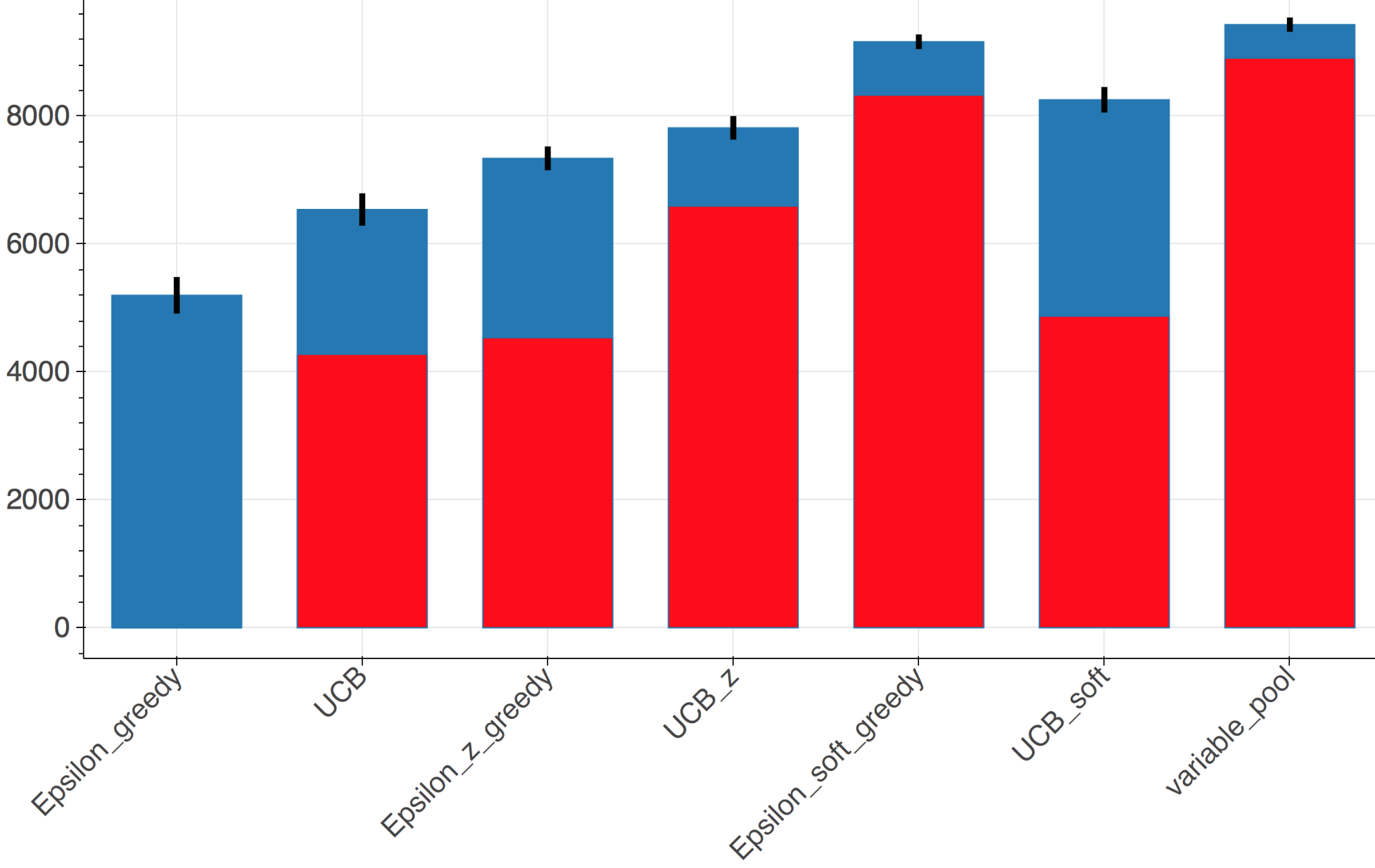} }\label{Truncated_Normal_Christmas_50a_500t}}\;\;
		\end{subfloatrow}
	}
	\caption{Comparison of average final rewards in games with 50 arms, 500 turns, and a Christmas-type greed function.}\label{Figure::50a_500t_Christmas}
\end{figure}

\begin{figure}[H]%
	\makebox[\textwidth][c]{ %to center figures!
		\begin{subfloatrow}
			\subfloat[\small{Rewards from Bernoulli distributions. }]{{\includegraphics[width=6.7cm,height=4.0cm]{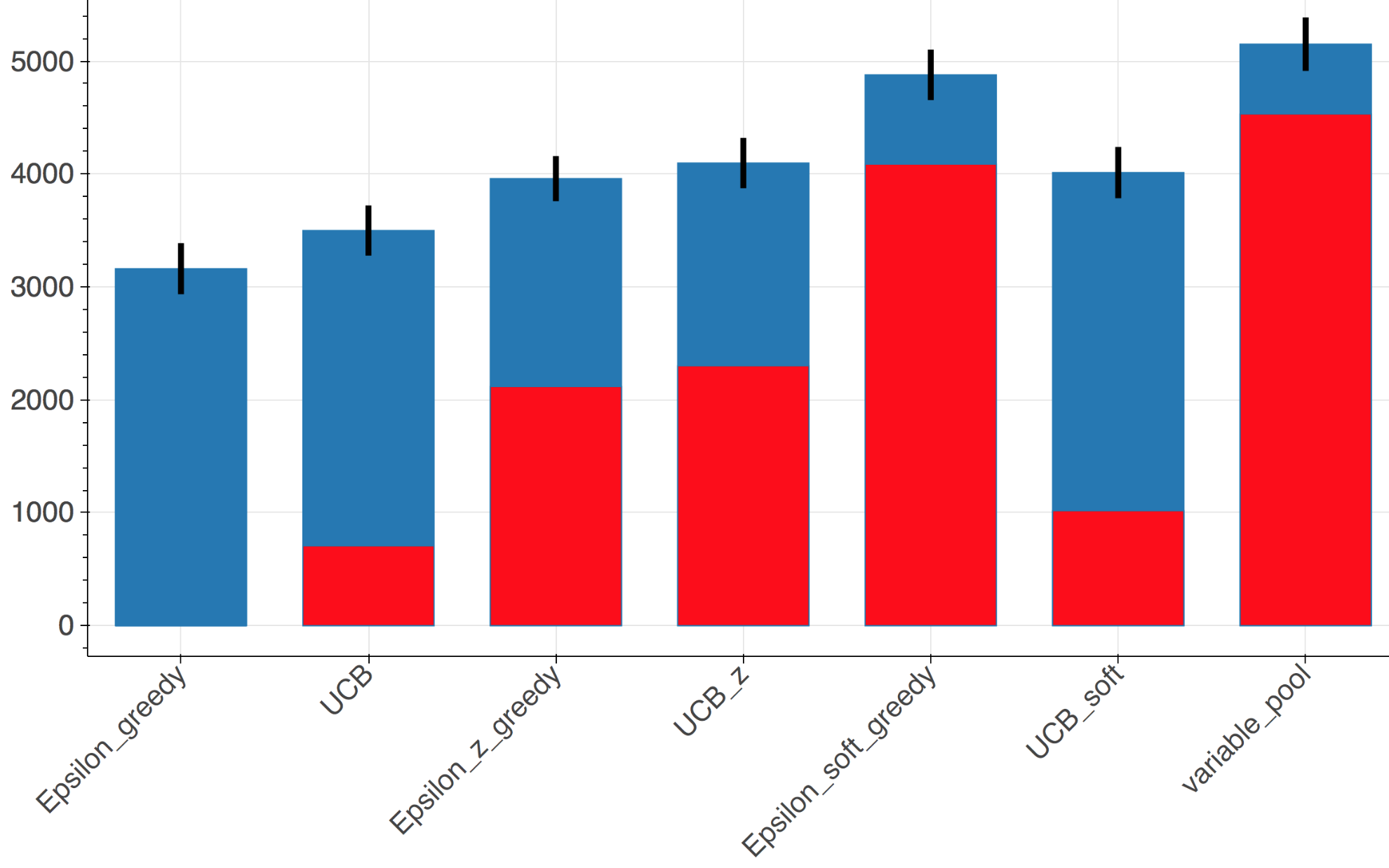} }\label{Bernoulli_Christmas_100a_500t}}%
			\;\;
			\qquad
			\subfloat[\small{Rewards from truncated Normal distributions.
			}]{{\includegraphics[width=6.7cm,height=4.0cm]{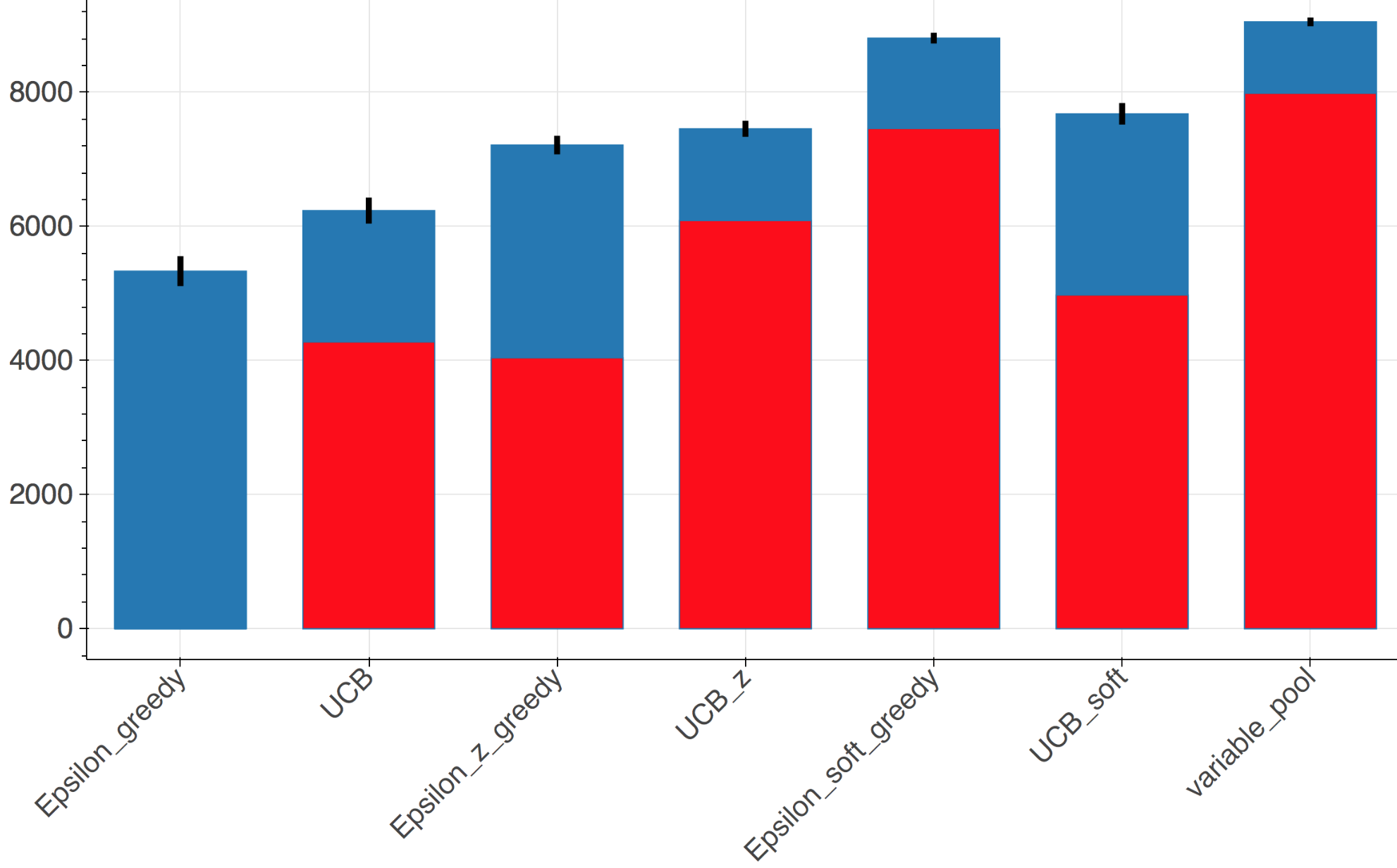} }\label{Truncated_Normal_Christmas_100a_500t}}\;\;
		\end{subfloatrow}
	}
	\caption{Comparison of average final rewards in games with 100 arms, 500 turns, and a Christmas-type greed function.}\label{Figure::100a_500t_Christmas}
\end{figure}

\begin{figure}[H]%
	\makebox[\textwidth][c]{ %to center figures!
		\begin{subfloatrow}
			\subfloat[\small{Rewards from Bernoulli distributions. }]{{\includegraphics[width=6.7cm,height=4.0cm]{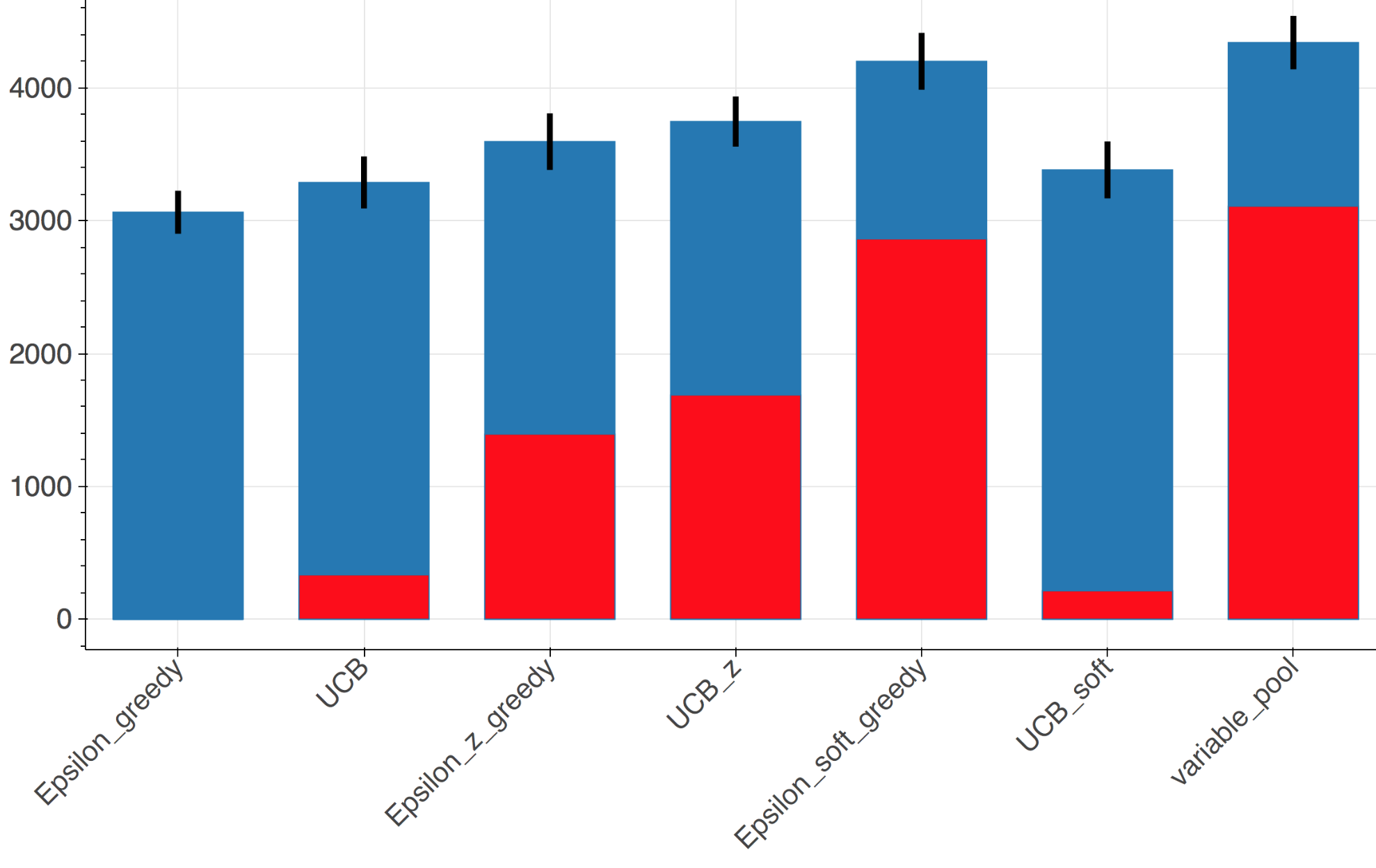} }\label{Bernoulli_Christmas_200a_500t}}%
			\;\;
			\qquad
			\subfloat[\small{Rewards from truncated Normal distributions.
			}]{{\includegraphics[width=6.7cm,height=4.0cm]{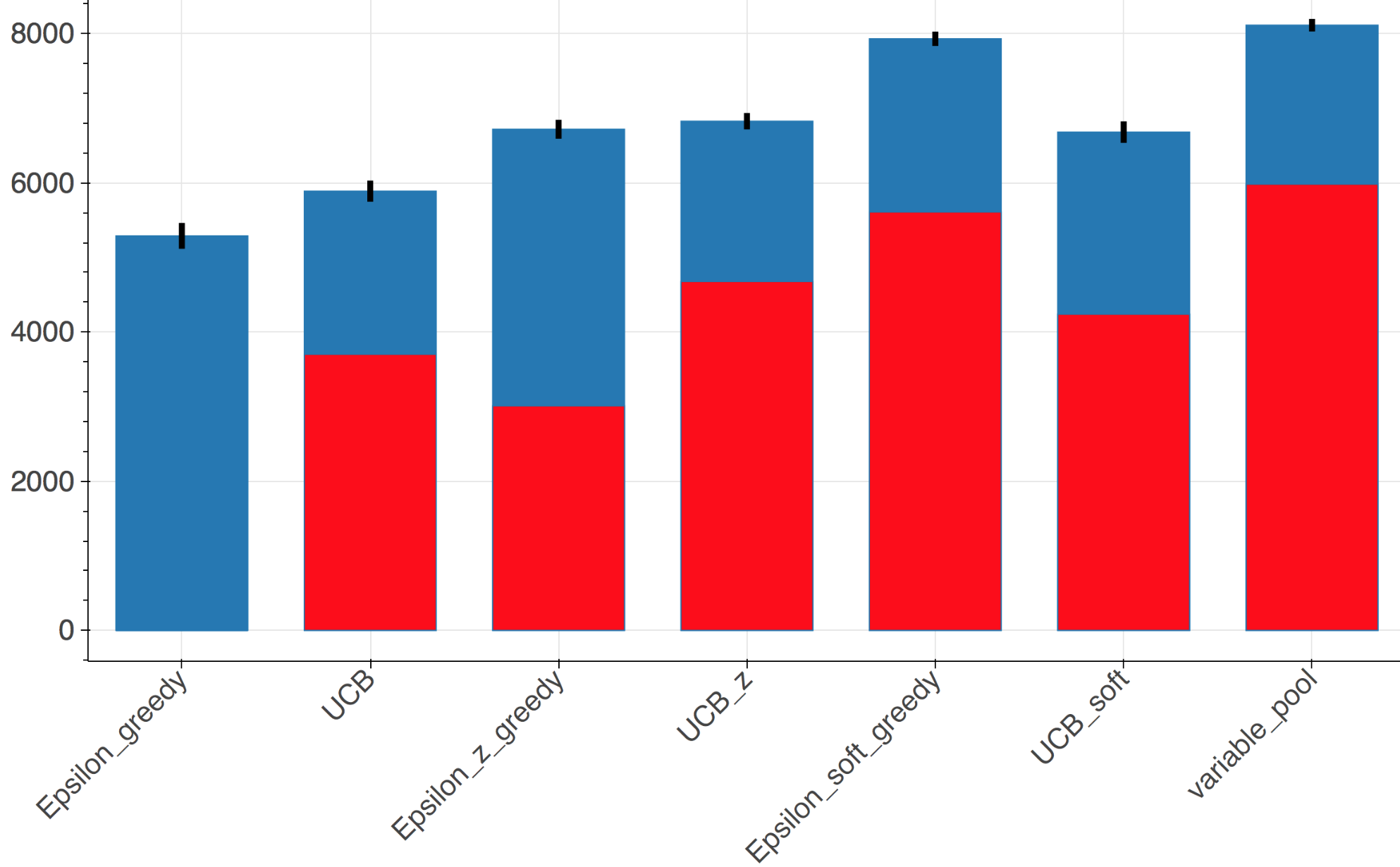} }\label{Truncated_Normal_Christmas_200a_500t}}\;\;
		\end{subfloatrow}
	}
	\caption{Comparison of average final rewards in games with 200 arms, 500 turns, and a Christmas-type greed function.}\label{Figure::200a_500t_Christmas}
\end{figure}

%\subsection{Christmas-type greed function with 1000 turns per game}

\begin{figure}[H]%
	\makebox[\textwidth][c]{ %to center figures!
		\begin{subfloatrow}
			\subfloat[\small{Rewards from Bernoulli distributions. }]{{\includegraphics[width=6.7cm,height=4.0cm]{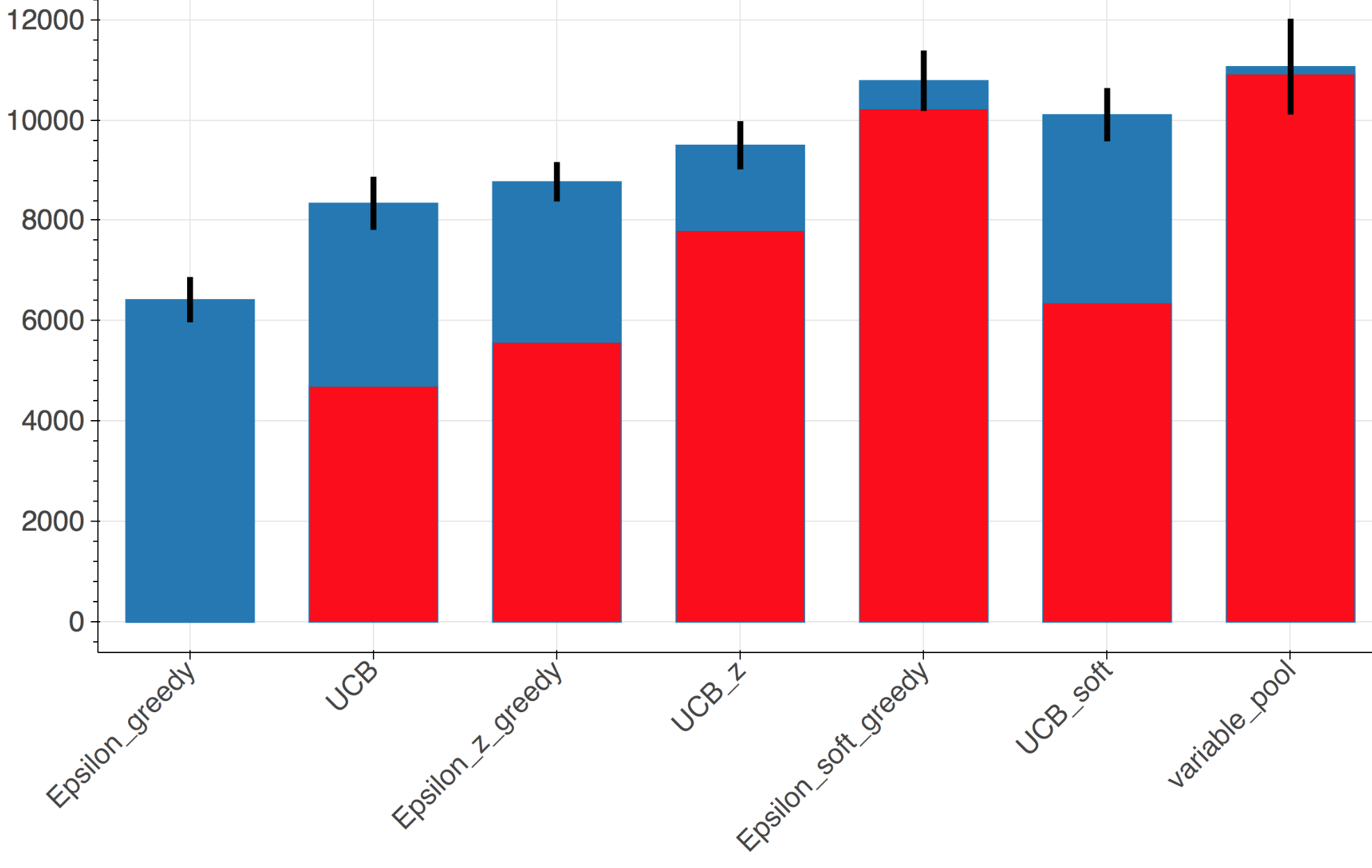} }\label{Bernoulli_Christmas_25a_1000t}}%
			\;\;
			\qquad
			\subfloat[\small{Rewards from truncated Normal distributions.
			}]{{\includegraphics[width=6.7cm,height=4.0cm]{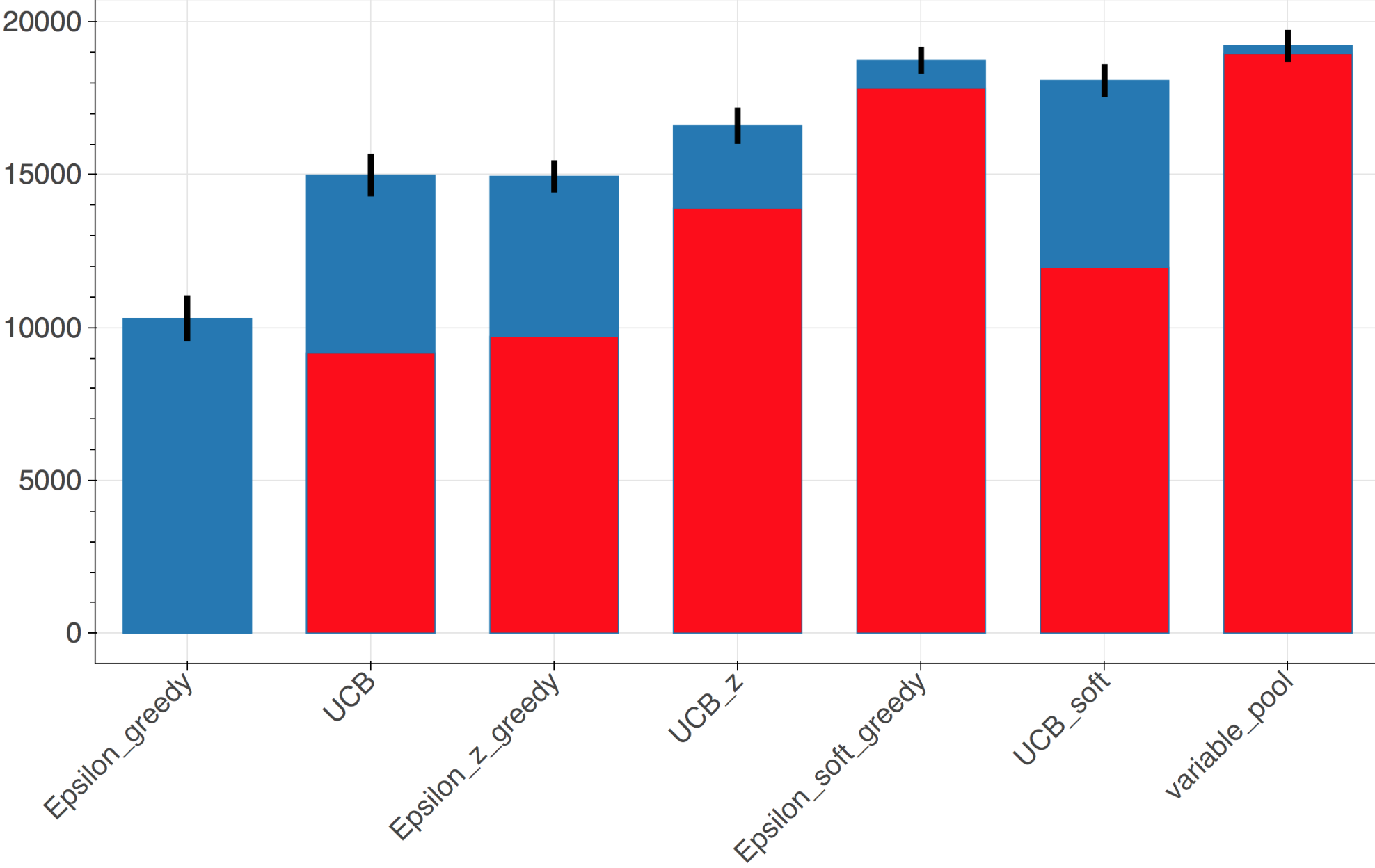} }\label{Truncated_Normal_Christmas_25a_1000t}}\;\;
		\end{subfloatrow}
	}
	\caption{Comparison of average final rewards in games with 25 arms, 1000 turns, and a Christmas-type greed function.}\label{Figure::25a_1000t_Christmas}
\end{figure}

\begin{figure}[H]%
	\makebox[\textwidth][c]{ %to center figures!
		\begin{subfloatrow}
			\subfloat[\small{Rewards from Bernoulli distributions. }]{{\includegraphics[width=6.7cm,height=4.0cm]{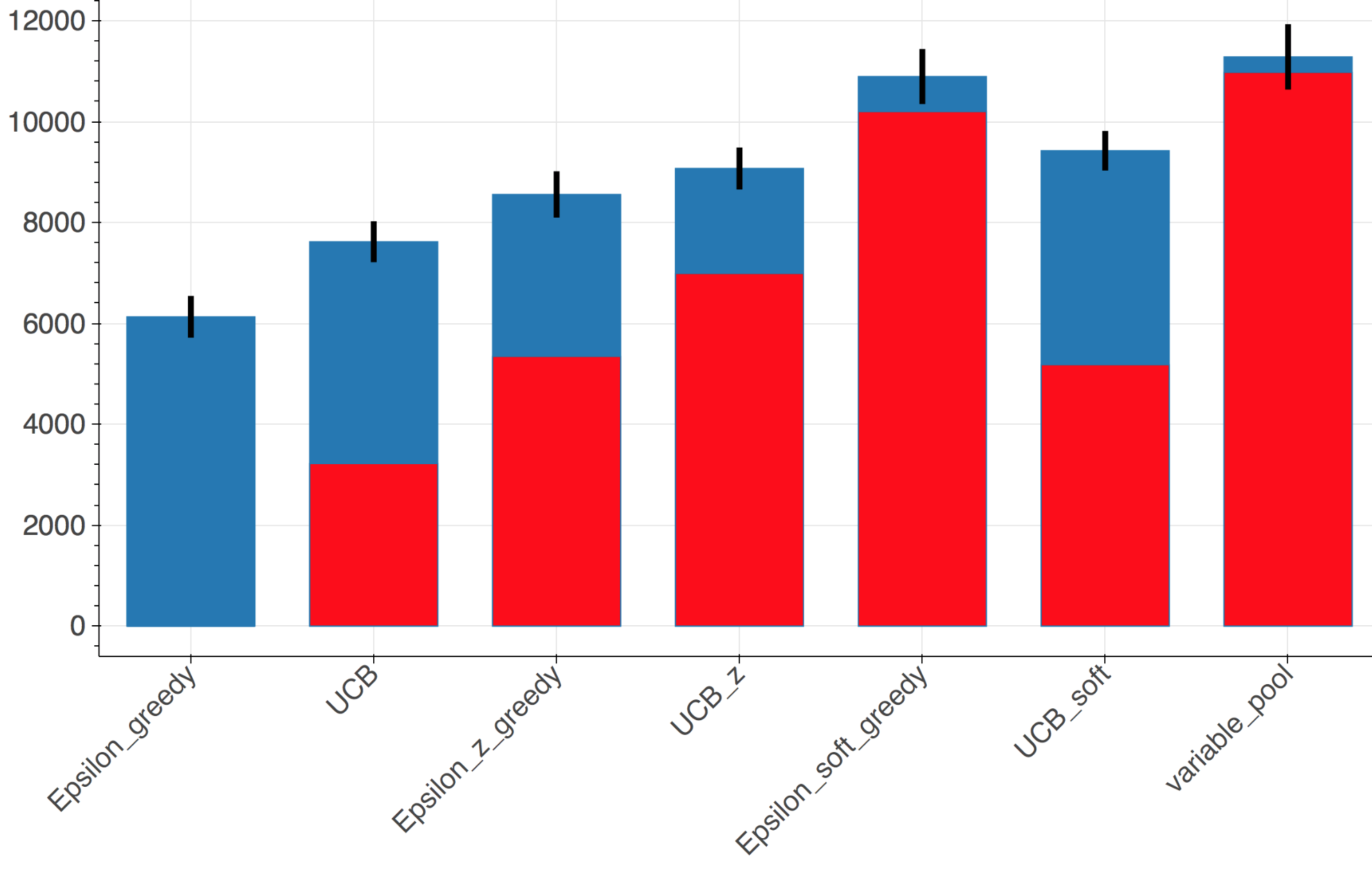} }\label{Bernoulli_Christmas_50a_1000t}}%
			\;\;
			\qquad
			\subfloat[\small{Rewards from truncated Normal distributions.
			}]{{\includegraphics[width=6.7cm,height=4.0cm]{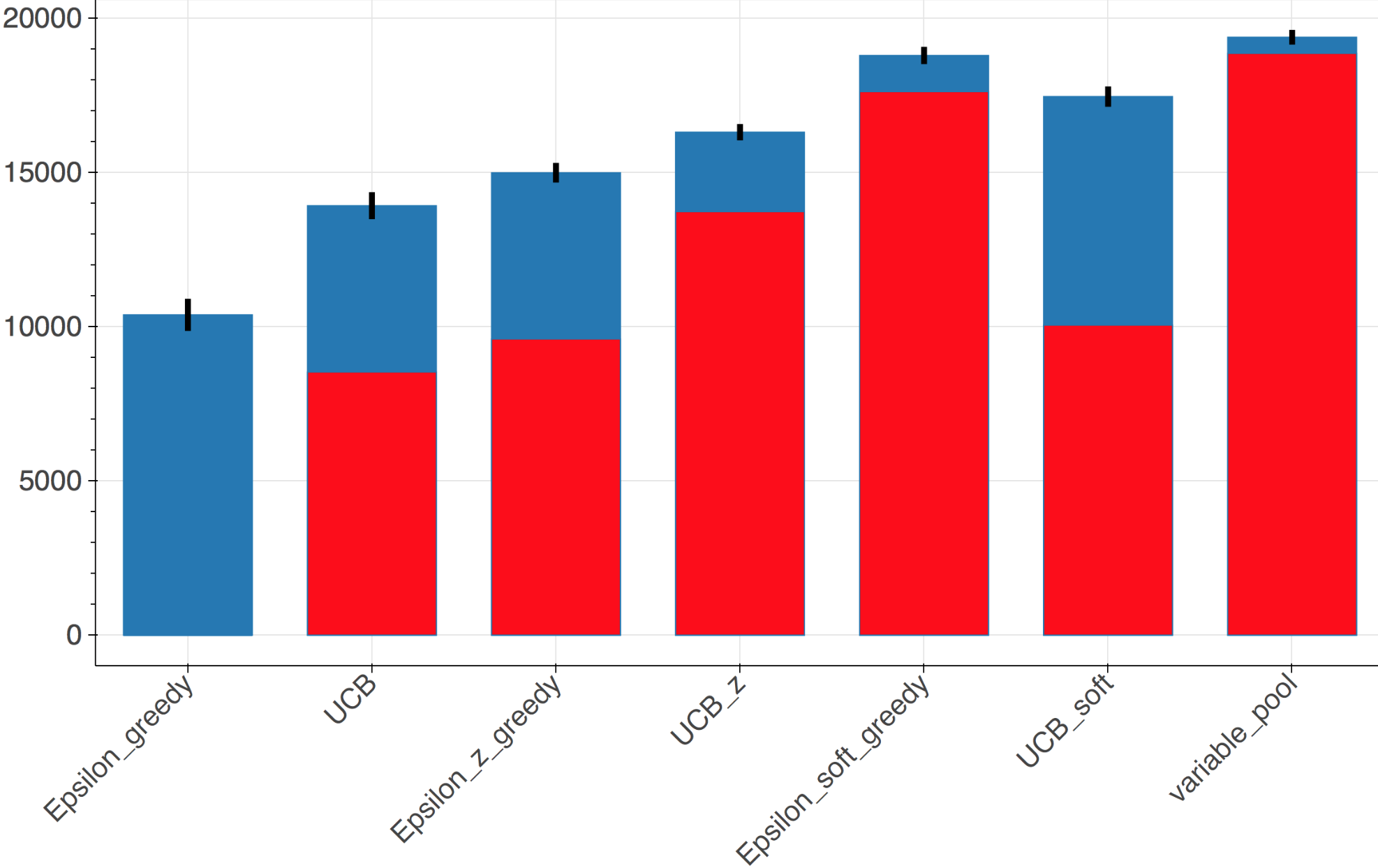} }\label{Truncated_Normal_Christmas_50a_1000t}}\;\;
		\end{subfloatrow}
	}
	\caption{Comparison of average final rewards in games with 50 arms, 1000 turns, and a Christmas-type greed function.}\label{Figure::50a_1000t_Christmas}
\end{figure}

\begin{figure}[H]%
	\makebox[\textwidth][c]{ %to center figures!
		\begin{subfloatrow}
			\subfloat[\small{Rewards from Bernoulli distributions. }]{{\includegraphics[width=6.7cm,height=4.0cm]{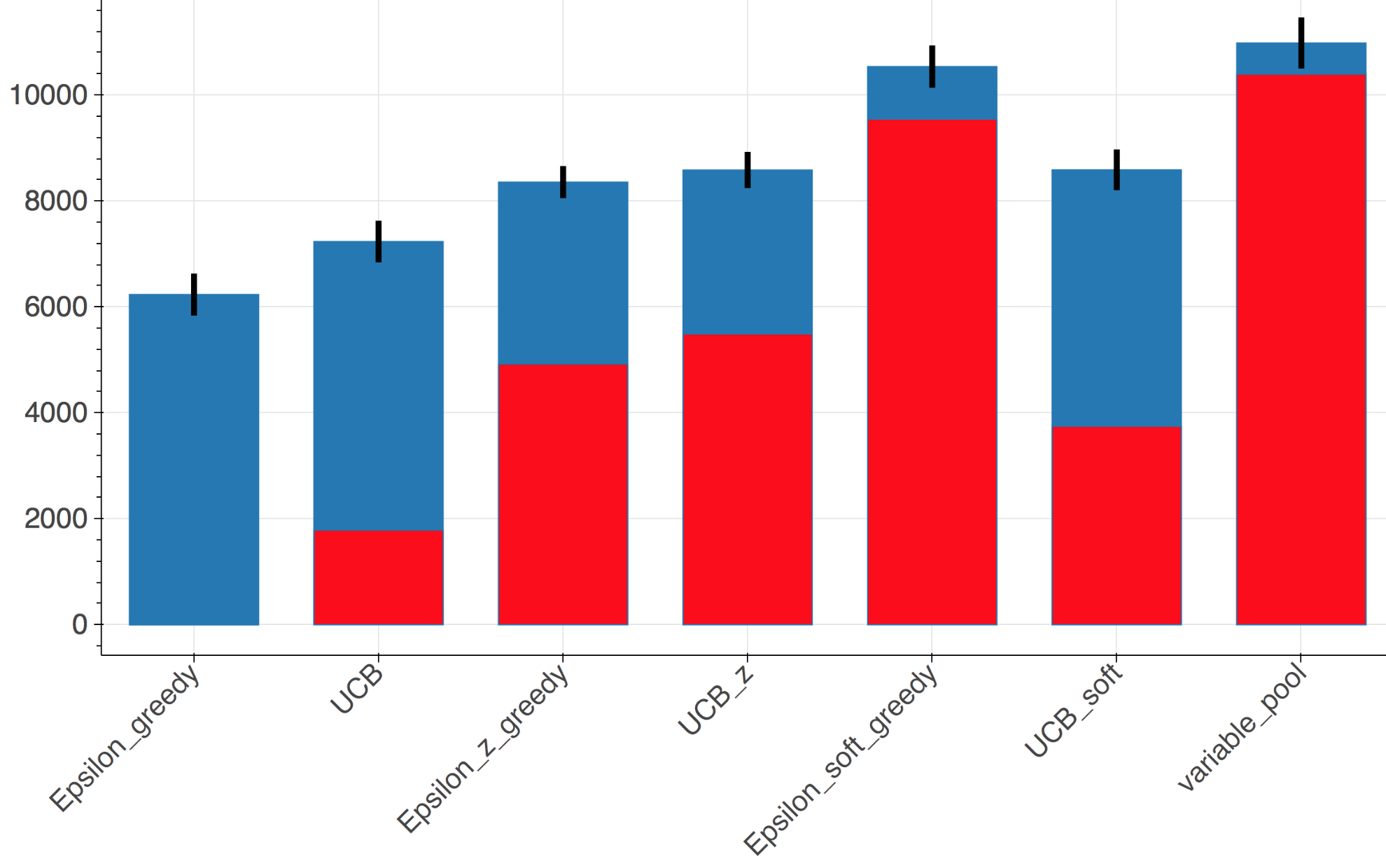} }\label{Bernoulli_Christmas_100a_1000t}}%
			\;\;
			\qquad
			\subfloat[\small{Rewards from truncated Normal distributions.
			}]{{\includegraphics[width=6.7cm,height=4.0cm]{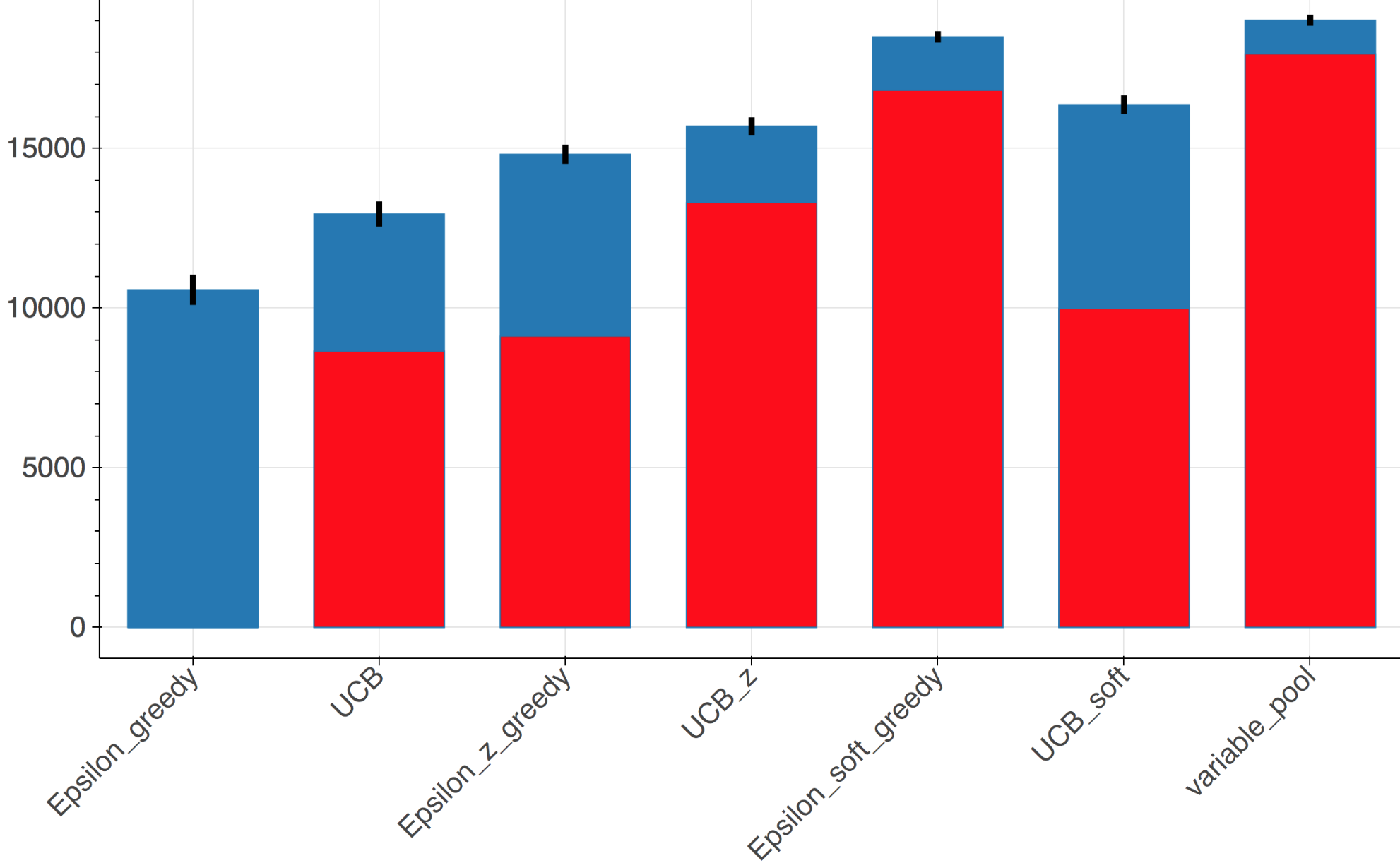} }\label{Truncated_Normal_Christmas_100a_1000t}}\;\;
		\end{subfloatrow}
	}
	\caption{Comparison of average final rewards in games with 100 arms, 1000 turns, and a Christmas-type greed function.}\label{Figure::100a_1000t_Christmas}
\end{figure}

\begin{figure}[H]%
	\makebox[\textwidth][c]{ %to center figures!
		\begin{subfloatrow}
			\subfloat[\small{Rewards from Bernoulli distributions. }]{{\includegraphics[width=6.7cm,height=4.0cm]{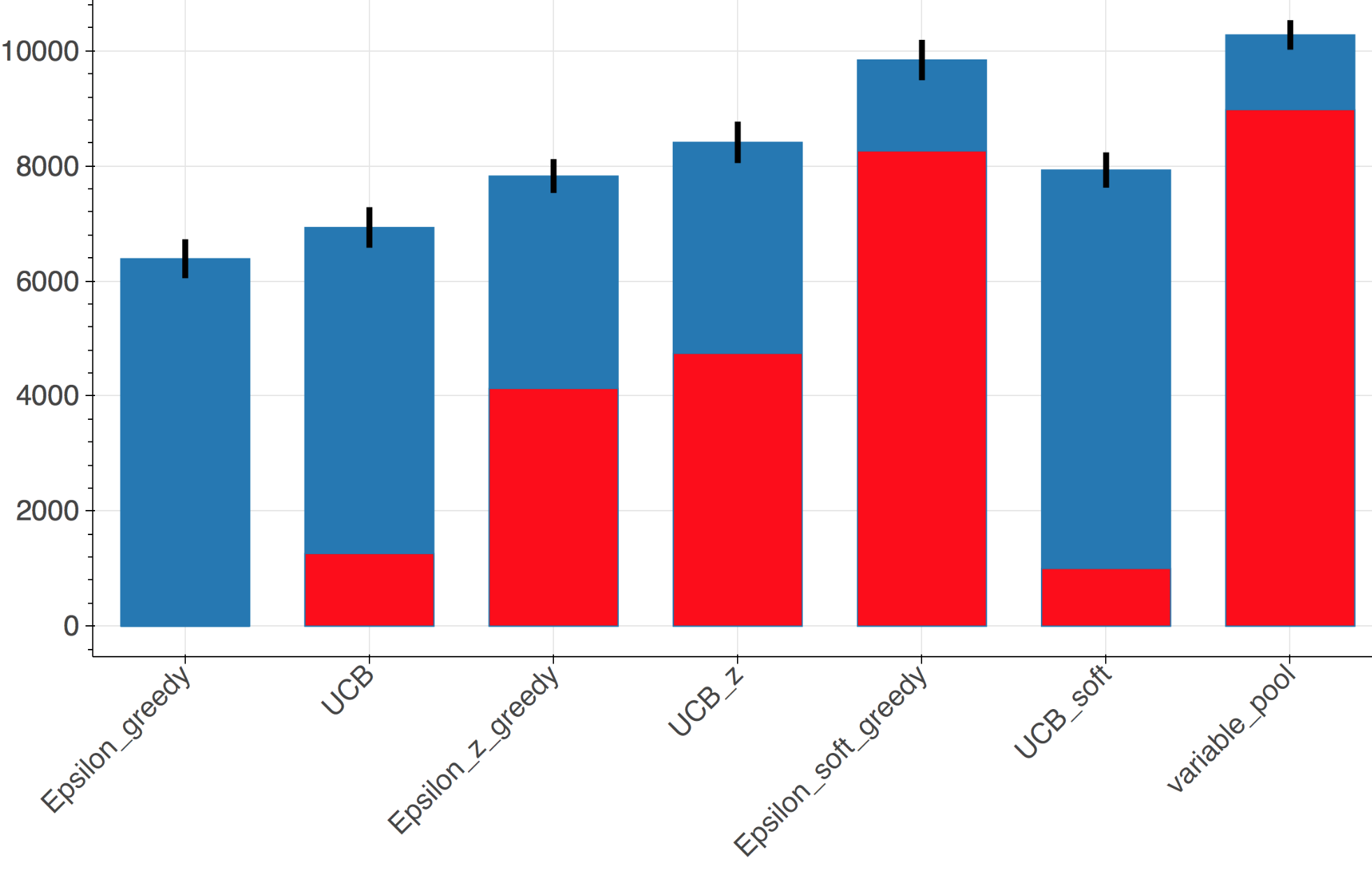} }\label{Bernoulli_Christmas_200a_1000t}}%
			\;\;
			\qquad
			\subfloat[\small{Rewards from truncated Normal distributions.
			}]{{\includegraphics[width=6.7cm,height=4.0cm]{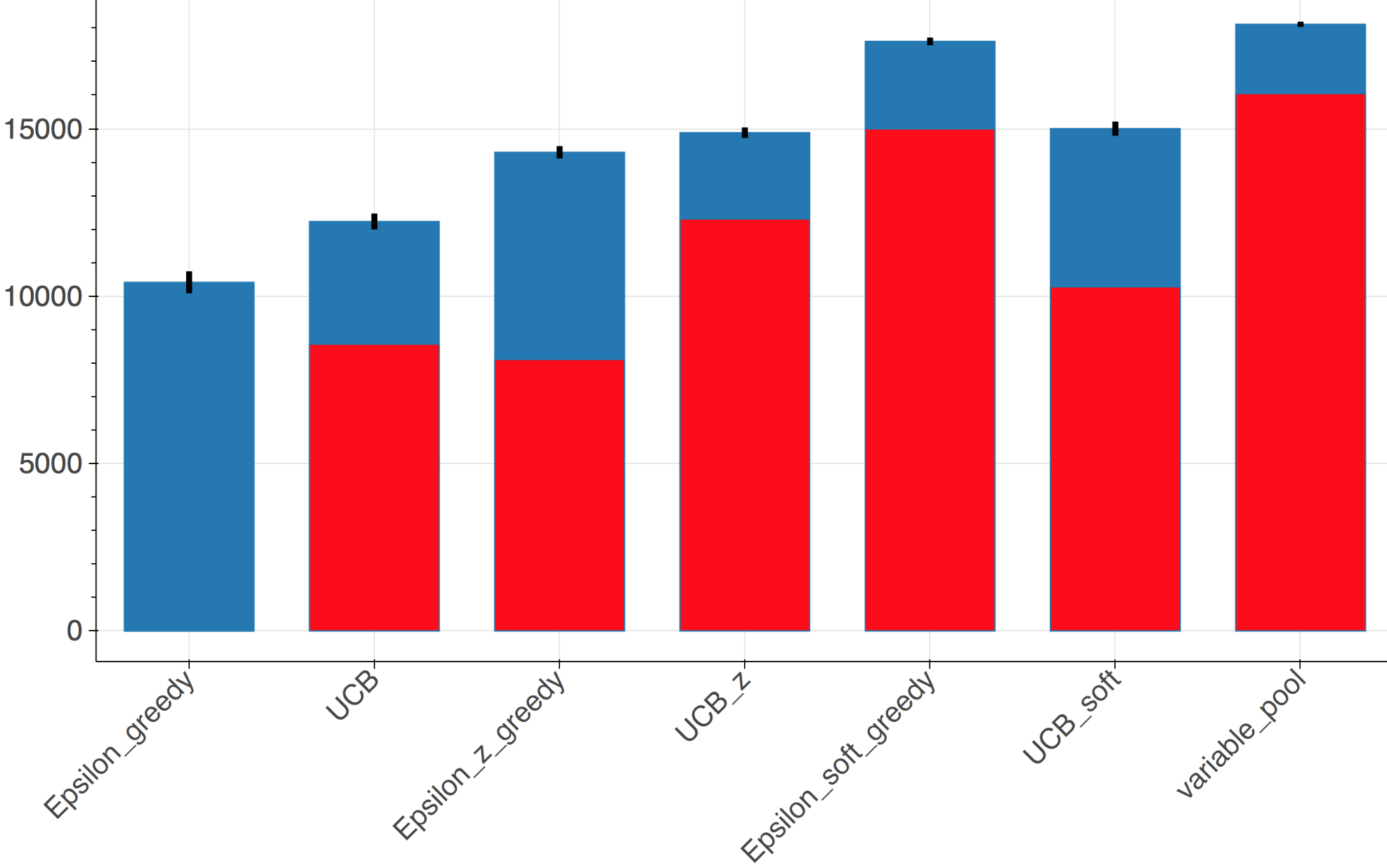} }\label{Truncated_Normal_Christmas_200a_1000t}}\;\;
		\end{subfloatrow}
	}
	\caption{Comparison of average final rewards in games with 200 arms, 1000 turns, and a Christmas-type greed function.}\label{Figure::200a_1000t_Christmas}
\end{figure}

%\subsection{Christmas-type greed function with 1500 turns per game}

\begin{figure}[H]%
	\makebox[\textwidth][c]{ %to center figures!
		\begin{subfloatrow}
			\subfloat[\small{Rewards from Bernoulli distributions. }]{{\includegraphics[width=6.7cm,height=4.0cm]{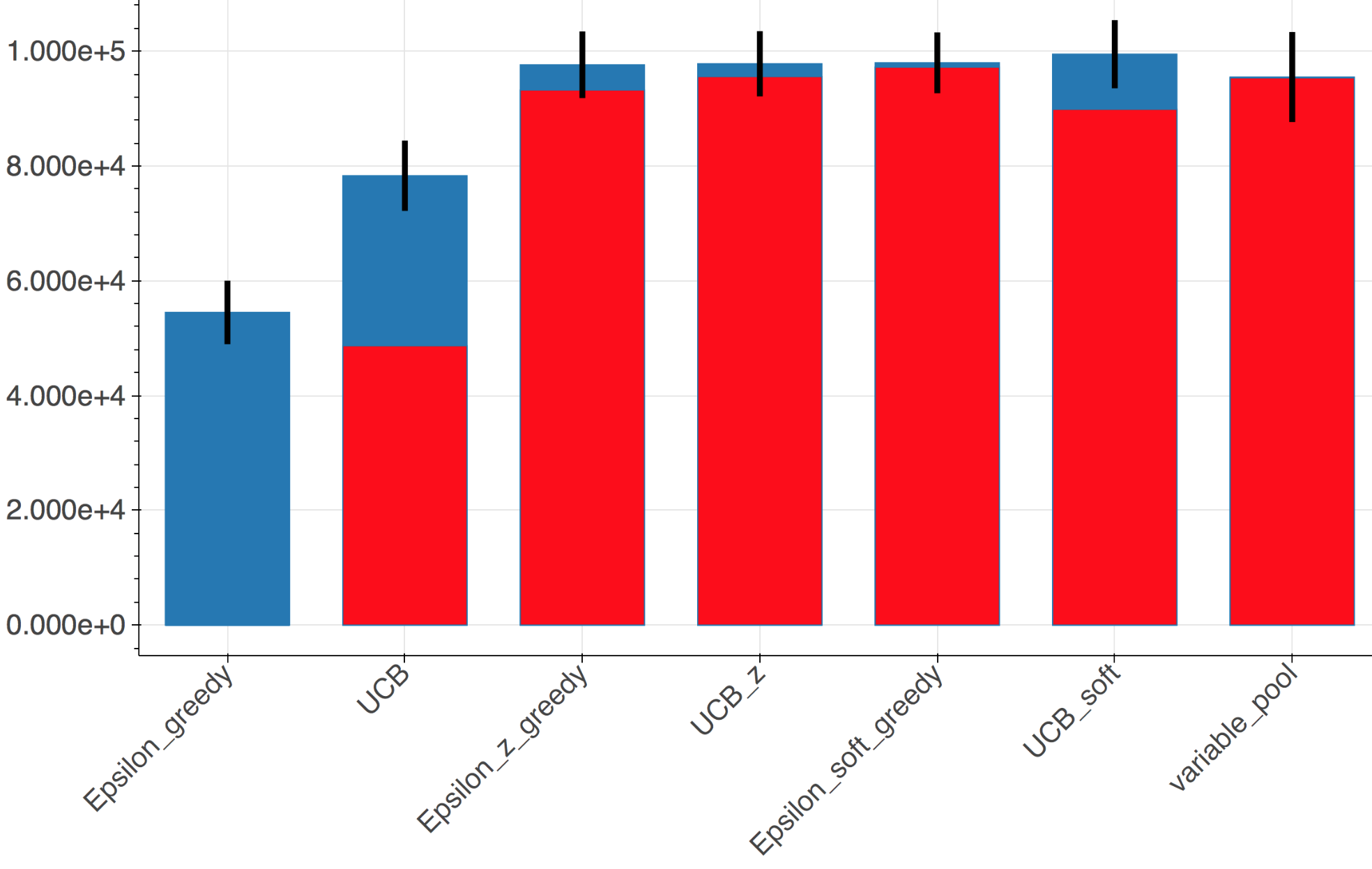} }\label{Bernoulli_Christmas_25a_1500t}}%
			\;\;
			\qquad
			\subfloat[\small{Rewards from truncated Normal distributions.
			}]{{\includegraphics[width=6.7cm,height=4.0cm]{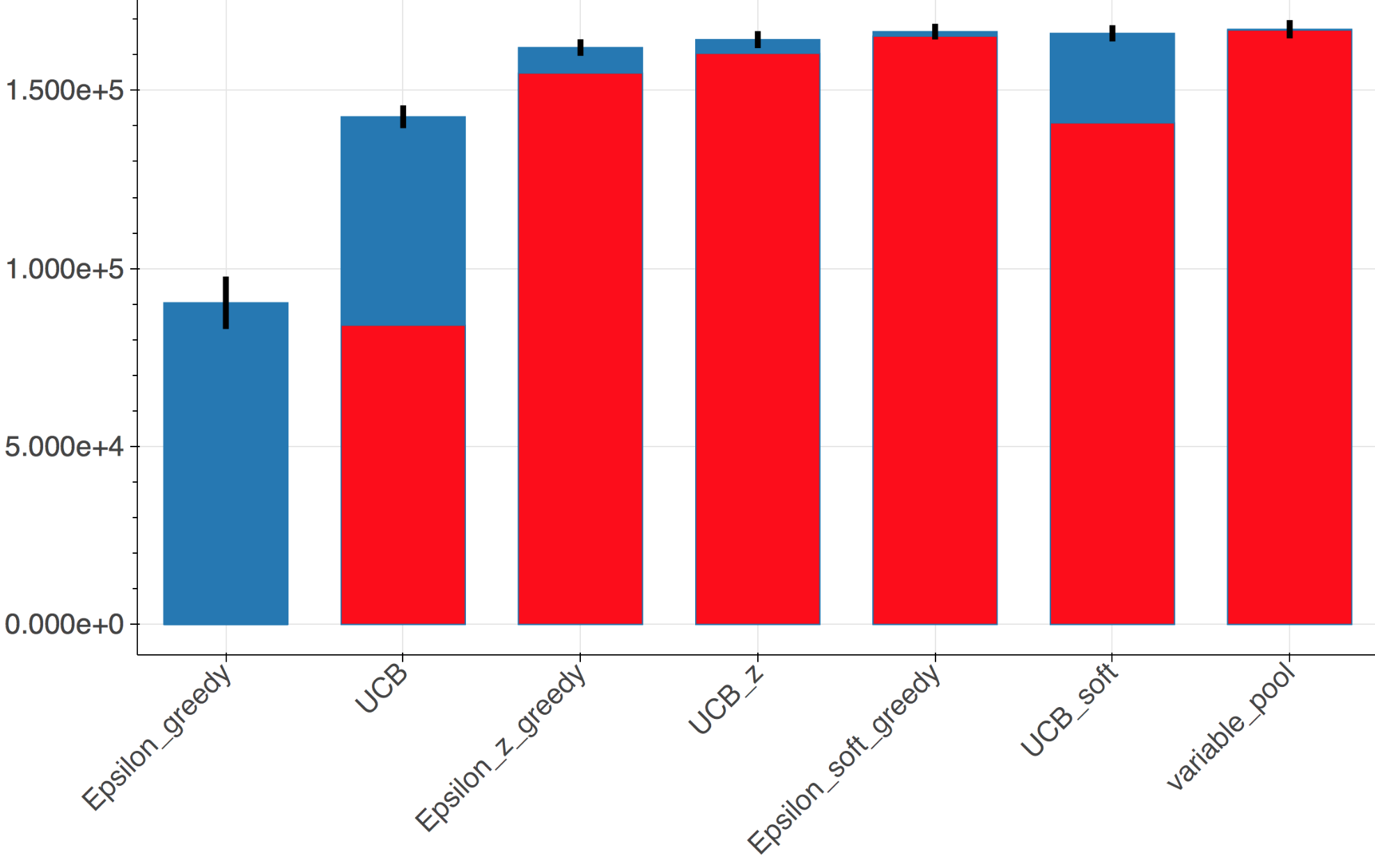} }\label{Truncated_Normal_Christmas_25a_1500t}}\;\;
		\end{subfloatrow}
	}
	\caption{Comparison of average final rewards in games with 25 arms, 1500 turns, and a Christmas-type greed function.}\label{Figure::25a_1500t_Christmas}
\end{figure}

\begin{figure}[H]%
	\makebox[\textwidth][c]{ %to center figures!
		\begin{subfloatrow}
			\subfloat[\small{Rewards from Bernoulli distributions. }]{{\includegraphics[width=6.7cm,height=4.0cm]{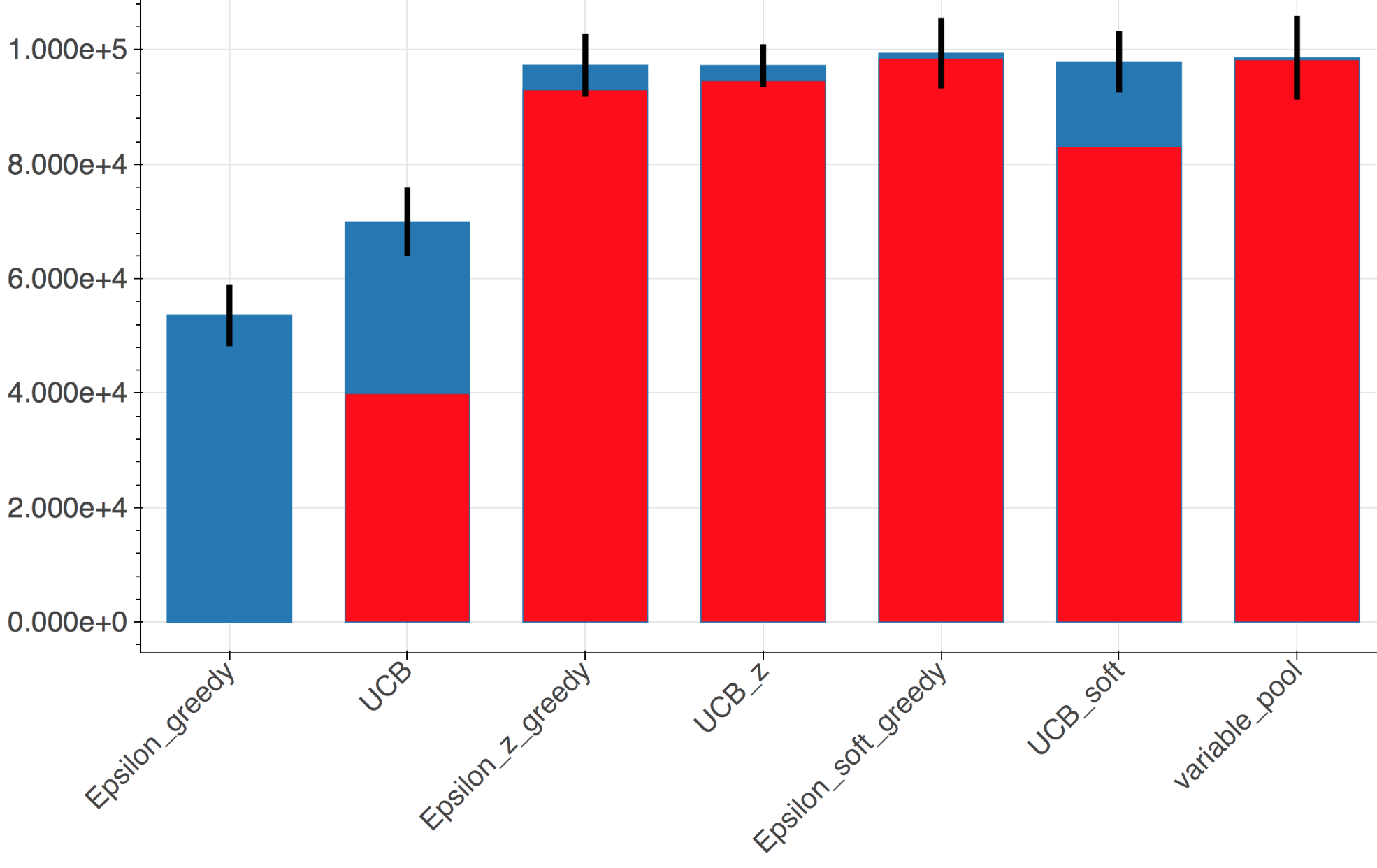} }\label{Bernoulli_Christmas_50a_1500t}}%
			\;\;
			\qquad
			\subfloat[\small{Rewards from truncated Normal distributions.
			}]{{\includegraphics[width=6.7cm,height=4.0cm]{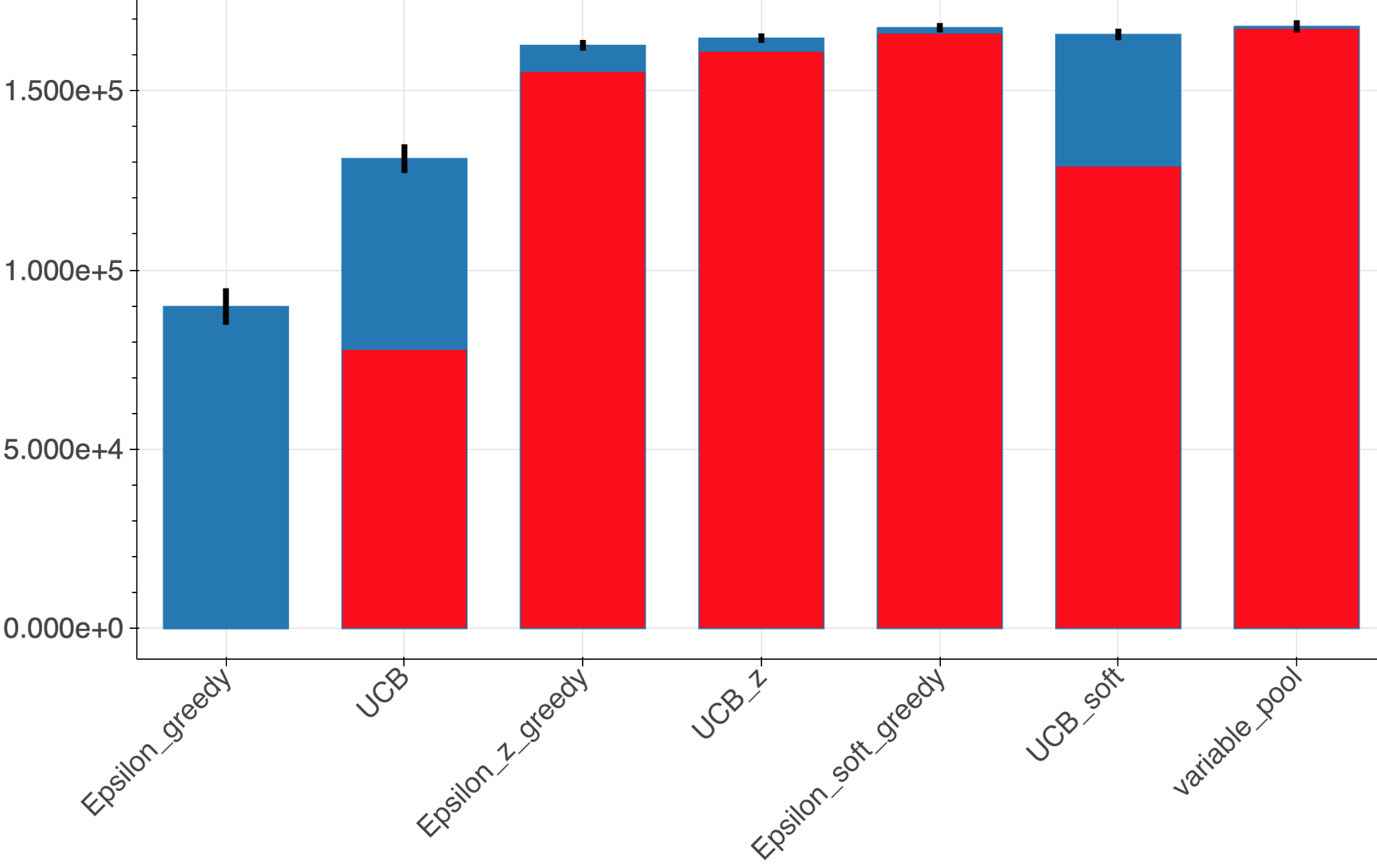} }\label{Truncated_Normal_Christmas_50a_1500t}}\;\;
		\end{subfloatrow}
	}
	\caption{Comparison of average final rewards in games with 50 arms, 1500 turns, and a Christmas-type greed function.}\label{Figure::50a_1500t_Christmas}
\end{figure}

\begin{figure}[H]%
	\makebox[\textwidth][c]{ %to center figures!
		\begin{subfloatrow}
			\subfloat[\small{Rewards from Bernoulli distributions. }]{{\includegraphics[width=6.7cm,height=4.0cm]{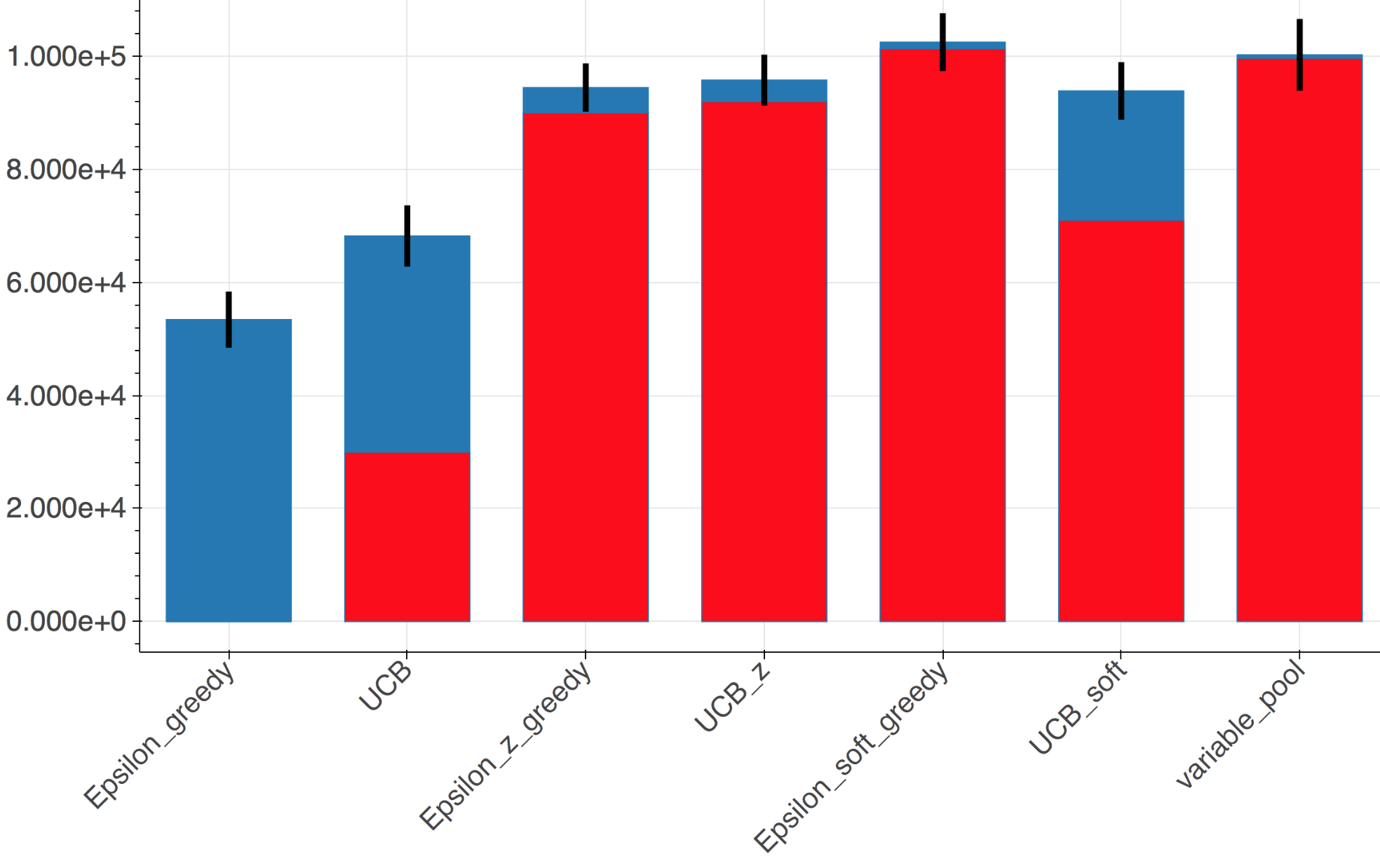} }\label{Bernoulli_Christmas_100a_1500t}}%
			\;\;
			\qquad
			\subfloat[\small{Rewards from truncated Normal distributions.
			}]{{\includegraphics[width=6.7cm,height=4.0cm]{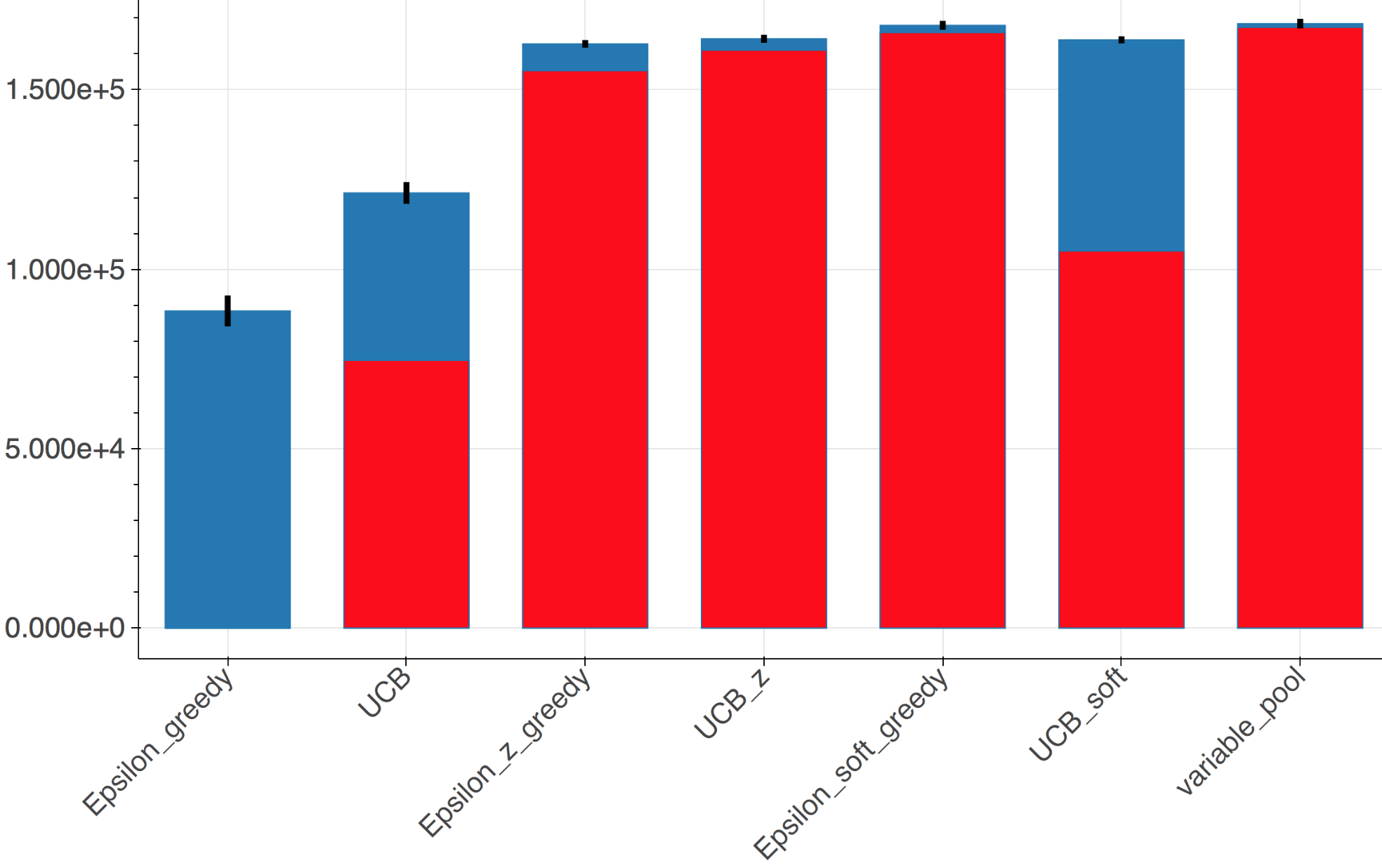} }\label{Truncated_Normal_Christmas_100a_1500t}}\;\;
		\end{subfloatrow}
	}
	\caption{Comparison of average final rewards in games with 100 arms, 1500 turns, and a Christmas-type greed function.}\label{Figure::100a_1500t_Christmas}
\end{figure}

\begin{figure}[H]%
	\makebox[\textwidth][c]{ %to center figures!
		\begin{subfloatrow}
			\subfloat[\small{Rewards from Bernoulli distributions. }]{{\includegraphics[width=6.7cm,height=4.0cm]{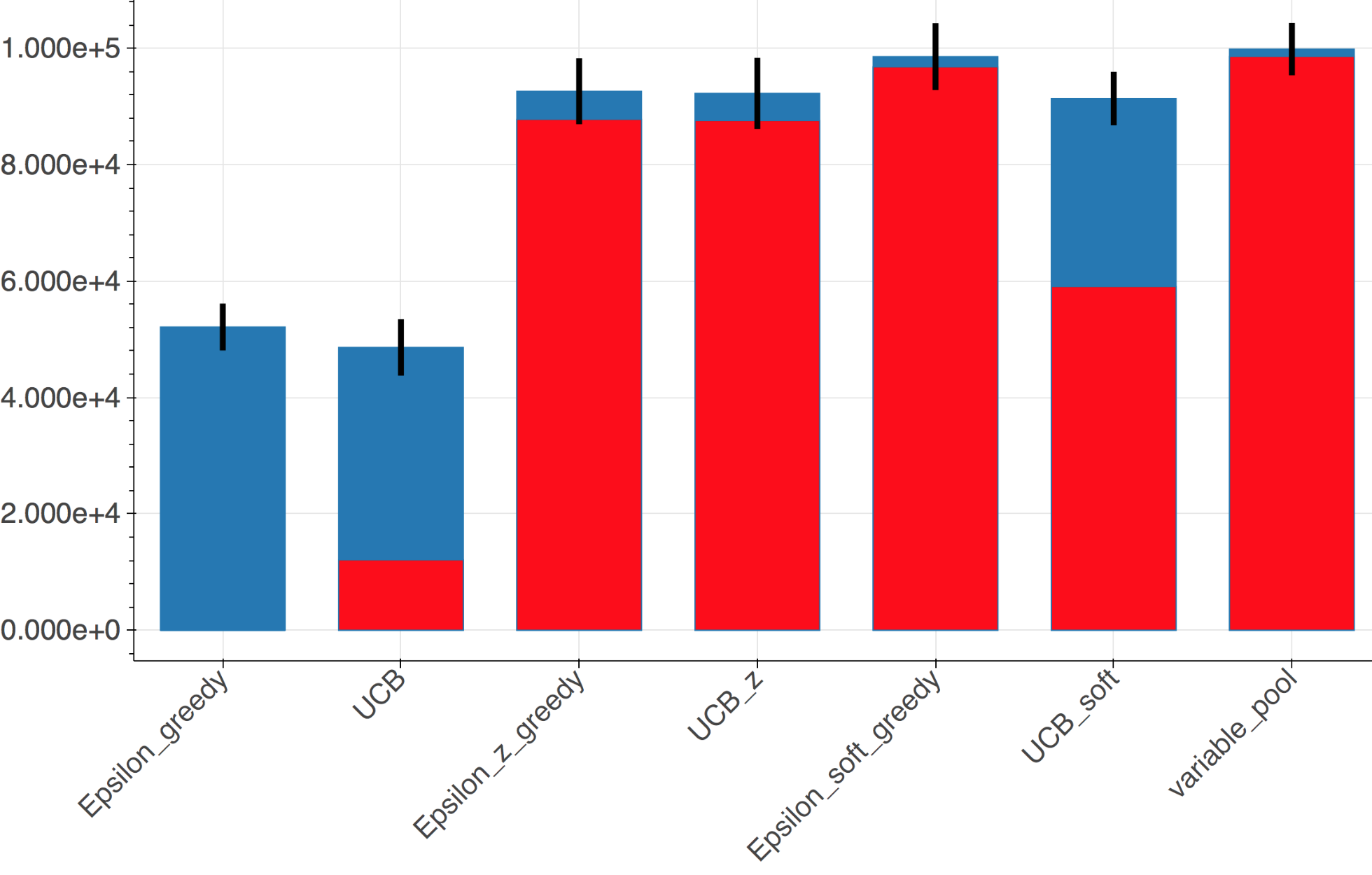} }\label{Bernoulli_Christmas_200a_1500t}}%
			\;\;
			\qquad
			\subfloat[\small{Rewards from truncated Normal distributions.
			}]{{\includegraphics[width=6.7cm,height=4.0cm]{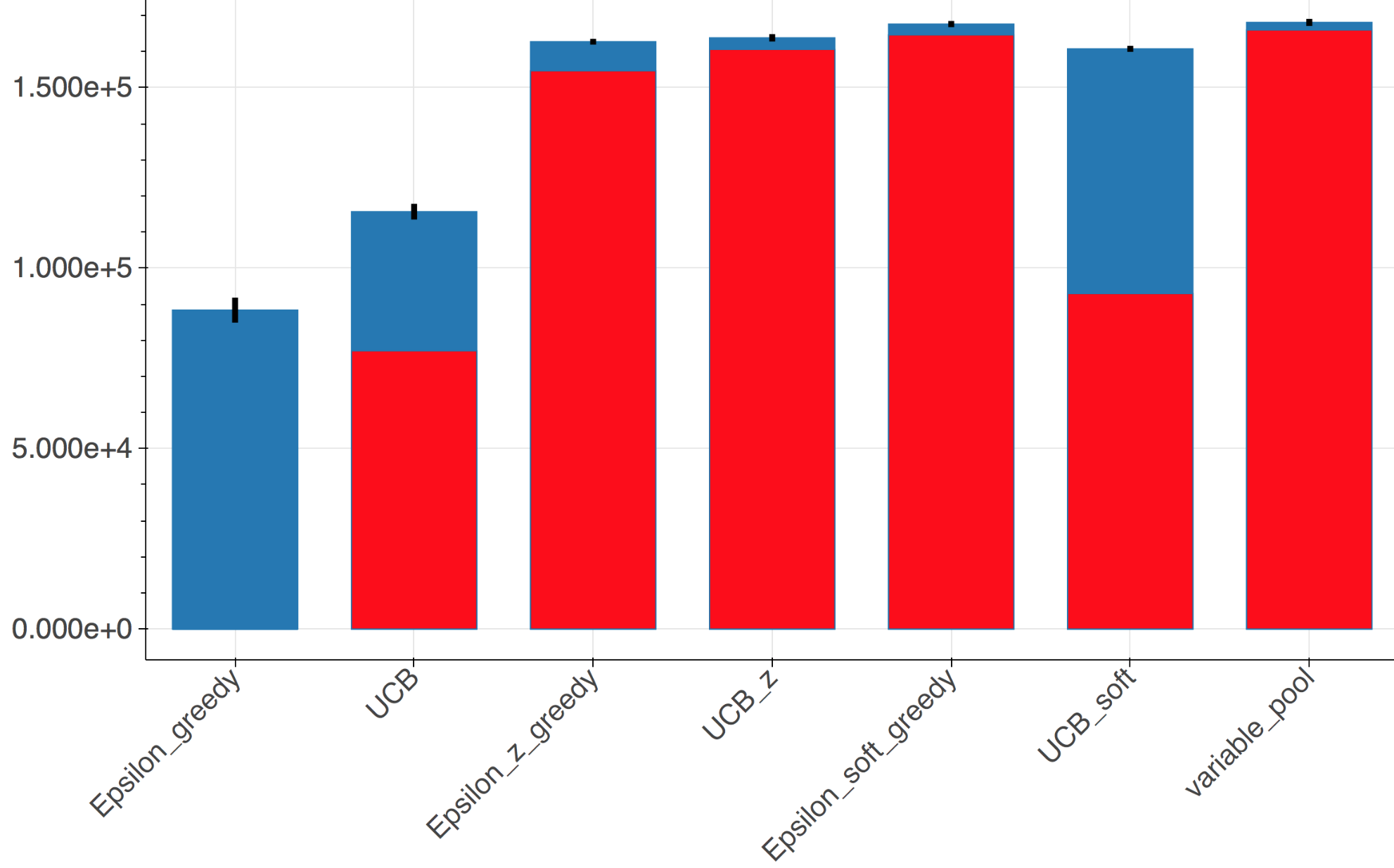} }\label{Truncated_Normal_Christmas_200a_1500t}}\;\;
		\end{subfloatrow}
	}
	\caption{Comparison of average final rewards in games with 200 arms, 1500 turns, and a Christmas-type greed function.}\label{Figure::200a_1500t_Christmas}
\end{figure}

%\subsection{Cumulative Reward increase when regulating greed over time compared to the (smarter) $\varepsilon$-greedy algorithm and the (smarter) UCB algorithm with a Christmas-type greed function.}

\begin{figure}[]
	\caption{Average increase in rewards (coming from Bernoulli distributions) compared to the (smarter) version of the $\varepsilon$-greedy algorithm (Algorithm \ref{Algorithm::epsilon_slightly_smarter}) with a Christmas-type greed function. }
	\centering
	\includegraphics[width=0.95\textwidth]{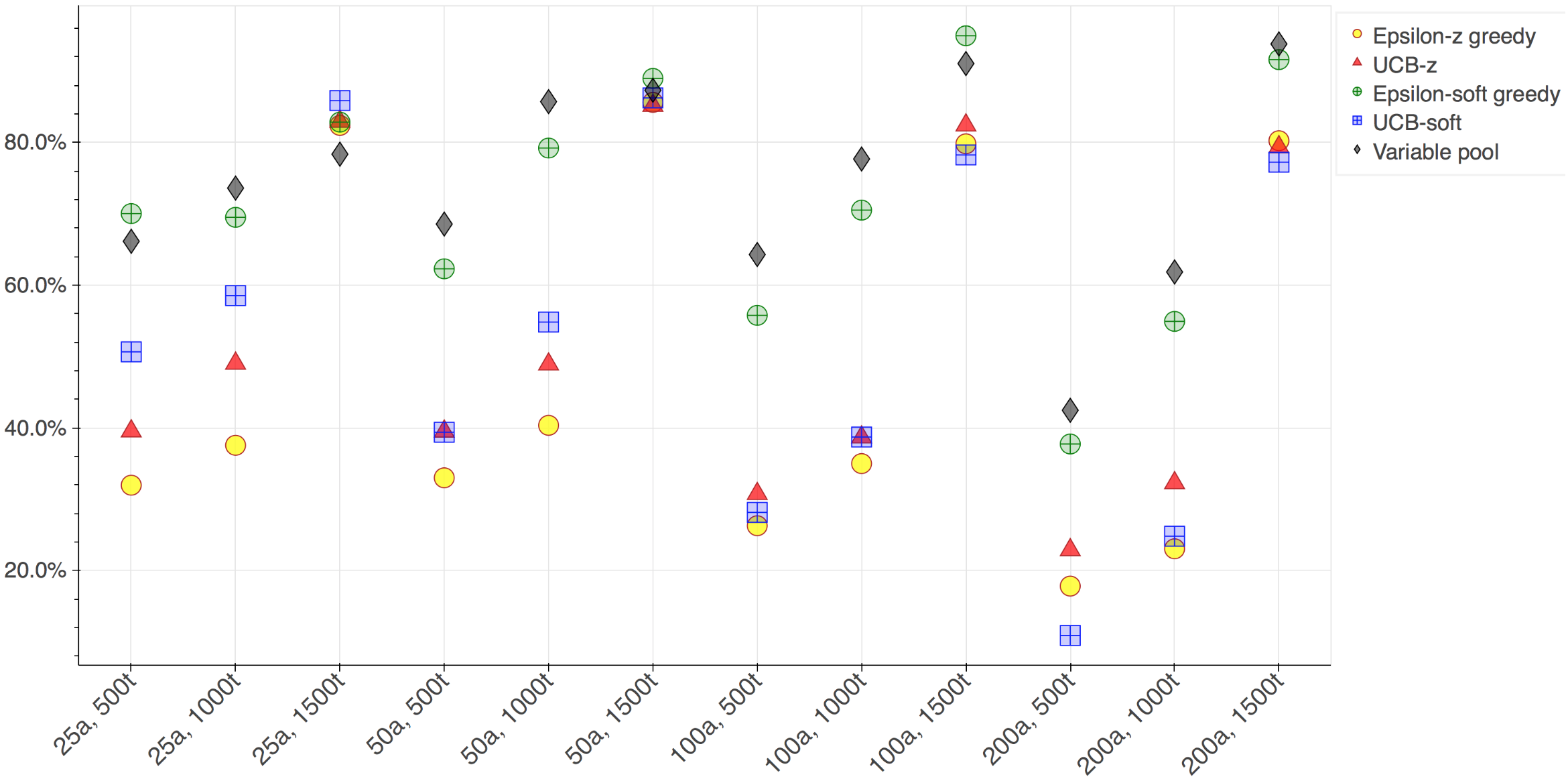}
	\label{BEC}
\end{figure}

\begin{figure}[]
	\caption{Average increase in rewards (coming from Bernoulli distributions) compared to the (smarter) version of the UCB algorithm (Algorithm \ref{Algorithm::UCB_slightly_smarter}) with a Christmas-type greed function. }
	\centering
	\includegraphics[width=0.95\textwidth]{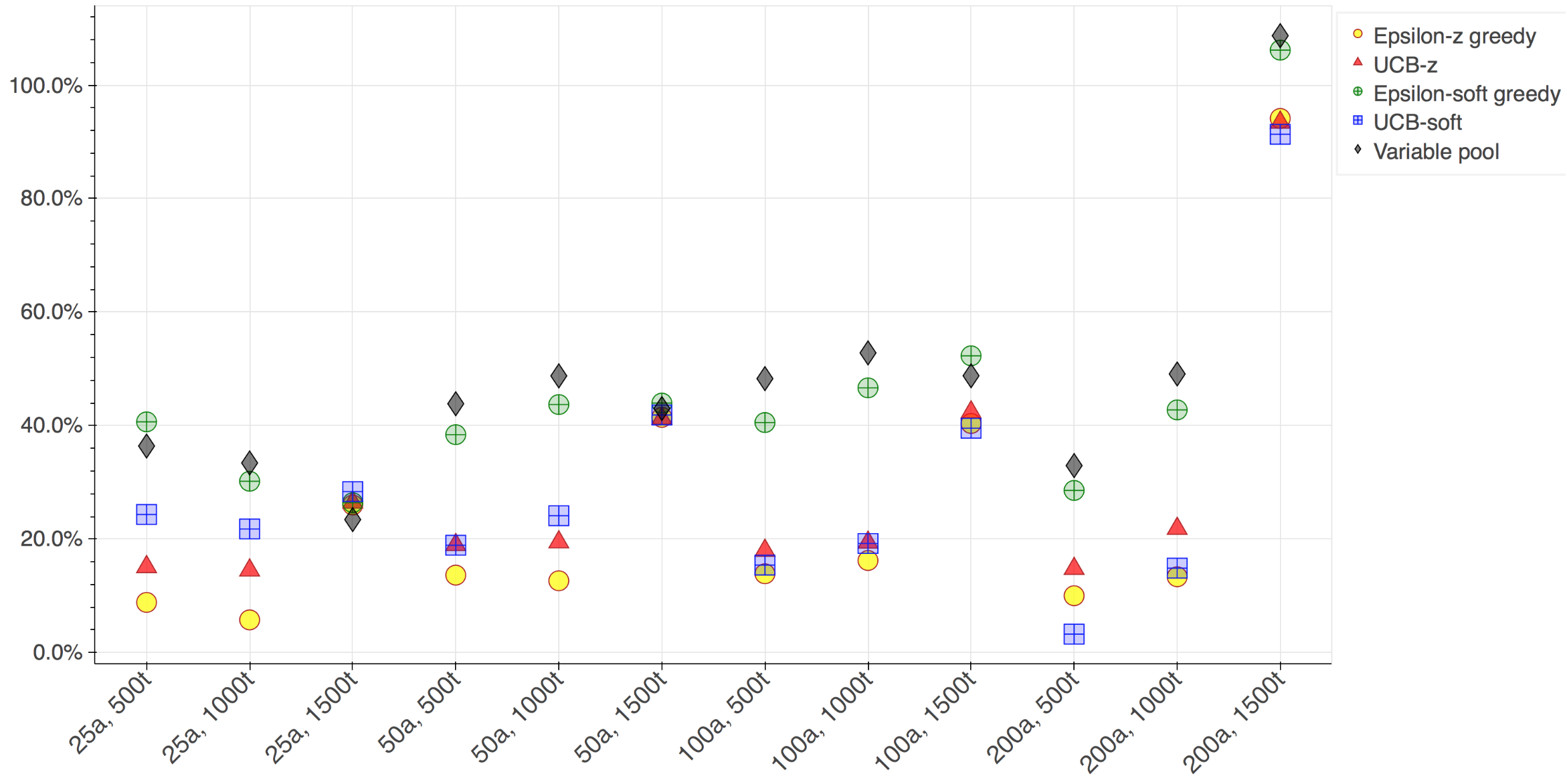}
	\label{BUC}
\end{figure}

\begin{figure}[]
	\caption{Average increase in rewards (coming from Truncated-Normal distributions) compared to the (smarter) version of the $\varepsilon$-greedy algorithm (Algorithm \ref{Algorithm::epsilon_slightly_smarter}) with a Christmas-type greed function. }
	\centering
	\includegraphics[width=0.99\textwidth]{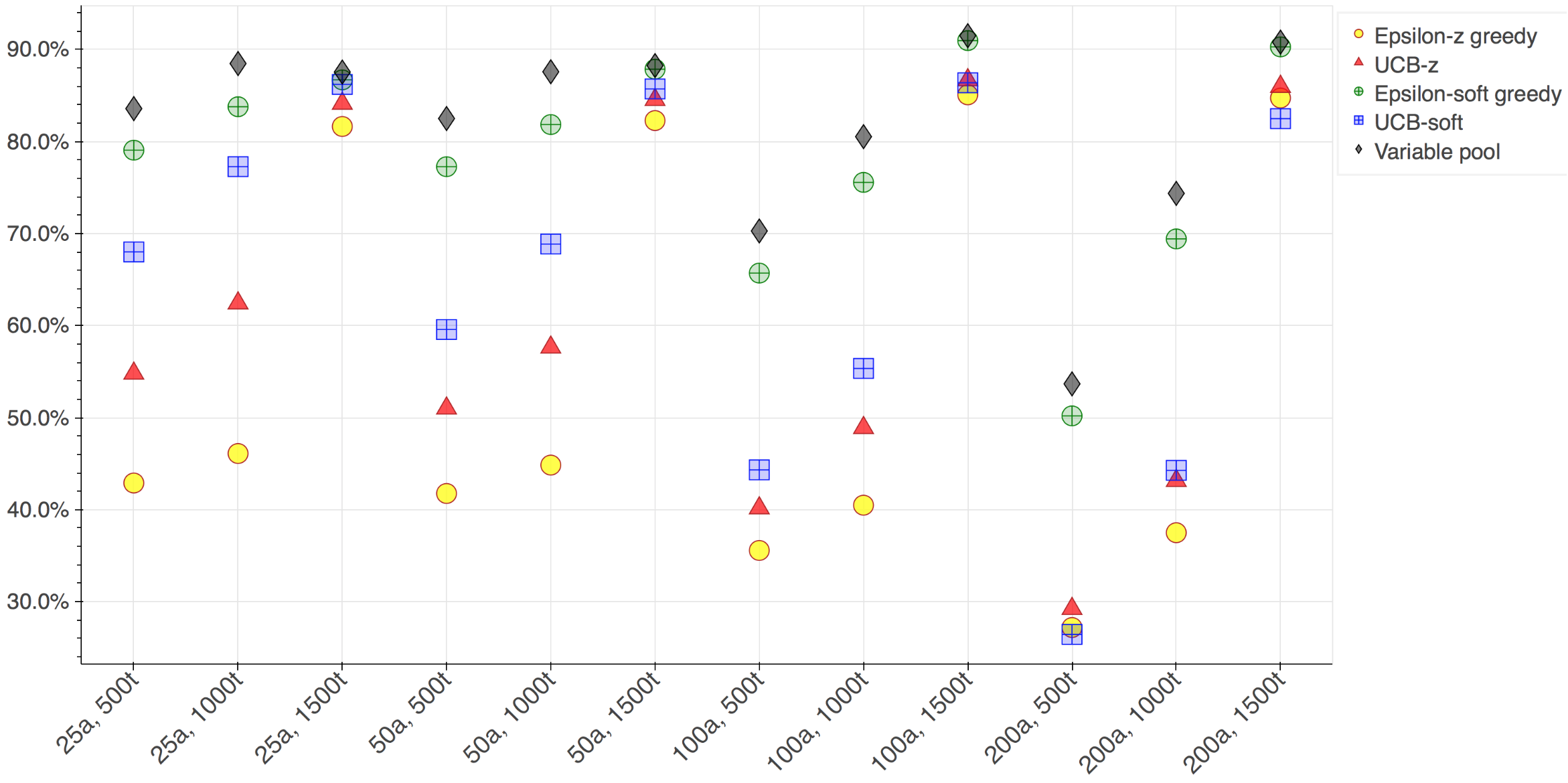}
	\label{NEC}
\end{figure}

\begin{figure}[]
	\caption{Average increase in rewards (coming from Truncated-Normal distributions) compared to the (smarter) version of the UCB algorithm (Algorithm \ref{Algorithm::UCB_slightly_smarter}) with a Christmas-type greed function. }
	\centering
	\includegraphics[width=0.95\textwidth]{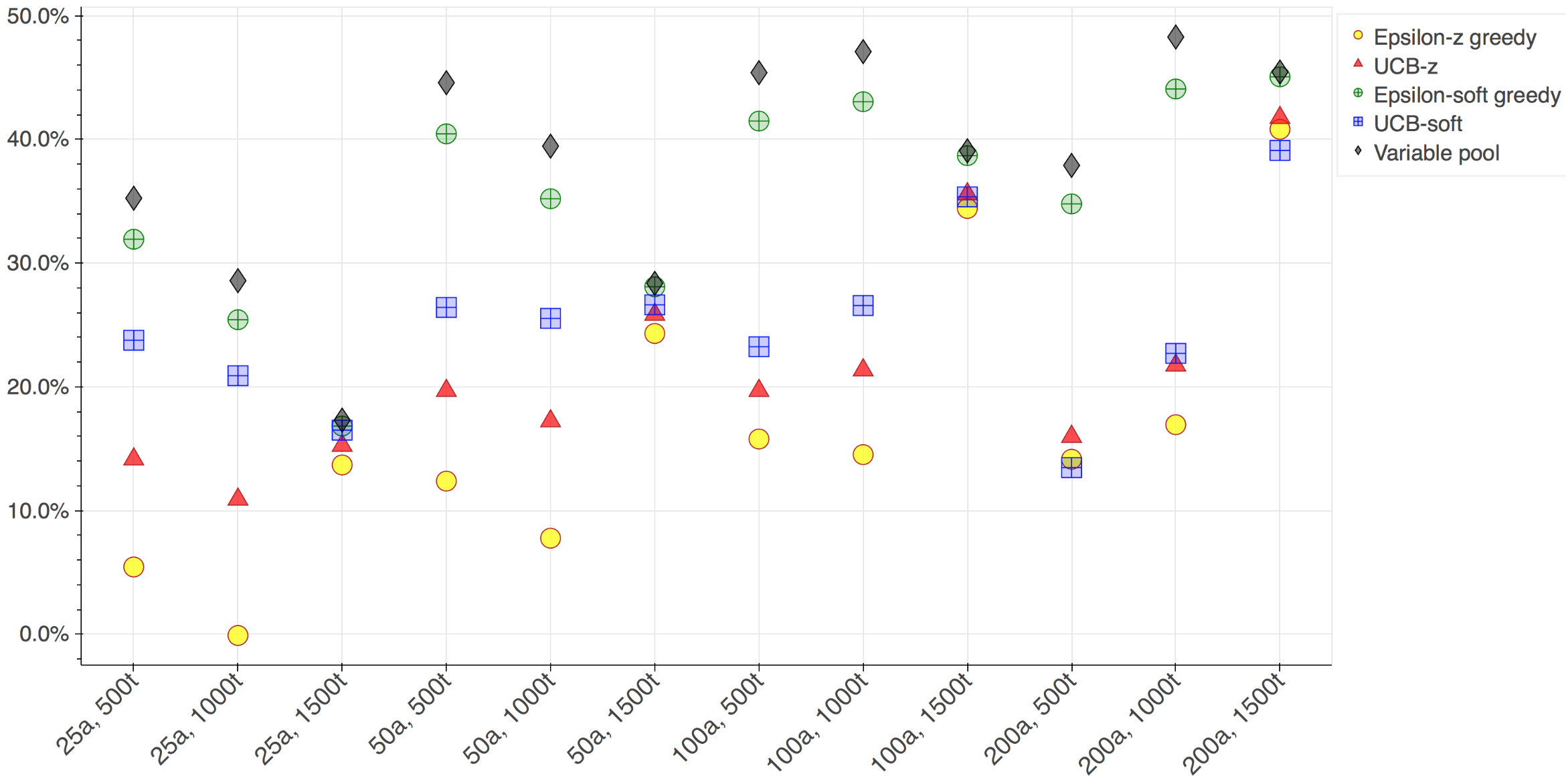}
	\label{NUC}
\end{figure}

\newpage

\section{Experiment details}
\label{Appendix::Experiment details}

\subsection{Offline evaluation methodology}
Each entry of the Yahoo! Webscope dataset contains information on:
\begin{itemize}
	\item the arm pulled (which is the article shown to the human viewing articles on Yahoo!);
	\item the outcome (whether the article was clicked or not);
	\item the pool of arms (articles) available at that time and the associated timestamp.
\end{itemize}
A unique property of this dataset is that the displayed article was chosen uniformly at random from a pool of available articles. Therefore, it is possible to use an unbiased offline evaluation method \citep[see][]{li2010contextual} to compare bandit algorithms in a reliable way. The key point of the method is to ignore the events in which the article chosen at random does not match the article chosen by the algorithm. This means that a large portion of the dataset gets discarded because it is very common that the random choice and the algorithm choice do not match. 
In the initialization phase, we made the simulation faster %(there are only 25 turns for the initialization so that we have more records to analyze the performance of the algorithms after initialization) 
and simpler: by pre-scanning the dataset, we can detect the articles that never match the recommendations chosen by the algorithm, and therefore we do not initialize such articles.% if the record feeds an algorithm an uninitialized article, initialize it; If the record feeds the algorithm an initialized article, ignore it.

% \ref{Initialization}, since initialization only plays 25 turns in a game and we care more about what happens later on.
% \begin{algorithm}
%     \label{Initialization}
%     \caption{Initialization}
%     \begin{algorithmic}
%         \STATE event stream $ Stream $
%         \STATE number of turns as initialization $ m $
%         \STATE $ i \gets 0 $
%         \WHILE{$ i<m $}
%         \STATE $ Record \gets Stream.next\_even()t $
%         \IF{$Record.article$ was not seen before}
%         \STATE update expectation of $Record.article$
%         \STATE $ i \gets i+1 $
%         \ENDIF
%         \ENDWHILE
%     \end{algorithmic}
% \end{algorithm}
A difference that the Yahoo! dataset has from our setting is that each visitor receives a different article recommendation (as opposed to customers receiving the same recommendation as is common in application like retail). Table \ref{actual dataset} shows how the data is structured in the Yahoo! dataset, while Table \ref{optimal dataset} shows how the data is expected by the setting described Section \ref{Section::problem-setup}. %The value of $G(t)$ is computed by binning records that appear within a second.
%Our algorithms assume each turn is associated with variant number of users, an optimal structure of the dataset is like Table \ref{optimal dataset}. However, the random arm-picker in the records considers each turn associated with one user, the actual structure of the dataset is like Table \ref{actual dataset}. We could first group the events by time bins, then treated all the recommendations in a time bin as one action in a turn. This would require the recommendations for all the users in one time bin to be the same. 
In order to reconcile this difference, we follow the structure of the event stream, treating each turn associated with one user, but we use $G(t)$ to decide at each turn how to balance exploration and exploitation according to the strategy of our algorithms. %That is, if $G(t)$ is high, do exploitation; if not, follow the standard exploration/exploitation strategy.

\begin{table}[]
	\centering
	\caption{Raw structure of the Yahoo! dataset. Each user receives a personalized recommendation. The value of $G(t)$ is computed by binning records that appear within a second.}
	\label{actual dataset}
	\begin{tabular}{ | c | c | c | c | c | c | c | } 
		\hline
		turn & timestamp & user & displayed article & cliced? & article pool & G(t)\\ 
		\hline
		1 & 1317513291 & user1 & id-560620 & 0 & id-560620, ... & 32\\
		\hline
		2 & 1317513291 & user2 & id-565648 & 0 & id-560620, ... & 32\\
		\hline
		3 & 1317513291 & user3 & id-563115 & 1 & id-560620, ... & 32\\
		\hline
		... & ... & ... & ... & ... & ... & ...\\
		\hline
		32 & 1317513291 & user32 & id-563115 & 0 & id-560620, ... & 32\\
		\hline
		33 & 1317513292 & user33 & id-552077 & 1 & id-552077, ... & 13\\
		\hline
		34 & 1317513292 & user34 & id-564335 & 0 & id-552077, ... & 13\\
		\hline
		... & ... & ... & ... & 0 & ... & ...\\
		\hline
		45 & 1317513292 & user45 & id-564335 & 1 & id-552077, ... & 13\\
		\hline
		... & ... & ... & ... & ... & ... & ...\\
		\hline
	\end{tabular}
\end{table}

\begin{table}[]
	\centering
	\caption{Expected structure of the dataset following the setting described in Section \ref{Section::problem-setup}. Visitors are binned in groups of size $G(t)$ and they all receive the same recommendation.}
	\label{optimal dataset}
	\begin{tabular}{ | c | c | c | c | c | c | c | } 
		\hline
		turn & timestamp & user & displayed article & clicked? & article pool & $G(t)$\\ 
		\hline
		1 & 1317513291 & user1 & id-560620 & 0 & id-560620, ... & 32\\
		&            & user2 &           &  &               & \\
		&            & user3 &           &  &               & \\
		&            & ...   &           &  &               & \\
		&            & user32 &           &  &               & \\
		\hline
		2 & 1317513292 & user33 & id-552077 & 1 &  id-552077, ...             & 13\\
		&            & user34 &           &  &               & \\
		&            & ... &           &  &               & \\
		&            & user45 &           &  &               & \\
		\hline
		... & ... & ... & ... & ... & ... & ...\\
		\hline
	\end{tabular}
\end{table}

\subsection{Choice of the parameters}

The algorithms have some tunable parameters. Table \ref{hyperparameters} lists the parameters chosen for each algorithm. They were chosen by tuning them on a small sample of the data. In practice, it is always useful to have data that resembles the one that the algorithm will use so that parameters can be tuned. 
The value of $z = 31$ corresponds to the 75th quantile of $G(t)$.

\begin{table}[]
	\centering
	\caption{Parameters of the algorithms used in the simulation.}
	\label{hyperparameters}
	\begin{tabular}{ | c | c | } 
		\hline
		algorithm & hyperparameters \\
		\hline
		$\varepsilon$-z Greedy & $z=31$ \\
		Variable Pool & $c=10$, $z=31$ \\
		UCB-L & $c=0.011$ \\
		UCB-$z$ & $z=31$ \\
		\hline
	\end{tabular}
\end{table}

\newpage

\section{When $G(t)$ is not known exactly}
In this Appendix we show how to change the theorems when $G(t)$ is not known. We also show that the final regret did not change considerably in the simulations when using some standard predictive methods to estimate $G(t+1)$ by using the information available in turn $t$.  

\subsection{Adapting regret bound theorems for the case of estimated $G(t)$}\label{Appendix::Unknown_G_th}
The theorems regarding the regret bound on the final cumulative reward are similar to the case when $G(t)$ is known.
Suppose we estimate $G(t)$ with $H(t)$, where $H(t)$ is estimated using any method. Then, for example,
%Then the regret bound will depend on both the real underlying multiplier function $G(t)$ and its estimator used to balance exploration and exploitation.
%For example 
Theorem \ref{Theorem::epsilon_z} becomes:
\tcbset{colback=blue!2!white}
\begin{tcolorbox}
	\begin{theorem}[$\varepsilon$-greedy algorithm with hard threshold and estimated multiplier function]\label{Theorem::epsilon_threshold_estimatedG}
		The bound on the mean regret $\E[R_n]$ at time $n$ is given by\footnotesize
		\begin{eqnarray}
			\E[R_n]\displaystyle  &\leq&  \displaystyle\sum_{j=1}^m G(j) \Delta_j  \nonumber\\
			& + &  \displaystyle\sum_{t=m+1}^n G(t)\ONE_{\{G(t)< z \hspace{0.1cm}\text{and}\hspace{0.1cm} H(t)<z\}}\sum_{j : \mu_j<\mu_*} \Delta_j \left( \varepsilon_t\frac{1}{m}+(1-\varepsilon_t) \beta_j(\tilde{t}) \right) \nonumber\\
			& + & \displaystyle\sum_{t=m+1}^n G(t)\ONE_{\{G(t) \geq z \hspace{0.1cm}\text{and}\hspace{0.1cm} H(t)<z\}}\sum_{j : \mu_j<\mu_*} \Delta_j \left( \varepsilon_t\frac{1}{m}+(1-\varepsilon_t) \beta_j(\tilde{t}) \right) \label{low_epsilon_threshold_estimatedH}\\
			& + &\displaystyle\sum_{t=m+1}^n G(t)\ONE_{\{G(t) < z\hspace{0.1cm}\text{and}\hspace{0.1cm} H(t)\geq z\}} \sum_{j : \mu_j<\mu_*} \Delta_j \beta_j(\tilde{t}) , \label{high_epsilon_threshold_estimatedH}\\
			& + &\displaystyle\sum_{t=m+1}^n G(t)\ONE_{\{G(t)\geq z\hspace{0.1cm}\text{and}\hspace{0.1cm} H(t)\geq z\}} \sum_{j : \mu_j<\mu_*} \Delta_j \beta_j(\tilde{t}) ,\nonumber
		\end{eqnarray}
		\begin{equation*}\label{beta_threshold_estimatedG}
			\text{where}\hspace{0.3cm}\beta_j(\tilde{t})=k\left( \frac{\tilde{t}}{m k e }\right) ^{-\frac{k}{10}}\log \left( \frac{\tilde{t}}{m k e } \right)   + \frac{4}{\Delta_j^2} \left( \frac{\tilde{t}}{m k e } \right)^{-\frac{k \Delta_j^2 }{4}}
		\end{equation*}\normalsize
		and $\tilde{t}$ is the number of times the estimated multiplier $H(t)$ has been under the threshold.
	\end{theorem}
\end{tcolorbox}
Summations \eqref{low_epsilon_threshold_estimatedH} and \eqref{high_epsilon_threshold_estimatedH} are the extra addends that appear when the estimator and the true multiplier function are not both either under or above the threshold $z$:
\begin{itemize}
	\item the term in \eqref{low_epsilon_threshold_estimatedH} represents the regret incurred when the algorithm balances exploration and exploitation (because the estimated $H(t)$ is below the threshold $z$) when it should have been exploiting (because $G(t)$ is actually above the threshold $z$);
	\item the term in \eqref{high_epsilon_threshold_estimatedH} represents the regret incurred when the algorithm is exploiting (because the estimated $H(t)$ is above the threshold $z$) when it should have been balancing exploration and exploitation (because $G(t)$ is actually below the threshold $z$).
\end{itemize}
Similarly, it is possible to rewrite the regret bound theorems of all the other algorithms for the case when $G(t)$ is estimated by $H(t)$:
\begin{itemize}
	\item for the UCB-$z$ algorithm, the regret bound would have two extra terms for when $G(t)$ and $H(t)$ are not above or below the threshold $z$ at the same time;
	\item for the Soft $\varepsilon$-greedy algorithm, the regret bound still depends on $G(t)$, but $\psi(t)$ depends on $H(t)$;
	\item for the Soft UCB algorithm, the regret bound still depends on $G(t)$, but $\xi(t)$ depends on $H(t)$;
	\item for the variable pool algorithm, the regret bound still depends on $G(t)$, but $\lambda_t$ depends on $H(t)$;
	\item for the Soft UCB mortal algorithm, the regret bound still depends on $G(t)$, but  $\psi_{\text{future}}(j,t)$ and $\xi_{\text{present}}(t)$ depend on $H(t)$.
\end{itemize}

%If $G(t)$ is not known, it may be possible to estimate. $G(t)$ could be modeled by seasonal trends, or by estimating $G(t)$ by $G(t-1)$ if $G(t)$ is a martingale, or by using data coming from related problems (e.g., number of customers buying gloves rather than scarves). In Appendix \ref{Appendix::Unknown_G} we use the following simple methods to estimate the multiplier function:
%\begin{itemize}
%	\item simple random walk;
%	\item simple average;
%	\item moving average;
%	\item moving weighted average;
%	\item simple exponential smoothing;
%	\item double exponential smoothing.
%\end{itemize}

\subsection{Rewards change when estimating $G(t+1)$ in turn $t$}\label{Appendix::Unknown_G}

We used some standard methods for predicting $G(t+1)$ by using the information available in turn $t$:

\begin{itemize}
	\item simple random walk;
	\item simple average;
	\item moving average;
	\item moving weighted average;
	\item simple exponential smoothing;
	\item double exponential smoothing.
\end{itemize}

\begin{figure}[]
	\caption{Final rewards when predicting $G(t)$ step by step compared to when knowing $G(t)$, with Bernoulli rewards and Wave-type greed function.}
	\centering
	\includegraphics[width=0.95\textwidth]{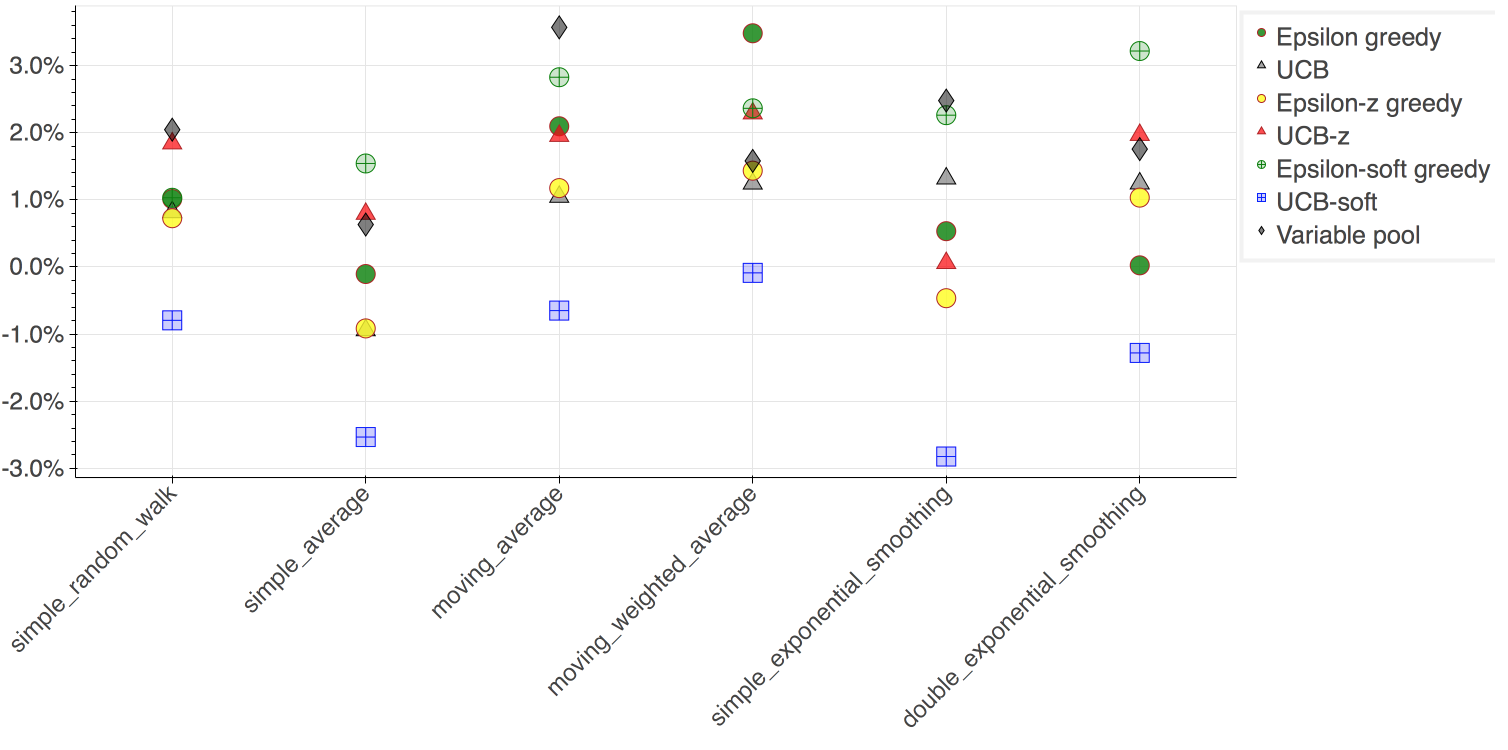}
	\label{BW}
\end{figure}

\begin{figure}[]
	\caption{Final rewards when predicting $G(t)$ step by step compared to when knowing $G(t)$, with Bernoulli rewards and Christmas-type greed function.}
	\centering
	\includegraphics[width=0.95\textwidth]{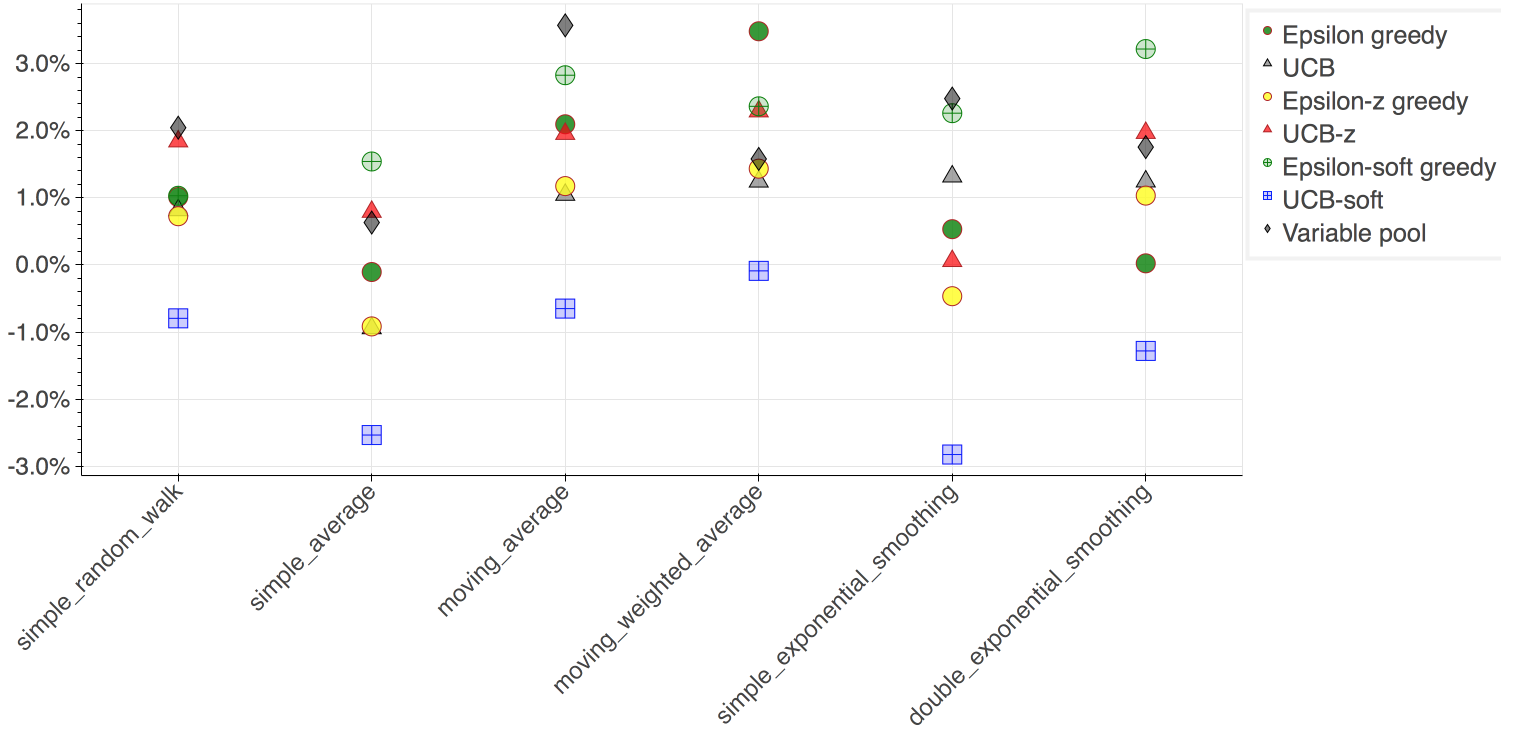}
	\label{BC}
\end{figure}

\begin{figure}[]
	\caption{Final rewards when predicting $G(t)$ step by step compared to when knowing $G(t)$, with Bernoulli rewards and Step-type greed function.}
	\centering
	\includegraphics[width=0.95\textwidth]{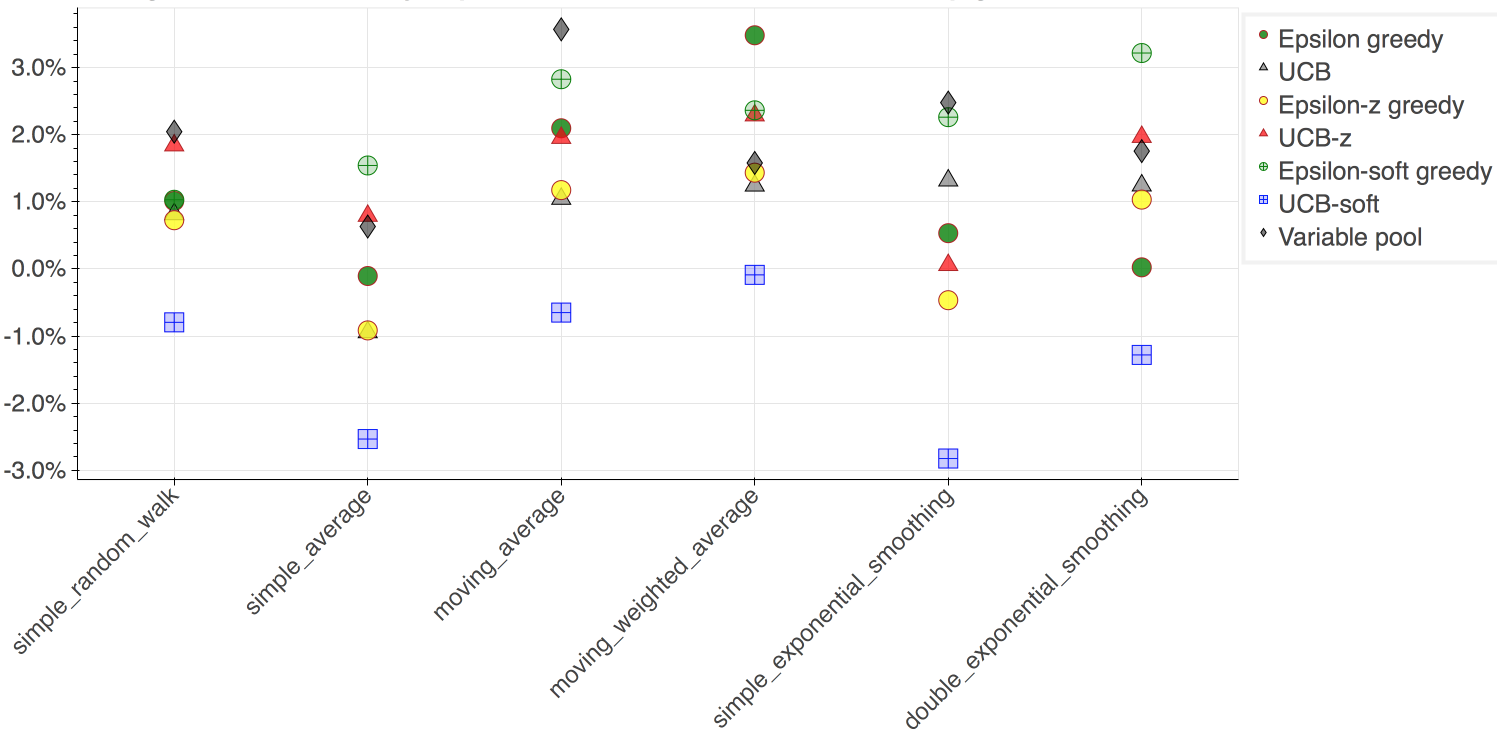}
	\label{BS}
\end{figure}

\begin{figure}[]
	\caption{Final rewards when predicting $G(t)$ step by step compared to when knowing $G(t)$, with Truncated-Normal rewards and Wave-type greed function.}
	\centering
	\includegraphics[width=0.95\textwidth]{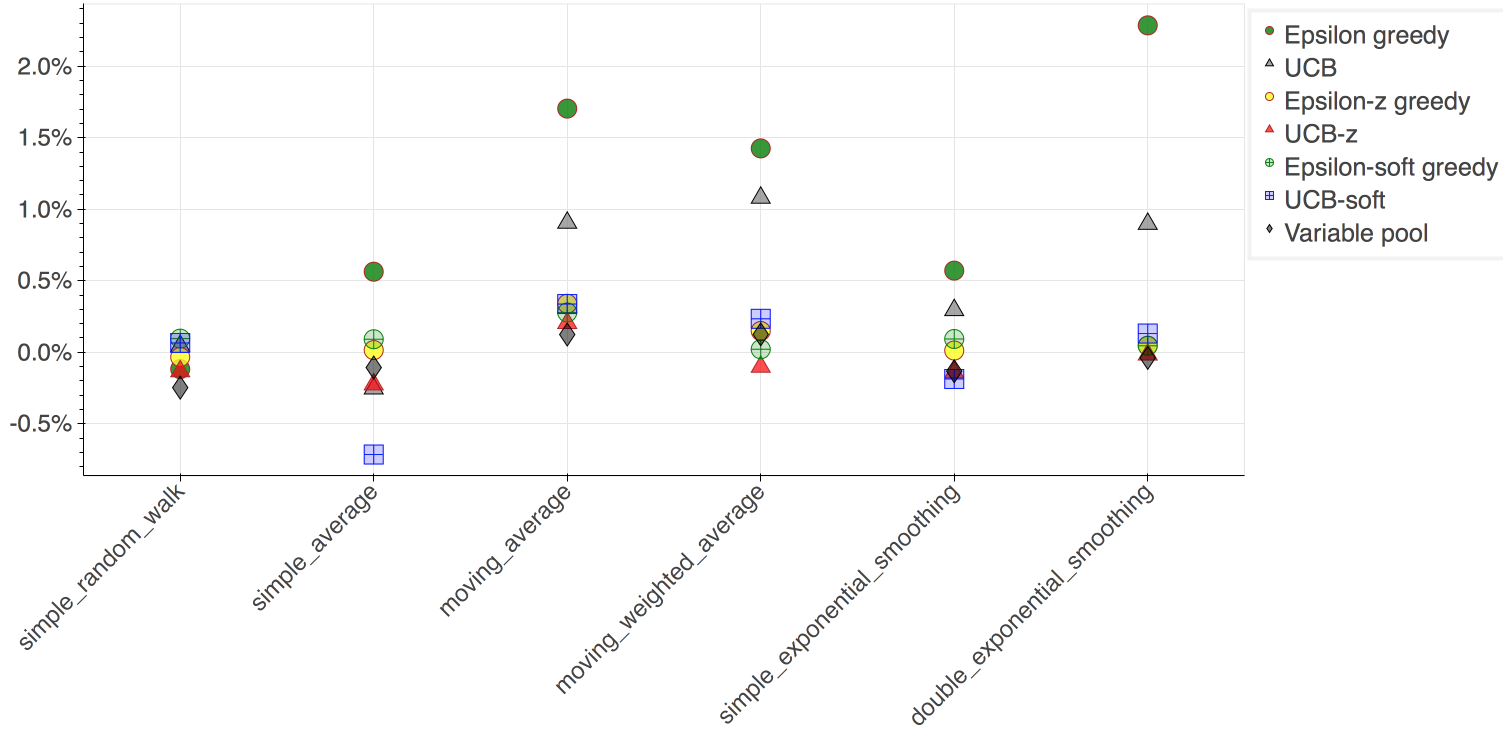}
	\label{TW}
\end{figure}

\begin{figure}[]
	\caption{Final rewards when predicting $G(t)$ step by step compared to when knowing $G(t)$, with Truncated-Normal rewards and Christmas-type greed function.}
	\centering
	\includegraphics[width=0.95\textwidth]{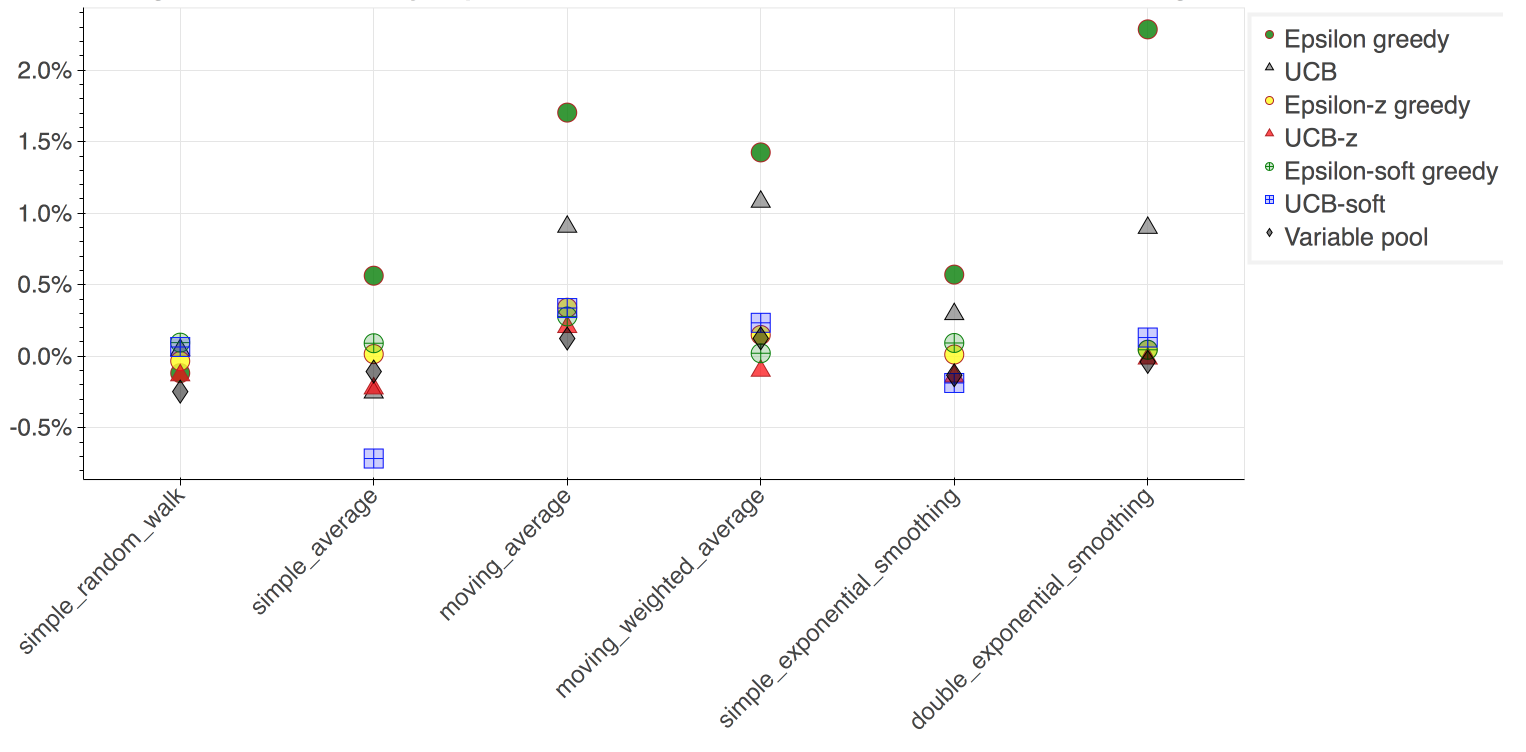}
	\label{TC}
\end{figure}

\begin{figure}[]
	\caption{Final rewards when predicting $G(t)$ step by step compared to when knowing $G(t)$, with Truncated-Normal rewards and Step-type greed function.}
	\centering
	\includegraphics[width=0.95\textwidth]{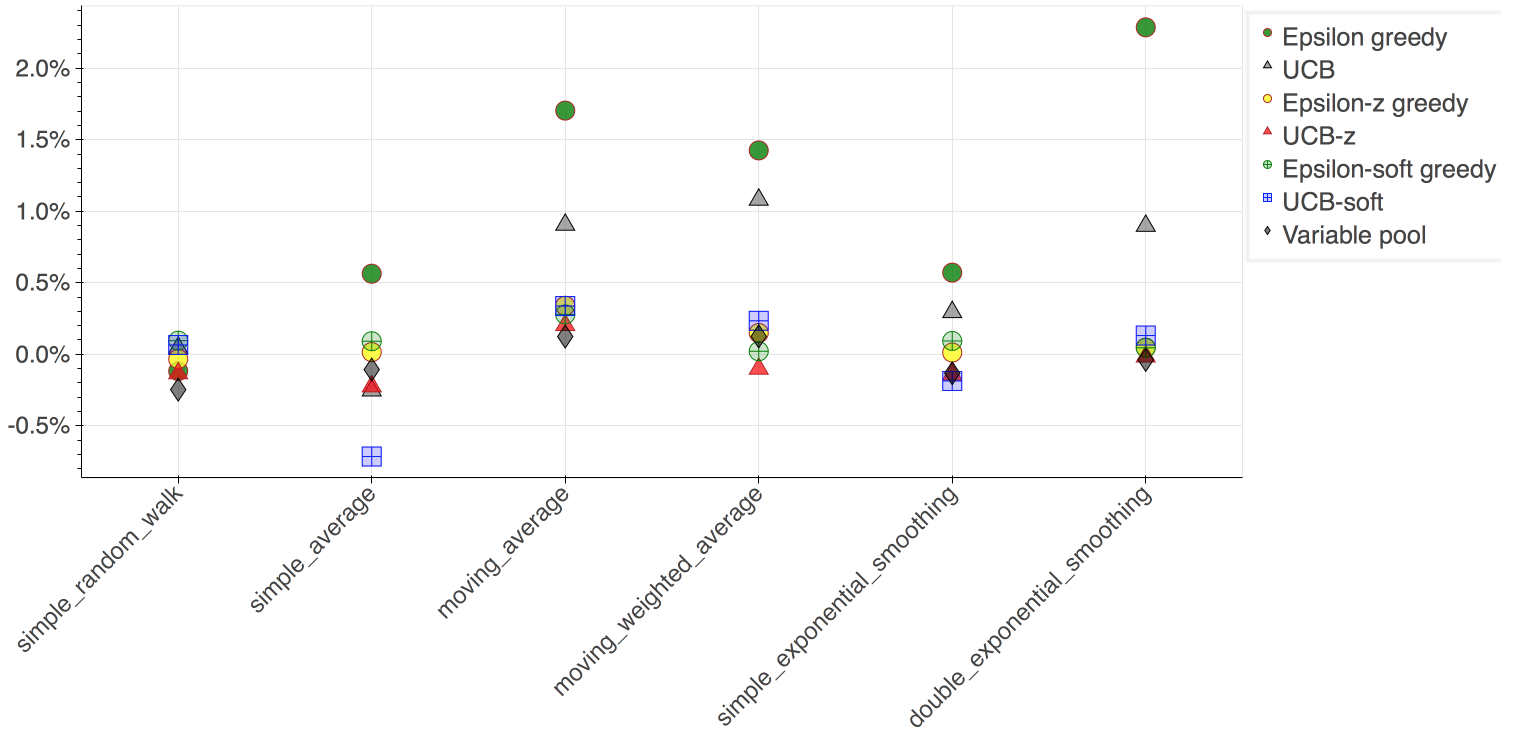}
	\label{TS}
\end{figure}

\noindent The difference in final rewards is not significantly different from when $G(t)$ is known. Figures \ref{BW}-\ref{TS} show the percentage change in final regret from the case of known $G(t)$ in the simulation. 

\newpage

\newpage
\section{Notation summary}\nllabel{Section::notation}
%\tcbset{colback=gray!5!white}
%\begin{tcolorbox}\footnotesize
\begin{itemize}
	\item $\beta_j(\tilde{t})$: upper bound on the probability of considering arm $j$ being the best arm at round $\tilde{t}$ when using Algorithm \ref{Algorithm::epsilon_threshold};
	\item $\beta_j^S(t)$: upper bound on the probability of considering arm $j$ being the best arm at round $t$ when using Algorithm \ref{Algorithm::epsilon_soft};
	\item $\beta_j^{\text{old}}(t)$: upper bound on the probability of considering arm $j$ being the best arm at round $t$ when using Algorithm \ref{Algorithm::epsilon_slightly_smarter};
	\item $\beta_j^U(t)$: upper bound on the probability of considering arm $j$ being the best arm at round $t$ when using Algorithm \ref{Algorithm::UCB_threshold};
	\item $\beta_j^{VP}(t)$: upper bound on the probability of considering arm $j$ being the best arm at round $t$ when using Algorithm \ref{Algorithm::z_pool};
	\item $\beta_j^M$: upper bound on the expected number of times suboptimal arm $j$ is pulled when using Algorithm \ref{Algorithm::UCB-LG};
	\item $B$: set of rounds when the ``high reward" zone is entered in Algorithm \ref{Algorithm::UCB_threshold} ($B=\{t: G(t-1)<z, G(t)>z\}$);
	\item $c$: a constant greater than $10$ in Algorithm \ref{Algorithm::epsilon_slightly_smarter};
	\item $\Delta_j$: difference between the mean reward of the optimal arm and the mean reward of arm $j$ ($\Delta_j=\mu_*-\mu_j$);
	\item $\Delta_{j,z}$: difference between the mean reward of the optimal arm in epoch $z$ and the mean reward of arm $j$;
	\item $d$: a constant such that $d<\min_j \Delta_j$ and $0<d<1$ in Algorithm \ref{Algorithm::epsilon_slightly_smarter};
	\item $\varepsilon_t$: probability of exploration at turn $t$ (used in Algorithm \ref{Algorithm::epsilon_threshold} and Algorithm \ref{Algorithm::epsilon_soft});
	\item $E_j$: number of epochs in which $L_j$ is partitioned depending on $i^*_t$;
	\item $\gamma$: lowest value of $\psi(t)$ ($\gamma=\min_{s\in\{m+1,\cdots,n\}}\psi(s)$);
	\item $G: \{1,\cdots,n\}\rightarrow \R^+$: known multiplier function; 
	\item $I_t$: arm played at turn $t$; 
	\item $i^*_t$: best arm in turn $t$ (mortal case);
	\item $k$: a constant greater than $10$ such that $k>4/(\min_j \Delta_j^2)$ in Algorithm \ref{Algorithm::epsilon_threshold} and Algorithm \ref{Algorithm::epsilon_soft};
	\item $L_j$: set of turns during which arm $j$ is available;
	\item $L_j^z$: epoch $z$ of the set $L_j$ depending on $i^*_t$;
	\item $l_j$: last turn in $L_j$; %$l_j^z$: last turn in epoch $L_j^z$;
	\item $l_j^z$: last turn in epoch $L_j^z$;
	\item $\mu_{*}$: mean reward of the optimal arm ($\mu_{*} = \max_{1\leq j \leq m} \mu_j$);
	\item $m$: number of arms;
	\item $m_t$: size of the arms pool in Algorithm \ref{Algorithm::z_pool};
	\item $M_t$: set of arms available in turn $t$ (mortal case);
	\item $n$: number of rounds;
	\item $\tilde{n}$: number of rounds under the threshold $z$ by the end of the game;
	\item $n'$: particular time defined as $km$ in the comparison between Algorithm \ref{Algorithm::epsilon_soft} and Algorithm \ref{Algorithm::epsilon_slightly_smarter} in Section \ref{Section::soft_espilon};
	\item $\psi(t)$: smoothing function used to define the probabilities of exploration $\varepsilon_t$ in Algorithm \ref{Algorithm::epsilon_soft} (see Figure \ref{plot_probabilities});
	\item $\psi_{\text{future}}(j,t)$: function that compares the scores of available arms in Algorithm \ref{Algorithm::UCB-LG};
	\item $R_n$: total regret at round $n$;
	\item $\text{score}_j$: score given by the sum of the values of $G(t)$ over the lifespan of arm $j$ (mortal case); 
	\item $s_j$: first turn in $L_j$; %$s_j^z$: first turn in epoch $L_j^z$;
	\item $s_j^z$: first turn in epoch $L_j^z$;
	\item $T_j(t-1)$: number of times arm $j$ has been played before round $t$ starts;
	\item $\tilde{t}$: number of rounds under the threshold $z$ up to time $t$;
	%\item $\Lambda_k=\max_{t\in Y_k} G(t)$: highest value of $G(t)$ on $Y_k$ in Algorithm \ref{Algorithm::UCB_threshold};
	%\item $R_k= \Lambda_k |Y_k| $: the maximum regret of the $k$th high reward zone in Algorithm \ref{Algorithm::UCB_threshold};
	\item $w$: first round such that $f(s) = km/s$ is less than $\gamma$ (i.e., $w = \min\{s: f(s) < \gamma \}$);
	\item $\xi(t)$: smoothing function used to define the decision rule in Algorithm \ref{Algorithm::UCB_soft};
	\item $\xi_{\text{present}}(t)$: smoothing function used to define the decision rule in Algorithm \ref{Algorithm::UCB-LG};
	\item $X_j(t)$: unscaled random reward for playing arm $j$ in turn $t$;
	%\item $X_j(t)G(t)$: actual reward for playing arm $j$ in turn $t$; 
	\item $\widehat{X}_j$: current estimate of $\mu_j$;
	\item $Y_k=\{t: t\geq y_k, G(t)>z, t<y_{k+1}\}$: set of rounds in the high-reward period entered at time $y_k$;
	\item $z$: regulating threshold (used in Algorithm \ref{Algorithm::epsilon_threshold} and Algorithm \ref{Algorithm::UCB_threshold});
\end{itemize}
%\end{tcolorbox}
\normalsize

\vskip 0.2in
\nocite{*}
\bibliography{mybib.bib}

\end{document}